\documentclass{article}

\PassOptionsToPackage{numbers, compress}{natbib}

\usepackage{microtype}
\usepackage{graphicx}
\usepackage{subcaption}
\usepackage{booktabs} % for professional tables
\usepackage{amsmath}
\DeclareMathOperator*{\argmax}{arg\,max}

\DeclareMathOperator*{\argtopk}{arg\,top-k}

\usepackage{amsfonts}
\usepackage{multirow}
\usepackage[table]{xcolor}
\usepackage[flushleft]{threeparttable}
\usepackage{chngcntr}
\usepackage{algorithm,algpseudocode}

\DeclareCaptionLabelFormat{supplementary}{#1 S#2}
\DeclareCaptionLabelFormat{algos}{Algorithm S#2}

\usepackage{hyperref}

\usepackage[preprint]{neurips_2021}

\title{B\textsuperscript{2}EA: An Evolutionary Algorithm Assisted by Two Bayesian Optimization Modules for Neural Architecture Search}

\author{%
	 	Hyunghun Cho \\
	 	GSCST,\\
	 	Seoul National University\\
	 	\texttt{webofthink@snu.ac.kr} \\
	 	 \And
		Jungwook Shin \\
		GSCST, Seoul National University\\
		\& SK Telecom Co., Ltd.\\
		\texttt{jungwook.shin@snu.ac.kr}	 	
	 	 \And
	 	 Wonjong Rhee \\
	 	GSCST \& GSAI \& AIIS,\\
	 	Seoul National University\\
	 	 \texttt{wrhee@snu.ac.kr}
}

\begin{document}

\maketitle

\begin{abstract}

The early pioneering Neural Architecture Search (NAS) works were multi-trial methods applicable to any general search space. The subsequent works took advantage of the early findings and developed weight-sharing methods that assume a structured search space typically with pre-fixed hyperparameters. 
Despite the amazing computational efficiency of the weight-sharing NAS algorithms, it is becoming apparent that multi-trial NAS algorithms are also needed for identifying very high-performance architectures, especially when exploring a general search space. 
%to overcome the limitations of the specialized NAS algorithms.
%
In this work, we carefully review the latest multi-trial NAS algorithms and identify the key strategies including Evolutionary Algorithm (EA), Bayesian Optimization (BO), diversification, input and output transformations, and lower fidelity estimation. To accommodate the key strategies into a single framework, we develop B\textsuperscript{2}EA that is a surrogate assisted EA with two BO surrogate models and a mutation step in between. To show that B\textsuperscript{2}EA is robust and efficient, we evaluate three performance metrics over 14 benchmarks with general and cell-based search spaces. Comparisons with state-of-the-art multi-trial algorithms reveal that B\textsuperscript{2}EA is robust and efficient over the 14 benchmarks for three difficulty levels of target performance.
The B\textsuperscript{2}EA code is publicly available at \url{https://github.com/snu-adsl/BBEA}.

\end{abstract}

% =============================================================
\section{Introduction}
\label{Introduction}

% 1. History of NAS
% 2. NAS types
% 3. One-shot problematic
% 4. Benchmarks and the NAS results (and we show Figure 1)
% * Recent benchmarks are not covered: NLP benchmarks, multi-fidelity benchmarks? 
% * Why DNNbench - new search space; not very well covered. Included so that HEBO looks bad?(answer: still this is DNN bench. not typical synthetic ML tasks)

% 5. Therefore, multi-trial NAS is also important.
% In general, multi-trial NAS is important for new problem, new application, new layer, ...
% 6. Why no multi-fidelity?
% - new ones are ...
% - we do not consider them because ... not possible for completely new problems...
% - We use ETR: we need defense for that
% 7. Our contributions
% - We follow the lineage of general multi-trial NAS: Zoph and Le's NAS, REA
% - we show that w/o using multi-fidelity or specialized encoding, we beat BOHB and BANANAS
% - Why we have chosen SAEA? - Robust

% Discussion:
% - Multi-fidelity is a future work 
% - Limitations of how we use EA (c.f. BANANAS' encoding;) we use no encoding. 

% Worries/Downsides: 
% 1. Figure 1
% - Yellow points look too bad: Add good points like ICLR2021
% - Multi-fidelity like DEHB, and?
%
% 2. Some metrics not covered: to be covered in metric section
% 3. Why validation performance instead of test performance?

% Related works:
% - SAEA
% - ?

\begin{figure}[t]
	\centering
	\includegraphics[width=\columnwidth]{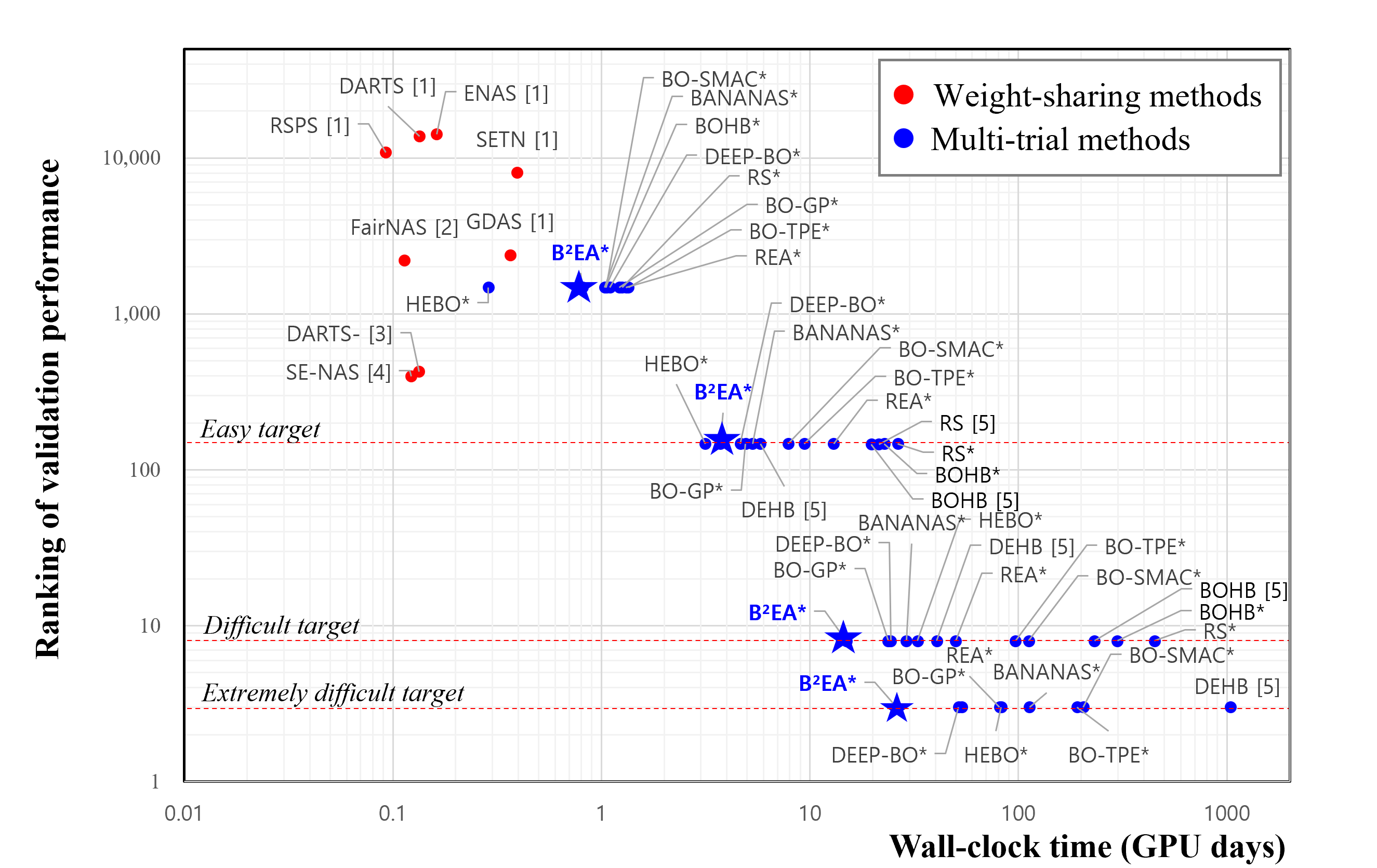}
	\caption{Comparison of NAS methods for NAS-Bench201-ImageNet16-120 task. Algorithm names with \textbf{*} indicate the evaluation results of this work. The results reported in five existing works are shown together: [1]~\citet{dong2020nasbench201}, [2]~\citet{chu2021fairnas}, [3]~\citet{chu2020darts}, [4]~\citet{hu2021improving}, and [5]~\citet{awad2021dehb}.}
	\label{fig:nas}
	%\vspace{-0.3cm}
\end{figure}

% % 1. History of NAS
The historical success of AlexNet~\cite{alex2012imagenet} triggered an exploding interest in deep learning 
and identification of high-performance neural architectures emerged as a crucial task. As the manual search continued as the dominant approach, the importance of automatic Neural Architecture Search~(NAS) has become evident. The early works focused on traditional Hyper-Parameter Optimization (HPO) methods such as Bayesian Optimization (BO)~\cite{snoek2012practical,hutter2011sequential,bergstra2011algorithms}, where the search space was general and included both basic hyperparameters (e.g., learning rate, regularization parameters~\cite{snoek2012practical}) and partial architectural elements (e.g., number of neurons in a layer, filter size, pooling option~\cite{bergstra2013making}). Following the early efforts, intensive architecture search was performed over entire-structured search spaces~\cite{zoph2016neural,baker2016designing,real2017large} where neural architectures achieving state-of-the-art performance were discovered at the cost of a substantially large computation. To reduce the computational burden, the research community quickly moved into a new paradigm where NAS algorithms focus on special search spaces such as cell-based, hierarchical, or morphism-based~\cite{he2021automl}, where pre-fixed hyperparameters are used. The most recent studies, however, recognized several downsides of the new paradigm and it is becoming apparent that a few distinct types of NAS algorithms are needed~\cite{li2020random,chu2020fair,wang2021rethinking}.

% 2. NAS types
%
% 2-1
NAS methods can be categorized in a few different ways, and comprehensive surveys were provided by~\citet{elsken2019neural} and~\citet{he2021automl}. An essential factor is the number and type of training that needs to be performed to complete a search because they directly affect the computational burden. 
% 2-2
\textit{One-shot} NAS trains an overparameterized supernet, and considers only the subnets as the search space~\cite{brock2017smash,bender2018understanding,zhang2020one,guo2020singlePO,hu2021improving}. Repeated full training can be avoided in this way, and \citet{liu2018darts} showed that efficient gradient descent methods can be utilized with a proper search space design. Some of the previous studies reported that one-shot 
methods can cause instability problems in practice~\cite{zela2019understanding,chu2020darts,wang2021rethinking}. 
%2-3 Weight-sharing vs multi-trial
\citet{dong2021nats} divides NAS methods into \textit{weight-sharing} and \textit{multi-trial}. 
Weight-sharing includes not only differentiable methods but also other approaches that utilize the concepts like 
%fine-tuning~\cite{bender2018understanding} 
sharing weights between models~\cite{brock2017smash,pham2018efficient} 
and progressive training~\cite{jaderberg2017population,liu2018progressive}. 
Multi-trial methods perform multiple full training over different configurations, and it is often believed that multi-trial requires a prohibitive computational cost. Multi-trial, however, is known to excel at identifying very high-performance architectures while weight-sharing can fail to do so.

% 3. Benchmarks and the NAS results (and we show Figure 1)
%
In Figure~\ref{fig:nas}, performance of NAS methods is shown for ImageNet16-120 task in NAS-Bench201~\cite{dong2020nasbench201}. The benchmark contains a total of 15,625 candidate configurations. The ranking of the neural architecture search result is shown because accuracy or error rate can be misleading depending on the performance distribution over the candidate configurations. For our results that are marked with \textbf{*}, the average performance was calculated over 500 runs with random seeds. 
NAS-Bench201-ImageNet16-120 is one of the 14 benchmarks that we study in this work, and its search space is cell-based. Therefore, it is possible to show the results of weight-sharing methods together. Note that most of the weight-sharing or one-shot methods cannot be evaluated for the benchmarks with a general search space~(e.g., benchmarks in HPO-Bench~\cite{klein2019tabular} and DNN-Bench~\cite{cho2020basic}).  

The one-shot methods marked with \textbf{[1]} in Figure~\ref{fig:nas} (RSPS~\cite{li2020random}, DARTS~\cite{liu2018darts}, ENAS~\cite{pham2018efficient}, SETN~\cite{dong2019one}, GDAS~\cite{dong2019searching}) show poor performance. The cell design of NAS-Bench201 includes skip-connection and the difference in search space is known to be the reason of the performance degradation~\cite{wang2021rethinking}. 
Recent robustified methods marked with \textbf{[2]},\textbf{[3]}, and \textbf{[4]} perform better, and DARTS-~\cite{chu2020darts} and SE-NAS~\cite{hu2021improving} achieve the best performance for the category of search budget below one GPU day. When the search budget is larger and a search for a high-performance architecture is desired, however, clearly multi-trial NAS algorithms can be imperative.

\begin{figure*}[!t]
	\centering
	\begin{subfigure}[b]{.3\textwidth}
		\includegraphics[width=\columnwidth]{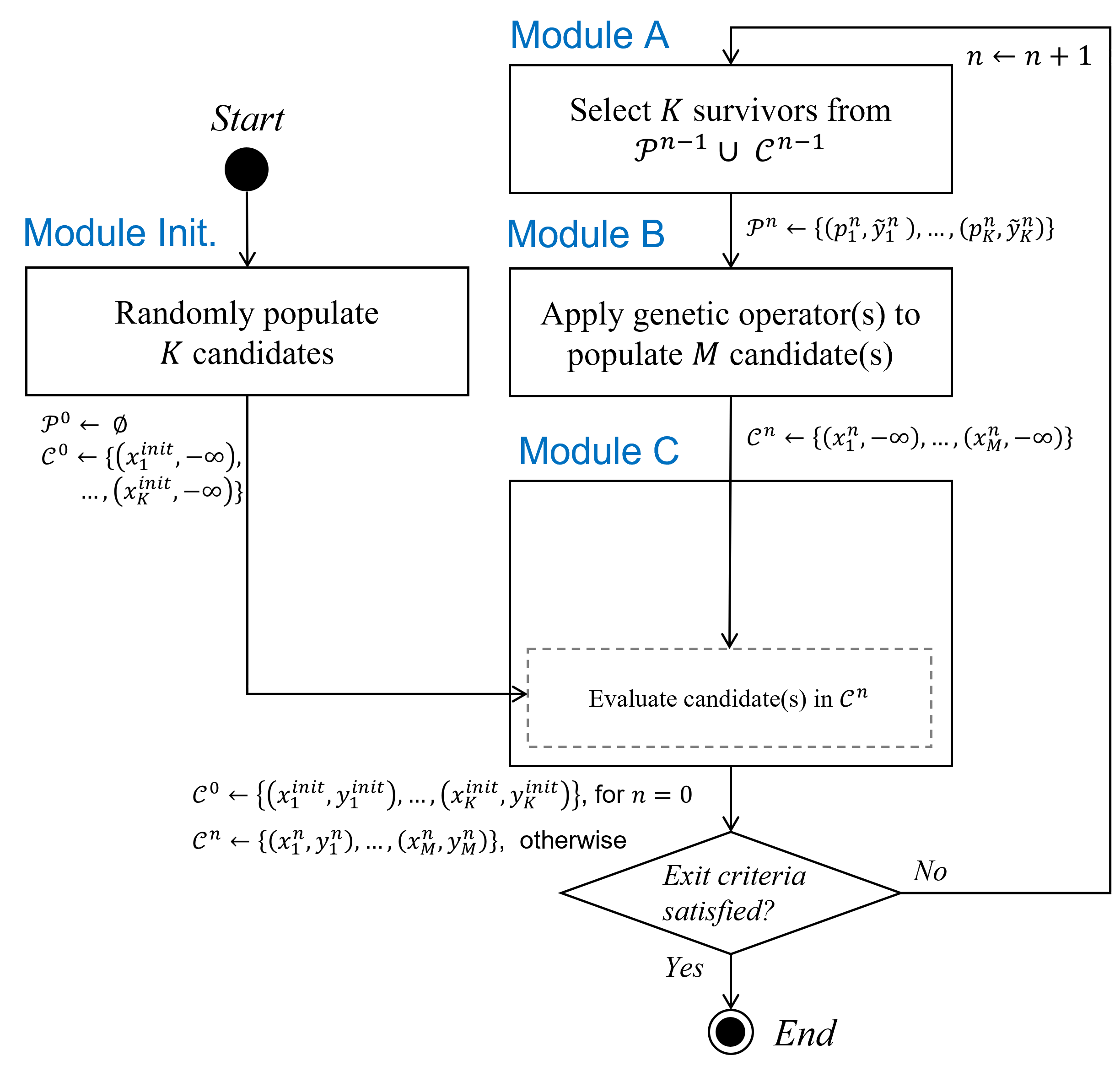}
		\caption{Evolutionary Algorithm}
		\label{fig0a}
	\end{subfigure}
	\hfill
	\begin{subfigure}[b]{.31\textwidth}
		\includegraphics[width=\columnwidth]{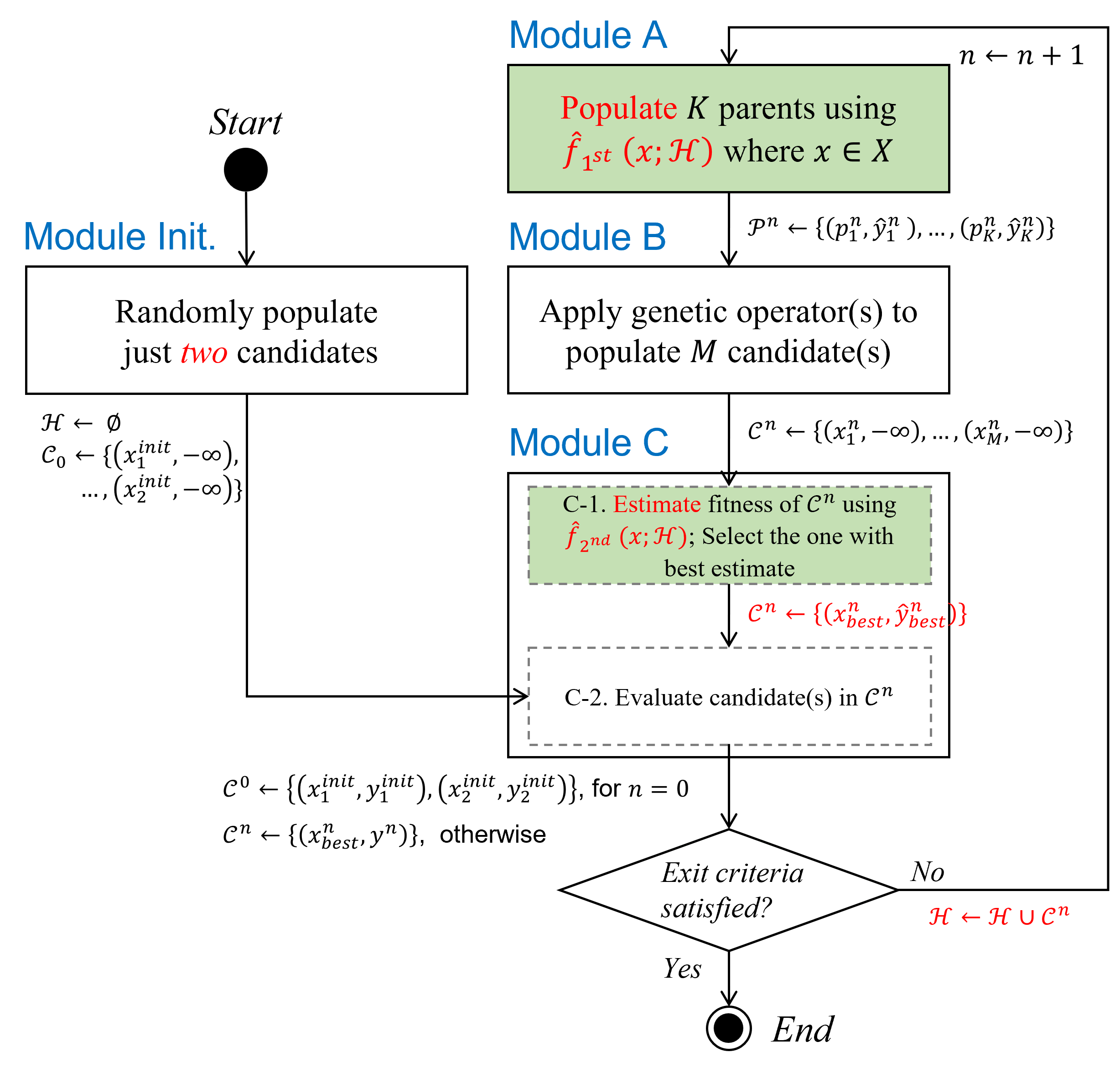}
		\caption{SAEA with two BO surrogate models} %\caption{SAEA with two BOs}
		\label{fig0b}
	\end{subfigure}
	\hfill
	\begin{subfigure}[b]{.34\textwidth}
		\includegraphics[width=\columnwidth]{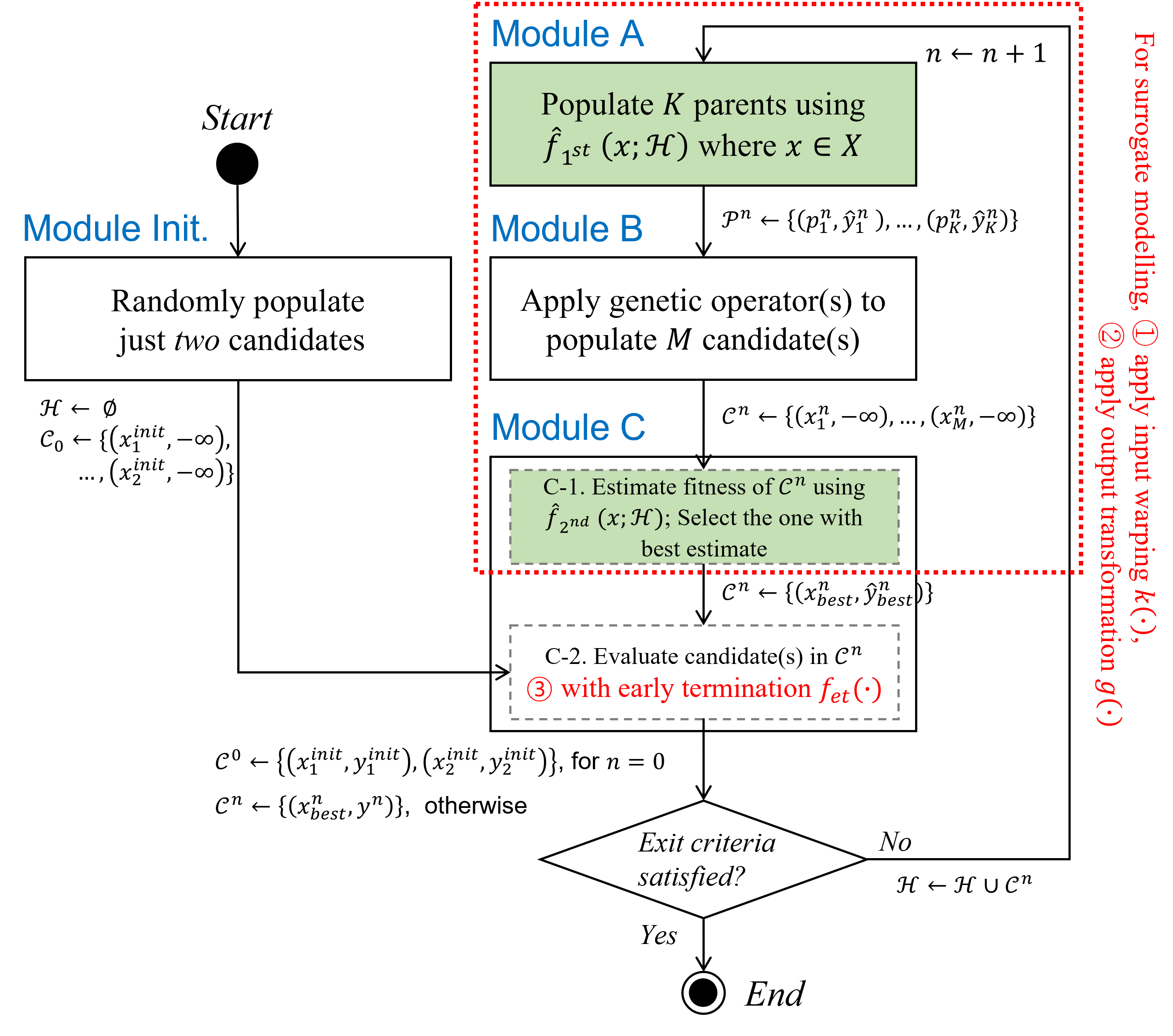}
		\caption{B\textsuperscript{2}EA}
		\label{fig0c}
	\end{subfigure}
	%\\[-1ex]
	\caption{Process flow diagrams of (a) basic EA, (b) enhanced EA with two BO surrogate models, and (c) further enhanced EA~(B\textsuperscript{2}EA) with an input warping strategy $k(\cdot)$, an output transformation strategy $g(\cdot)$, and a conservative early termination strategy $f_{et}(\cdot)$. 
		%Further details can be found in Section~\ref{sec:method}.
	}
	\label{fig0}
	%\vspace{-0.3cm}
\end{figure*}

In this study, we develop a multi-trial NAS algorithm that can be useful when searching for a high-performance architecture or when the search space is general. Our algorithm is named as \textbf{B\textsuperscript{2}EA} where the name comes from the use of two Bayesian Optimization (BO) surrogate models within an Evolutionary Algorithm (EA). For the `\textit{Difficult target}' in Figure~\ref{fig:nas}, it can be seen that our algorithm can complete the search within 15 GPU days while random search (RS) needs around 450 GPU days. The process flow diagram of an EA is shown in Figure~\ref{fig0a}. 
While EA is a search method, it has an interesting analogy with the three dimensions of NAS that were identified by \citet{elsken2019neural} - \textit{search space}, \textit{search method}, and \textit{evaluation method}.
In Figure~\ref{fig0a}, Module A performs the role of controlling \textit{search space} and Module C performs the role of \textit{evaluating} the selected candidate(s). 
This inspired us to adopt two BOs as shown in Figure~\ref{fig0b} and to come up with a new type of Surrogate Assisted Evolutionary Algorithm (SAEA)~\cite{jin2011surrogate}. The first BO in Module A is responsible for adaptively controlling the search space. The second BO in Module C is used for estimating the performance of many candidate configurations without any DNN training. To our best knowledge, this is the first time to have two BO models used together with an EA. While several options are available for the genetic operator in Module B, we choose mutation because it is simple and can complement BO models that can become biased~\cite{park2021robust}. 
Besides the main idea of combining two BO surrogate models and an EA, B\textsuperscript{2}EA embraces three key strategies from the latest works as shown in Figure~\ref{fig0c}. 
\section{Related works}
\label{sec:relatedworks}

\textbf{EA and BO:} 
For the conventional black-box optimization problems, Evolutionary Algorithm (EA) has been favored for the functions with cheap evaluation costs while Bayesian Optimization (BO) has been preferred for the functions with expensive evaluation costs. EA tends to provide a robust performance over a wide variety of optimization tasks, but it requires a large number of function evaluations. BO tends to be an efficient solution in terms of computational burden, but it works well only if the choice of modeling happens to match the problem characteristics and also a catastrophic modeling bias does not occur~\cite{cho2020basic}. 
While the early NAS works often considered BO as the search method, the extremely high cost of deep neural network training discouraged researchers from studying EA. Surprisingly, however, a simple EA method known as REA~\cite{real2019regularized} has been shown to be effective for automatically designing neural architectures and surpassed the performance of human designs~\cite{real2017large}.

\textbf{SAEA:} 
Surrogate-assisted evolutionary algorithm~(SAEA) is a subclass of EA, where a surrogate model is used to augment the prediction of fitness during the search. The motivation of SAEA is to reduce the function evaluation cost of the expensive problems~\cite{jin2011surrogate}.
%
% Simplify... 
%[HH]
%The existing works are typically identified two approaches: iterative sampling and %evolution control. Unlike iterative sampling methods that do not require evaluations of true expensive function, evolution control methods still require function evaluation while a surrogate model can be beneficial in offspring generation~\cite{emmerich2006single} or lazy evaluation~\cite{mlakar2015gp}.
%[HH]
% HH:simplified
A surrogate model can be used for offspring generation~\cite{emmerich2006single} or for lazy evaluation~\cite{mlakar2015gp}.
%
% Explain benchmark tasks in SAEA
In the EA community, typically SAEA works have been evaluated with synthetic object functions known as COCO benchmark~\cite{hansen2021coco}. 
%For real-world problems, computational fluid dynamics simulations or game related problems have been considered~\cite{volz2020bench}. 
% 
In our work, the target application of multi-trial NAS is distinct because of the exceptionally high cost of function evaluation. Compared to the computational cost of a full DNN training, the computational cost of surrogate modeling becomes negligible. This is one of our motivations for adopting two BO surrogate models into B\textsuperscript{2}EA. 
%With our best knowledge, B\textsuperscript{2}EA is the first SAEA algorithm that is assisted by two BO surrogate models. 
%To be precise, we adopt two \textit{diversified} BO models as will be explained in Section~\ref{sec:cooperation}. 

\textbf{Input and output transformations:} 
HEBO is the winner of the black-box optimization (BBO) challenge at NeurIPS 2020~\cite{turner2020Bayesian}, where the challenge focused on evaluating derivative-free optimizers for tuning the hyperparameters of machine learning models.
To deal with the complexities associated with the competition datasets, both input and output nonlinear transformations were used~\cite{cown2020emprical}. 
More specifically, they utilized an input warped GP~\cite{snoek14} to handle non-stationary functions and output transformations~\cite{rios2019compositionally} to model non-Gaussian data.
DEEP-BO also utilizes an output transformation and explains it as one of the four basic BO enhancement strategies~\cite{cho2020basic}. In our work, we integrate input warping and output transformation as in HEBO.

\textbf{Multi-fidelity:}
While multi-fidelity techniques~\cite{domhan2015speeding,falkner18bohb} are becoming popular, they have been under-utilized in the NAS community~\cite{yan2021bench}. This is because the estimation of validation performance based on lower fidelity observations can be inaccurate~\cite{golovin2017google,elsken2019neural}. For instance, it has been pointed out that the use of batch-norm can cause a large gap between a low fidelity observation based estimation and the corresponding high fidelity evaluation~\cite{choi2018difficulty,cho2020basic}. For B\textsuperscript{2}EA, we choose to stay conservative and integrate the \textit{early termination rule}~\cite{cho2020basic} that is known to have a minimal influence on NAS performance, especially when searching for an architecture of high performance.

\section{Cooperation between two BO models}
\label{sec:cooperation}

\begin{figure*}[t]	
	\begin{subfigure}[b]{0.19\textwidth}
		\includegraphics[width=\columnwidth]{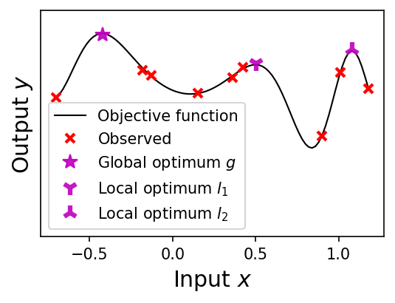}
		\caption{Search history}
		\label{fig2:a}
	\end{subfigure}
	\hfill			
	\begin{subfigure}[b]{0.193\textwidth}
		\includegraphics[width=\columnwidth]{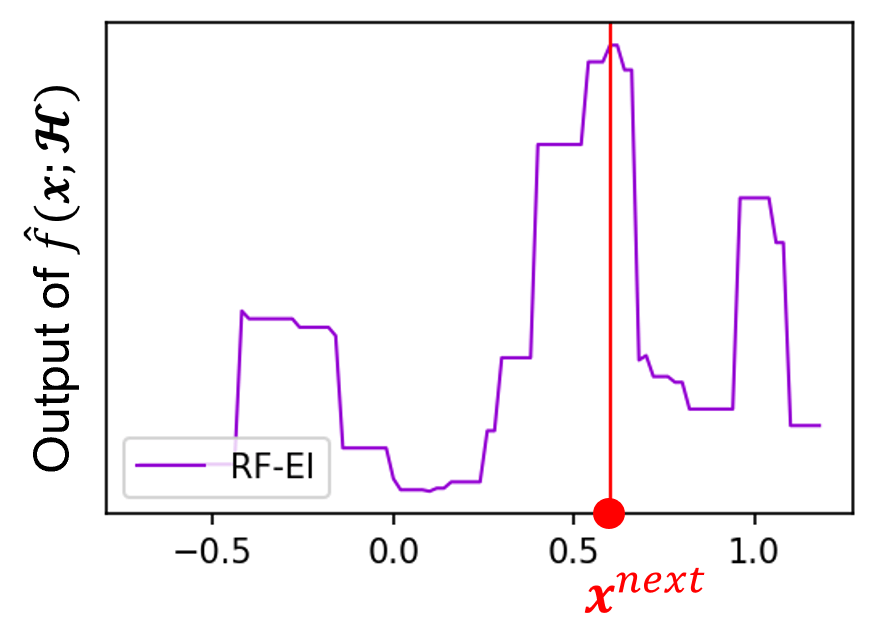}
		\caption{Selected by RF-EI}
		\label{fig2:b}
	\end{subfigure}
	\hfill
	\begin{subfigure}[b]{0.18\textwidth}
		\centerline{\includegraphics[width=\columnwidth]{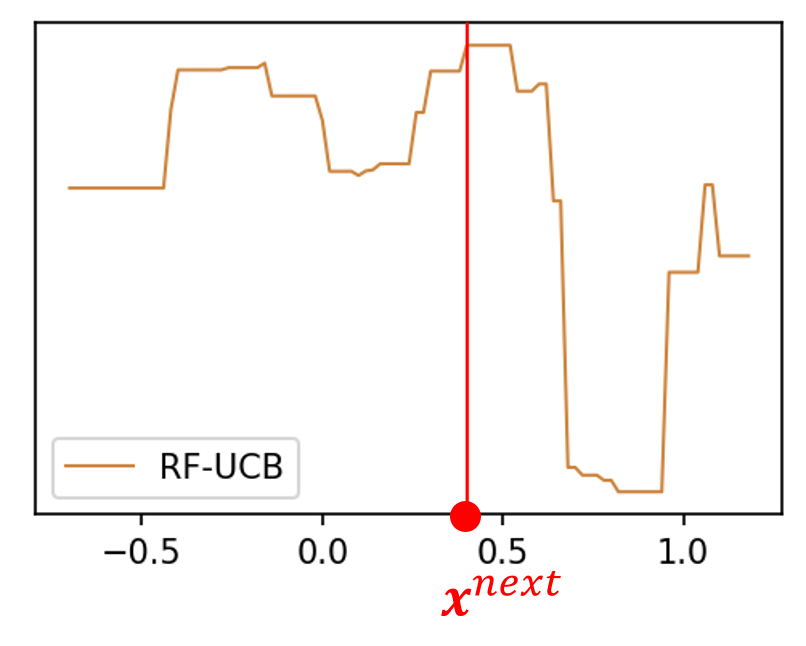}}
		\caption{Selected by RF-UCB}
		\label{fig2:c}
	\end{subfigure}
	\hfill	
	\begin{subfigure}[b]{0.18\textwidth}
		\centerline{\includegraphics[width=\columnwidth]{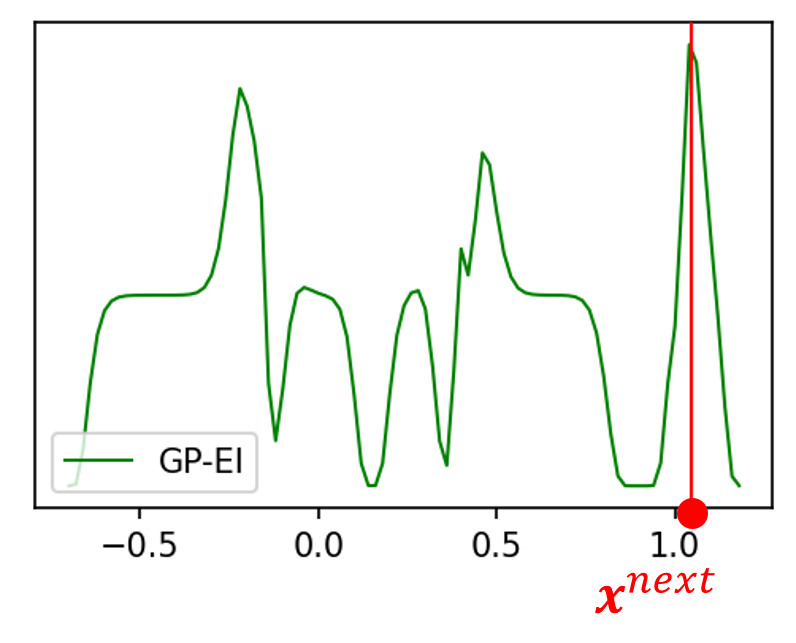}}
		\caption{Selected by GP-EI}
		\label{fig2:d}
	\end{subfigure}
	\hfill
	\begin{subfigure}[b]{0.18\textwidth}
		\centerline{\includegraphics[width=\columnwidth]{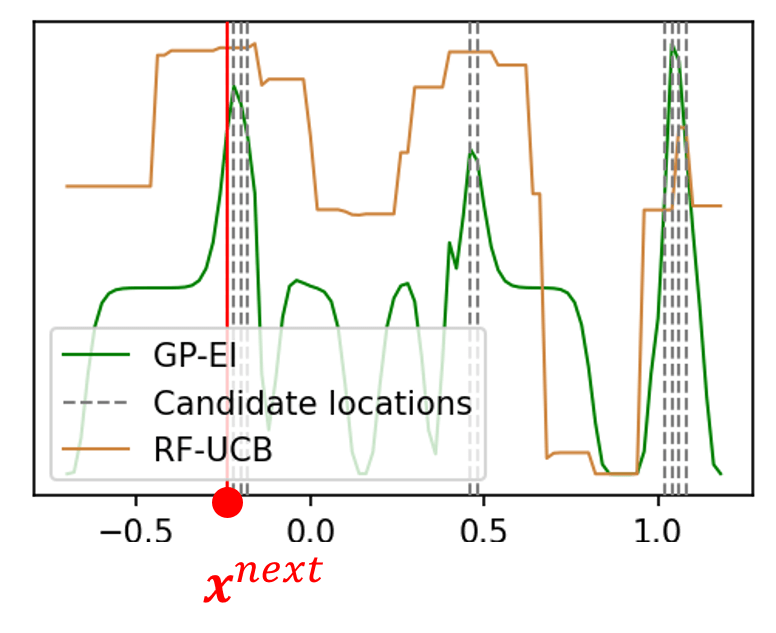}}
		\caption{Cooperation}
		\label{fig2:e}
	\end{subfigure}
	%\\[-3ex]	
	\caption{
		Enhancement through cooperation. An optimization problem with one-dimensional search space is shown in (a), where nine candidates have been evaluated so far. Global and local optima are shown as well. Compared to the individual decisions in (b), (c), and (d), a cooperative decision can lead to a better candidate selection, as shown in (e). The cooperation is achieved by the first BO model (GP-EI in this example) constraining the search space to its top $M\mathtt{=}10$ points and the second BO model (RF-UCB in this example) choosing its best only from these $M$ candidate points. Because the two models have different modeling characteristics, they end up compromising on the final candidate selection.} 
	\label{fig2}
	%\vspace{-0.4cm}	
\end{figure*}

Before moving forward, we explain the key design concepts of B\textsuperscript{2}EA's two BO surrogate models. 

\textbf{Diversified BO:} While BO can be an efficient strategy even when the cardinality of search history $|\mathcal{H}|$ is small, it can also be  biased and catastrophic. 
A typical BO can be myopic because it considers no more than a single step in the future~\cite{gonzalez2016glasses}, and its common acquisition functions can lead to an overly greedy optimization~\cite{shahriari2016taking}. To address this drawback of BO, we adopt a diversified BO~\cite{cho2020basic}. Diversified BO relies on multiple BO models, as in the ensemble, and it can escape from a local optimum simply by rotating over a set of BO models. Once an informative sample is added to the search history $\mathcal{H}$, the previously catastrophic models can become effective as well. Because adaptive diversification was shown to be not so helpful in~\citet{cho2020basic}, we adopt round-robin diversification for the first BO module and uniform random diversification for the second BO module. 
%As in~\citet{cho2020basic}, we use Gaussian Process (GP)~\cite{snoek2012practical} and Random Forest (RF)~\cite{hutter2011sequential} as the base regression models and utilize PI, EI, and UCB~\cite{kushner1964new,mockus1978toward,srinivas2010gaussian} as the acquisition strategies to come up with six BO surrogate models for diversification. %
An example of catastrophic behavior can be found in Figure~S2 of Supplementary D.

\textbf{Cooperation between two BO surrogate models:} In Figure~\ref{fig0b}, the first BO, $\hat{f}_{1^{st}}$, is responsible for dynamic adjustment of the search space and the second BO $\hat{f}_{2^{nd}}$ is responsible for the performance estimation of the $M$ candidates. While the two modules have been designed for two distinct goals, they can also be understood as a group. In Figure~\ref{fig2}, it can be seen that the first BO's candidate selections are regularized by the second BO, which has a different opinion from the first BO. In addition to the diversification of BO, this cooperation plays a key role in making B\textsuperscript{2}EA robust over a variety of optimization tasks.

% =============================================================
\section{Proposed method: B\textsuperscript{2}EA}
\label{sec:method}

% A NAS problem with a general search space can be formulated as a black-box optimization problem below.
% \begin{equation*}
%     \mathbf{x}^* = \underset{\mathbf{x}\ \in\ \mathcal{X}}{\argmin} {f(\mathbf{x})},  
% \end{equation*}
% $\mathbf{x}$ can be decomposed into $\mathbf{x} = (\mathbf{x}_A, \mathbf{x}_H)$, $\mathbf{x}_A$ is the set of architecture variables, $\mathbf{x}_H$ is the set of non-architecture variables, $\mathcal{X}$ is the search space, and $f$ is the function to be maximized. 

Compared to the weight-sharing NAS algorithms, multi-trial NAS algorithms require far more computational time. This means a failed search can be much more painful for a multi-trial NAS. Therefore, we have chosen \textit{robustness} over a broad range of complicated black-box optimization problems as the most important goal when developing B\textsuperscript{2}EA. To achieve high robustness while relying on BO for computational efficiency, we adopt three strategies - diversification at a single BO level, cooperation at multiple BO level, and mutation at EA level. In this section, we first explain the SAEA framework that implements the three robustness strategies and then briefly explain the additional enhancement strategies. 

\subsection{SAEA framework}
\label{subsec:method_SAEA}

The insights behind B\textsuperscript{2}EA includes dynamic search space control with the first diversified BO and cheap function output estimation with the second diversified BO. The process flow diagram of B\textsuperscript{2}EA is shown in Figure~\ref{fig0c}, and the pseudo-code is shown in Algorithm S1 of Supplementary B. The main steps of Algorithm S1 can be explained as follows:

\textbf{Initialization:} For the basic EA in Figure~\ref{fig0a}, the initial population $\mathcal{C}^0$ consists of randomly selected $K$ candidates. This simple initialization scheme poses two main problems for functions with high evaluation costs. First, all $K$ candidates must be evaluated. 
Second, the random selection of $K$ candidates can be highly inefficient from BO's perspective because BO can perform much better by sequentially selecting the $K$ candidates. 
Therefore, we modify the initialization step of EA to choose only two candidates $\mathbf{x}^{init}_{1}$ and $\mathbf{x}^{init}_{2}$ as is usually done in BO. For a typical BO with a continuous search space, one candidate is chosen near the center, and the other is chosen near the boundary corner. In our implementation, however, we have randomly chosen the two candidates because some of the search dimensions are not continuous and it can be ambiguous to identify near-center or near-boundary candidates.

\textbf{Module A:} For the genetic algorithm, which is a type of EA, optimization is performed over many generations, and the building-block hypothesis is often believed to be the reason for its frequent success. 
In the building-block hypothesis, \textit{``short, low order, and highly fit schemata are sampled, recombined, and resampled to form strings of potentially higher fitness"}~\cite{golberg1989genetic}. In Figure~\ref{fig0a}, the population in module A is iteratively updated using tournaments or other methods for selecting the survivors, and the population is expected to retain highly fit schemata as it evolves.
% Other regularization methods for update of $\mathcal{P}$
Traditional EA methods~\cite{zhu2019eena,real2019regularized} have a strategy to remove some candidates in $\mathcal{P}^n$ in consideration of limited computational resources.
In contrast, B\textsuperscript{2}EA dynamically generates a fresh set of $K$ candidates for each iteration $n$ using the first diversified BO $\hat{f}_{1^{st}}$. 
The history $\mathcal{H}$ is used for modeling $\hat{f}_{1^{st}}$, and the top $K$ performing candidates from the search space $\mathcal{X}$ are chosen based on the performance estimates of $\hat{f}_{1^{st}}$.
In this way, we are giving up on the building block hypothesis that may take a long time to be effective, and instead relying on the modeling capability of the BO, which can be useful even with a small number of evaluated candidates. 
Because diversified BO can be highly accurate even when $|\mathcal{H}|$ is small, this replacement can be a key enhancement factor for  functions with high evaluation costs.
%%% Update of recent works from autoML survey paper 
%In contrast, \citet{camero2021bayesian} utilize BO to gradually reduce the chance of infeasible candidates instead of choosing promising candidates.
%

\textbf{Module B:} We keep the basic functionality of the EA's genetic operation intact, but we consider only mutation in this work. There are many other possibilities, for instance, crossover or mutation followed by crossover. 
However, our main interest is to investigate the benefits and robustness of integrating two diversified BO modules into a basic EA framework, and thus we intentionally keep this part simple. In Figure~\ref{fig0b}, module B is responsible for generating $M$ candidates to form $\mathcal{C}^n$ by applying mutation to the $K$ parents in $\mathcal{P}^n$. Assuming $K \geq M$ for simplicity, B\textsuperscript{2}EA randomly selects $M$ candidates from $\mathcal{P}^n$ with a mutation applied to a randomly selected dimension of $\mathbf{x}$ for each candidate to generate $M$ offsprings that form $\mathcal{C}^n$. 
We follow the intensification strategy of \citet{hutter2011sequential} to handle mixed-type hyperparameters.

\textbf{Module C:} For the basic EA in Figure~\ref{fig0a}, module C simply evaluates all $M$ candidates. Obviously, this is not desirable for functions with high evaluation costs, including NAS. To alleviate this problem, we use the second BO module $\hat{f}_{2^{nd}}$ to cheaply estimate the true function values of the $M$ candidates. Upon completion of the estimation, the candidate with the best estimate is chosen as the final candidate and its true function value is evaluated. In an extreme scenario, modules A and B can be configured such that the entire search space $\mathcal{X}$ is contained in $\mathcal{C}^n$. In this case, modules A and B do not perform any meaningful operation, and the entire EA framework collapses to a BO algorithm (module C only). In this sense, B\textsuperscript{2}EA can be considered as a class of algorithms that include BO as a subclass. 

\textbf{Exit criteria and update of $\mathcal{H}$:} Exit criteria are configured in the same manner as in any EA or BO algorithm. In our experiments, we consider fixed budget targets and fixed performance targets. History $\mathcal{H}$ is updated by adding $(\mathbf{x}^{n}_{best}, \mathbf{y}^{n})$, as in any BO.

\subsection{Further enhancements}
\label{subsec:method_further}
Besides the main development of SAEA with two BO surrogate models, we adopt input warping~\cite{snoek14} and output transformation~\cite{rios2019compositionally} as in HEBO. Also, we adopt the early termination rule~\cite{cho2020basic} from DEEP-BO. The resulting B\textsuperscript{2}EA is shown in Figure~\ref{fig0c}.

\section{Experiments}
\label{sec:experiments}
%\vspace{-1mm}

% % Why we use validation performance instead of test performance.
% While test performance of the best architecture found with a given budget is typically used~\cite{ying2019bench}, validation performance is used as well~\cite{dong2020nasbench201} because optimization is guided by the validation performance. Especially, validation performance can be less biased to interpret the intermediate performance of multi-fidelity optimizers that dynamically control training epochs of candidates~\cite{falkner18bohb,awad2021dehb}.
% Furthermore, because the performance differences among top candidates tend to be highly reduced, the use of ranking is better to reveal the optimizer performance.
% % correlation coefficient between validation set and test set of ImageNet16-120: 0.997
% % 
% Unlike the total cost of a search is shown for single-trial methods, for multi-trial methods, the expected time to achieve four different target settings are shown that achieve very easy (10\%), easy (1\%), difficult (0.05\%), and extremely difficult (0.02\%) targets, respectively. 

%While B\textsuperscript{2}EA might work well for any given black-box optimization problem, we exclusively focus on DNN problems whose evaluation cost (i.e. training cost) is notoriously high. 

We evaluate three NAS benchmark datasets that are publicly accessible\footnote{The three datasets are the NAS benchmark datasets that were available at the time of our investigation.}. The three datasets include optimization problems with architecture parameters $\mathbf{x}_A$ only and a mixture of parameters with both $\mathbf{x}_A$ and hyperparameters $\mathbf{x}_H$.
We compare B\textsuperscript{2}EA with nine benchmark algorithms. To ensure that our results are not misleading due to the choice of performance metric, we evaluate three different performance metrics. All our experimental results are based on \textbf{500} repetitions with random seeds, and the wall-clock times of both black-box function evaluation time and the NAS algorithm run-time are reflected. The details for reproducing our results can be found in Supplementary C.

\begin{table*}[t]
	
	\centering
	\vspace{-2mm}
	\caption{Tabular benchmarks for NAS}
	\label{tab1}
	\scriptsize
	\resizebox{0.95\linewidth}{!}{
		\begin{threeparttable}
			
			\begin{tabular}{l ccc ccc ccc r}
				\toprule
				\multirow{2}{*}{\textbf{\begin{tabular}[c]{@{}l@{}}Benchmark\\ name\end{tabular}}} & \multicolumn{3}{c}{\textbf{Number of tasks}}  & \multicolumn{6}{c}{\textbf{Number of hyperparameters}} & \multicolumn{1}{c}{\multirow{2}{*}{\textbf{\begin{tabular}[r]{@{}r@{}}Full candidate\\set size\end{tabular}}}} \\
				\cmidrule{2-10}
				& MLP & CNN & RNN &  Categorical & Continuous & Discrete & $|\mathbf{x}_A|$ & $|\mathbf{x}_H|$ & $|\mathbf{x}|$ & \multicolumn{1}{c}{} \\
				\midrule
				HPO-Bench %~\cite{klein2019tabular}%
				& 4 & - & - & 3 & 3 & 3 & 4 & 5 & 9 & 62,208 \\
				NAS-Bench-101 %~\cite{ying2019bench}% 
				& - & 1 & - & 5 & - & 21 & 26 & - & 26 &  423k \\
				NAS-Bench-201 %~\cite{dong2020nasbench201}% 
				& - & 3 & - & 6 & - & - & 6 & - & 6 & 15,625 \\
				DNN-Bench %~\cite{cho2020basic}%
				& - & 5 & 1 & 0$\sim$3 & 2$\sim$5 & 2$\sim$6 & 1$\sim$7 & 3$\sim$7 & 7$\sim$10 & 20,000$\dagger$ \\
				\bottomrule
			\end{tabular}
			\begin{tablenotes}
				\item \textit{$\dagger$Except for CIFAR10-ResNet, which has 7,000 configurations.}
			\end{tablenotes}
			\vspace{-3mm}
		\end{threeparttable}	
	}
\end{table*}

\textbf{Pre-evaluated NAS benchmark datasets:} 
To facilitate scientific research on developing HPO and NAS algorithms, tabular benchmark datasets can be used~\cite{lindauer2020best,eggensperger2021hpobench}. 
Our experiments are also based on these pre-evaluated and multi-fidelity benchmarks, and the NAS benchmarks we used are summarized in Table~\ref{tab1}. 
The HPO-Bench~\cite{klein2019tabular} focuses on MLP regression problems with mixed variables $(\mathbf{x}_A, \mathbf{x}_H)$, and it was used as the main performance benchmark in \citet{lee2020efficient}.
The NAS-Bench includes NAS-Bench-101~\cite{ying2019bench} and NAS-Bench-201~\cite{dong2020nasbench201}, and they focus on the architecture variables $\mathbf{x}_A$. 
NAS-Bench-101 was used as the main performance benchmark in \citet{white2019bananas,wang2020nas} and as a benchmark in \citet{letham2020re,ru2020bayesian}. 
Recently, NAS-Bench-201 was used as a benchmark in \citet{zhang2020one,chu2020darts,chu2021fairnas,hu2021improving,awad2021dehb}.
DNN-Bench contains five CNN tasks and one RNN task with mixed variables $(\mathbf{x}_A, \mathbf{x}_H)$, and it was introduced and used in~\citet{cho2020basic}. 
Together, the four benchmarks include 14 DNN optimization tasks with a wide range of task characteristics.

\textbf{Benchmark algorithms:}
Random Search (RS)~\cite{bergstra2012random} is used as the baseline. GP~\cite{snoek2012practical}, SMAC~\cite{hutter2011sequential}, and TPE~\cite{bergstra2011algorithms} are BO algorithms known for their high performance for HPO. To emphasize that they are BO algorithms, we have referred to them as BO-GP, BO-SMAC, and BO-TPE in our work. BOHB~\cite{falkner18bohb} is a state-of-the-art multi-fidelity algorithm that combines hyperband~\cite{li2017hyperband} and TPE.
HEBO is the winner of BBO challenge at NeurIPS2020, and it is known to perform very well for ML tasks~\cite{turner2020Bayesian}. DEEP-BO is a diversified BO method developed for DNN tasks~\cite{cho2020basic}. 
BANANAS~\cite{white2019bananas} is a BO-based algorithm for NAS tasks that utilizes meta-neural networks. 
Regularized Evolution Algorithm (REA)~\cite{real2019regularized} is a state-of-the-art EA algorithm that discovered AmoebaNets.

\textbf{Three performance metrics:}
To minimize the risk of biased interpretation of the experimental results, we evaluate three different performance metrics. 
The first is \textit{intermediate regret} $r_t$ at wall clock time $t$, which is a popular and commonly used metric in the HPO and NAS research community. The definition is $r_t = |f(\mathbf{x}^*)-f(\tilde{\mathbf{x}}_t)|$, where  $\mathbf{x}^*$ is the optimal configuration in $\mathcal{X}$, and $\tilde{\mathbf{x}}_t$ is the best configuration found in time $t$. The optimal configuration $\mathbf{x}^*$ should be known and used only for the purpose of evaluating $r_t$.
The main disadvantage of $r_t$ is that its value is dependent on the actual value of the loss metric, making the analysis and interpretation of $r_t$ difficult because they are specific to each individual task.
Note that the validation performance is used as a proxy for the true performance $f(\mathbf{x})$ for the pre-evaluated NAS benchmark datasets. 

The second and third metrics are \textit{success rate} $\mathbb{P}(\tau \le t)$ and \textit{expected time} $\mathbb{E}[\tau]$, which were adopted in \citet{cho2020basic}. 
Consider a random variable $\tau$ that represents the time required to achieve the target performance $c$. 
Then, $\mathbb{E}[\tau]$ is simply the expected time to achieve $c$, and it becomes a fixed-target metric. Now, if we choose a time budget $t_b$, then $\mathbb{P}(\tau \le t_b)$ is the success rate of achieving performance $c$ before $t$ reaches $t_b$. 
Therefore, $\mathbb{P}(\tau \le t_b)$ becomes a fixed-budget metric, which is also associated with a fixed target $c$. 
While the two metrics are intuitive and easily interpretable, they can be sensitive to the choices of $c$ and $t_b$. 
To make the two metrics as neutral as possible, we determined $c$ and $t_b$ in a systematic and consistent manner over all 14 benchmarks. 
For $c$, we chose targets $c_{e}$, $c_{d}$ and $c_{x}$ in consideration of the difficulties for each pre-evaluated dataset, where $c_{e}$, $c_{d}$ and $c_{x}$ are chosen as the top 1 \%, top 0.05 \% and top 0.02 \% performances of all configurations in the pre-evaluated dataset, respectively. 
%
%For $c$, we chose an easy target $c_{e}$ and a difficult target $c_{d}$ for each pre-evaluated dataset, where $c_{e}$ is chosen as the top 1 \% performance and $c_{d}$ as the top 0.05 \% performance of all configurations in the pre-evaluated dataset.
%
%Between the two, $c_{d}$ provides a more important and realistic metric for common HPO and NAS tasks, where an extremely high performance is desired. 
For $t_b$ of each dataset, we systematically chose it as the time for the best-performing algorithm to achieve a success rate of 99 \%. 
Because $c_{e}$, $c_{d}$,  $c_{x}$ and $t_b$ are chosen with a fixed method independent of the dataset, we believe that the expected time $\mathbb{E}[\tau]$ and the success rate $\mathbb{P}(\tau \le t_b)$ serve as fair metrics. 
%Because $c_{e}$, $c_{d}$, and $t_b$ are chosen with a fixed method independent of the dataset, we believe that the expected time $\mathbb{E}[\tau]$ and success rate $\mathbb{P}(\tau \le t_b)$ serve as fair metrics. 

Finally, there are cases where an algorithm fails to achieve $c$ even after running for the maximum allowed run time $t_{max}$. 
In this case, it is obviously misleading to consider $t_{max}$ as the time to achieve $c$. 
To address this issue,
the expected time is adjusted to 
$ \frac{\mathbb{P}(\tau \le t_{max}) \mathbb{E}_s[\tau]   +       (1 - \mathbb{P}(\tau \le t_{max}))t_{max}}{\mathbb{P}(\tau \le t_{max})}$ as in \citet{auger2005}, 
where $\mathbb{E}_s[\tau]$ is the expected time of successful cases only.

%TODO: consider the experiments where $t_b=t_m$!!!

\textbf{Settings of B\textsuperscript{2}EA:}
We use $K\mathtt{=}M\mathtt{=}10$, and mutation is applied with the probability of $0.5$ for all of our experiments. This could have been extensively tuned for performance improvement, but we performed only a minimal amount of sanity checking with the goal of obtaining generalizable results.
For diversified BO, we followed \cite{cho2020basic} and used \{GP-EI, GP-PI, GP-UCB, RF-EI, RF-PI, RF-UCB\} as individual BO models (i.e., $\{\hat{f}_1,\dots \hat{f}_{6}\}$)
 with input warping and output power transformation in \citet{turner2020Bayesian}.
We also applied the early termination rule in \citet{cho2020basic} that is a very conservative multi-fidelity technique. 
For handling categorical variables, we used the adjacency matrix encoding of \citet{white2020study} for NAS-Bench and one-hot encoding for the rest.

\begin{figure*}[ht]
	%\vspace{-2mm}
	\centering
	\begin{subfigure}[b]{.8\textwidth}
		\includegraphics[width=.2\textwidth]{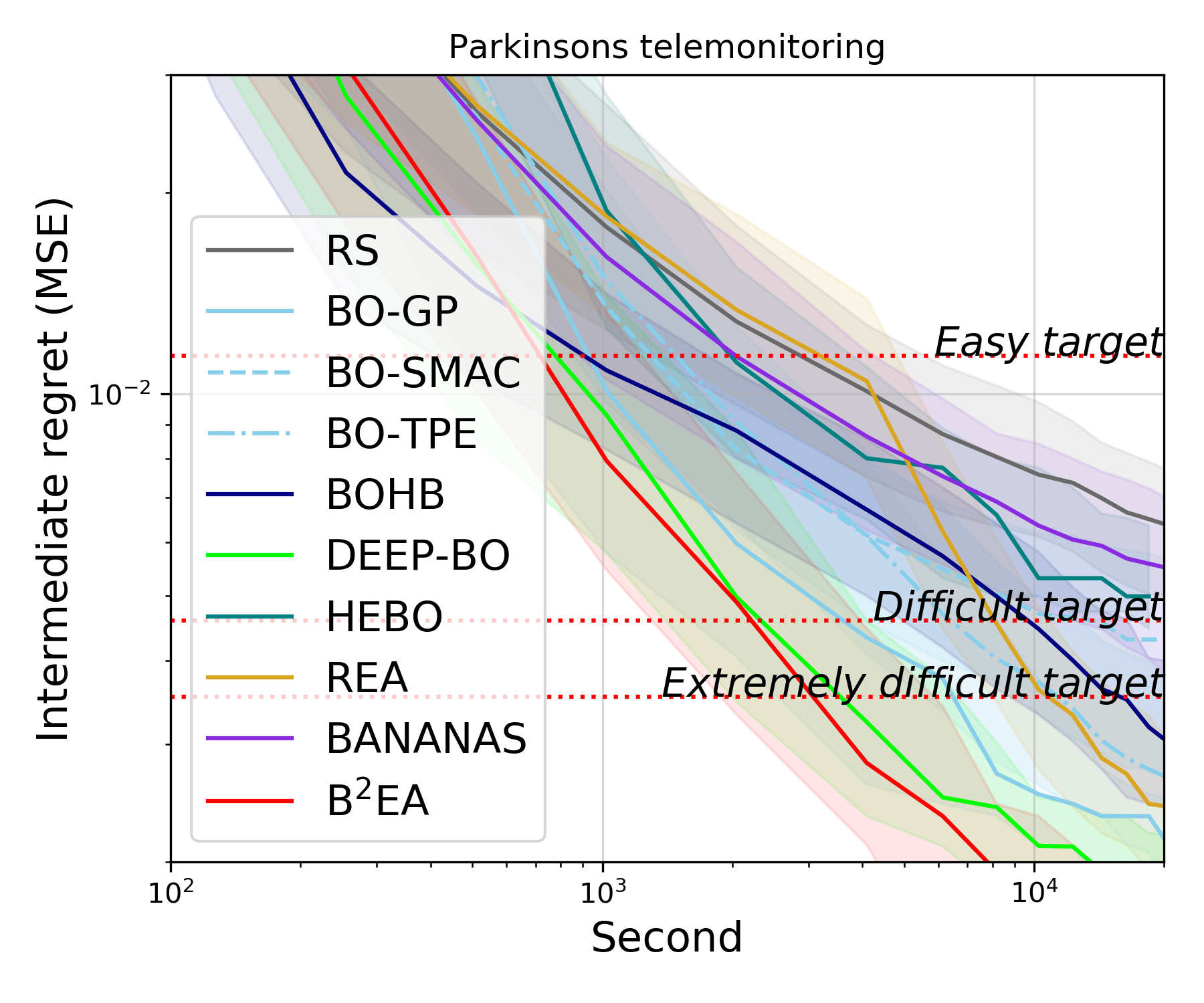}
		\hfill
		\includegraphics[width=.2\textwidth]{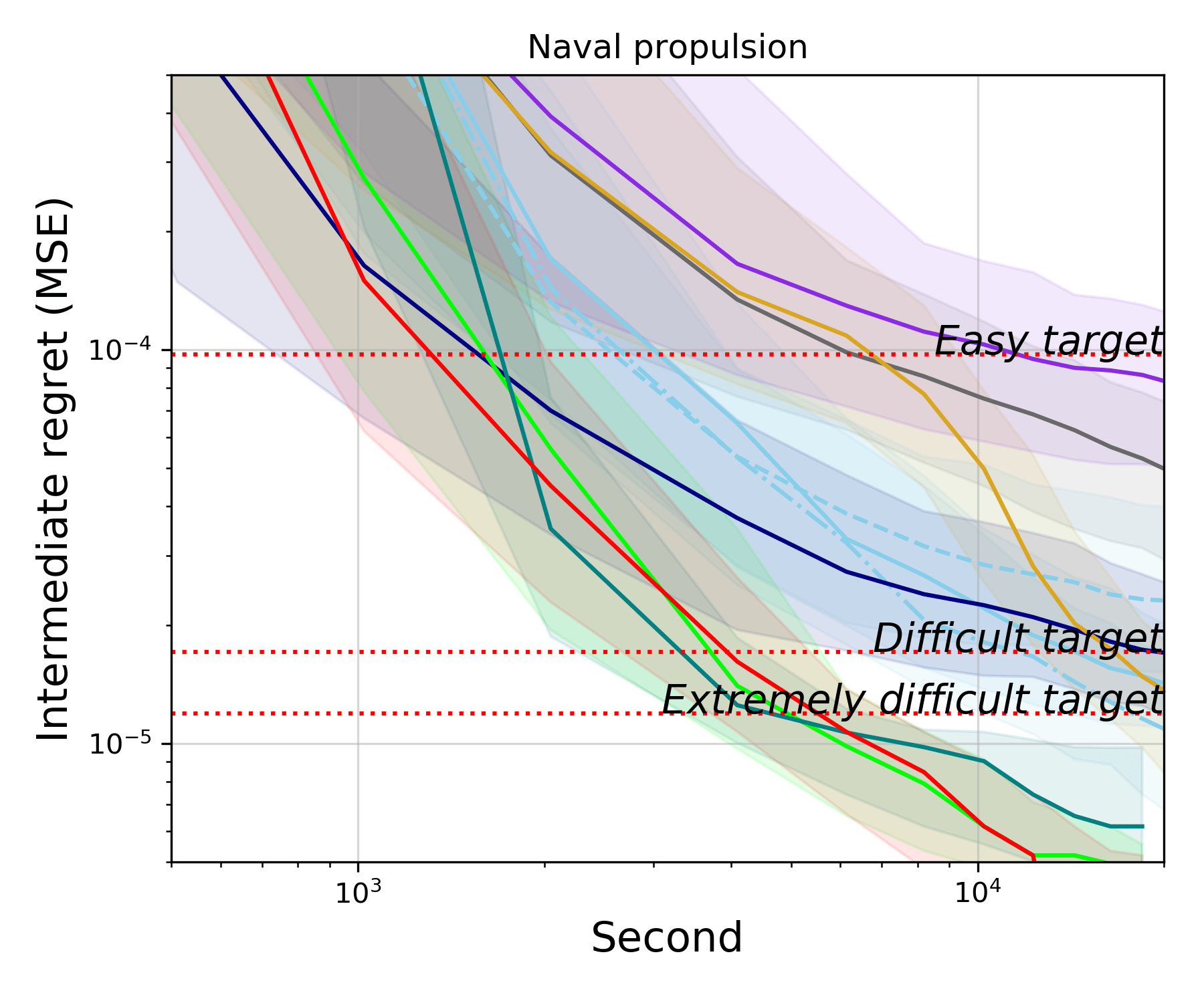}
		\hfill
		\includegraphics[width=.2\textwidth]{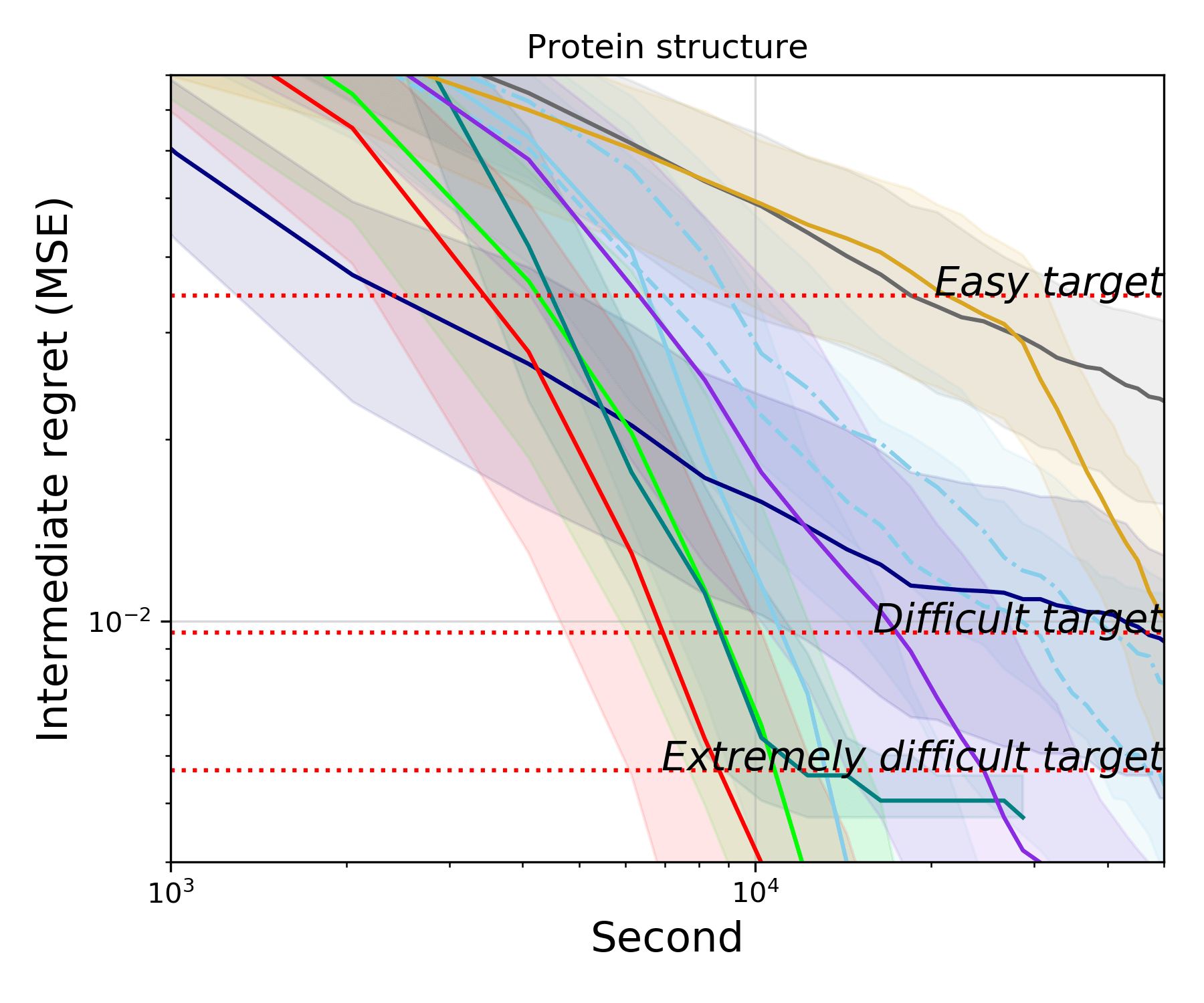}
		\hfill
		\includegraphics[width=.2\textwidth]{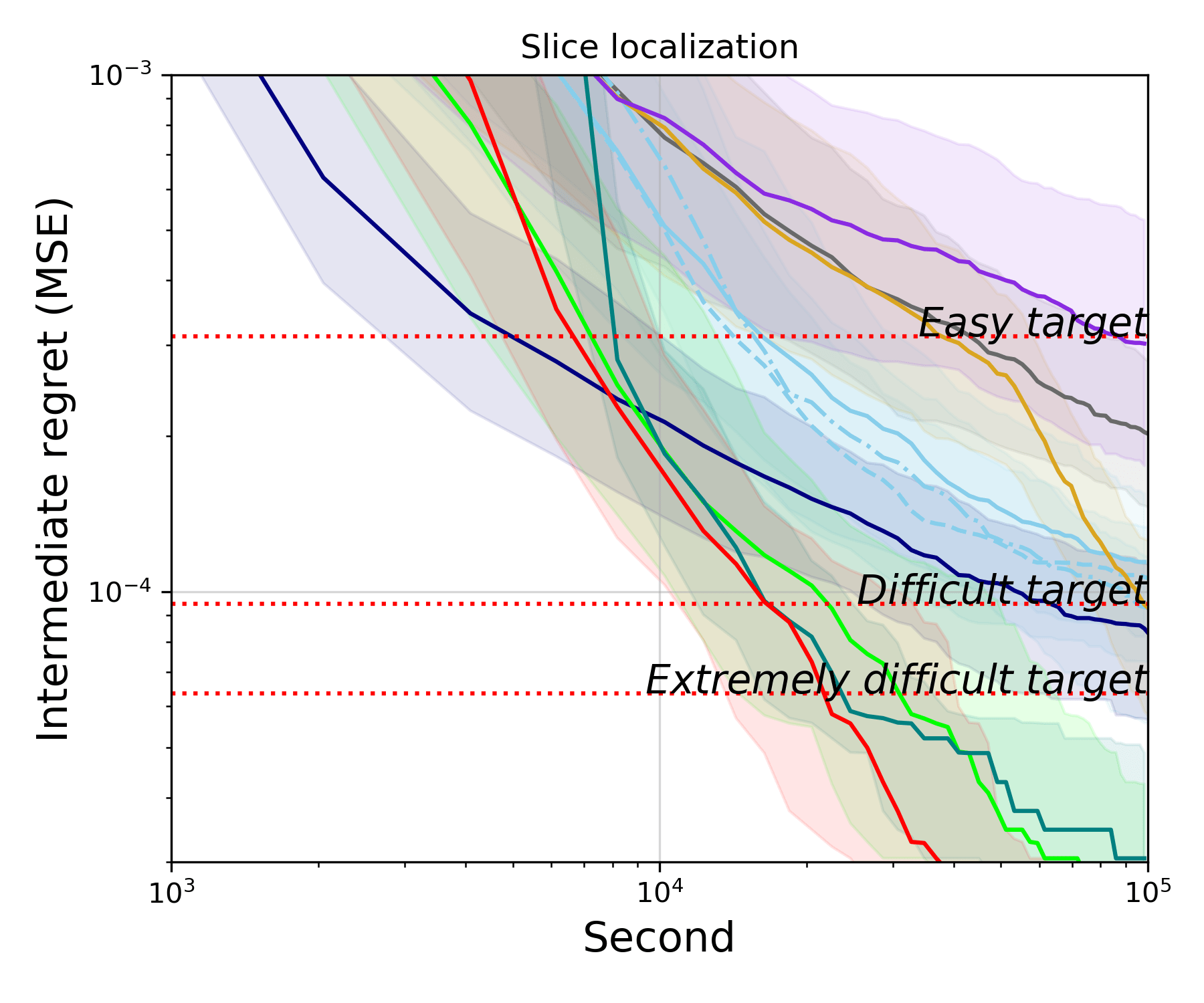}
		\hfill
		\caption{HPO-Bench tasks}
		\label{fig2b:a}
	\end{subfigure}
	\centering
	\begin{subfigure}[b]{.8\textwidth}		
		\includegraphics[width=.2\textwidth]{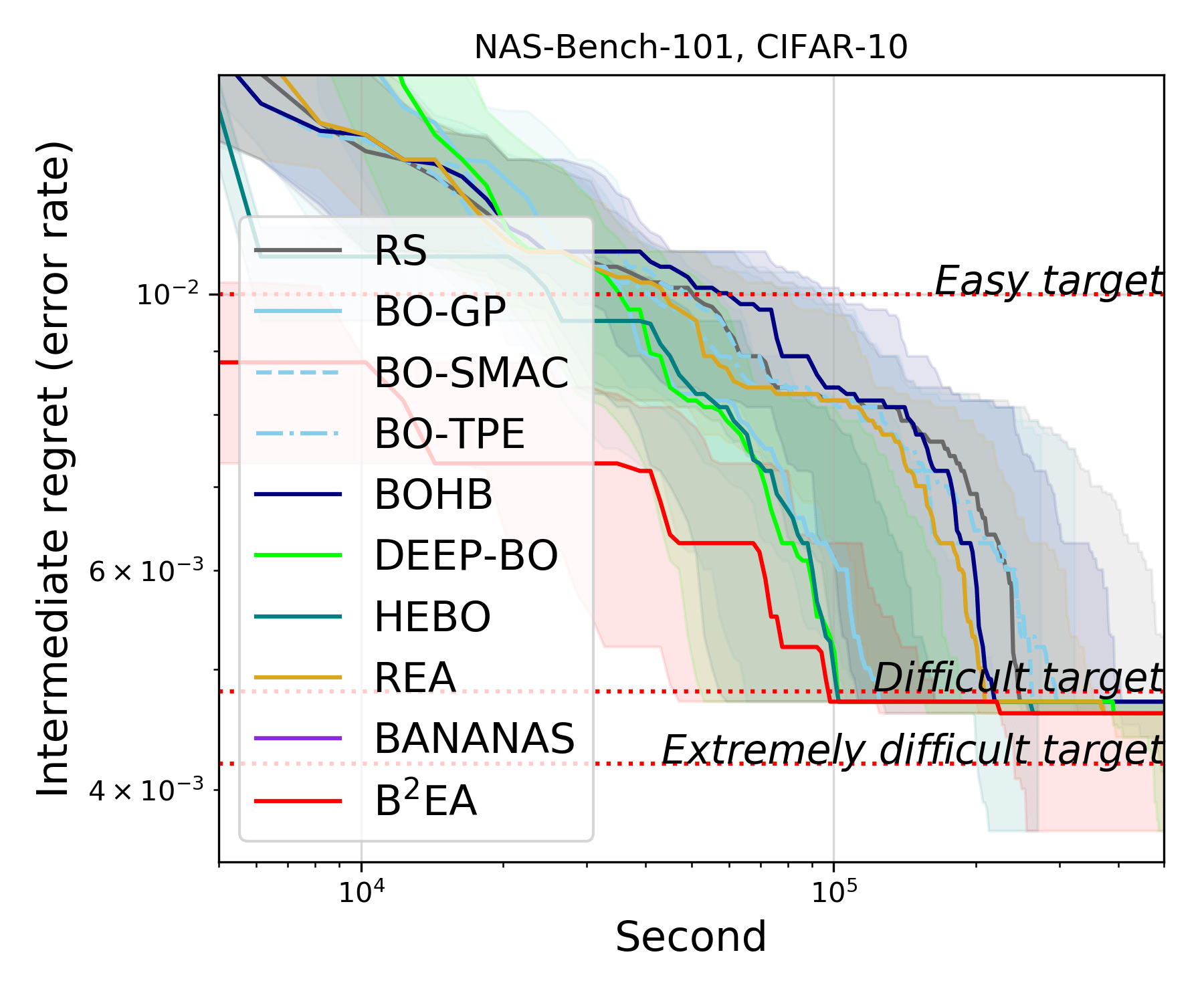}
		\hfill
		\includegraphics[width=.2\textwidth]{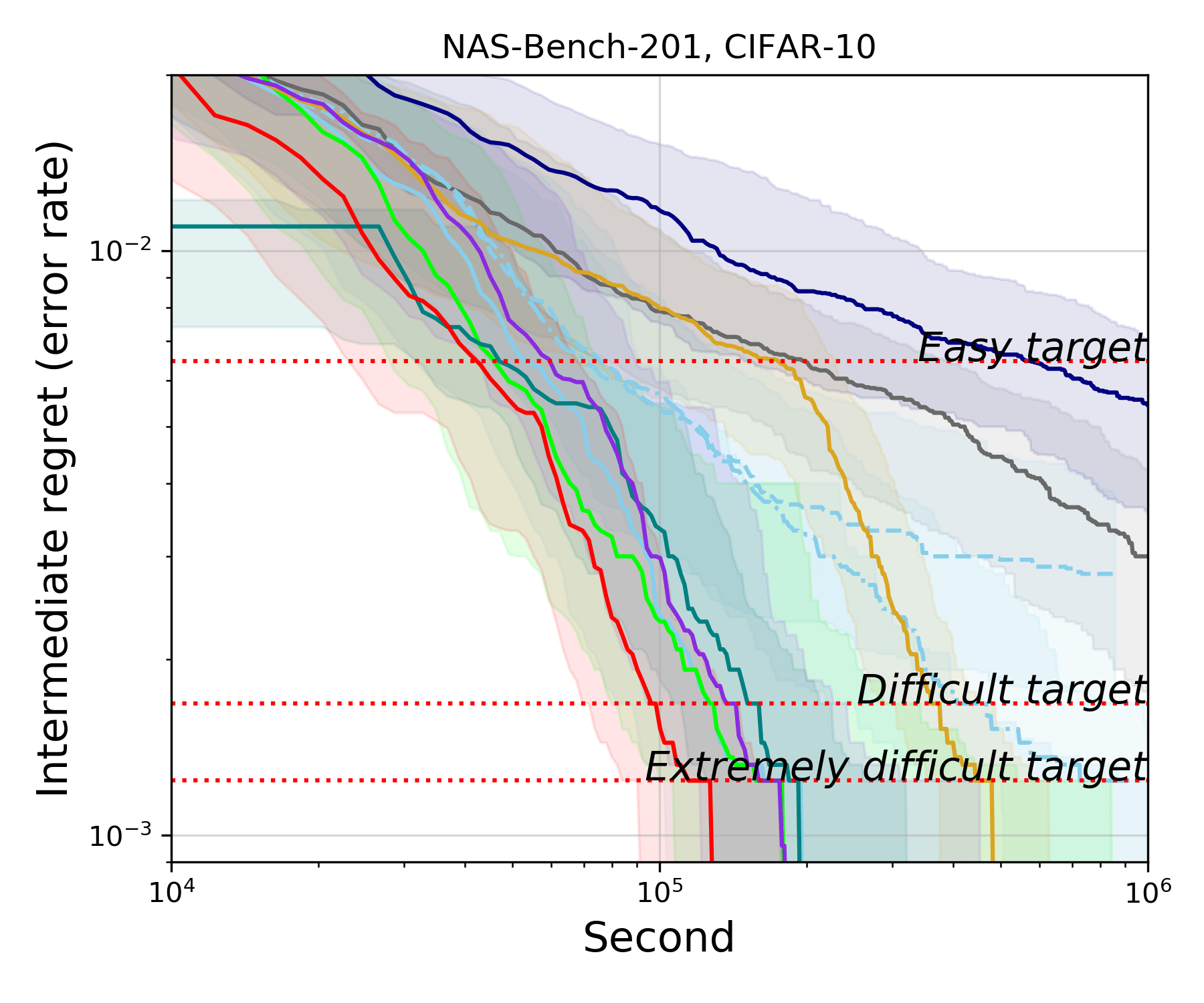}
		\hfill
		\includegraphics[width=.2\textwidth]{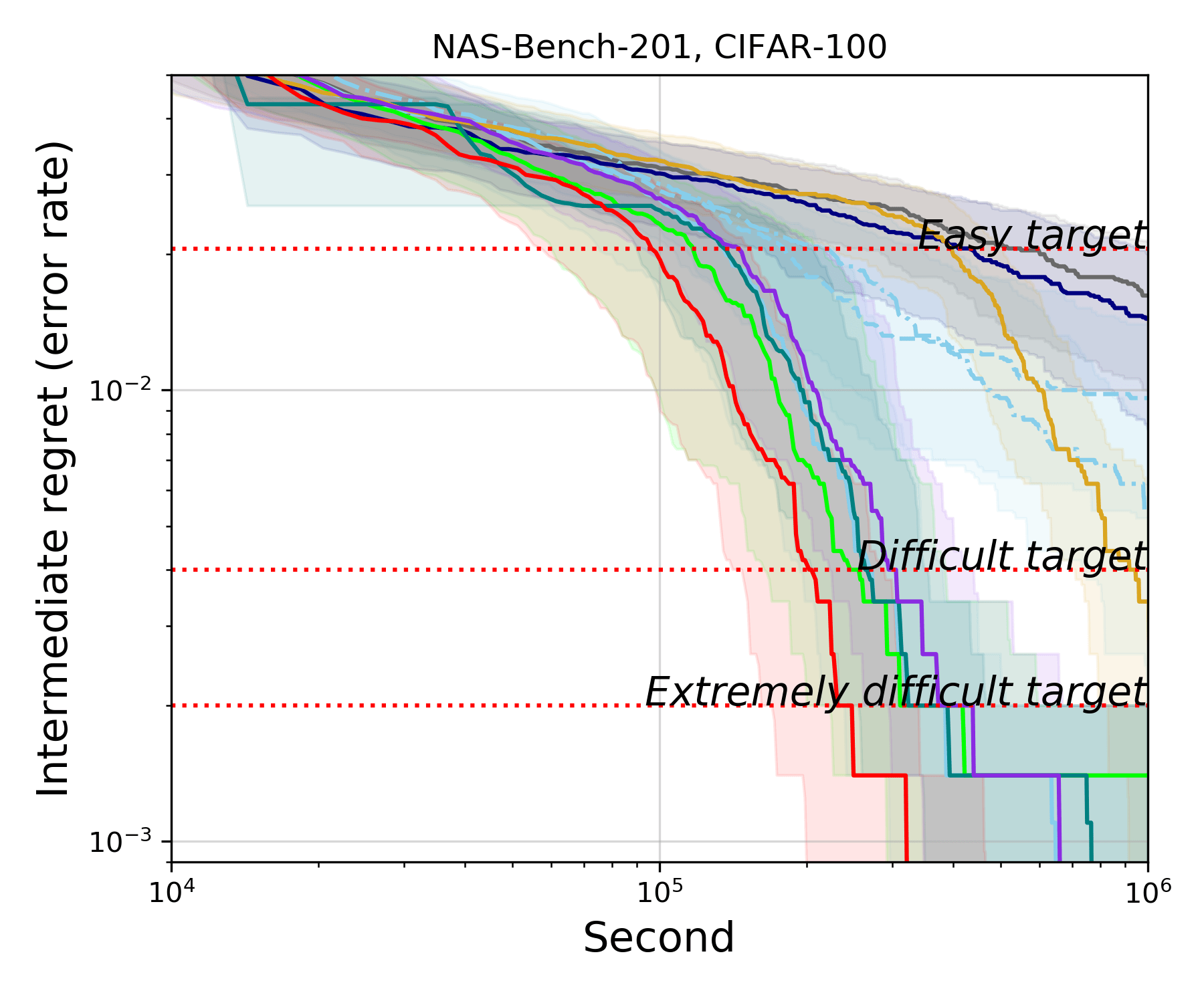}
		\hfill
		\includegraphics[width=.2\textwidth]{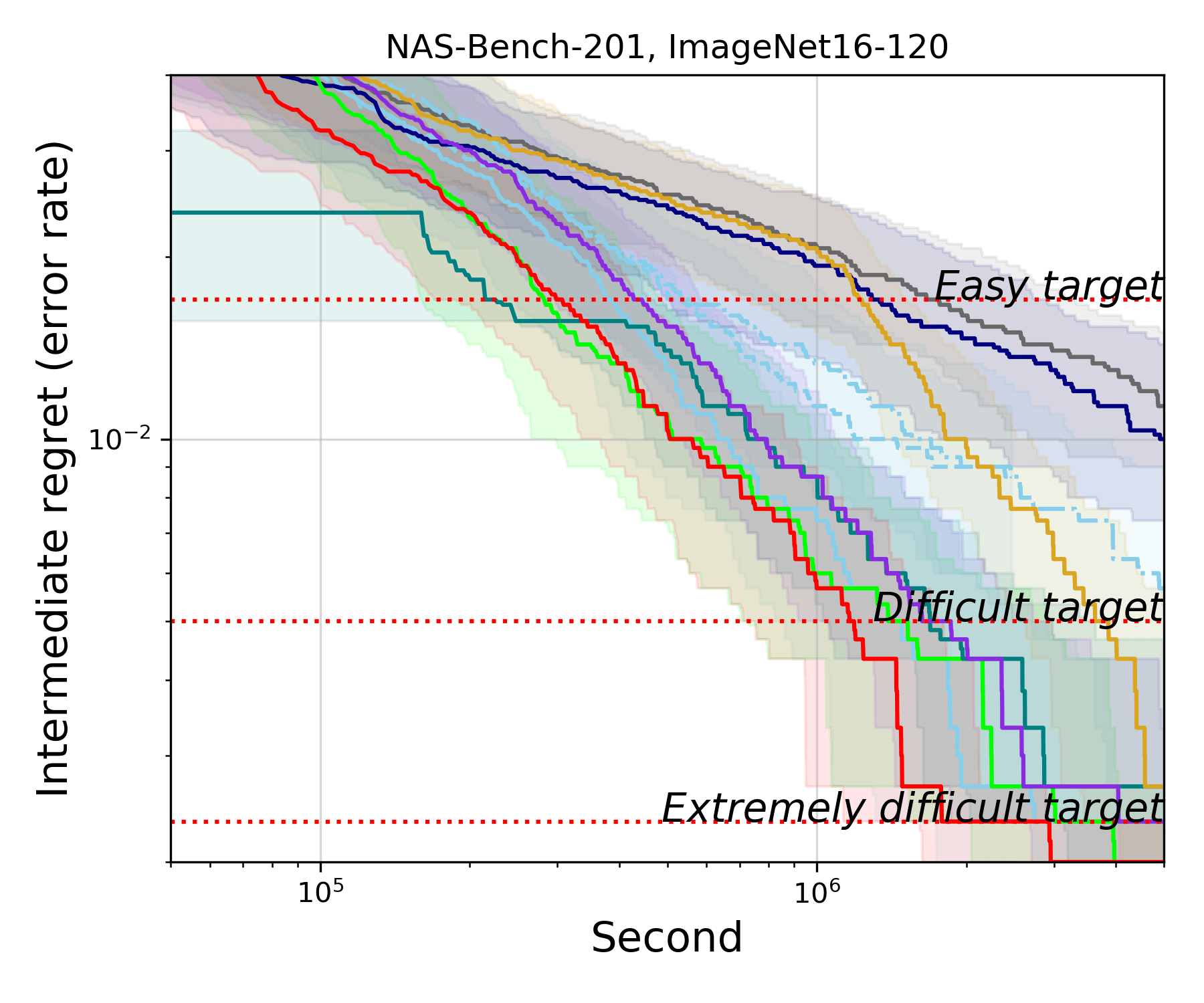}
		\hfill
		\caption{NAS-Bench tasks}
		\label{fig2b:b}	
	\end{subfigure}
	\begin{subfigure}[b]{\textwidth}		
		\includegraphics[width=.16\textwidth]{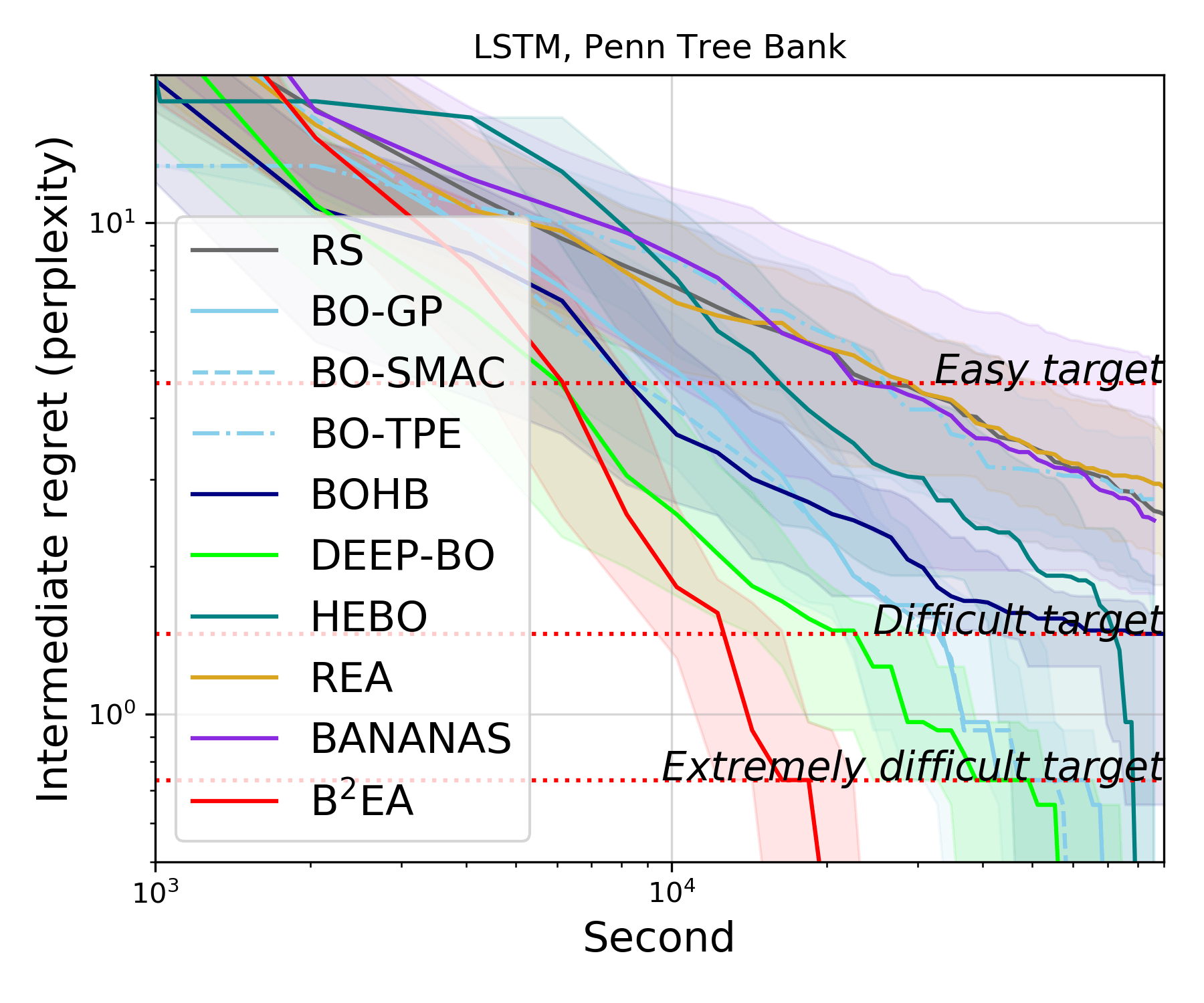}
		\hfill
		\includegraphics[width=.16\textwidth]{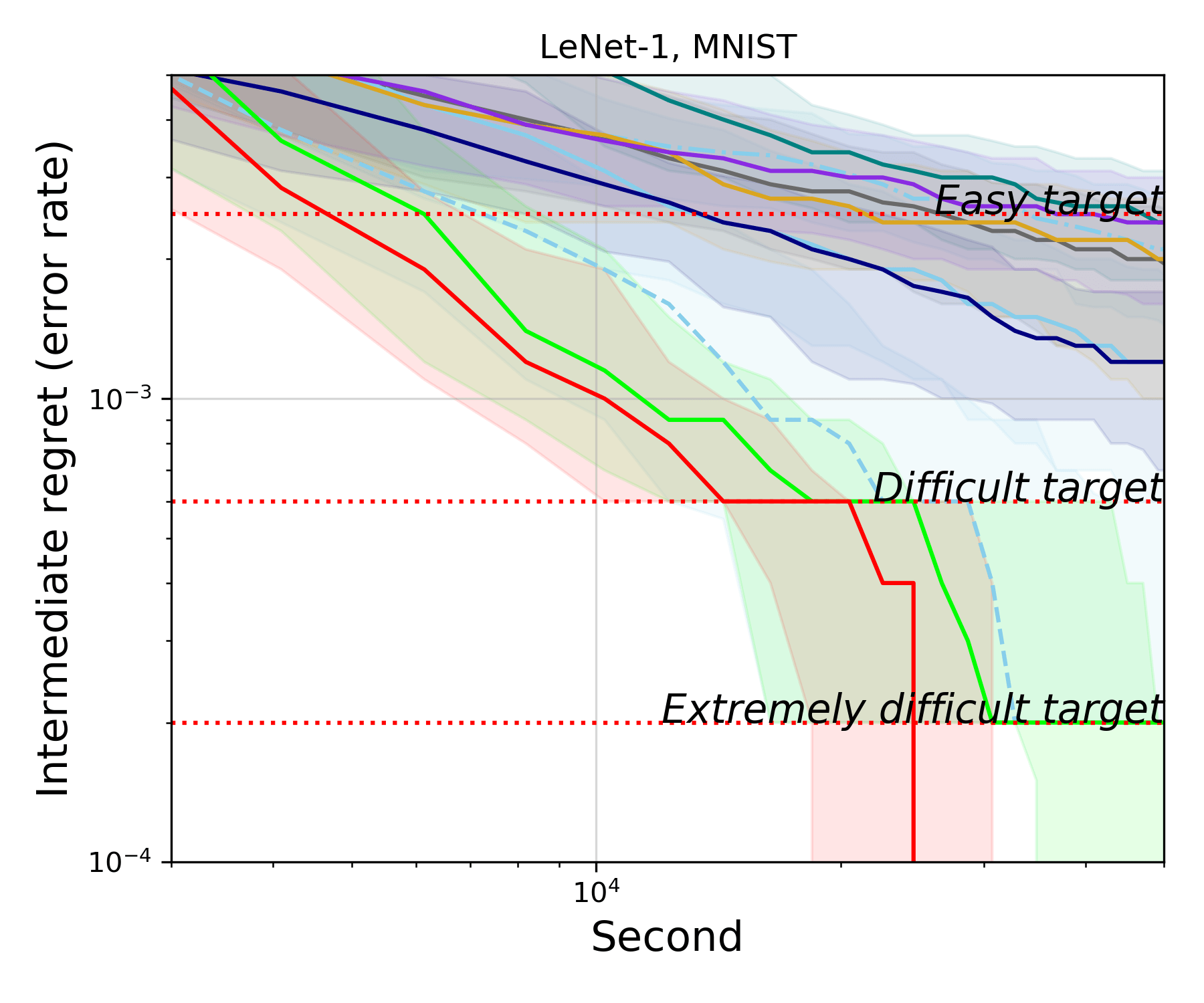}
		\hfill
		\includegraphics[width=.16\textwidth]{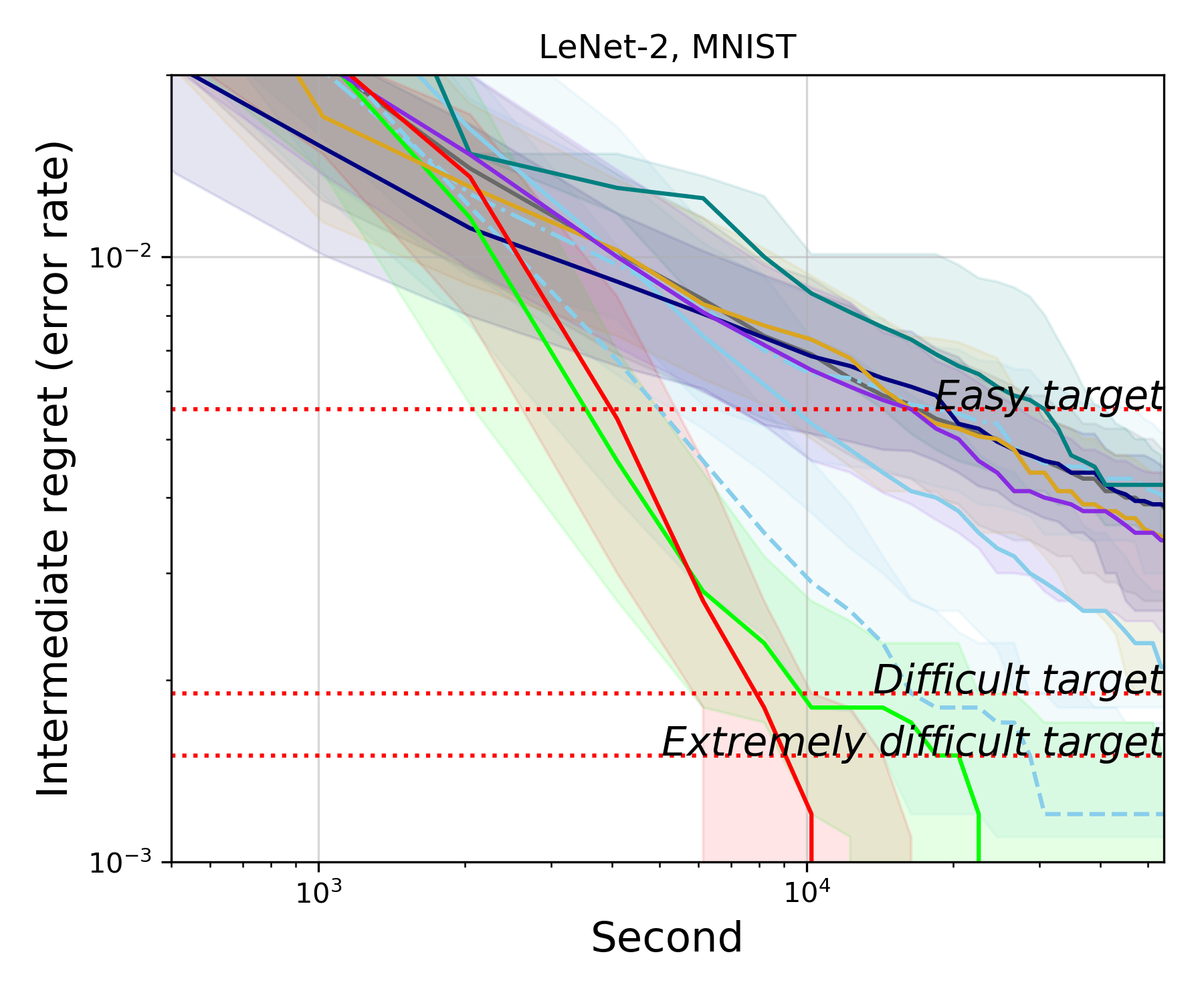}
		\hfill
		\includegraphics[width=.16\textwidth]{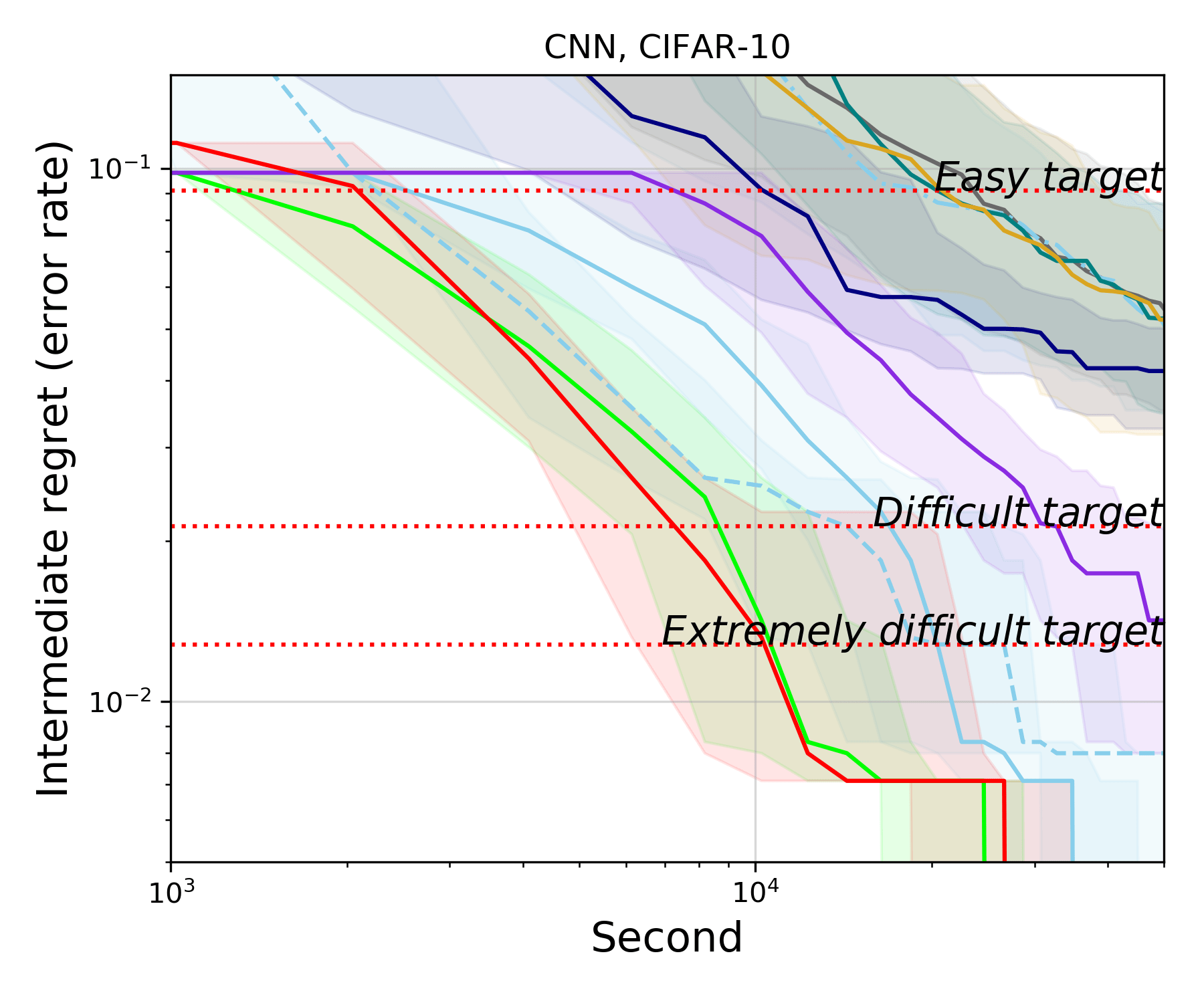}
		\hfill
		\includegraphics[width=.16\textwidth]{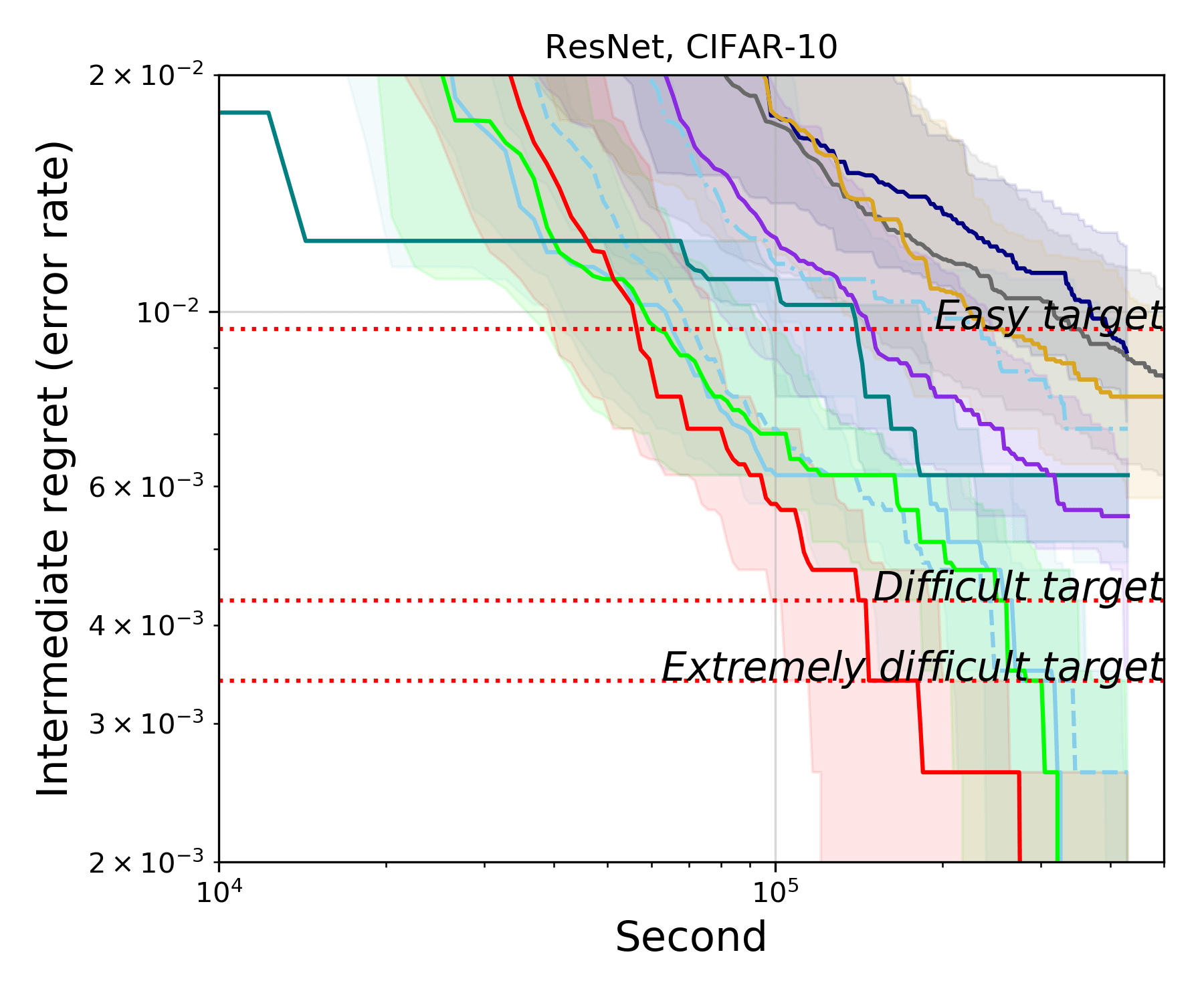}
		\hfill
		\includegraphics[width=.16\textwidth]{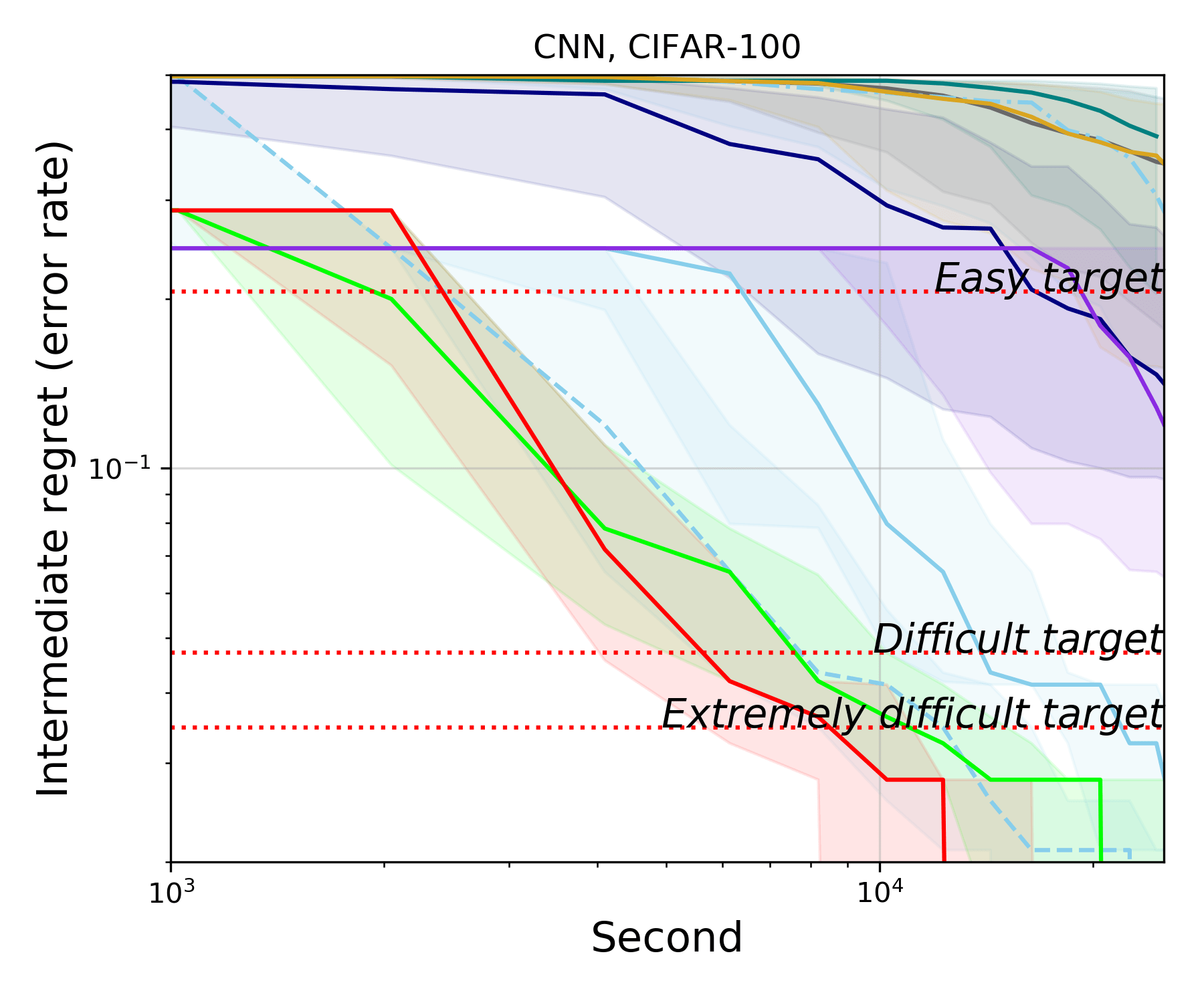}
		\caption{DNN-Bench tasks}
		\label{fig2b:c}	
	\end{subfigure}
	%\\[-3ex]
	%\vspace{-2mm}
	\caption{Performance in terms of intermediate regret metric $r_t$. 
		The solid lines show the median $r_t$ value of 500 runs, and the shaded areas show the interquartile range. The two axes are in logarithmic scales.
		%We set three different target goals: easy, difficult, and extremely difficult as the top 1 \%, top 0.05 \% and 0.02 \% performance, respectively.
		%of the full search space $\mathcal{X}$.
	}
	\label{fig2b}
	%\vspace{-4mm}
\end{figure*}

\begin{table*}[ht]
	\centering
	\caption{Success rate $\mathbb{P}(\tau \le t)$ performance (in \%) for all benchmarks. The best performer in each task is shown in \textbf{bold}. 
		The $\mathbb{P}(\tau \le t)$ for $c_e$ and $c_x$ are presented in Table S1 of Supplementary A.}
	\scriptsize 
	\label{tab1e:success_rate}
	\resizebox{.99\textwidth}{!}{
		\begin{threeparttable}
			\begin{tabular}{lll rrrrrrr rrr}
				\toprule
				\multicolumn{3}{c}{\textbf{Target setting}} & \multicolumn{9}{c}{\textbf{Benchmark algorithms}} & {\begin{tabular}[r]{@{}r@{}}\textbf{Proposed}\\\textbf{algorithm}\end{tabular}}  \\
				\midrule
				\textbf{Goal} & \textbf{Source} & \textbf{Task name} & RS & {\begin{tabular}[r]{@{}r@{}}BO-\\GP\end{tabular}} & {\begin{tabular}[r]{@{}r@{}}BO-\\SMAC\end{tabular}} & {\begin{tabular}[r]{@{}r@{}}BO-\\TPE\end{tabular}} & BOHB & DEEP-BO & HEBO & BANANAS & REA &  B\textsuperscript{2}EA \\
				\midrule
				\multirow{15}{*}{\begin{tabular}[c]{@{}l@{}}Difficult\\ target\\$c_d$\end{tabular}} & \multirow{4}{*}{HPO-Bench} & Parkinsons & 14.2 & 94.0 & 51.8 & 78.2 & 70.2 & \textbf{99.6} & 15.4 & 27.8 & 88.8 & 98.0 \\
				&  & Naval & 4.4 & 38.4 & 24.0 & 47.4 & 32.6 & 97.4 & \textbf{99.4} & 3.6 & 13.8 & 98.6 \\
				&  & Protein & 3.8 & 81.0 & 38.2 & 24.0 & 40.6 & 92.2 & 97.8 & 63.0 & 3.6 & \textbf{99.4} \\
				&  & Slice & 5.8 & 31.6 & 39.2 & 39.2 & 52.8 & 84.6 & \textbf{99.6} & 5.4 & 23.4 & 90.4 \\
				\cmidrule{2-13}
				& NAS-Bench-101 & CIFAR-10 & 71.6 & 95.0 & 88.2 & 64.4 & 79.0 & 97.6 & \textbf{99.2} & 93.8 & 82.0 & 98.4 \\
				\cmidrule{2-3}
				& \multirow{3}{*}{NAS-Bench-201} & CIFAR-10 & 7.4 & 74.4 & 16.0 & 47.8 & 3.4 & 78.4 & 96.6 & 86.0 & 67.8 & 99.2 \\
				&  & CIFAR-100 & 3.0 & 95.4 & 12.4 & 21.8 & 4.8 & 93.8 & 97.8 & 90.8 & 21.8 & \textbf{99.0} \\
				&  & ImageNet16-120 & 5.2 & 86.4 & 27.0 & 36.8 & 9.6 & 87.4 & 78.8 & 83.3 & 58.2 & \textbf{99.0} \\
				\cmidrule{2-13}
				& \multirow{6}{*}{DNN-Bench} & PTB-LSTM & 4.2 & 41.2 & 43.8 & 3.0 & 14.0 & 68.2 & 16.6 & 6.4 & 5.8 & \textbf{99.8} \\
				&  & MNIST-LeNet1 & 6.8 & 23.4 & 75.0 & 4.0 & 12.0 & 90.0 & 1.8 & 2.8 & 6.6 & \textbf{99.4} \\
				&  & MNIST-LeNet2 & 7.0 & 21.2 & 65.4 & 5.0 & 0.0 & 77.2 & 3.0 & 6.2 & 7.4 & \textbf{99.0} \\
				&  & CIFAR10-CNN & 5.0 & 85.2 & 82.2 & 8.4 & 0.0 & 96.0 & 6.4 & 50.8 & 9.2 & \textbf{99.2} \\
				&  & CIFAR10-ResNet & 4.0 & 63.6 & 67.4 & 14.8 & 1.0 & 63.4 & 14.4 & 12.6 & 4.6 & \textbf{99.0} \\
				&  & CIFAR100-CNN & 2.4 & 58.8 & 87.6 & 2.0 & 2.0 & 95.8 & 2.4 & 7.2 & 1.4 & \textbf{99.2} \\
				\cmidrule{2-13}
				& \multirow{2}{*}{Overall success rate} & Mean   & 10.3 & 63.5 & 51.3 & 28.3 & 23.0 & 87.3 & 52.1 & 38.6 & 28.2 & \textbf{98.4}  \\
				& & Std. deviation & 0.2 & 0.3 & 0.3 & 0.2 & 0.3 & 0.1 & 0.5 & 0.4 & 0.3 & \textbf{0.0} \\
				\cmidrule{2-13}
				& \multirow{2}{*}{Overall rank} & Mean & 8.6 & 4.0 & 5.3 & 7.0 & 7.6 & 2.6 & 4.8 & 6.4 & 7.1 & \textbf{1.3} \\
				& & Std. deviation & 1.5 & 1.1 & 2.2 & 1.8 & 2.1 & 0.9 & 3.3 & 2.2 & 1.7 & \textbf{0.5} \\
				\bottomrule	
			\end{tabular}
			%\begin{tablenotes}
			%	\small
			%	\item \textit{$\dagger$ Performance measured at the maximum budget (i.e. $t_b = t_{max}$).}
			%\end{tablenotes}
			%\vspace{-5mm}
		\end{threeparttable}		
	}
\end{table*}

\begin{table*}[ht]
	\centering
	\caption{Expected time $\mathbb{E}[\tau]$ performance (in \textit{minutes}) for all benchmarks. The $\mathbb{E}[\tau]$ for $c_e$ and $c_x$ are presented in Table S2 of Supplementary A.} 
	\label{tab1e:expected_time}
	\resizebox{0.99\textwidth}{!}{
		\begin{threeparttable}
			\begin{tabular}{lll rrrrrrr rrr}
				\toprule
				\multicolumn{3}{c}{\textbf{Target setting}} & \multicolumn{9}{c}{\textbf{Benchmark algorithms}} & {\begin{tabular}[r]{@{}r@{}}\textbf{Proposed}\\\textbf{algorithm}\end{tabular}}  \\
				\midrule
				\textbf{Goal} & \textbf{Source} & \textbf{Task name} & RS & {\begin{tabular}[r]{@{}r@{}}BO-\\GP\end{tabular}} & {\begin{tabular}[r]{@{}r@{}}BO-\\SMAC\end{tabular}} & {\begin{tabular}[r]{@{}r@{}}BO-\\TPE\end{tabular}} & BOHB & DEEP-BO & HEBO & BANANAS & REA &  B\textsuperscript{2}EA \\
				\midrule
				\multirow{13}{*}{\begin{tabular}[c]{@{}l@{}}Difficult\\ target\\$c_d$\end{tabular}} & \multirow{4}{*}{HPO-Bench} & Parkinsons & 1,156.1 & 92.9 & 435.6 & 170.2 & 196.9 & 54.4 & 1,329.1 & 835.3 & 153.5 & \textbf{52.1} \\
				&  & Naval & 1,450.9 & 298.0 & 711.3 & 242.5 & 385.9 & 67.0 & \textbf{54.9} & 5,840.2 & 330.4 & 67.2 \\
				&  & Protein & 4,111.6 & 263.3 & 599.7 & 935.2 & 711.0 & 178.4 & 161.9 & 355.2 & 3,181.2 & \textbf{125.5} \\
				&  & Slice & 9,549.0 & 3,511.4 & 2,491.4 & 2,068.3 & 1,705.7 & 661.3 & \textbf{326.5} & 24,380.9 & 2,152.8 & 472.7 \\
				\cmidrule{2-13}
				& NAS-Bench-101 & CIFAR-10 & 5,727 & 2,768 & 7,492 & 7,286 & 4,819 & 2,204 & \textbf{1,969} & 2,529 & 4,656 & 1,973 \\
				\cmidrule{2-3}
				& \multirow{3}{*}{NAS-Bench-201} & CIFAR-10 & 58,474 & 6,153 & 74,382 & 11,031 & 128,338 & 5,108 & 3,147 & 3,700 & 6,492 & \textbf{1,986} \\
				&  & CIFAR-100 & 132,392 & 5,133 & 75,683 & 32,784 & 98,750 & 5,049 & 4,719 & 6,241 & 20,905 & \textbf{3,727} \\
				&  & ImageNet16-120 & 646,652 & 34,178 & 161,390 & 139,496 & 428,714 & 35,196 & 47,285 & 41,763 & 71,998 & \textbf{21,523} \\
				\cmidrule{2-13}
				& \multirow{6}{*}{DNN-Bench} & PTB-LSTM & 4,640 & 574 & 610 & 10,200 & 1,679 & 380 & 1,385 & 4,777 & 5,096 & \textbf{213} \\
				&  & MNIST-LeNet1 & 7,256 & 2,687 & 535 & 10,417 & 3,866 & 391 & 78,876 & 24,944 & 7,736 & \textbf{284} \\
				&  & MNIST-LeNet2 & 3,917 & 1,333 & 387 & 6,833 & 8,566 & 294 & 12,799 & 5,129 & 2,313 & \textbf{134} \\
				&  & CIFAR10-CNN & 5,248 & 318 & 307 & 4,576 & 84,194 & \textbf{187} & 6,832 & 636 & 7,739 & 195 \\
				&  & CIFAR10-ResNet & 86,421 & 5,118 & 4,499 & 37,691 & 197,037 & 4,688 & 40,752 & 31,657 & 150,634 & \textbf{2,325} \\
				&  & CIFAR100-CNN & 4,957 & 231 & 150 & 3,313 & 2,541 & 121 & 7,207 & 1,127 & 4,256 & \textbf{96} \\
				\cmidrule{2-13}
				& \multirow{2}{*}{\begin{tabular}[c]{@{}l@{}}Normalized \\$\mathbb{E}[\tau]$ (in \%) \end{tabular}} & Mean  & 55.5 & 7.8 & 23.2 & 31.2 & 48.2 & 4.7 & 34.9 & 33.6 & 30.0 & \textbf{3.6}  \\
				& & Std. deviation   & 33.1 & 9.1 & 29.5 & 31.9 & 37.4 & 7.3 & 43.4 & 33.7 & 28.0 & \textbf{6.6}  \\
				\bottomrule	
			\end{tabular}
			%\vspace{-8mm}	
		\end{threeparttable}		
	}
\end{table*}

\begin{table}[ht]
	\centering
	\caption{Top-5 performers for each benchmark in terms of $\mathbb{P}(\tau \le t)$ performance. They are ordered according to the ranking. %Note that the \% number is the mean success rate. 
	Table for $\mathbb{E}[\tau]$ can be found in Table S3 of Supplementary A, and the top two rankers are exactly the same as in the below table. %with the same ordering as in the below table.
	}
	%For $c_e$ and $c_d$, we measured $\mathbb{P}(\tau \le t)$ at time $t$ when an algorithm is over 99\%. For $c_x$, performance on some tasks measured at the maximum budget (i.e. $t_m = t_{max}$). Refer to Table A1 in supplementary materials. }
	\scriptsize 
	\label{tab1as}
	\resizebox{0.8\linewidth}{!}{
		\begin{tabular}{l rrr}
			\toprule
			\multirow{2}{*}{\textbf{Benchmark}} &  \multicolumn{3}{c}{\textbf{Algorithm (Mean success rate $\mathbb{P}(\tau \le t)$)}} \\
			\cmidrule{2-4}
			&  Easy target $c_e$ & Difficult target $c_d$ & Extremely difficult target $c_x$ \\
			\midrule
			\multirow{5}{*}{HPO-Bench}  & B\textsuperscript{2}EA (98.4 \%) & B\textsuperscript{2}EA (96.6 \%) & B\textsuperscript{2}EA (97.7 \%) \\
			& DEEP-BO (94.8 \%) & DEEP-BO (93.5 \%) & DEEP-BO (92.3 \%) \\
			& HEBO (94.5 \%) & HEBO (78.1 \%) & HEBO (71.0 \%) \\
			& BOHB (92.3 \%) & BO-GP (61.3 \%) & BO-GP (54.9 \%)   \\
			& BO-SMAC (81.7 \%) & BOHB (49.1 \%) & BO-TPE (39.3 \%)   \\
			\midrule
			\multirow{5}{*}{NAS-Bench}  & B\textsuperscript{2}EA (99.1 \%) & B\textsuperscript{2}EA (98.9 \%) & B\textsuperscript{2}EA (83.1 \%) \\
			& HEBO (95.5 \%) & HEBO (93.1 \%) & HEBO (71.8 \%) \\
			& BANANAS (93.3 \%) & DEEP-BO (89.3 \%) & BANANAS (68.1 \%) \\
			& BO-GP (89.8 \%) & BANANAS (88.5 \%) & DEEP-BO (59.4 \%)   \\
			& DEEP-BO (89.3\%) & BO-GP (87.8 \%) & BO-GP (56.7 \%)   \\
			\midrule
			\multirow{5}{*}{DNN-Bench}  & B\textsuperscript{2}EA (99.1 \%) & B\textsuperscript{2}EA (99.3 \%) & B\textsuperscript{2}EA (98.4 \%) \\
			& DEEP-BO (95.5 \%) & DEEP-BO (81.8 \%) & DEEP-BO (70.4 \%) \\
			& BO-SMAC (82.8 \%) & BO-SMAC (70.2 \%) & BO-SMAC (62.6 \%) \\
			& BO-GP (66.4 \%) & BO-GP (48.9 \%) & BO-GP (37.5 \%)   \\
			& BOHB (38.7 \%) & BANANAS (14.3 \%) & BANANAS (8.4 \%)  \\
			\bottomrule
		\end{tabular}
	}
	%\vspace{-8mm}
\end{table}

\textbf{Experiment results:}
Figure~\ref{fig2b} shows the intermediate regret $r_t$ performance of the 500 runs.
%A multi-fidelity method, BOHB, performs well with a small budget (e.g., $t \le 20,000$ s) in HPO tasks, but it is eventually outperformed by B\textsuperscript{2}EA.
%On HPO-Bench and NAS-Bench tasks, the high performance of DEEP-BO is interesting because it was not even in the benchmark list of the previous works.
%~\cite{klein2019tabular}. 
%
For success rate $\mathbb{P}(\tau \le t)$ and expected time $\mathbb{E}[\tau]$, the results for the difficult target $c_d$ are shown in Table~\ref{tab1e:success_rate} and Table~\ref{tab1e:expected_time}, respectively. The full results including easy target $c_e$ and extremely difficult target $c_x$ can be found in Supplementary A.
From the three evaluation results, it can be observed that B\textsuperscript{2}EA is the only method that shows a robust performance over all 14 benchmarks and all three difficulty levels. 
To understand the specialty of each algorithm, we have calculated the average $\mathbb{P}(\tau \le t)$ performance over the three benches and the three difficulty levels. The results are shown in Table~\ref{tab1as}, and B\textsuperscript{2}EA is the top ranker for all of the nine categories. DEEP-BO is the second ranker for HPO-Bench and DNN-Bench. HEBO is the second ranker for NAS-Bench. Considering that HEBO was not designed for NAS, this is an interesting result. 
Per-algorithm performance can be found in Table S2 of Supplementary A.

% =============================================================
\section{Discussion}
\label{sec:discussion}

\begin{table}[h]
	\centering
	\caption{Ablation study results of B\textsuperscript{2}EA: relative performance of $\mathbb{E}[\tau]$ with respect to B\textsuperscript{2}EA is shown. Smaller values are better. For `SAEA off', we have used GP-EI instead.
	%(otherwise, the performance degradation is quite significant). 
	%The bold ones indicate the cases where performance is improved by turning off an enhancement technique.
	}
	\label{table:ablation_b2ea}
	\small	
	\resizebox{\linewidth}{!}{
	\begin{tabular}{lc cccc cccc ccc}
		\toprule
		\multirow{2}{*}{\textbf{\begin{tabular}[c]{@{}l@{}}Enhancement\\ techniques\end{tabular}}} & & \multicolumn{3}{c}{\textbf{HPO-Bench}} & & \multicolumn{3}{c}{\textbf{NAS-Bench}} & & \multicolumn{3}{c}{\textbf{DNN-Bench}} \\
		\cmidrule{3-5} \cmidrule{7-9} \cmidrule{11-13}
		& & $c_e$ & $c_d$ & $c_x$ & & $c_e$ & $c_d$ & $c_x$ & & $c_e$ & $c_d$ & $c_x$ \\
		\midrule
		B\textsuperscript{2}EA (everything on) & & 100.0 & 100.0 & 100.0 & & 100.0 & 100.0 & 100.0 & & 100.0 & 100.0 & 100.0 \\
		\midrule
		SAEA off & & 142.5 & 123.0 & 116.9 &  & 186.4 & 123.9 & 173.3 &  & 156.0 & 119.8 & 130.5 \\
		Input warping off & & \textbf{99.6} & \textbf{93.6} & \textbf{95.8} & & 116.0 & 107.8 & 107.4 &  & 104.2 & 102.7 & 108.6 \\
		Output transformation off & & 148.3 & 299.7 & 455.1 & & \textbf{96.8} & 179.2 & 227.6 & & 129.9 & 227.7 & 306.6 \\
		Early termination off & & 101.6 & 101.7 & 102.0 & & 104.3 & 107.1 & 103.2 & & 124.3 & 117.4 & 112.6 \\
		\bottomrule
	\end{tabular}
	\vspace{-8mm}
	}
\end{table}

\begin{table}[h]
	\centering
	\caption{Ablation study results of SAEA: relative performance of $\mathbb{E}[\tau]$ with respect to B\textsuperscript{2}EA is shown. Smaller values are better. `DEEP-BO like' can be regarded as a DEEP-BO improved with input warping and output transformation.}
	\label{table:ablation_saea}
	\scriptsize
	\resizebox{0.55\linewidth}{!}{
		\begin{threeparttable}		
			\begin{tabular}{c ccc r}
				\toprule
				\multirow{2}{*}{\textbf{\begin{tabular}[c]{@{}c@{}}Module\\A/B/C\end{tabular}}} & \multicolumn{3}{c}{\textbf{Standardized $\mathbb{E}[\tau]$}} & \multirow{2}{*}{\textbf{\begin{tabular}[c]{@{}c@{}}Remark\end{tabular}}}\\
				\cmidrule{2-4}
				& \multicolumn{1}{c}{$c_e$} & \multicolumn{1}{c}{$c_d$} & \multicolumn{1}{c}{$c_x$} & \\
				\midrule
				on/on/on & 100.0 & 100.0 & 100.0 &  B\textsuperscript{2}EA \\
				\midrule		
				on/\textbf{off}/on & \textbf{93.0} & 100.4 & 124.4 & \\
				%\textbf{off}/on/on & 133.4 & 184.2 & 179.3 \\ % NOTE:invalid condition
				\textbf{off}/\textbf{off}/on & 101.5 & 103.4 & 125.9 & DEEP-BO like \\		 
				\midrule
				on/on/\textbf{off} & 550.6 & 6423.3$\dagger$ & 5138.3$\dagger$ & \\			
				on/\textbf{off}/\textbf{off} & 472.8 & 2629.3 & 3618.1 & \\
				%\textbf{off} & on & \textbf{off} & 731.6 & 1926.6 & 2920.4 \\% NOTE:invalid condition
				\textbf{off}/\textbf{off}/\textbf{off} & 861.0 & 2764.7 & 4426.8 & RS \\		
				\bottomrule
			\end{tabular}
			\begin{tablenotes}
				\item \textit{$\dagger$ Failed experiments were excluded.}
			\end{tablenotes}
		\end{threeparttable}
		%\vspace{-8mm}
	}
\end{table}

\textbf{Ablation studies}: 
Ablation study results of B\textsuperscript{2}EA's enhancement techniques are shown in Table~\ref{table:ablation_b2ea}. When SAEA is turned off, the performance degrades over all nine cases. Therefore, we can observe that SAEA is a robust technique for the 14 benchmark tasks. Early termination is also a robust technique thanks to its conservative design, but its positive influence is also limited. 
On the contrary, HPO-Bench can benefit by having input warping turned off and NAS-Bench's easy-target case can benefit by having output transformation turned off. The corresponding cases are shown in bold. Therefore, we can say that the two features are less robust over the 14 benchmarks. 
For the input warping, HPO-Bench's tasks are less complicated than NAS-Bench's or DNN-Bench's tasks, and it can be speculated that input warping's handling of non-stationary functions caused a negative effect instead of creating a positive effect. 
For the output transformation, it is unclear why it is helpful to have it off for the NAS-Bench's easy-target case. In our experiments, generally output transformation turned out to be essential as can be seen from the three cases with extremely difficult target ($c_x$). For the extremely difficult target, the positive effect is understandable because NAS needs to be able to differentiate even a very small improvement in neural architecture performance when searching for an extremely well performing candidate. 
%
% On the contrary, HPO-Bench can benefit by having input warping turned off and NAS-Bench-easy-target can benefit by having output transformation turned off. The corresponding cases are shown in bold. Therefore, we can say that the two features are less robust over the 14 benchmarks. 
% However, output transformation can have a significant effect for the cases with extremely difficult target ($c_x$). This is understandable because NAS needs to be able to differentiate even a very small improvement in neural architecture performance when searching for an extremely well performing candidate. 
%
Ablation study results of SAEA's three modules are shown in Table~\ref{table:ablation_saea}. We have listed only the meaningful combinations of on and off. 
It can be observed that all three modules are generally helpful. 
When the target performance is easy ($c_e$), however, removing mutation was helpful. This is in line with the common understanding of BO performing better than EA when the computation budget is small. This can be also confirmed in Figure~\ref{fig:nas} where REA's performance is the worst for the extremely easy target.

% Ablation study results of SAEA's three modules are shown in Table~\ref{table:ablation_saea}. It can be observed that all three modules are generally helpful, but mutation can be turned off for an improvement when the target performance is easy ($c_e$).

\textbf{Acquisition functions of BO:} 
HEBO uses a multi-objective acquisition ensemble that aims to find  Pareto-optimal points~\cite{cown2020emprical}. DEEP-BO simply uses a round-robin of BO models where EI, PI, and UCB take turns. When we compared the two approaches, both performed comparably (see Figure~S3 of Supplementary D). Therefore, we have chosen to use simple round-robin or uniform random schemes in B\textsuperscript{2}EA.
As for the surrogate models, we can expect increasing diversity to be helpful and B\textsuperscript{2}EA uses both GP and RF (see the results in Figure~S4).  

\textbf{Modeling cost of BO models:} Some of the BO models can have prohibitively high modeling costs when $|\mathcal{H}|$ is large. For instance, the complexity of GP is known to be $\mathbb{O}(|\mathcal{H}|^3)$. When the black-box optimization relies solely on the accuracy of a single BO model, evaluating all the samples in $\mathcal{H}$ can be an essential goal.
B\textsuperscript{2}EA, however, utilizes a number of techniques including diversification, and it can be prudent to avoid such an expensive modeling or its work-arounds. Therefore, when $|\mathcal{H}|$ is larger than 200, we randomly sample only 200 samples from $\mathcal{H}$ in each iteration and use them for GP modeling.

% WR TODO (5/12; do not erase): 
% Limitations:
% 1. To use HPO and NAS algorithms, the selection of search space is extremely important. This part of work still belongs to human, but perhaps an extended research would be possible to show that B^2EA can be used for a coarse search to determine search space followed by a full search. What we have shown is that "for a fixed search space, B^2EA can perform robustly and efficiently over the existing state of the art solutions.". 
% 2. We did not consider multi-fidelity or freeze-and-thaw type of approaches. 

% discussion required regarding potential societal impacts, especially for negative perspectives  
%\textbf{Broader impacts:}
%Repeatedly training large DNNs can consume a significant amount of energy. 
%It might be possible to reduce the overall amount of training if the advanced HPO and NAS algorithms became the standard for tuning, thereby reducing the energy consumption of the research community. 
%Therefore, we proposed a robust optimization algorithm, B\textsuperscript{2}EA. 
%It might be possible to develop more energy-efficient optimization techniques based on our work.  

\textbf{Modern neural architecture design:} 
NAS algorithms have become an essential part of modern neural architecture design, and apparently there is a synergy effect over a variety of design methods including manual design, brute-force search, weight-sharing NAS, and multi-trial NAS. 
Once a neural architecture of high performance is published, a highly efficient NAS algorithm can be configured to optimize over a search space near the published neural architecture. The search space design will be dependent on the type of NAS. In fact, the success of recent NAS algorithms might be due to the highly constrained search space that was chosen based on the previously found high-performance architectures~\cite{liu2017hierarchical,zoph2018learning,liu2018progressive}. This is not necessarily a problem because NAS can benefit from any discovery of high-performance architectures through human effort, multi-trial NAS with a very large computational budget, or anything else.

% \textbf{Modern network design:} Modern network design has been generating an echoing synergy with NAS research. Once a network architecture of a high performance is published, a highly efficient NAS algorithm can be configured to optimize over a highly constrained search space near the published network architecture. In fact, the success of recent NAS algorithms might be due to the highly constrained search space that was chosen based on the previously found high-performance architectures~\cite{liu2017hierarchical,zoph2018learning,liu2018progressive}. This is not necessarily a problem, because NAS can benefit from any discovery of high-performance architecture through either human effort or multi-trial NAS with a very large computation budget. 
% %
% Considering the opposite direction, network design can also benefit from NAS. For instance, EfficientNet was developed by leveraging a multi-objective NAS in the search space specialized for mobile devices~\cite{tan2019mnas} and a low-dimensional network design space of RegNet was created by combining manual design and NAS~\cite{radosavovic2020designing}.

% =============================================================
\section{Conclusions and Limitations}
\label{sec:conclusion}
\vspace{-1mm}
B\textsuperscript{2}EA is a multi-trial NAS algorithm that is efficient because of the BO surrogate models and robust because of the diversification, cooperation, and mutation. Input warping, output transformation, and a conservative early termination turned out to be beneficial, too. 
We also note a few limitations of our work. We have performed experiments over 14 NAS tasks, but there are new benchmarks that have been published recently~\cite{klyuchnikov2020bench,mehrotra2020bench,li2021hw,hirose2021hpo}. We have adopted a conservative multi-fidelity technique, but less conservative multi-fidelity techniques are desired as long as robustness can be maintained. Parallelization is a fundamental technique for speeding up a search, and integrating parallelization techniques with B\textsuperscript{2}EA remains as a future work.

% HH:Implications about weight-sharing vs. multi-trial
%
% HH:Missing limitations and possible future works 
% No parallelization performance (compared to DEEP-BO)
% Performance evaluation on the other recent benchmark tasks~\cite{klyuchnikov2020bench,mehrotra2020bench,li2021hw,hirose2021hpo}
% Research direction w.r.t. the other related works using surrogate model (e.g. PNAS~\cite{liu2018progressive}) or progressive training (e.g. PBT~\cite{jaderberg2017population}) in NAS 

% =============================================================
\section{Acknowledgements}
\label{sec:ack}
% To notice some experiments performed using computational resouces of SK Telecom Co., Ltd
% Jungwook Shin is an AI Research Scientist in the SK Telecom Co., Ltd. 
% Copy from https://ieeexplore.ieee.org/search/searchresult.jsp?newsearch=true&queryText=Short-term%20Traffic%20Prediction%20with%20Deep%20Neural%20Networks:%20A%20Survey
This work was supported by an NRF grant~(MSIT) under Grant NRF-2020R1A2C2007139, an ETRI grant~[21ZR1100, A Study of Hyper-Connected Thinking Internet Technology by autonomous connecting, controlling and evolving ways], and an IITP grant~(MSIT) [NO.2021-0-01343, Artificial Intelligence Graduate School Program (Seoul National University)].

\bibliography{main}

%=============================================
%HH:build the supplementary included
\clearpage

\newpage

% To set caption style in supplementary
\captionsetup[figure]{labelformat=supplementary, labelfont=it, labelsep=period}
\captionsetup[subfigure]{labelformat=parens, labelfont=small}
\captionsetup[table]{labelformat=supplementary, labelfont=it, labelsep=period}
\captionsetup[algorithm]{labelformat=algos, labelfont=it, labelsep=period}		
\setcounter{figure}{0}
\setcounter{table}{0}
\setcounter{algorithm}{0}

\onecolumn
\begin{center}
	\textbf{\Large Supplementary materials}
\end{center}

\appendix

\section{Full Benchmark Results}

%\begin{figure*}[!ht]
%	
%	\begin{subfigure}[b]{\textwidth}
%		
%		%\vspace*{5ex}
%		
%		\includegraphics[width=.3\textwidth]{figures/f2c1mi.png}
%		\hfill
%		\includegraphics[width=.3\textwidth]{figures/f2c2mi.png}
%		\hfill
%		\includegraphics[width=.3\textwidth]{figures/f2c3mi.png}
%		\hfill
%		
%		%\vspace*{5ex}
%		
%		\includegraphics[width=.3\textwidth]{figures/f2c4mi.png}
%		\hfill
%		\includegraphics[width=.3\textwidth]{figures/f2c5mi.png}
%		\hfill
%		\includegraphics[width=.3\textwidth]{figures/f2c6mi.png}
%		\hfill
%		%\caption*{(c) DNN-Bench tasks}
%		%\label{fig2b-a:c}	
%	\end{subfigure}
%	%
%	%\\[5ex]
%	%
%	\caption*{Figure A1: Comparison of the intermediate regret $r_t$ performance. We repeated 500 randomly seeded runs. Note that the solid line denotes the median $r_t$ value, and the shaded area is the IQR of $r_t$ values. 
%		The plots show the experimental results on \textit{DNN-Bench} tasks, and the other plots of the HPO and NAS benchmarks are shown in Figure 4 of the paper.}
%	\label{fig2b-a}
%\end{figure*} 

\begin{figure*}[h]
	\begin{subfigure}[b]{\textwidth}
		%\vspace*{3ex}
		\includegraphics[width=.23\textwidth]{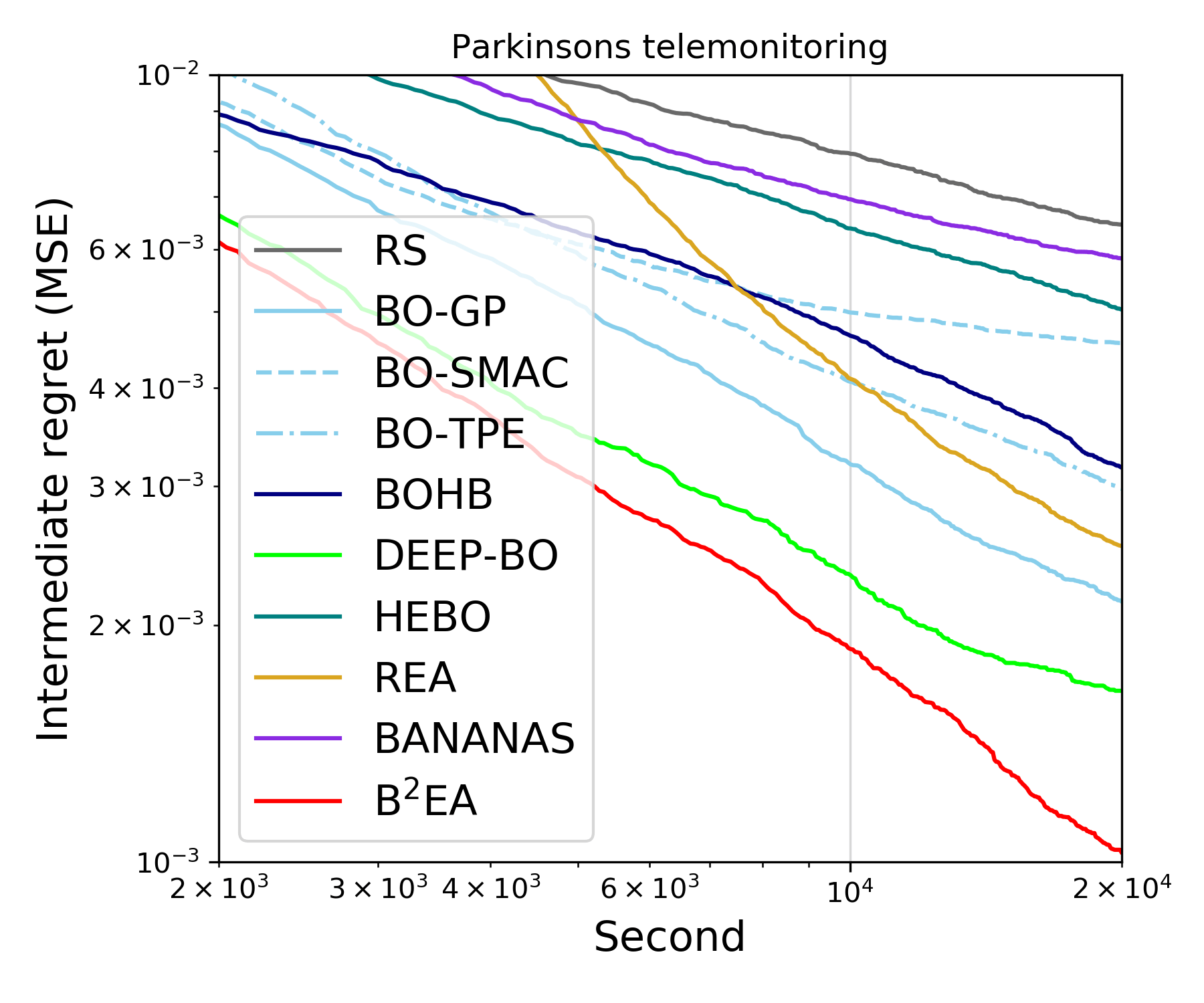}
		\hfill
		\includegraphics[width=.23\textwidth]{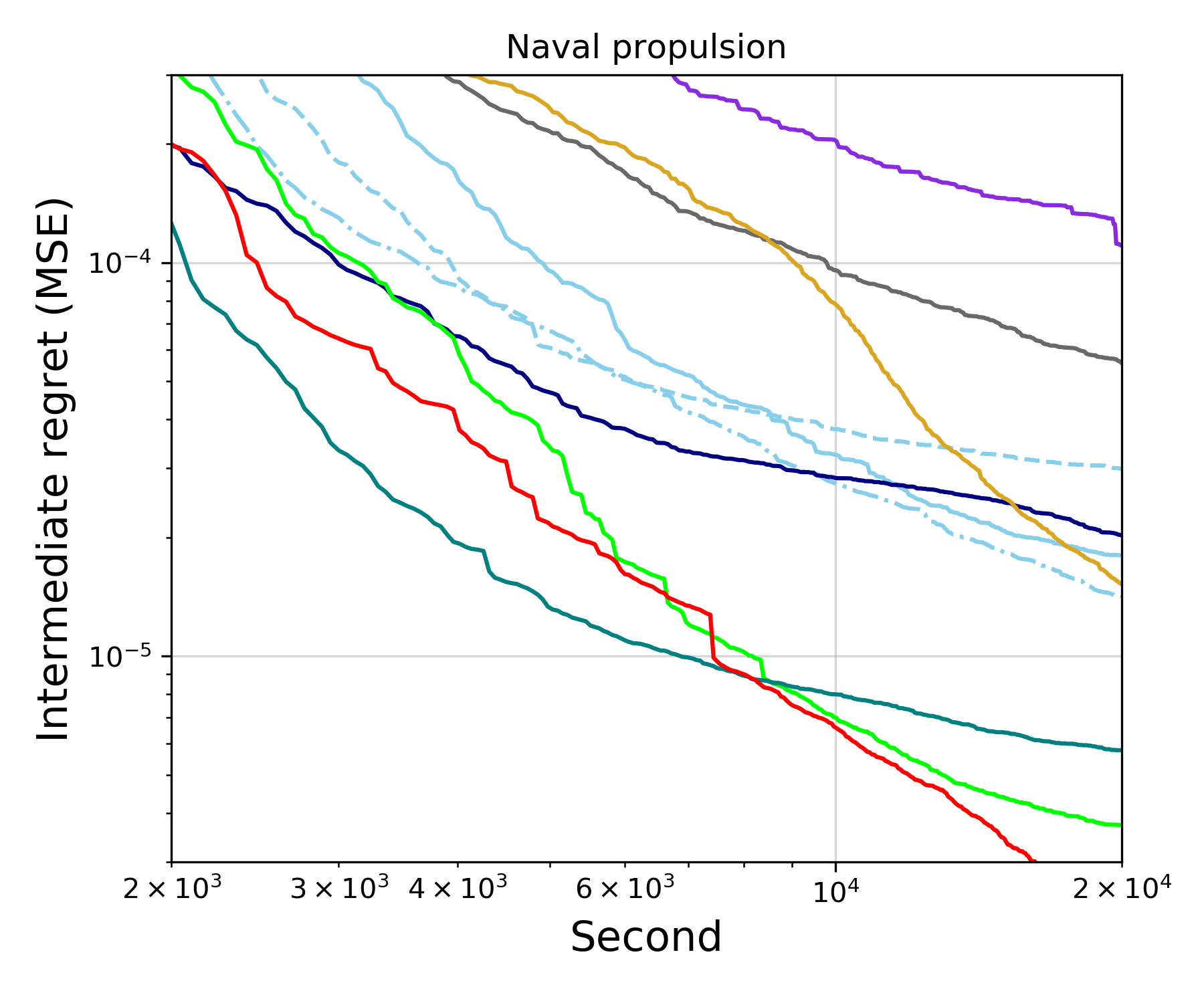}
		\hfill
		\includegraphics[width=.23\textwidth]{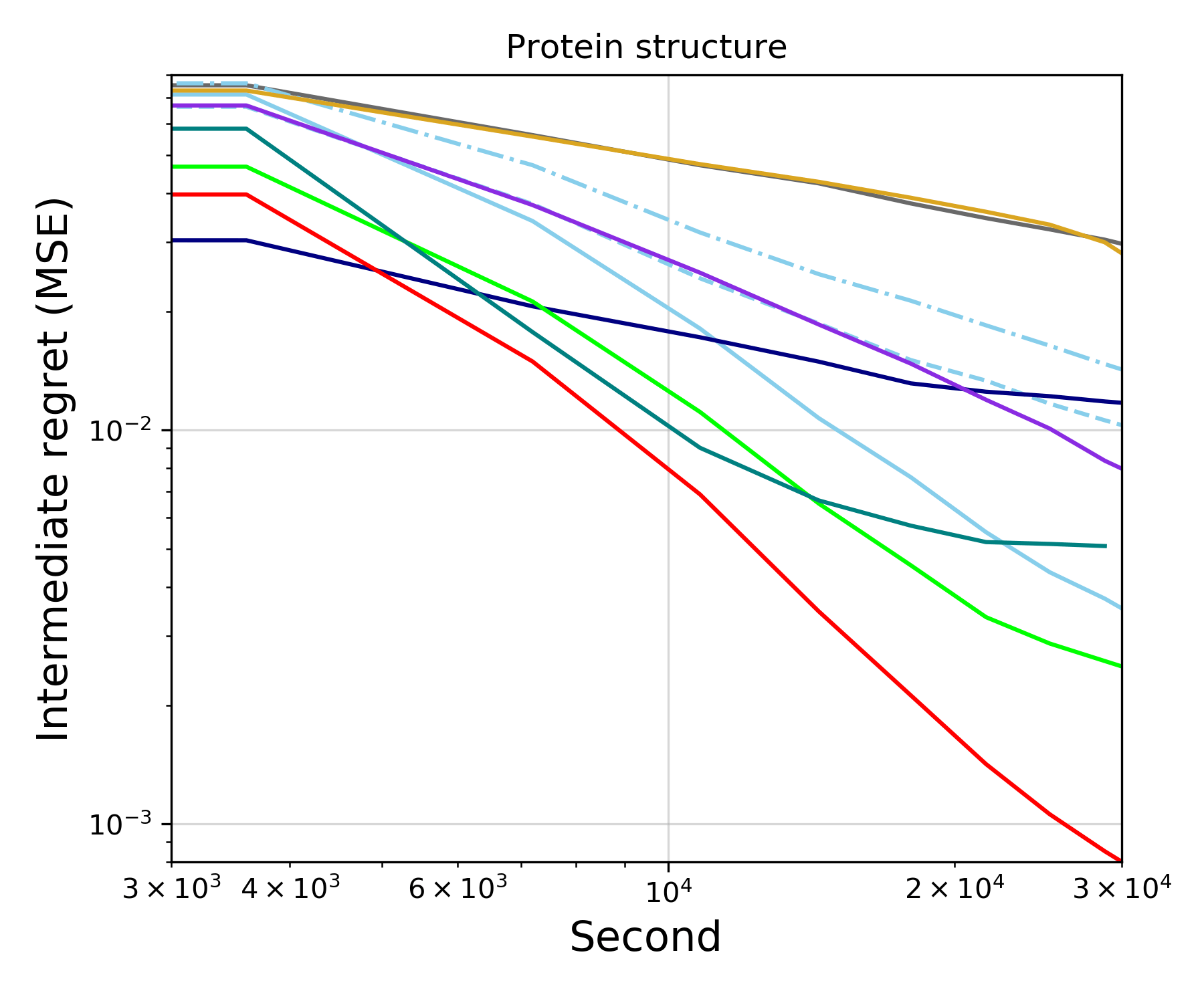}
		\hfill
		\includegraphics[width=.23\textwidth]{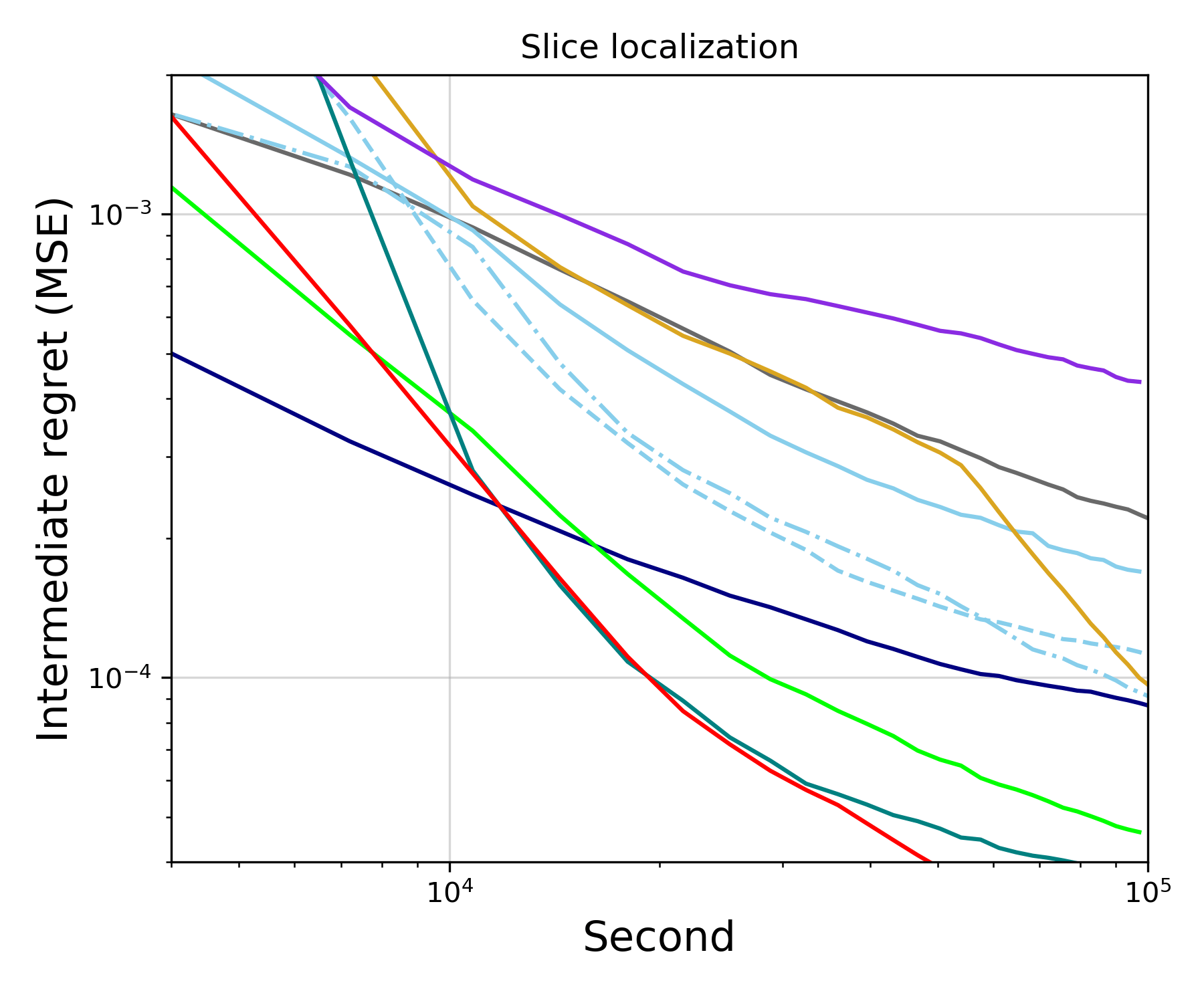}
		\hfill
		\caption{HPO-Bench tasks.}
		
	\end{subfigure}
	\\[3ex]
	\begin{subfigure}[b]{\textwidth}		
		\includegraphics[width=.23\textwidth]{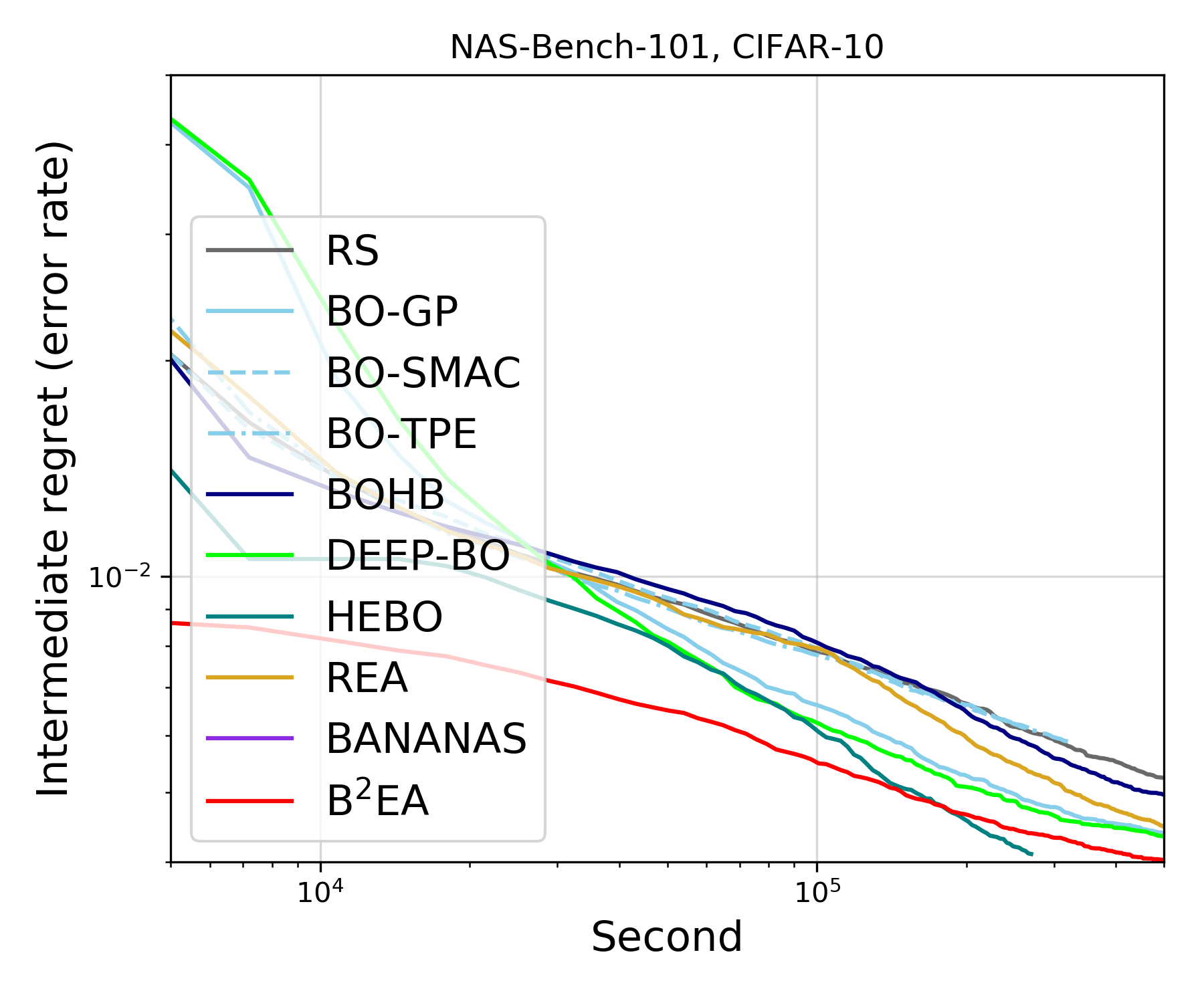}
		\hfill
		\includegraphics[width=.23\textwidth]{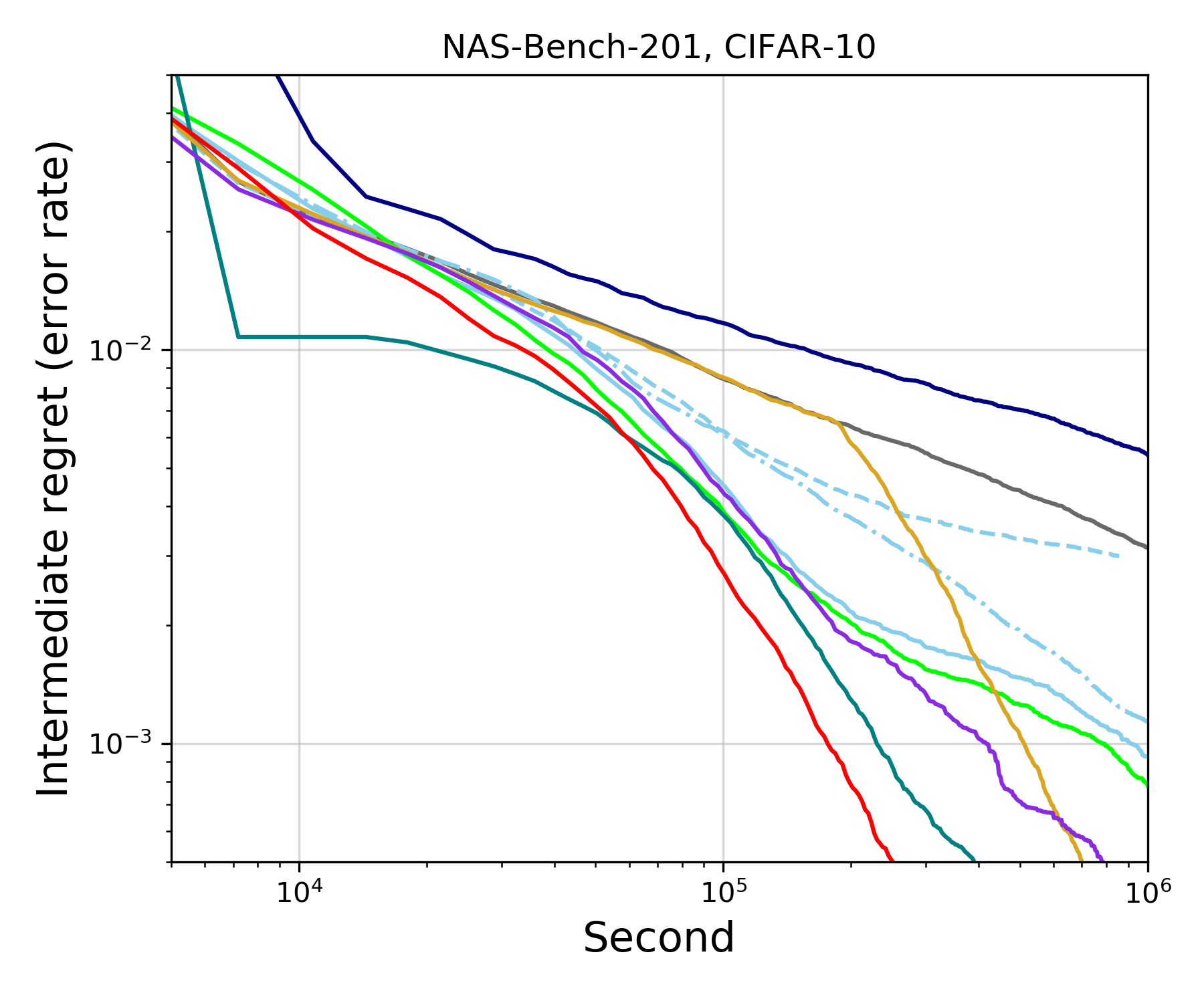}
		\hfill
		\includegraphics[width=.23\textwidth]{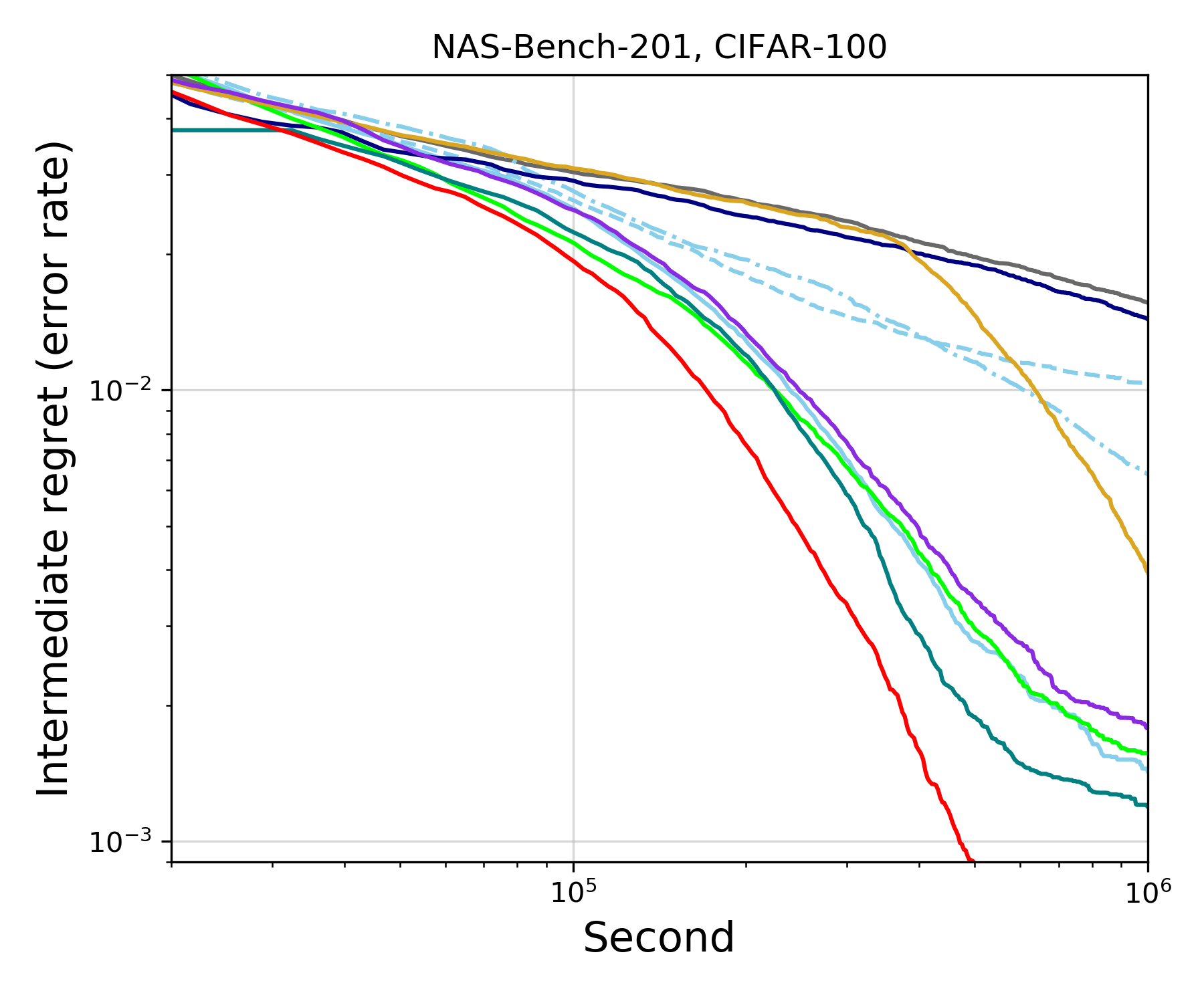}
		\hfill
		\includegraphics[width=.23\textwidth]{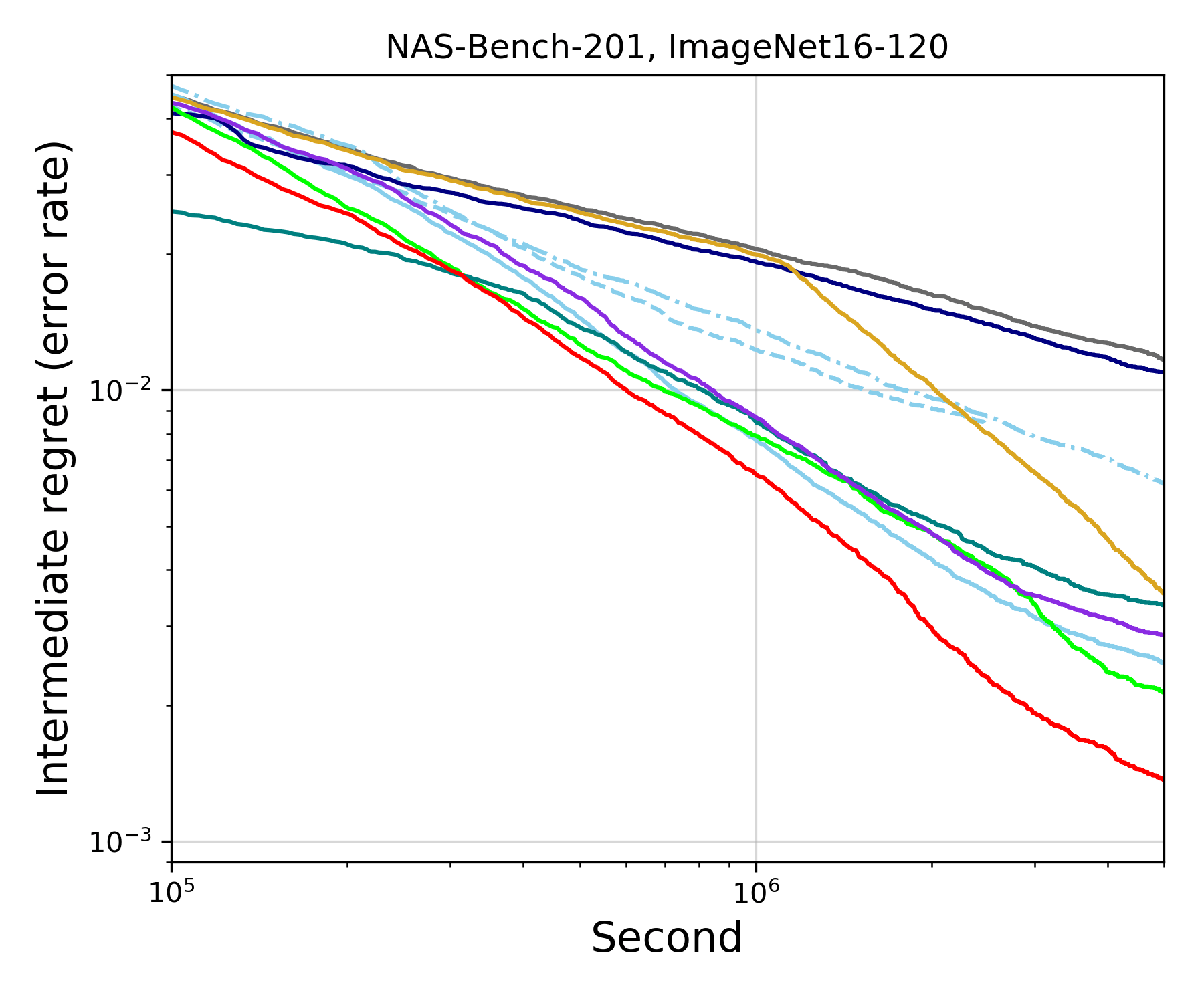}
		\hfill
		\caption{NAS-Bench tasks.}
		
	\end{subfigure}
	\\[3ex]
	\begin{subfigure}[b]{\textwidth}		
		\includegraphics[width=.23\textwidth]{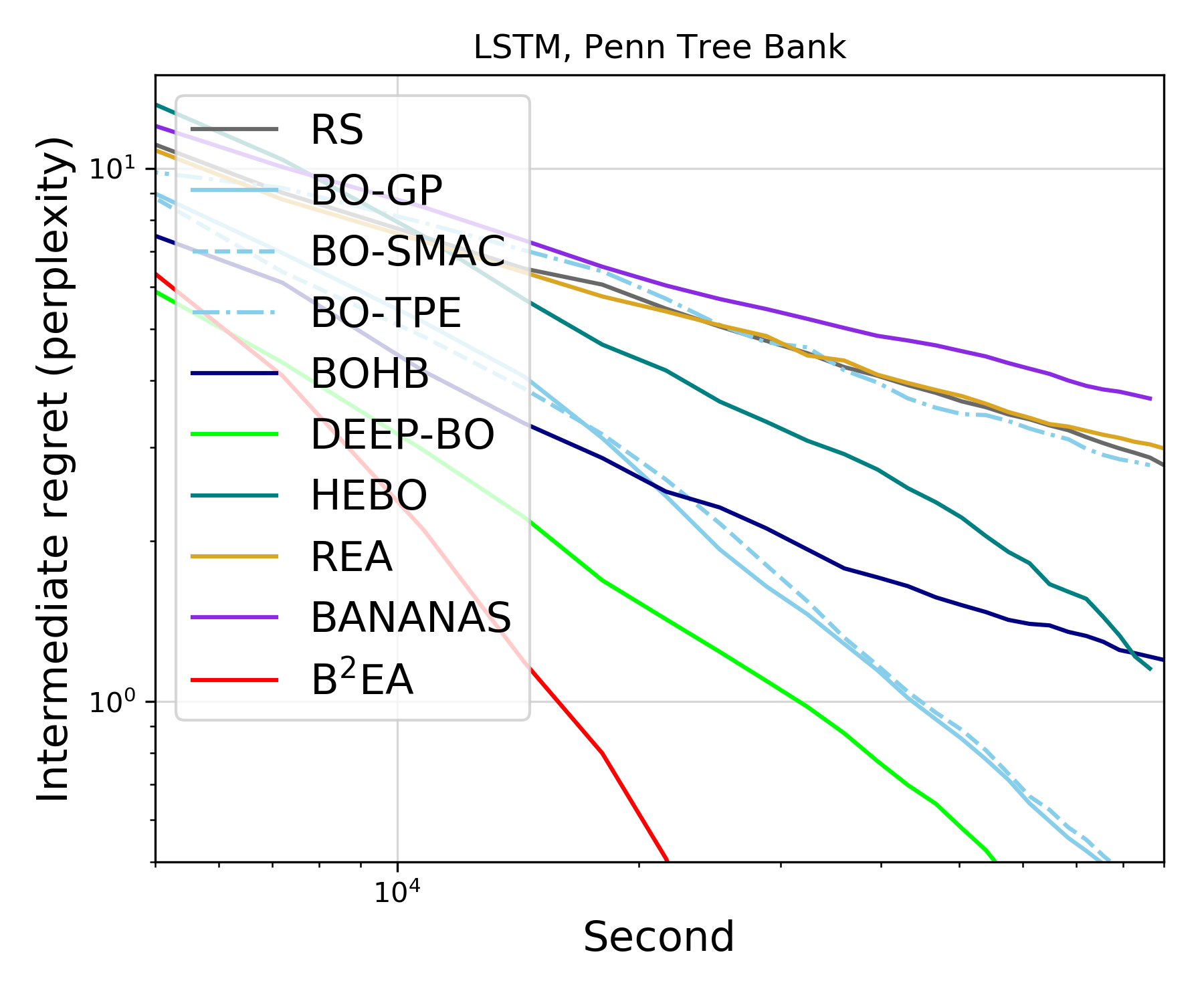}
		\hfill
		\includegraphics[width=.23\textwidth]{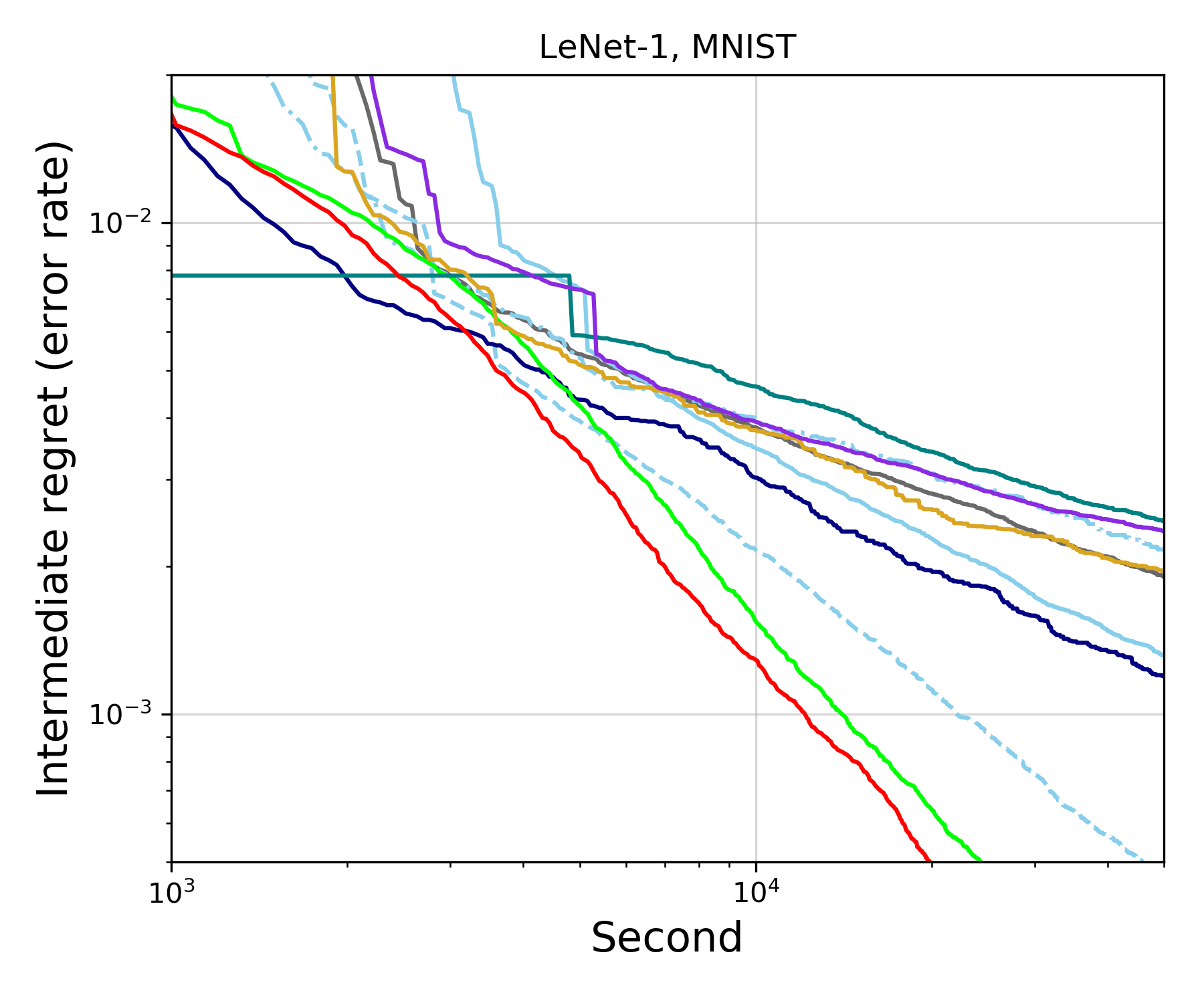}
		\hfill
		\includegraphics[width=.23\textwidth]{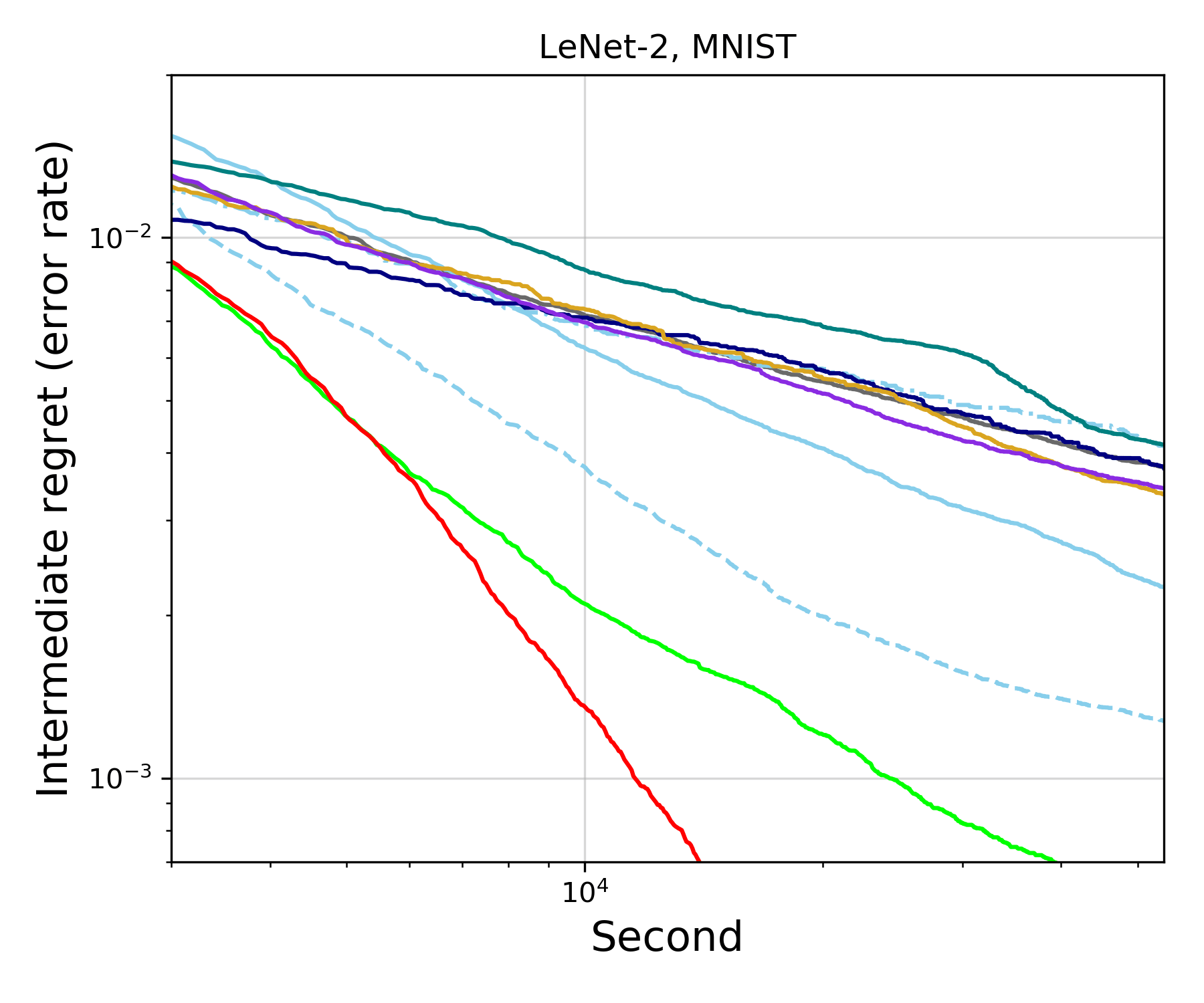}
		\hfill
		
		\vspace*{3ex}
		
		\includegraphics[width=.23\textwidth]{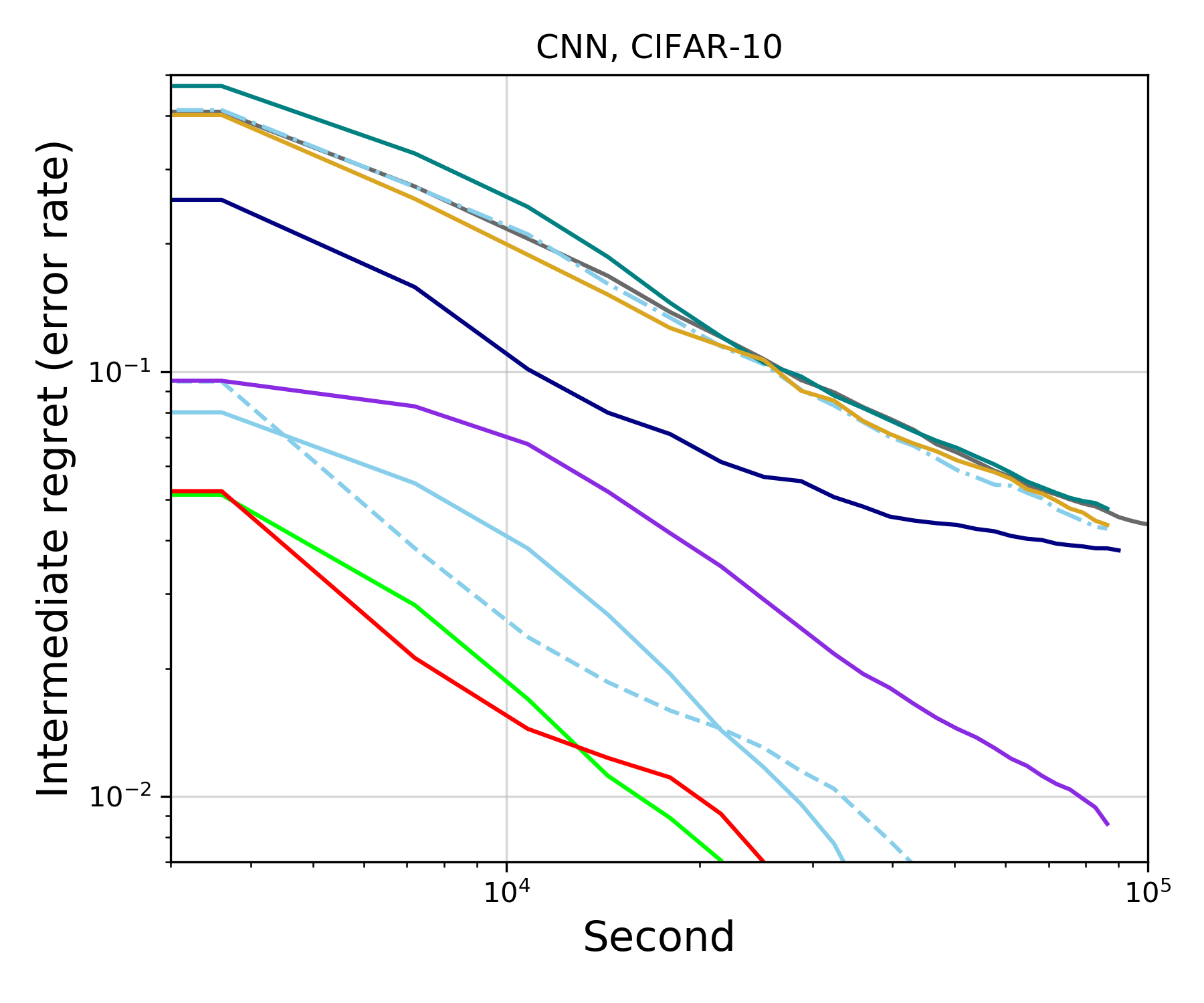}
		\hfill
		\includegraphics[width=.23\textwidth]{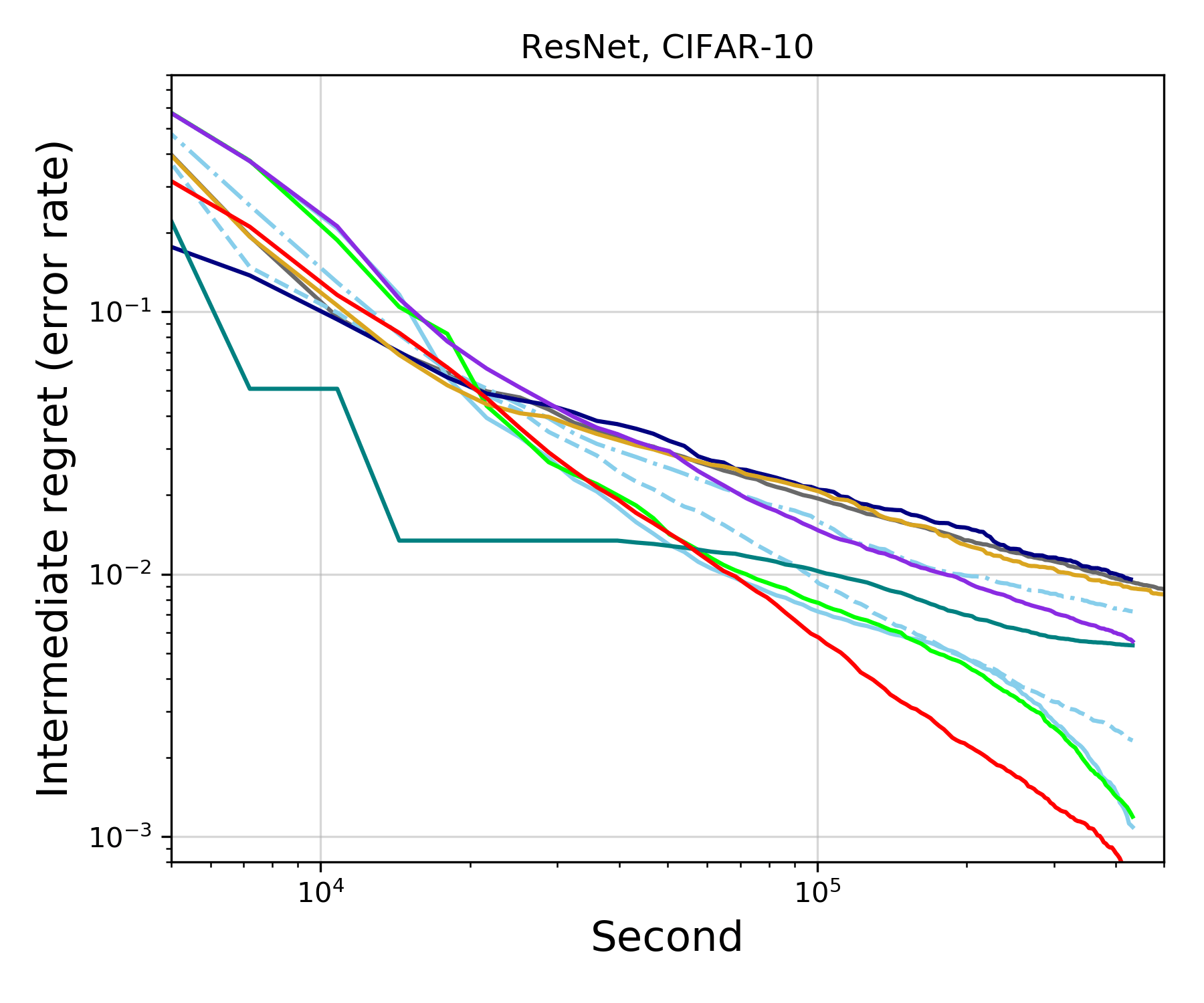}
		\hfill
		\includegraphics[width=.23\textwidth]{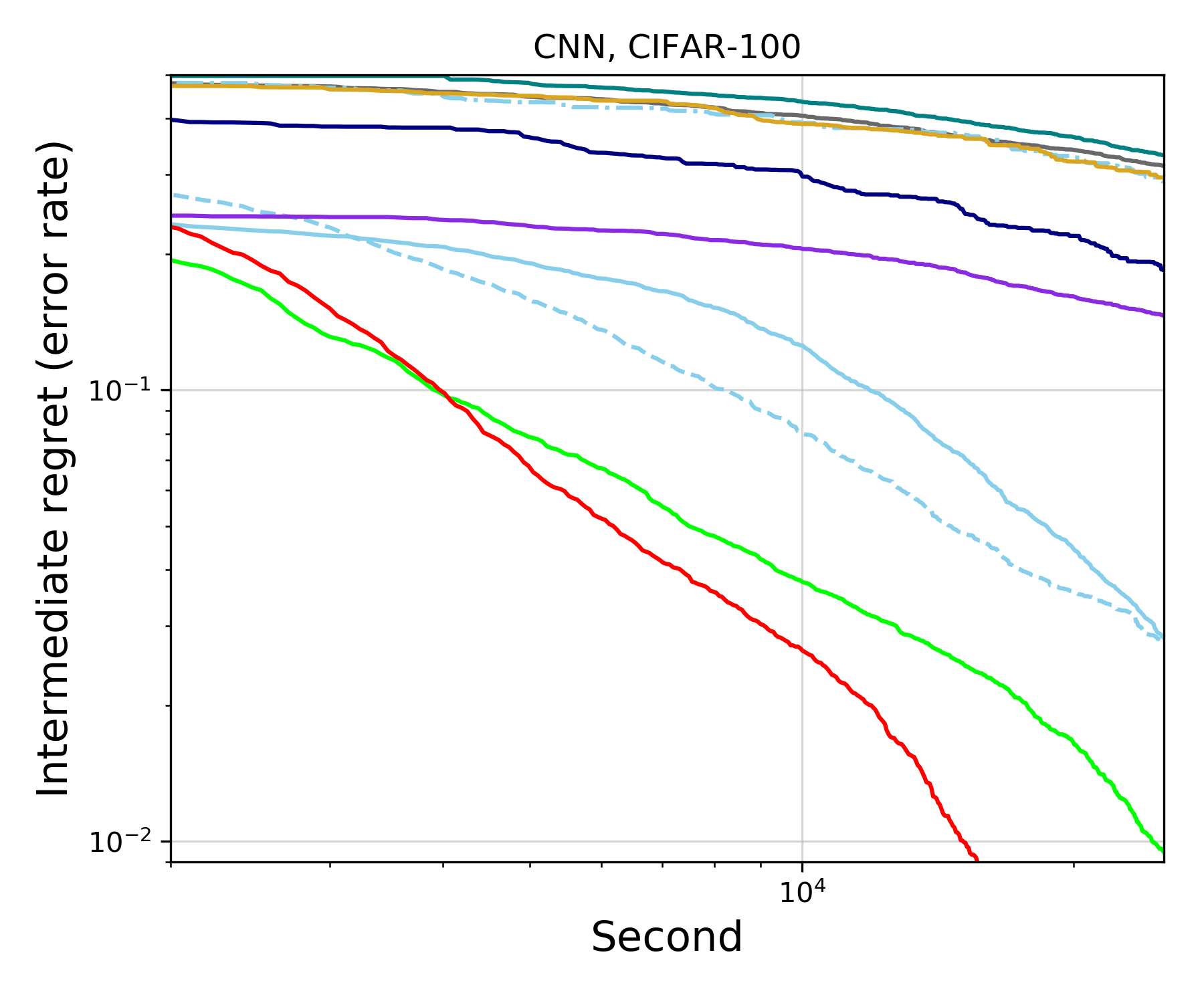}
		\hfill
		\caption{DNN-Bench tasks.}
		%\label{fig2a-a:c}	
	\end{subfigure}
	\\[3ex]
	\caption
	{Comparison of intermediate regret $r_t$ performance, but plotted for \textit{mean} $r_t$ value of 500 repeated runs. }
	\label{fig2a-a}
\end{figure*}

\begin{table*}[t]
	\centering
	%\caption*
	\caption{Success rate $\mathbb{P}(\tau \le t)$ performance (in \%) for all benchmarks. We measured $\mathbb{P}(\tau \le t)$ at time $t$ when the best-performing algorithm achieved a success rate of 99\%. The best performer in each task is shown in \textbf{bold}.}
	\scriptsize 
	%\label{tab1a}
	\resizebox{\textwidth}{!}{
		\begin{threeparttable}
			\begin{tabular}{lll rrrrrrr rrr}
				\toprule
				\multicolumn{3}{c}{\textbf{Target setting}} & \multicolumn{9}{c}{\textbf{Benchmark algorithms}} & {\begin{tabular}[r]{@{}r@{}}\textbf{Proposed}\\\textbf{algorithm}\end{tabular}}  \\
				\midrule
				\textbf{Goal} & \textbf{Source} & \textbf{Task name} & RS & {\begin{tabular}[r]{@{}r@{}}BO-\\GP\end{tabular}} & {\begin{tabular}[r]{@{}r@{}}BO-\\SMAC\end{tabular}} & {\begin{tabular}[r]{@{}r@{}}BO-\\TPE\end{tabular}} & BOHB & DEEP-BO & HEBO & BANANAS & REA &  B\textsuperscript{2}EA \\
				\midrule
				\multirow{15}{*}{\begin{tabular}[c]{@{}l@{}}Easy\\ target\\$c_e$\end{tabular}} & \multirow{4}{*}{HPO-Bench} & Parkinsons & 71.8 & 95.8 & 96.0 & 91.8 & 97.8 & 99.0 & 80.8 & 78.2 & 74.6 & \textbf{99.4} \\	
				&  & Naval  & 35.6 & 65.8 & 76.8 & 76.0 & 84.2 & 95.0 & \textbf{99.0} & 29.0 & 29.2 & 97.0  \\
				&  & Protein & 34.6 & 81.2 & 79.8 & 65.2 & 94.8 & 93.8 & \textbf{99.2} & 78.0 & 30.6 & 98.6 \\
				&  & Slice & 30.0 & 61.2 & 74.2 & 68.2 & 92.2 & 91.2 & \textbf{99.0} & 27.2 & 32.6 & 98.6 \\
				\cmidrule{2-13}
				& NAS-Bench-101 & CIFAR-10 & 55.6 & 74.6 & 56.4 & 63.0 & 49.0 & 76.0 & 86.0 & 88.4 & 65.0 & \textbf{99.4} \\
				\cmidrule{2-3}
				& \multirow{3}{*}{NAS-Bench-201} & CIFAR-10 & 56.6 & 95.8 & 85.0 & 90.8 & 33.2 & 96.8 & \textbf{99.6} & 95.2 & 78.0 & 98.8 \\
				&  & CIFAR-100 & 39.6 & 92.8 & 78.8 & 74.4 & 48.6 & 92.0 & 98.0 & 93.4 & 48.2 & \textbf{99.0} \\
				&  & ImageNet16-120 & 36.0 & 95.8 & 81.0 & 75.2 & 43.2 & 92.4 & 98.4 & 96.0 & 37.8 & \textbf{99.0} \\
				\cmidrule{2-13}
				& \multirow{6}{*}{DNN-Bench} & PTB-LSTM & 31.8 & 63.2 & 68.6 & 26.0 & 80.0 & 92.0 & 44.4 & 36.0 & 27.0 & \textbf{99.0} \\
				&  & MNIST-LeNet1 & 35.8 & 56.4 & 83.6 & 24.0 & 63.0 & 96.8 & 23.0 & 30.6 & 44.0 & \textbf{99.2} \\
				&  & MNIST-LeNet2 & 35.8 & 56.8 & 87.6 & 39.0 & 29.0 & 96.8 & 17.0 & 38.6 & 36.0 & \textbf{99.2}  \\
				&  & CIFAR10-CNN & 13.2 & 74.6 & 87.8 & 16.0 & 21.0 & 99.2 & 4.0 & 15.8 & 13.6 & \textbf{99.6} \\
				&  & CIFAR10-ResNet & 22.8 & 95.4 & 87.6 & 32.0 & 13.0 & 89.2 & 44.6 & 45.6 & 22.8 & \textbf{99.0} \\
				&  & CIFAR100-CNN & 8.0 & 52.2 & 81.8 & 9.0 & 26.0 & \textbf{99.0} & 4.0 & 14.4 & 6.8 & 98.6 \\
				\cmidrule{2-13}
				& \multirow{2}{*}{Overall success rate} & Mean   & 36.2 & 75.8 & 80.4 & 53.6 & 55.4 & 93.5 & 64.1 & 54.7 & 39.0 & \textbf{98.9}  \\
				& & Std. deviation   & 16.7 & 16.8 & 9.6 & 28.3 & 29.7 & 6.0 & 31.9 & 31.5 & 21.3 & \textbf{0.6}  \\
				\cmidrule{2-13}
				& \multirow{2}{*}{Overall rank} & Mean  & 8.7 & 4.6 & 4.8 & 6.9 & 6.1 & 3.0 & 5.0 & 6.2 & 8.2 & \textbf{1.4} \\
				& & Std. deviation   & 0.9 & 1.3 & 1.6 & 1.3 & 2.9 & 1.2 & 3.8 & 2.5 & 1.2 & \textbf{0.5} \\
				\midrule
				\multirow{15}{*}{\begin{tabular}[c]{@{}l@{}}Difficult\\ target\\$c_d$\end{tabular}} & \multirow{4}{*}{HPO-Bench} & Parkinsons & 14.2 & 94.0 & 51.8 & 78.2 & 70.2 & \textbf{99.6} & 15.4 & 27.8 & 88.8 & 98.0 \\
				&  & Naval & 4.4 & 38.4 & 24.0 & 47.4 & 32.6 & 97.4 & \textbf{99.4} & 3.6 & 13.8 & 98.6 \\
				&  & Protein & 3.8 & 81.0 & 38.2 & 24.0 & 40.6 & 92.2 & 97.8 & 63.0 & 3.6 & \textbf{99.4} \\
				&  & Slice & 5.8 & 31.6 & 39.2 & 39.2 & 52.8 & 84.6 & \textbf{99.6} & 5.4 & 23.4 & 90.4 \\
				\cmidrule{2-13}
				& NAS-Bench-101 & CIFAR-10 & 71.6 & 95.0 & 88.2 & 64.4 & 79.0 & 97.6 & \textbf{99.2} & 93.8 & 82.0 & 98.4 \\
				\cmidrule{2-3}
				& \multirow{3}{*}{NAS-Bench-201} & CIFAR-10 & 7.4 & 74.4 & 16.0 & 47.8 & 3.4 & 78.4 & 96.6 & 86.0 & 67.8 & 99.2 \\
				&  & CIFAR-100 & 3.0 & 95.4 & 12.4 & 21.8 & 4.8 & 93.8 & 97.8 & 90.8 & 21.8 & \textbf{99.0} \\
				&  & ImageNet16-120 & 5.2 & 86.4 & 27.0 & 36.8 & 9.6 & 87.4 & 78.8 & 83.3 & 58.2 & \textbf{99.0} \\
				\cmidrule{2-13}
				& \multirow{6}{*}{DNN-Bench} & PTB-LSTM & 4.2 & 41.2 & 43.8 & 3.0 & 14.0 & 68.2 & 16.6 & 6.4 & 5.8 & \textbf{99.8} \\
				&  & MNIST-LeNet1 & 6.8 & 23.4 & 75.0 & 4.0 & 12.0 & 90.0 & 1.8 & 2.8 & 6.6 & \textbf{99.4} \\
				&  & MNIST-LeNet2 & 7.0 & 21.2 & 65.4 & 5.0 & 0.0 & 77.2 & 3.0 & 6.2 & 7.4 & \textbf{99.0} \\
				&  & CIFAR10-CNN & 5.0 & 85.2 & 82.2 & 8.4 & 0.0 & 96.0 & 6.4 & 50.8 & 9.2 & \textbf{99.2} \\
				&  & CIFAR10-ResNet & 4.0 & 63.6 & 67.4 & 14.8 & 1.0 & 63.4 & 14.4 & 12.6 & 4.6 & \textbf{99.0} \\
				&  & CIFAR100-CNN & 2.4 & 58.8 & 87.6 & 2.0 & 2.0 & 95.8 & 2.4 & 7.2 & 1.4 & \textbf{99.2} \\
				\cmidrule{2-13}
				& \multirow{2}{*}{Overall success rate} & Mean   & 10.3 & 63.5 & 51.3 & 28.3 & 23.0 & 87.3 & 52.1 & 38.6 & 28.2 & \textbf{98.4}  \\
				& & Std. deviation & 17.9 & 27.6 & 26.5 & 24.5 & 27.4 & 11.5 & 45.7 & 37.5 & 31.6 & \textbf{2.3} \\
				\cmidrule{2-13}
				& \multirow{2}{*}{Overall rank} & Mean & 8.6 & 4.0 & 5.3 & 7.0 & 7.6 & 2.6 & 4.8 & 6.4 & 7.1 & \textbf{1.3} \\
				& & Std. deviation & 1.5 & 1.1 & 2.2 & 1.8 & 2.1 & 0.9 & 3.3 & 2.2 & 1.7 & \textbf{0.5} \\
				\midrule
				\multirow{15}{*}{\begin{tabular}[c]{@{}l@{}}Extremely\\ difficult\\ target\\$c_x$\end{tabular}} & \multirow{4}{*}{HPO-Bench} & Parkinsons$\dagger$ & 9.2 & 89.6 & 30.8 & 67.0 & 58.4 & 96.8 & 14.0 & 17.6 & 82.4 & \textbf{98.8} \\	
				&  & Naval  & 3.2 & 28.6 & 14.4 & 42.0 & 20.4 & 98.2 & 96.6 & 1.6 & 22.6 & \textbf{99.0}  \\
				&  & Protein & 1.8 & 82.8 & 33.0 & 15.8 & 23.6 & 91.4 & 78.0 & 62.0 & 3.0 & \textbf{98.4} \\
				&  & Slice$\dagger$ & 4.8 & 18.4 & 20.2 & 32.4 & 33.8 & 82.6 & \textbf{95.4} & 2.2 & 28.4 & 94.4  \\
				\cmidrule{2-13}
				& NAS-Bench-101 & CIFAR-10$\dagger$ & 7.0 & 24.4 & 78.2 & 11.4 & 6.4 & 27.0 & 72.8 & \textbf{84.2} & 25.0 & 41.0 \\
				\cmidrule{2-3}
				& \multirow{3}{*}{NAS-Bench-201} & CIFAR-10 & 8.4 & 75.4 & 16.2 & 51.8 & 3.6 & 79.6 & 97.2 & 76.8 & 76.2 & \textbf{99.2} \\
				&  & CIFAR-100 & 1.6 & 66.4 & 5.0 & 16.6 & 2.2 & 64.0 & 68.2 & 76.8 & 17.2 & \textbf{99.0} \\
				&  & ImageNet16-120$\dagger$ & 3.2 & 60.6 & 17.2 & 24.0 & 5.8 & 67.0 & 48.8 & 34.4 & 45.2 & \textbf{93.0} \\
				\cmidrule{2-13}
				& \multirow{6}{*}{DNN-Bench} & PTB-LSTM & 2.0 & 34.2 & 31.8 & 0.0 & 9.0 & 40.2 & 14.8 & 2.6 & 3.4 & \textbf{99.2} \\
				&  & MNIST-LeNet1$\dagger$ & 3.4 & 15.6 & 62.8 & 1.0 & 4.0 & 70.8 & 0.0 & 1.2 & 5.2 & \textbf{94.8} \\
				&  & MNIST-LeNet2 & 5.6 & 14.2 & 62.4 & 4.0 & 0.0 & 69.8 & 3.0 & 4.8 & 7.0 & \textbf{99.0}  \\
				&  & CIFAR10-CNN & 3.0 & 86.4 & 76.6 & 1.2 & 0.0 & 97.2 & 0.4 & 28.6 & 5.6 & \textbf{99.0} \\
				&  & CIFAR10-ResNet & 2.4 & 47.4 & 58.0 & 7.2 & 1.0 & 56.6 & 3.2 & 9.0 & 2.6 & \textbf{99.0} \\
				&  & CIFAR100-CNN & 0.4 & 27.0 & 83.8 & 1.0 & 3.0 & 87.8 & 1.0 & 4.2 & 0.6 & \textbf{99.2} \\
				\cmidrule{2-13}
				& \multirow{2}{*}{Overall success rate} & Mean   &  4.0 & 47.9 & 42.2 & 19.7 & 12.2 & 73.5 & 42.4 & 29.0 & 23.2 & \textbf{93.8}  \\
				& & Std. deviation   & \textbf{2.7} & 28.1 & 27.1 & 21.3 & 16.7 & 21.6 & 40.7 & 32.1 & 27.0 & 15.3  \\
				\cmidrule{2-13}
				& \multirow{2}{*}{Overall rank} & Mean  & 8.6 & 4.0 & 5.3 & 7.0 & 7.6 & 2.6 & 4.8 & 6.4 & 7.1 & \textbf{1.3} \\
				& & Std. deviation   & 1.5 & 1.1 & 2.2 & 1.8 & 2.1 & 0.9 & 3.3 & 2.2 & 1.7 & \textbf{0.5} \\
				\bottomrule	
			\end{tabular}
			\begin{tablenotes}
				\small
				\item \textit{$\dagger$ Target performance could not be achieved even after using the maximum budget.  Therefore, performance was measured for the maximum budget (i.e. $t_x = t_{max}$).}
			\end{tablenotes}
			%\vspace{-5mm}
		\end{threeparttable}		
	}
\end{table*}

\newpage

\clearpage

\newpage

%\subsection{Expected time}

%Success rate $\mathbb{P}(\tau \le t)$ performance (in \%) for all benchmarks. We measured $\mathbb{P}(\tau \le t)$ at time $t$ when the best-performing algorithm achieved a success rate of 99\%. The best performer in each task is shown in \textbf{bold}.

\begin{table*}
	\centering	
	\caption
	{Expected time  $\mathbb{E}[\tau]$ performance (in minutes) for all benchmarks. The best performer in each task is shown in \textbf{bold}.} 
	%\label{tab1b}
	\resizebox{\textwidth}{!}{
		\begin{threeparttable}
			\begin{tabular}{lll rrrrrrr rrr}
				\toprule
				\multicolumn{3}{c}{\textbf{Target setting}} & \multicolumn{9}{c}{\textbf{Benchmark algorithms}} & {\begin{tabular}[r]{@{}r@{}}\textbf{Proposed}\\\textbf{algorithm}\end{tabular}}  \\
				\midrule
				\textbf{Goal} & \textbf{Source} & \textbf{Task name} & RS & {\begin{tabular}[r]{@{}r@{}}BO-\\GP\end{tabular}} & {\begin{tabular}[r]{@{}r@{}}BO-\\SMAC\end{tabular}} & {\begin{tabular}[r]{@{}r@{}}BO-\\TPE\end{tabular}} & BOHB & DEEP-BO & HEBO & BANANAS & REA &  B\textsuperscript{2}EA \\
				\midrule
				\multirow{13}{*}{\begin{tabular}[c]{@{}l@{}}Easy\\ target\\$c_e$\end{tabular}} & \multirow{4}{*}{HPO-Bench} & Parkinsons & 68.7 & 25.4 & 26.0 & 33.3 & 19.7 & 15.9 & 47.5 & 61.7 & 54.7 & \textbf{14.6} \\	
				&  & Naval & 149.1 & 54.1 & 54.6 & 49.8 & 34.6 & 26.7 & 23.5 & 276.1 & 108.7 & \textbf{22.8}  \\
				&  & Protein & 418.6 & 123.1 & 134.9 & 189.9 & \textbf{61.0} & 86.0 & 75.2 & 151.6 & 369.5 & 64.2 \\
				&  & Slice & 944.2 & 434.4 & 302.1 & 331.1 & 124.7 & 170.9 & 134.4 & 1,970.4 & 630.6 & \textbf{115.5} \\
				\cmidrule{2-13}
				& NAS-Bench-101 & CIFAR-10 & 1,062 & 714 & 1,362 & 1,088 & 1,214 & 678 & 460 & 439 & 893 & \textbf{101}  \\
				\cmidrule{2-3}
				& \multirow{3}{*}{NAS-Bench-201} & CIFAR-10 & 4,391 & 1,206 & 2,238 & 1,727 & 13,407 & 1,010 & 919 & 1,264 & 2,570 & \textbf{848} \\
				&  & CIFAR-100 & 12,220 & 2,739 & 4,898 & 4,900 & 10,012 & 2,648 & 2,479 & 2,751 & 6,288 & \textbf{1,842} \\
				&  & ImageNet16-120 & 37,946 & 7,066 & 11,313 & 13,526 & 32,713 & 6,674 & \textbf{4,529} & 7,646 & 18,680 & 5,349 \\
				\cmidrule{2-13}
				& \multirow{6}{*}{DNN-Bench} & PTB-LSTM & 586 & 191 & 183 & 535 & 142 & 109 & 362 & 644 & 607 & \textbf{102} \\
				&  & MNIST-LeNet1 & 594 & 348 & 158 & 757 & 259 & 104 & 1,219 & 1,114 & 664 & \textbf{79} \\
				&  & MNIST-LeNet2 & 378 & 215 & 100 & 469 & 392 & \textbf{60} & 557 & 331 & 353 & 66  \\
				&  & CIFAR10-CNN & 506 & 56 & 50 & 453 & 213 & \textbf{28} & 480 & 135 & 456 & 31 \\
				&  & CIFAR10-ResNet & 7,118 & 1,040 & 1,212 & 6,496 & 10,244 & 1,082 & 2,351 & 3,370 & 6,572 & \textbf{887} \\
				&  & CIFAR100-CNN & 880 & 119 & 82 & 727 & 367 & \textbf{32} & 1,205 & 445 & 889 & 40 \\
				\cmidrule{2-13}
				& \multirow{2}{*}{\begin{tabular}[c]{@{}l@{}}Normalized \\$\mathbb{E}[\tau]$ (in \%) \end{tabular}} & Mean  & 75.9 & 24.2 & 27.1 & 52.9 & 50.4 & 15.3 & 46.4 & 54.1 & 61.8 & \textbf{10.4}  \\
				& & Std. deviation   & 23.4 & 12.7 & 23.5 & 26.1 & 35.5 & 11.8 & 38.8 & 34.4 & 22.5 & \textbf{5.2}  \\
				\midrule
				\multirow{13}{*}{\begin{tabular}[c]{@{}l@{}}Difficult\\ target\\$c_d$\end{tabular}} & \multirow{4}{*}{HPO-Bench} & Parkinsons & 1,156.1 & 92.9 & 435.6 & 170.2 & 196.9 & 54.4 & 1,329.1 & 835.3 & 153.5 & \textbf{52.1} \\
				&  & Naval & 1,450.9 & 298.0 & 711.3 & 242.5 & 385.9 & 67.0 & \textbf{54.9} & 5,840.2 & 330.4 & 67.2 \\
				&  & Protein & 4,111.6 & 263.3 & 599.7 & 935.2 & 711.0 & 178.4 & 161.9 & 355.2 & 3,181.2 & \textbf{125.5} \\
				&  & Slice & 9,549.0 & 3,511.4 & 2,491.4 & 2,068.3 & 1,705.7 & 661.3 & \textbf{326.5} & 24,380.9 & 2,152.8 & 472.7 \\
				\cmidrule{2-13}
				& NAS-Bench-101 & CIFAR-10 & 5,727 & 2,768 & 7,492 & 7,286 & 4,819 & 2,204 & \textbf{1,969} & 2,529 & 4,656 & 1,973 \\
				\cmidrule{2-3}
				& \multirow{3}{*}{NAS-Bench-201} & CIFAR-10 & 58,474 & 6,153 & 74,382 & 11,031 & 128,338 & 5,108 & 3,147 & 3,700 & 6,492 & \textbf{1,986} \\
				&  & CIFAR-100 & 132,392 & 5,133 & 75,683 & 32,784 & 98,750 & 5,049 & 4,719 & 6,241 & 20,905 & \textbf{3,727} \\
				&  & ImageNet16-120 & 646,652 & 34,178 & 161,390 & 139,496 & 428,714 & 35,196 & 47,285 & 41,763 & 71,998 & \textbf{21,523} \\
				\cmidrule{2-13}
				& \multirow{6}{*}{DNN-Bench} & PTB-LSTM & 4,640 & 574 & 610 & 10,200 & 1,679 & 380 & 1,385 & 4,777 & 5,096 & \textbf{213} \\
				&  & MNIST-LeNet1 & 7,256 & 2,687 & 535 & 10,417 & 3,866 & 391 & 78,876 & 24,944 & 7,736 & \textbf{284} \\
				&  & MNIST-LeNet2 & 3,917 & 1,333 & 387 & 6,833 & 8,566 & 294 & 12,799 & 5,129 & 2,313 & \textbf{134} \\
				&  & CIFAR10-CNN & 5,248 & 318 & 307 & 4,576 & 84,194 & \textbf{187} & 6,832 & 636 & 7,739 & 195 \\
				&  & CIFAR10-ResNet & 86,421 & 5,118 & 4,499 & 37,691 & 197,037 & 4,688 & 40,752 & 31,657 & 150,634 & \textbf{2,325} \\
				&  & CIFAR100-CNN & 4,957 & 231 & 150 & 3,313 & 2,541 & 121 & 7,207 & 1,127 & 4,256 & \textbf{96} \\
				\cmidrule{2-13}
				& \multirow{2}{*}{\begin{tabular}[c]{@{}l@{}}Normalized \\$\mathbb{E}[\tau]$ (in \%) \end{tabular}} & Mean  & 55.5 & 7.8 & 23.2 & 31.2 & 48.2 & 4.7 & 34.9 & 33.6 & 30.0 & \textbf{3.6}  \\
				& & Std. deviation   & 33.1 & 9.1 & 29.5 & 31.9 & 37.4 & 7.3 & 43.4 & 33.7 & 28.0 & \textbf{6.6}  \\
				\midrule
				\multirow{13}{*}{\begin{tabular}[c]{@{}l@{}}Extremely\\ difficult\\ target\\$c_x$\end{tabular}} & \multirow{4}{*}{HPO-Bench} & Parkinsons & 2,049 & 150 & 1,049 & 294 & 316 & 94 & 2,456 & 1,834 & 218 & \textbf{78} \\	
				&  & Naval & 3,222 & 588 & 1,464 & 401 & 722 & 88 & \textbf{82} & 14,839 & 522 & 88  \\
				&  & Protein & 7,677 & 306 & 914 & 1,808 & 1,247 & 217 & 315 & 487 & 6,077 & \textbf{160} \\
				&  & Slice & 18,538 & 7,829 & 5,947 & 3,323 & 3,353 & 849 & \textbf{524} & 74,631 & 3,253 & 551 \\
				\cmidrule{2-13}
				& NAS-Bench-101 & CIFAR-10 & 66,742 & 28,286 & 114,334 & 52,367 & 78,461 & 20,009 & 7,402 & \textbf{4,170} & 26,365 & 15,089  \\
				\cmidrule{2-3}
				& \multirow{3}{*}{NAS-Bench-201} & CIFAR-10 & 138,779 & 8,824 & 184,487 & 21,380 & 267,942 & 7,340 & 5,527 & 6,857 & 8,500 & \textbf{2,452} \\
				&  & CIFAR-100 & 255,135 & 7,844 & 185,194 & 45,525 & 174,796 & 8,215 & 7,474 & 9,259 & 26,393 & \textbf{4,148} \\
				&  & ImageNet16-120 & 1,427,033 & 76,921 & 294,908 & 275,107 & 785,983 & 74,433 & 118,953 & 162,487 & 116,940 & \textbf{37,488} \\
				\cmidrule{2-13}
				& \multirow{6}{*}{DNN-Bench} & PTB-LSTM & 8,756 & 772 & 883 & $\infty$ & 3,262 & 676 & 1,623 & 9,975 & 7,732 & \textbf{263} \\
				&  & MNIST-LeNet1 & 14,743 & 4,505 & 793 & 47,326 & 9,089 & 643 & $\infty$ & 59,236 & 9,300 & \textbf{458} \\
				&  & MNIST-LeNet2 & 6,991 & 2,665 & 702 & 9,924 & 18,524 & 529 & 24,555 & 12,159 & 10,969 & \textbf{186}  \\
				&  & CIFAR10-CNN & 9,684 & 373 & 392 & 19,031 & $\infty$ & \textbf{224} & 42,028 & 1,175 & 12,440 & 233 \\
				&  & CIFAR10-ResNet & 140,961 & 7,203 & 5,811 & 66,749 & 197,037 & 5,652 & 212,288 & 49,325 & 233,875 & \textbf{2,431} \\
				&  & CIFAR100-CNN & 15,568 & 368 & 209 & 14,047 & 4,950 & 196 & 17,548 & 2,030 & 11,557 & \textbf{139} \\
				\cmidrule{2-13}
				& \multirow{2}{*}{\begin{tabular}[c]{@{}l@{}}Normalized \\$\mathbb{E}[\tau]$ (in \%) \end{tabular}} & Mean  & 61.0 & 6.7 & 25.2 & 31.4 & 43.6 & 3.7 & 40.9 & 41.9 & 33.9 & \textbf{2.3}  \\
				& & Std. deviation   & 32.0 & 6.0 & 32.6 & 25.9 & 33.0 & 4.4 & 47.3 & 43.1 & 33.4 & \textbf{3.3}  \\
				\bottomrule

			\end{tabular}
			
		\end{threeparttable}		
	}
\end{table*}

\newpage

\clearpage

\newpage

\clearpage

\newpage

\begin{table*}[h]
	\centering
	\caption
	{Top-5 performers for each benchmark in terms of normalized expected time $\mathbb{E}[\tau]$ performance. They are ordered according to the ranking.} 
		%Note that the normalized $\mathbb{E}[\tau]$ denotes the relative performance of an algorithm to the $\mathbb{E}[\tau]$ of the worst algorithm in each task: smaller the $\mathbb{E}[\tau]$, the better is the result. 
	\scriptsize 
	%\small 
	%\label{tab1bs}
	%\resizebox{\linewidth}{!}{
		\begin{tabular}{l rrr}
			\toprule
			\multirow{2}{*}{\textbf{Benchmark}} &  \multicolumn{3}{c}{\textbf{Algorithm (Normalized mean $\mathbb{E}[\tau]$)}} \\
			\cmidrule{2-4}
			&  Easy target $c_e$ & Difficult target $c_d$ & Extremely difficult target $c_x$ \\
			\midrule
			\multirow{5}{*}{HPO-Bench}  & B\textsuperscript{2}EA (12.67) & B\textsuperscript{2}EA (2.52) & B\textsuperscript{2}EA (1.65) \\
			& DEEP-BO (15.51) & DEEP-BO (3.07) & DEEP-BO (2.09) \\
			& BOHB (15.54) & BO-GP (8.23) & BO-GP (6.13) \\
			& HEBO (25.62) & BOHB (11.43) & BO-TPE (10.67)    \\
			& BO-SMAC (26.31) & HEBO (26.55) & HEBO (26.34)  \\
			\midrule
			\multirow{5}{*}{NAS-Bench}  & B\textsuperscript{2}EA (10.73) & B\textsuperscript{2}EA (8.51) & B\textsuperscript{2}EA (1.65) \\
			& HEBO (18.20) & HEBO (9.90) & HEBO (4.95) \\
			& BANANAS (21.08) & DEEP-BO (10.66) & BANANAS (5.31) \\
			& DEEP-BO (24.15) & BANANAS (11.95) & DEEP-BO (7.17)   \\
			& BO-GP (25.61)  & BO-GP (12.73) & BO-GP (9.12)   \\
			\midrule
			\multirow{5}{*}{DNN-Bench}  & B\textsuperscript{2}EA (8.72) & B\textsuperscript{2}EA (1.04) & B\textsuperscript{2}EA (1.09) \\
			& DEEP-BO (9.14) & DEEP-BO (1.80) & DEEP-BO (2.35) \\
			& BO-SMAC (14.64) & BO-SMAC (2.40) & BO-SMAC (2.94) \\
			& BO-GP (21.32) & BO-GP (4.27) & BO-GP (5.38)   \\
			& BOHB (47.75) & BANANAS (25.17) & BANANAS (47.50)  \\
			\bottomrule
		\end{tabular}
%	}
\end{table*}

\begin{table*}[h]
	\centering
	\caption
	{Summary of normalized expected time $\mathbb{E}[\tau]$ performance for each benchmark. Performance worse than random search is shown in {\color[HTML]{FF0000}red} color.} 
	%\label{tab1bs}
	\resizebox{\textwidth}{!}{
		\begin{threeparttable}
			\begin{tabular}{l l rrrrrrrrrr}
				\toprule
				\textbf{Target} & \textbf{Benchmark} & \textbf{RS} & \textbf{BO-GP} & \textbf{BO-SMAC} & \textbf{BO-TPE} & \textbf{BOHB} & \textbf{DEEP-BO} & \textbf{HEBO} & \textbf{BANANAS} & \textbf{REA} & \textbf{B\textsuperscript{2}EA} \\
				\midrule
				& HPO-Bench & 75.5 & 27.0 & 26.3 & 32.2 & 15.5 & 15.5 & 25.6 & 81.5 & 59.8 & \textbf{12.7} \\
				& NAS-Bench & 77.7 & 25.6 & 46.6 & 42.1 & {\color[HTML]{FF0000} 89.3} & 24.1 & 18.2 & 21.1 & 46.3 & \textbf{10.7} \\
				\multirow{-3}{*}{\begin{tabular}[c]{@{}l@{}}Easy\\ target $c_e$\end{tabular}} & DNN-Bench & 75.0 & 21.3 & 14.6 & 73.8 & 47.7 & 9.1 & {\color[HTML]{FF0000} 79.0} & 57.9 & 73.4 & \textbf{8.7} \\
				\midrule
				& HPO-Bench & 62.75 & 8.23 & 17.44 & 12.05 & 11.43 & 3.07 & 26.55 & 67.87 & 25.85 & \textbf{2.52} \\
				& NAS-Bench & 80.50 & 12.73 & 60.02 & 38.05 & 76.30 & 10.66 & 9.90 & 11.95 & 23.53 & \textbf{8.51} \\
				\multirow{-3}{*}{\begin{tabular}[c]{@{}l@{}}Difficult\\ target $c_d$\end{tabular}} & DNN-Bench & 34.03 & 4.27 & 2.40 & {\color[HTML]{FF0000} 39.52} & {\color[HTML]{FF0000} 53.92} & 1.80 & {\color[HTML]{FF0000} 57.06} & 25.17 & 37.09 & \textbf{1.04} \\
				\midrule
				& HPO-Bench & 57.50 & 6.13 & 18.11 & 10.67 & 9.62 & 2.09 & 26.34 & {\color[HTML]{FF0000} 70.25} & 23.97 & \textbf{1.65} \\
				& NAS-Bench & 77.54 & 9.12 & 65.53 & 22.73 & 73.05 & 7.17 & 4.95 & 5.31 & 11.19 & \textbf{4.59} \\
				\multirow{-3}{*}{\begin{tabular}[c]{@{}l@{}}Extremely\\ difficult\\ target $c_x$\end{tabular}} & DNN-Bench$\dagger$ & 52.19 & 5.38 & 2.94 & \multicolumn{1}{r}{{\color[HTML]{FF0000} Fail}} & \multicolumn{1}{r}{{\color[HTML]{FF0000} Fail}} & 2.35 & \multicolumn{1}{r}{{\color[HTML]{FF0000} Fail}} & 47.50 & {\color[HTML]{FF0000} 55.56 } & \textbf{1.09} \\
				\bottomrule	
			\end{tabular}
			
			%\vspace{-8mm}	
			\begin{tablenotes}
				%\small
				\item \textit{$\dagger$ Compared with the algorithms that achieve the target at least once.}
			\end{tablenotes}
		\end{threeparttable}		
	}
\end{table*}

\newpage

\clearpage

\newpage

\section{Implementation Details}

First, we note that the code is provided as a part of the supplementary material. 
% TODO:reveal github URL here when it is released as arXiv version
To produce the experiment results in our work, the proposed algorithms and the benchmark algorithms were merged into a unified package. 
Because the software licenses included in this project are all GPL-compatible, the code will be open-sourced under the GPL-3.0 license.

\begin{algorithm}[t]
	
	\caption
	{B\textsuperscript{2}EA}
	%\label{alg:B2EA}
	\small
	%\scriptsize
	
	\begin{algorithmic}
		\State {\bfseries Inputs:} black-box function $f$, diversified BO models $\{\hat{f}_1,\dots, \hat{f}_{B}\}$, full search space $\mathcal{X}$
		
		\State \textcolor{blue}{\# --- \textit{Initialization}}
		\State Choose $\mathbf{x}^{init}_{1}, \mathbf{x}^{init}_{2} \in \mathcal{X}$
		\State Evaluate $\mathbf{y}^{init}_{1} \leftarrow f(\mathbf{x}^{init}_{1})$, $\mathbf{y}^{init}_{2} \leftarrow f(\mathbf{x}^{init}_{2})$ 
		\State Set {$\mathcal{H}\leftarrow \{ (\mathbf{x}^{init}_{1}, \mathbf{y}^{init}_{1}), (\mathbf{x}^{init}_{2}, \mathbf{y}^{init}_{2}) \} $}
		
		\For{$n=1, 2, \dotsc$}
		\State \textcolor{blue}{\# --- \textit{Module A (Populate with first BO)}}
		\State Set $\hat{f}_{1^{st}}$ as $\hat{f}_m$ where $m=\mbox{mod}(n,B)+1$
		
		\State Set $\mathcal{P}^n$ as $\underset{\mathbf{x} \in \mathcal{X}}{\argtopk} \hat{f}_{1^{st}}(\mathbf{x};\mathcal{H})$
		
		\State \textcolor{blue}{\# --- \textit{Module B (Apply genetic operator)}}
		\State Start with $K$ parents. Apply mutation with the probability of $0.5$ to form $\mathcal{C}^n$ that contains $M$ offspring.
		
		\State \textcolor{blue}{\#--- \textit{Module C (1. Estimate and choose the best with second BO; 2. Evaluate)}}
		
		\State Set $\hat{f}_{2^{nd}}$ as one of $\{\hat{f}_1,\dots, \hat{f}_{B}\}$. Choose randomly but exclude  $\hat{f}_{1^{st}}$ ($=\hat{f}_m$).  
		
		\State Select $\mathbf{x}^{n} \in \underset{\mathbf{x} \in \mathcal{C}^n}{\argmax} \hat{f}_{2^{nd}}(\mathbf{x};\mathcal{H})$
		
		\State Evaluate ${y}_{n} \leftarrow f(\mathbf{x}_{n})$ 
		
		\State \textcolor{blue}{\#--- \textit{Stop Criteria}}		
		\State Check for exit criteria
		
		\State \textcolor{blue}{\#--- \textit{Update $\mathcal{H}$}}			
		\State Update {$\mathcal{H}\leftarrow\mathcal{H}\cup \{(\mathbf{x}_{n}, {y}_{n})\}$}
		\EndFor
		\State {\bfseries *Note:} For simplicity, we omitted $k(\cdot)$ and $g(\cdot)$ for surrogate $\hat{f}$ modeling, and also early termination strategy $f_{et}(\cdot)$. 
	\end{algorithmic}
	
\end{algorithm}

\textbf{Proposed algorithms} 
are designed to be a meta-heuristic algorithm framework where the Evolutionary Algorithm (EA) and Bayesian Optimization (BO) can harmonize in a unified manner. In this regard, we implemented the following components:  

\begin{itemize}

	\item \textbf{EA components}: While we followed the workflow similar to the implementation\footnote{\url{https://colab.research.google.com/github/google-research/google-research/blob/master/evolution/regularized_evolution_algorithm/regularized_evolution.ipynb}} of \citet{real2019regularized}, the mutation for the mixed-type parameters was newly implemented.

	\item \textbf{BO models}: To utilize two different models and three acquisition functions, we started from publicly available codes and partially implemented the Gaussian process (GP) model, random forest (RF) model, and the associated utility functions. 
	The GP model was based on the \textsf{InputWarpedGP} class of the GPy library~\cite{gpy2014}
	%The GP model was based on the Snoek's \textit{Spearmint} version~\footnote{\url{github.com/JasperSnoek/spearmint/tree/master/spearmint}} released under the GPL-3.0 license,  
	and our RF model was based on the \textsf{RandomForestRegressor} class of the scikit-learn library~\cite{pedregosa2011scikit}, which is open-sourced under a BSD license.
	More specifically, the kernel design in our GP model is almost the same as HEBO~\cite{cown2020emprical}, where the linear and Mat'ern 3/2 kernels were used with automatic relevance determination. However, its hyper-parameters were solely sampled using Monte Carlo estimation of the acquisition function values to reduce the modeling cost of GP regression. 
	Second, the number of trees in our RF model has been set to 50, and the minimum number of elements in a split has been set at two. 
	For the three acquisition functions, Snoek's implementation~\footnote{\url{github.com/JasperSnoek/spearmint/tree/master/spearmint}} was used with the noise $epsilon$ set to 0.0001.
	
	\item \textbf{Input transformation}: For our input warped GP model, the cumulative distribution function of the Kumaraswamy distribution was used to handle non-stationary functions~\cite{snoek14}.
	
	\item \textbf{Output transformation}: To stabilize variance and minimize skewness in the power transformation, the box-cox and yeo-johnson transformations were employed in the same way as in HEBO~\cite{cown2020emprical}. For each transformation, the  \textsf{power\_transform} implementation from the scikit-learn was used.

	%\item \textbf{Sampling methods}: We applied various samplers, such as Sobol sequence, Latin hypercube, and Cartesian grids provided by Spearmint and PyDOE (BSD-3 license). 

	\item \textbf{Early termination rule}: We adopted the basic heuristics of \citet{cho2020basic} for the early termination based on DEEP-BO\footnote{\url{github.com/snu-adsl/DEEP-BO}}, released under a GPL-3.0 license.
	Based on the prior assumption of the correlation between validation performance during training and the best one found after completion, the early termination rule percentile $\beta$ controls the proper value of thresholds in earlier and later training epochs.
	Similar to \citet{cho2020basic}, $\beta$ was set to 0.1 for joint HPO-NAS tasks (HPO-Bench and DNN-Bench). 
	Because the NAS-Bench tasks have been carefully tuned for training-related parameters, $\beta$ was set to 0.25 for NAS-Bench tasks.   

\end{itemize}

\textbf{Benchmark algorithms} 
are officially implemented in the following packages with the associated software license: \textit{Spearmint}~\cite{snoek2012practical} (GPL-3), \textit{Hyperopt}~\cite{bergstra2013making} (proprietary), \textit{HpBandSter}~\cite{falkner18bohb} (BSD-3), \textit{naszilla}~\cite{white2019bananas} (Apache-2.0), and \textit{HEBO}~\cite{cown2020emprical} (MIT).
%Note that some implementations of these algorithms have been integrated into the \textbf{/algo} directory of the code.
%
%For the hyperparameter setting of the benchmark algorithms, we used the same settings that have been used in the previous studies. 
Specifically, the following references were used in our benchmark experiments:
\begin{itemize}
	\item AutoML NAS experiments\footnote{\url{https://github.com/automl/nas_benchmarks/tree/development/experiment_scripts}} (BSD-3)
	
	\item NAS201 experiments\footnote{\url{https://github.com/D-X-Y/AutoDL-Projects/blob/master/docs/NAS-Bench-201.md}} (MIT)
	
	\item BANANAS experiments\footnote{\url{https://github.com/naszilla/naszilla}} (Apache-2.0) 
\end{itemize}
%
%To compare the performance of the algorithms executed in the different experimental scripts above, we saved them in a simple report that contain the search trajectories. More specifically, the experimental script was slightly modified to report the intermediate results as a simple JSON file.  

\clearpage

\newpage

\section{Experimental Details}

\begin{table*}[h]
	\centering
	\caption
	{Summary of target and budget settings. To fairly compare the overall benchmark tasks, two target goals $c_e$, $c_d$ and, $c_x$ were set as the top 1 \%, 0.05 \%  and 0.02 \% best configuration performances, respectively. 
		In addition, the values of $t_e$, $t_d$ and $t_x$ were determined when any algorithm was over 99 \% for $\mathbb{P}(\tau \le t)$ according to the target performance setting.
		If no algorithm could successfully achieve the above condition, the time was set to be the same as the maximum budget.
	}
	%\label{tab-a2}
	\small
	\resizebox{\linewidth}{!}{
		\begin{threeparttable}	
			\begin{tabular}{ll ll ll ll r}
				\toprule
				\multirow{2}{*}{\textbf{Source}} & \multicolumn{1}{l}{\multirow{2}{*}{\textbf{Task}}} & \multicolumn{2}{c}{\textbf{Easy target}} & \multicolumn{2}{c}{\textbf{Difficult target}} & \multicolumn{2}{c}{\textbf{Extremely difficult target}} & \multicolumn{1}{c}{\multirow{2}{*}{\textbf{Max budget} $t_{max}$}}\\
				\cmidrule{3-8}
				& \multicolumn{1}{c}{} & \textbf{Regret} $c_e$ & \textbf{Budget} $t_e$ & \textbf{Regret} $c_d$ & \textbf{Budget} $t_d$ & \textbf{Regret} $c_x$ & \textbf{Budget} $t_x$ & \multicolumn{1}{c}{} \\
				\midrule
				\multirow{4}{*}{HPO-Bench} & Parkinsons$\dagger$ & 0.0114 & 80m & 0.00459 & 240m & 0.00353 & 300m & 300m \\
				& Naval & 9.74109E-05 & 90m & 1.71E-05 & 180m & 1.19E-05 & 252m & 300m \\
				& Protein & 0.034537211 & 190m & 0.00957 & 360m & 0.00568 & 480m & 500m \\
				& Slice$\dagger$ & 0.000312 & 6h & 9.46E-05 & 21h & 9.35E-05 & 28h & 28h \\
				\midrule
				\multicolumn{1}{l}{NAS-101-Bench} & CIFAR-10$\dagger$ & 0.01 & 16h & 0.0048 & 130h & 0.0041 & 138h & 138h \\
				\cmidrule{1-2}
				\multirow{3}{*}{NAS-201-Bench} & CIFAR-10 & 0.00648 & 65h & 0.00168 & 120h & 0.00168 & 132h & 138h \\
				& CIFAR-100 & 0.0206 & 108h & 0.004 & 192h & 0.002 & 205h & 277h \\
				& ImageNet16-120$\dagger$ & 0.017 & 294h & 0.005 & 1124h & 0.002 & 1388h & 1388h \\
				\midrule
				\multirow{6}{*}{DNN-Bench} & PTB-LSTM & 4.7 & 5h & 1.456 & 8h & 0.734 & 9h & 12h \\
				& MNIST-LeNet1$\dagger$ & 0.0025 & 280m & 0.0006 & 12h & 0.0002 & 12h & 12h \\
				& MNIST-LeNet2 & 0.0056 & 3h & 0.0019 & 7h & 0.0015 & 12h & 15h \\
				& CIFAR10-CNN & 0.0908 & 80m & 0.0213 & 9h & 0.0128 & 10h & 14h \\
				& CIFAR10-ResNet & 0.0155 & 45h & 0.0055 & 108h & 0.0034 & 108h & 120h \\
				& CIFAR100-CNN & 0.2059 & 110m & 0.0471 & 4h & 0.03477 & 5h & 7h \\
				\bottomrule
			\end{tabular}
			\begin{tablenotes}
				\small
				\item \textit{\textbf{m} and \textbf{h} refer to minute and hour, respectively.}
				\item \textit{$\dagger$ No algorithm could achieve over 99 \% of success rate for $t_{x}$ until the maximum budget $t_{max}$ was used.}
			\end{tablenotes}
		\end{threeparttable}
	}
\end{table*}

\textbf{Tabular benchmark datasets}: 
As summarized in the experiment section of the paper, we used four benchmark datasets saved as database files. 
Specifically in NAS-Bench, we used the files as following: \textit{nasbench\_full.tfrecord} and \textit{NAS-Bench-201-v1\_1-096897.pth}.
%

% Remarks on different hardware, versions of DL libraries, different runtimes for the different methods
\textbf{Computing environment}: 
In general, the training time of a DNN is highly dependent on the specific hardware and the version of DL library. In our experiments, however, the training time was consistently calculated thanks to the use of tabular benchmark datasets. 
We randomized the computing resource assignments among different computing machines to measure the modeling time of surrogate functions. Intel Core i7 processors are commonly employed in our local devices. C5 instances are utilized in Amazon EC2.
The modeling time is significantly less than DNN training time anyway.

%\textbf{Proxy metric}:
%
%While the mean test performance of $f(\textbf{x})$ of the maximum epoch was used as the final performance in the previous studies~\cite{ying2019bench,dong2020nasbench201} (even when a full training is not performed by the algorithm, the full training result of the chosen algorithm), we used a noisy validation performance of $f(\textbf{x})$ at the training epochs instead.
%While using the test performance could closely simulate real world computational constraints, % excerpted from~\cite{ying2019bench} this could mislead the HPO algorithm comparison.
%In this regard, we noticed a flaw in the experiment script of the previous works, where a multi-fidelity optimization method (i.e., BOHB) was benchmarked with the final test performance. To be specific, even though a configuration terminated early with shorter epoch lengths, the mean test performance at the final epochs was reported for comparison.  

\textbf{Invalid candidate handling}:
Even when we tested all the algorithms on a well-defined search space, some candidate configurations could not provide satisfactory results due to known limitations. For instance, a performance record of a valid configuration can be absent in the lookup table because of the constraints on the data generation process in NAS-Bench-101.
In this regard, we considered the following fail-over:
if the specification of a candidate was invalid, then we removed this case from the candidate set or chose a valid one instead. 
Moreover, if a candidate was valid but its result was not in the database, then we approximated it to the one whose evaluation result existed in the table.

%\textbf{Run environment setup}:
%
%To reproduce the benchmark results in our work, refer to the README.md of the aforementioned code. 

%\clearpage

%\newpage

\section{Further Explanations}

\begin{figure*}[h]
	\begin{subfigure}[b]{0.225\textwidth}
		\includegraphics[width=\columnwidth]{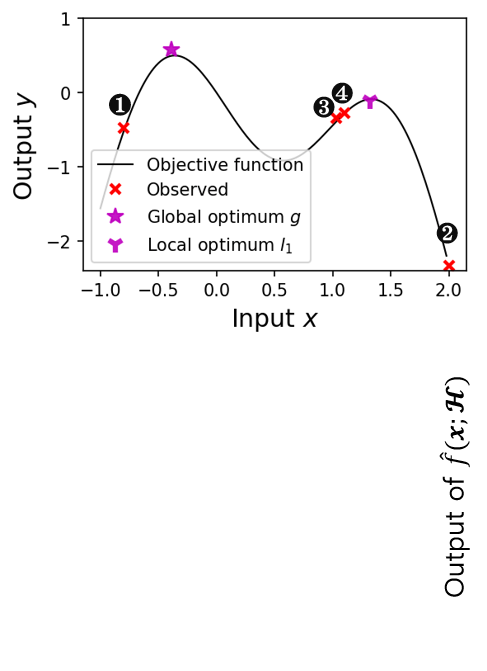}
		\caption{Search history}
		%\label{fig2:a}
	\end{subfigure}
	\hfill			
	\begin{subfigure}[b]{0.22\textwidth}
		\includegraphics[width=\columnwidth]{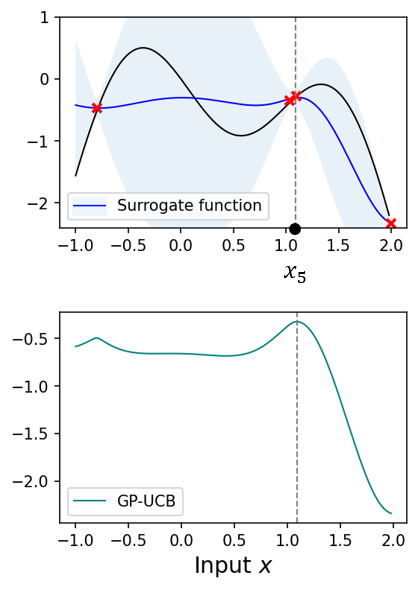}
		\caption{Selected by GP-UCB}
		%\label{fig2:b}
	\end{subfigure}
	\hfill
	\begin{subfigure}[b]{0.22\textwidth}
		\centerline{\includegraphics[width=\columnwidth]{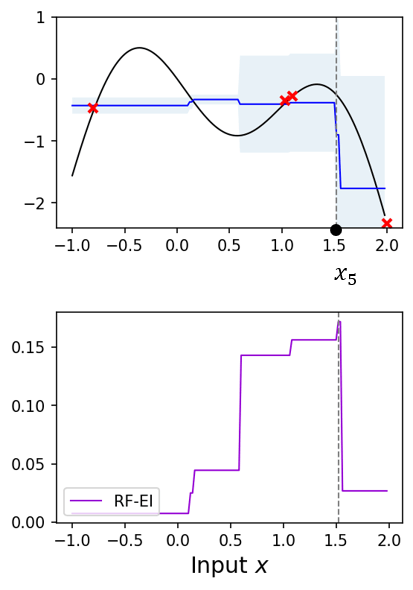}}
		\caption{Selected by RF-EI}
		%\label{fig2:c}
	\end{subfigure}
	\hfill	
	\begin{subfigure}[b]{0.215\textwidth}
		\centerline{\includegraphics[width=\columnwidth]{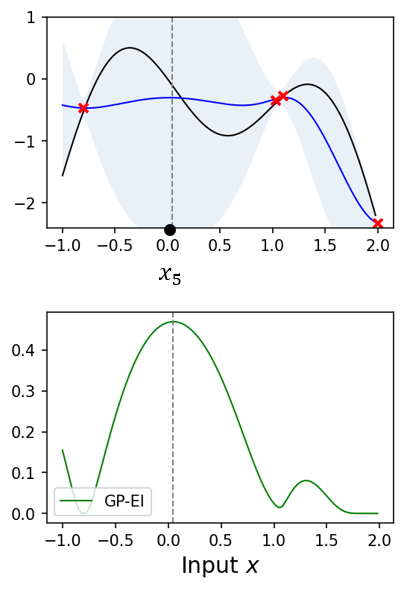}}
		\caption{Selected by GP-EI}
		%\label{fig2:d}
	\end{subfigure}		
	\caption
	{Catastrophic behavior of a single BO and diversification as a remedy. An optimization problem with one dimensional search space is shown in (a) where four candidates have been evaluated so far. In (b), GP-UCB selects the next candidate $x_5$ very close to the previously tried candidates $x_3$ and $x_4$, and subsequently becomes stuck near them. In (c), the situation is better but RF-EI also selects a point near the local optimum $l_1$. In (d), GP-EI has a different inductive bias and selects the next candidate near the global optimum $g$. 
	In general, it is impossible to tell which model will avoid a catastrophic behavior, but ensemble of multiple acquisition functions~\cite{cown2020emprical} or rotating over multiple BOs~\cite{cho2020basic} is known to be robust.
	}
	%\label{fig2}
	%\vspace{-2mm}
\end{figure*}

\begin{figure*}[h]
	\begin{subfigure}[b]{0.225\textwidth}
		\includegraphics[width=\columnwidth]{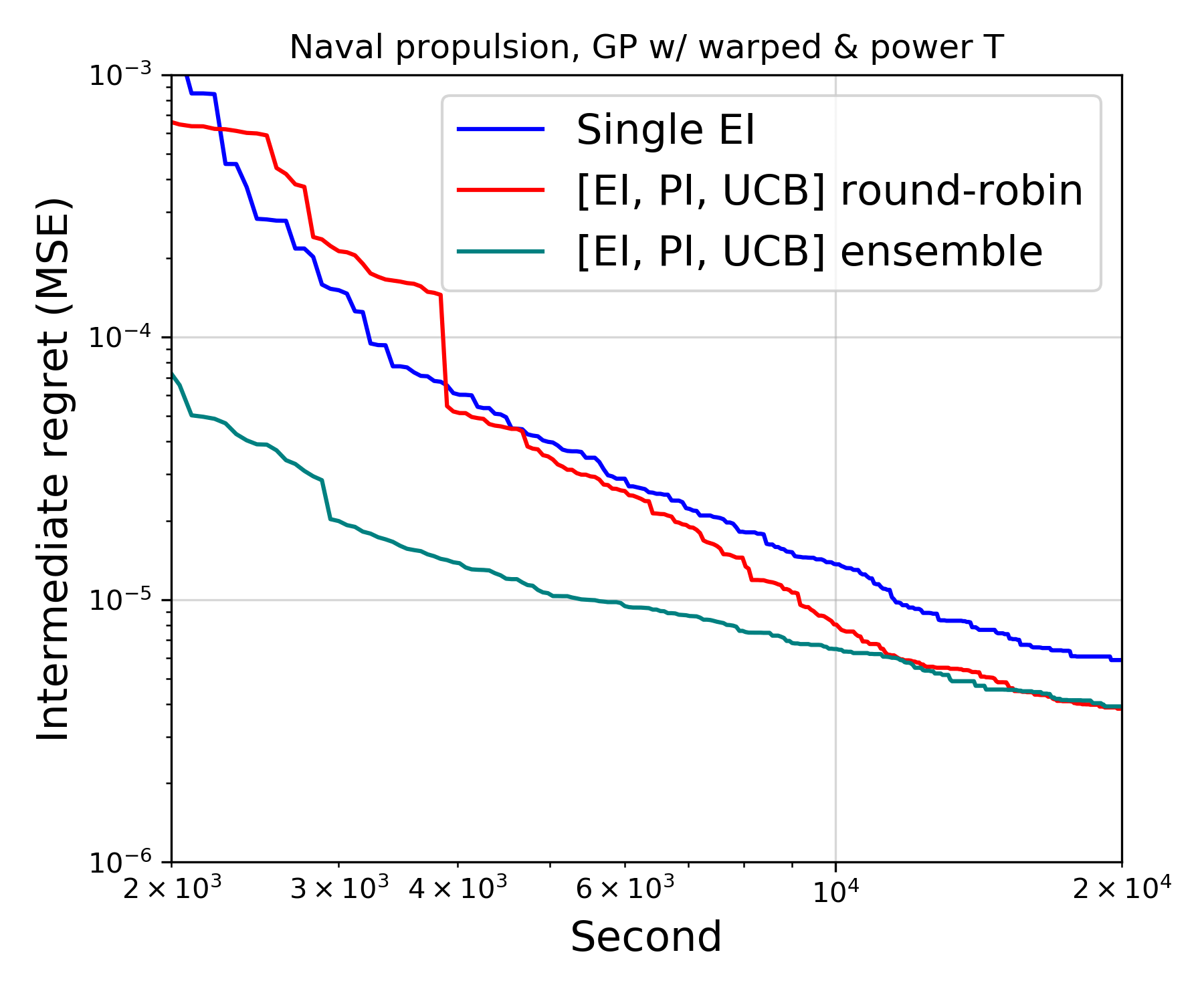}
		\caption{HPO-naval}		
	\end{subfigure}
	\hfill			
	\begin{subfigure}[b]{0.22\textwidth}
		\includegraphics[width=\columnwidth]{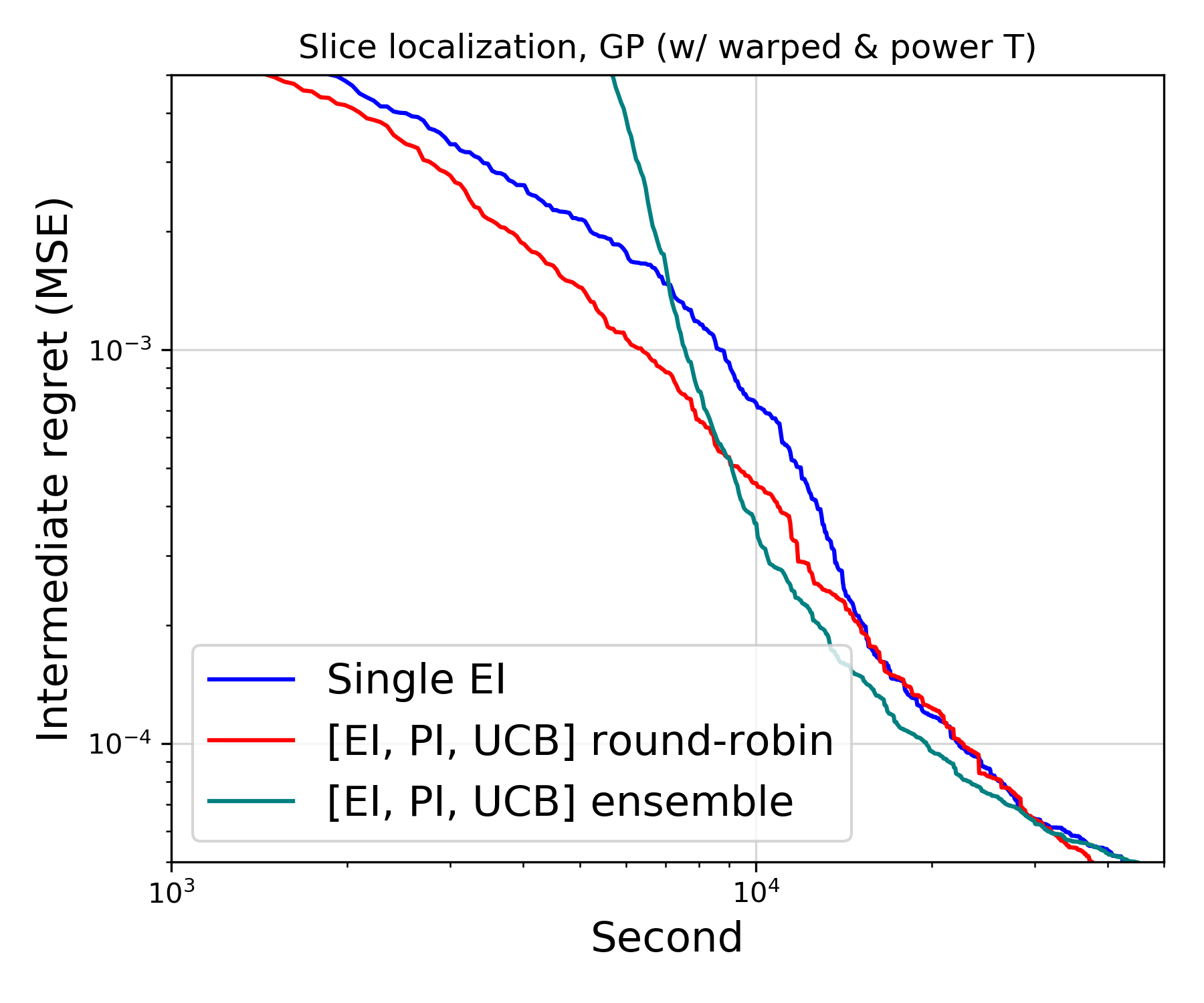}
		\caption{HPO-slice}
	\end{subfigure}
	\hfill
	\begin{subfigure}[b]{0.22\textwidth}
		\centerline{\includegraphics[width=\columnwidth]{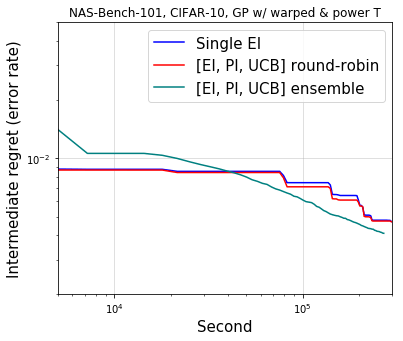}}
		\caption{NAS101-CIFAR10}
	\end{subfigure}
	\hfill	
	\begin{subfigure}[b]{0.215\textwidth}
		\centerline{\includegraphics[width=\columnwidth]{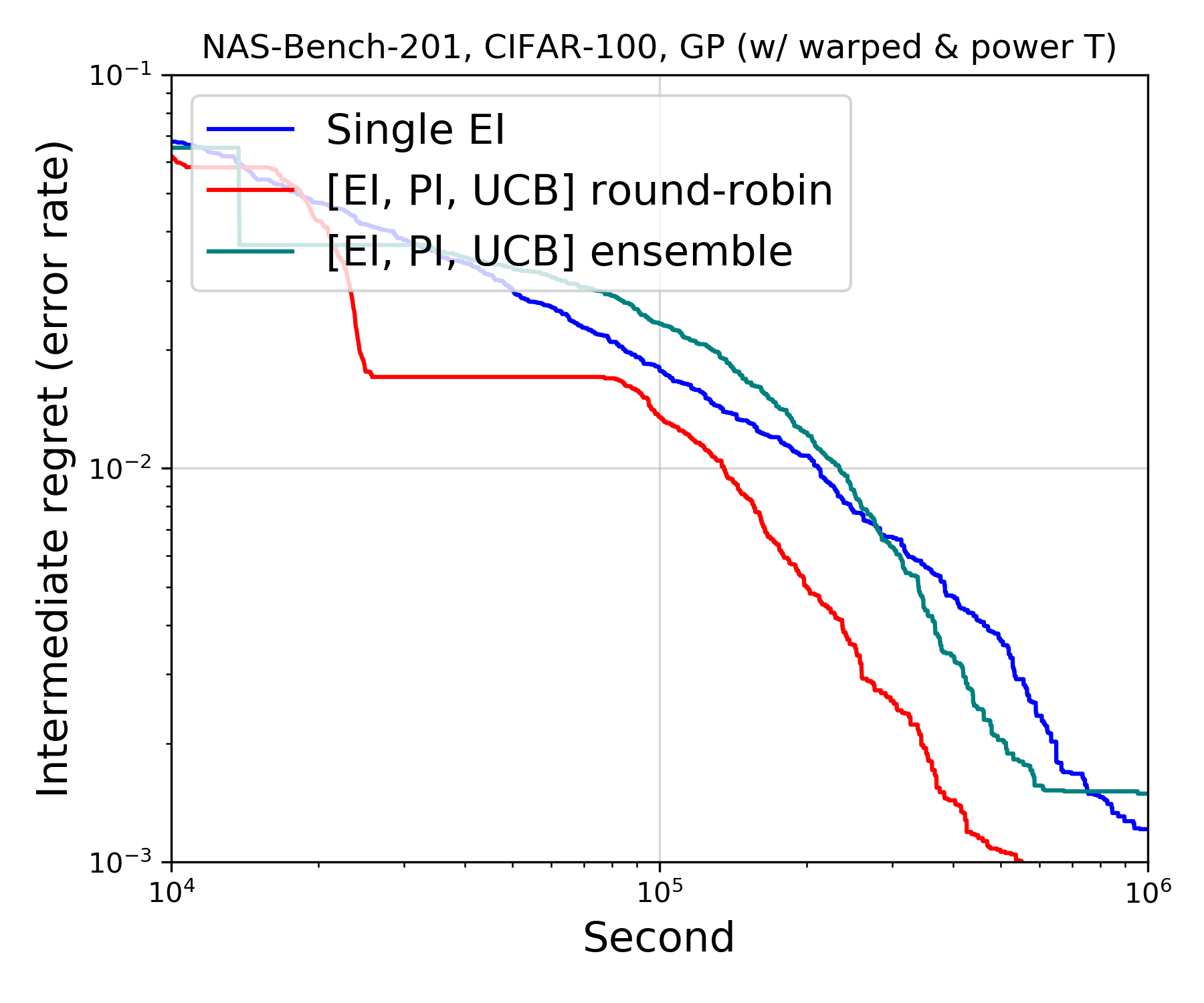}}
		\caption{NAS201-CIFAR100}
	\end{subfigure}		
	\caption
	{Comparison of mean intermediate regret performance $r_t$ when a different acquisition strategy is used in a typical BO. Utilizing multiple acquisition functions can be performed in many different ways. Here, the simple round-robin strategy is based on~\citet{cho2020basic} and the ensemble strategy is based on~\citet{cown2020emprical}. We also compare them with EI acquisition only~\cite{mockus1978toward}, that is commonly used.
	}
	%\label{fig2}
	%\vspace{-2mm}
\end{figure*}

\begin{figure*}[h]
	\begin{subfigure}[b]{0.225\textwidth}
		\includegraphics[width=\columnwidth]{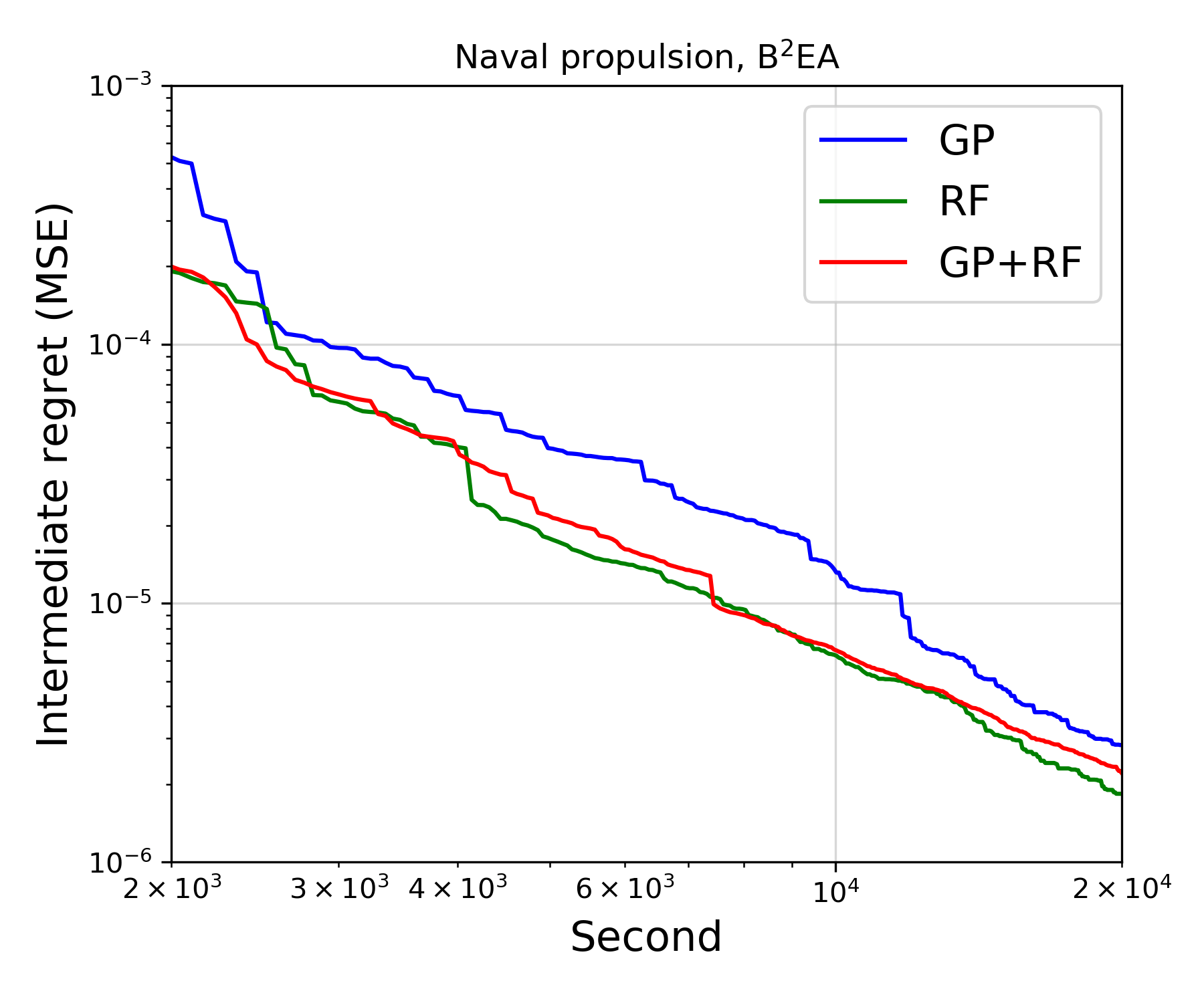}
		\caption{HPO-naval}		
	\end{subfigure}
	\hfill			
	\begin{subfigure}[b]{0.22\textwidth}
		\includegraphics[width=\columnwidth]{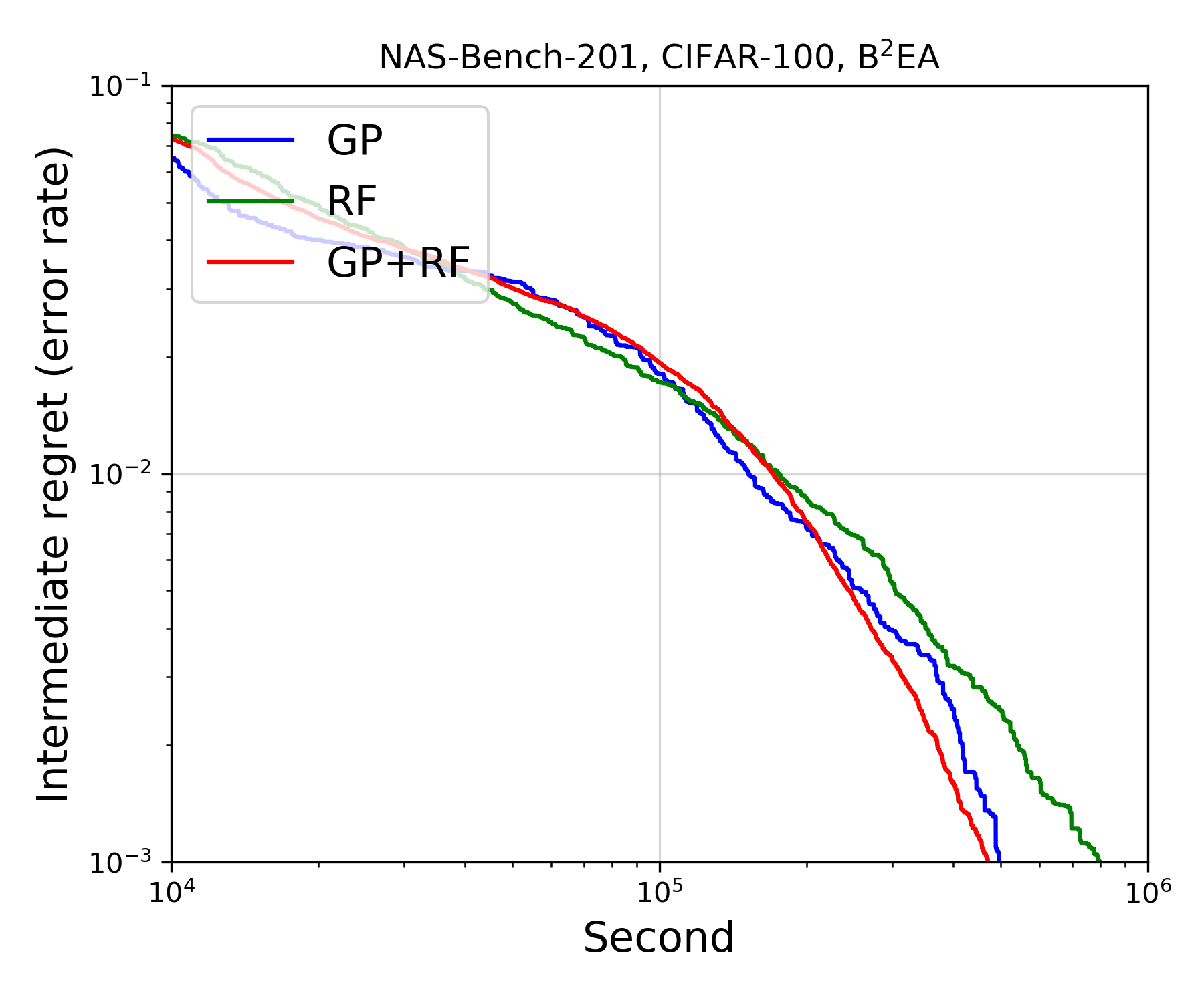}
		\caption{NAS201-CIFAR100}
	\end{subfigure}
	\hfill
	\begin{subfigure}[b]{0.22\textwidth}
		\centerline{\includegraphics[width=\columnwidth]{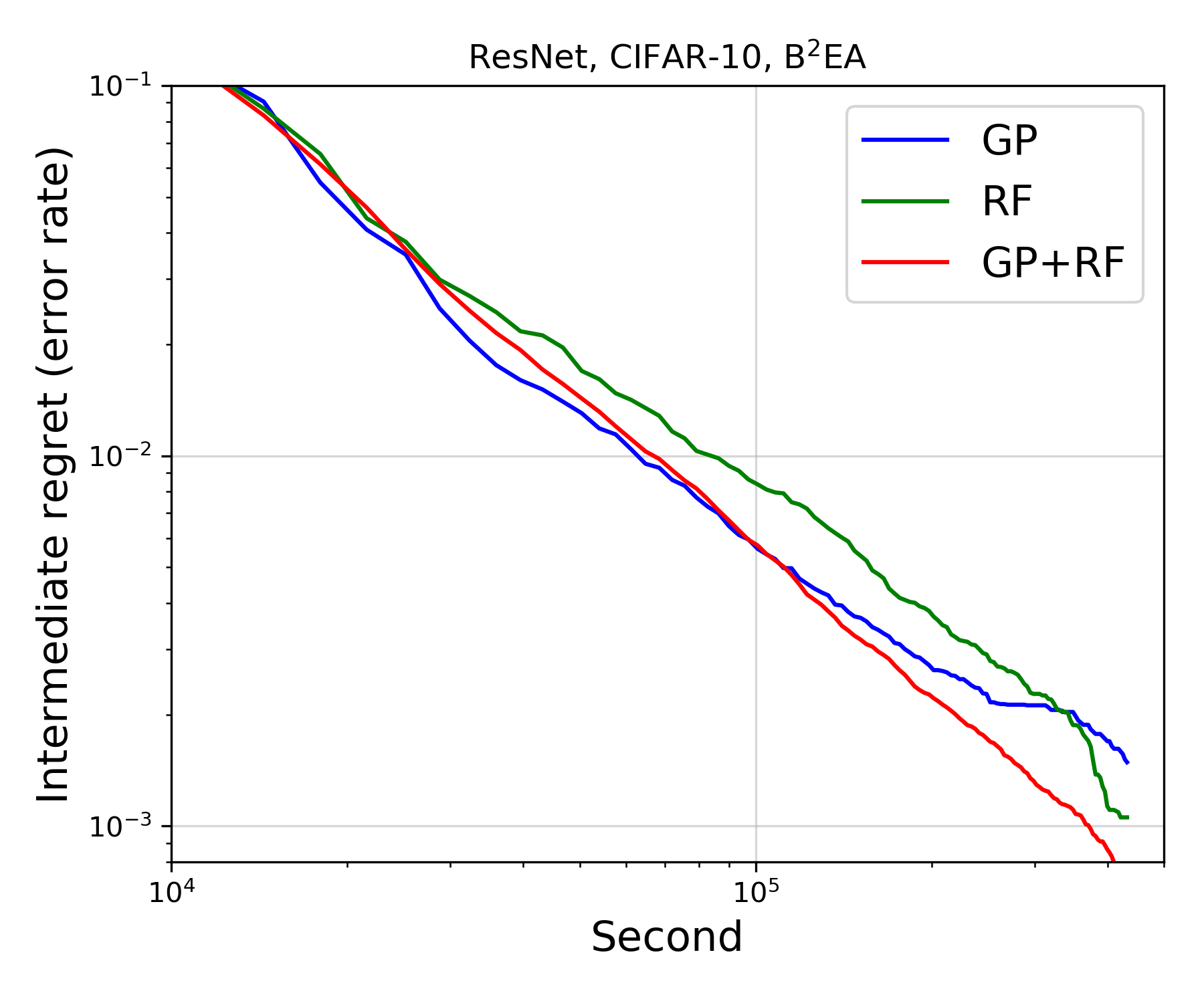}}
		\caption{CIFAR10-ResNet}
	\end{subfigure}
	\hfill	
	\begin{subfigure}[b]{0.215\textwidth}
		\centerline{\includegraphics[width=\columnwidth]{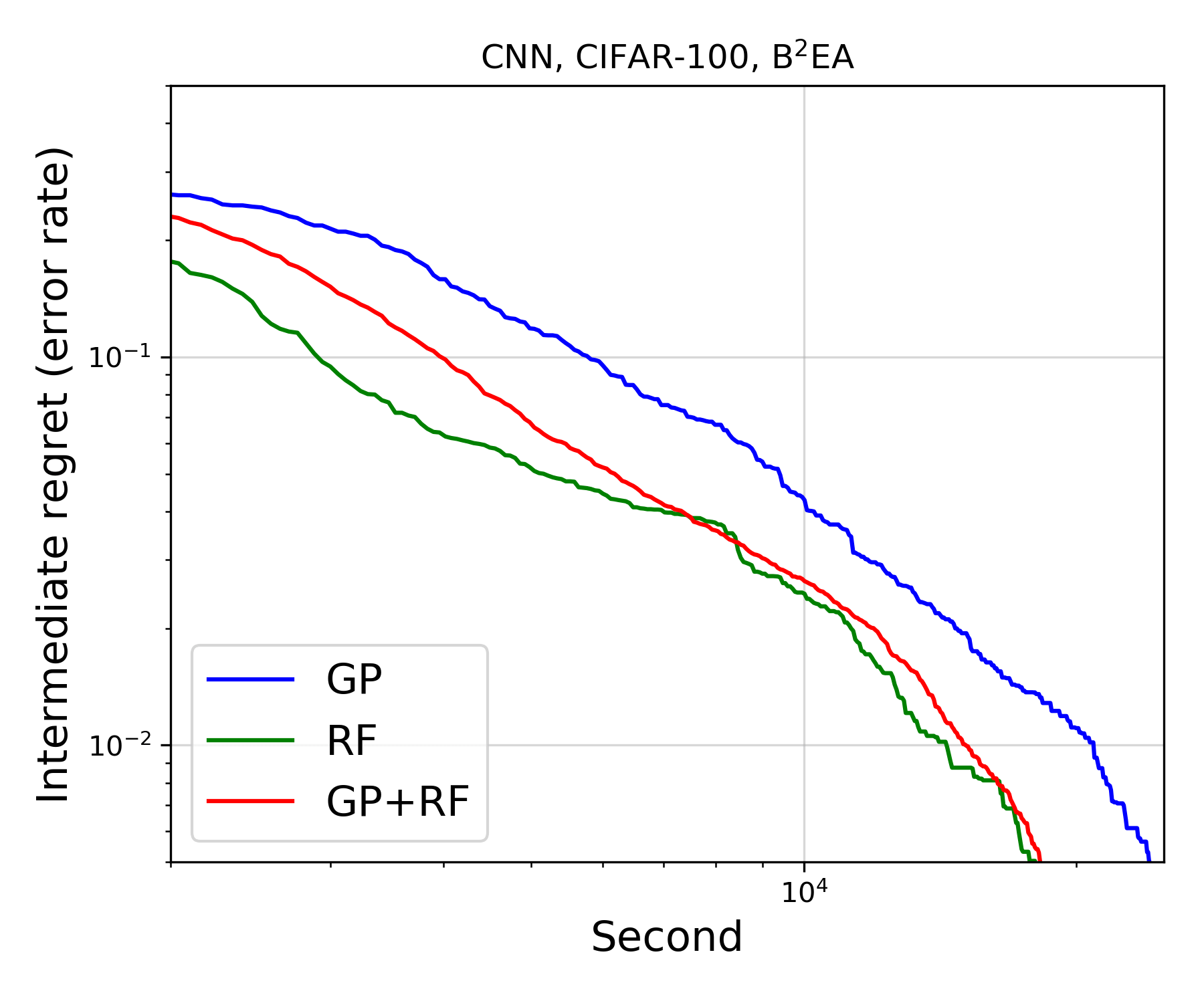}}
		\caption{CIFAR100-CNN}
	\end{subfigure}		
	\caption
	{Comparison of mean intermediate regret performance $r_t$ for three different choices of surrogate model in B\textsuperscript{2}EA (GP only, RF only, both GP and RF). We plotted the mean $r_t$ values of 100 repeated runs. 
	Traditionally, GP has been known to be effective for optimizing continuous variables and RF has been known to be effective for optimizing categorical variables. However, our experiment results over NAS tasks do not agree with the known wisdom. Therefore, we have chosen to use both GP and RF as the surrogate models in B\textsuperscript{2}EA.
	 }
	%\label{fig2}
	%\vspace{-2mm}
\end{figure*}

\newpage

\clearpage

\newpage

%\subsection{Intermediate regret}

%
%\begin{figure*}[h]
%	\begin{subfigure}[b]{\textwidth}
%		%\vspace*{5ex}
%		\includegraphics[width=.23\textwidth]{figures/f2a1.png}
%		\hfill
%		\includegraphics[width=.23\textwidth]{figures/f2a2.png}
%		\hfill
%		\includegraphics[width=.23\textwidth]{figures/f2a3.png}
%		\hfill
%		\includegraphics[width=.23\textwidth]{figures/f2a4.png}
%		\hfill
%		\caption{HPO-Bench tasks.}
%		
%	\end{subfigure}
%	%
%	\\[3ex]
%	%	
%	\begin{subfigure}[b]{\textwidth}		
%		\includegraphics[width=.23\textwidth]{figures/f2b0.png}
%		\hfill
%		\includegraphics[width=.23\textwidth]{figures/f2b1.png}
%		\hfill
%		\includegraphics[width=.23\textwidth]{figures/f2b2.png}
%		\hfill
%		\includegraphics[width=.23\textwidth]{figures/f2b3.png}
%		\hfill
%		\caption{NAS-Bench tasks.}
%		
%	\end{subfigure}
%	%
%	\\[3ex]
%	%		
%	\begin{subfigure}[b]{\textwidth}		
%		\includegraphics[width=.23\textwidth]{figures/f2c1.png}
%		\hfill
%		\includegraphics[width=.23\textwidth]{figures/f2c2.png}
%		\hfill
%		\includegraphics[width=.23\textwidth]{figures/f2c3.png}
%		\hfill
%		
%		\vspace*{3ex}
%		
%		\includegraphics[width=.23\textwidth]{figures/f2c4.png}
%		\hfill
%		\includegraphics[width=.23\textwidth]{figures/f2c5.png}
%		\hfill
%		\includegraphics[width=.23\textwidth]{figures/f2c6.png}
%		\hfill
%		\caption{DNN-Bench tasks.}
%		%\label{fig2a-a:c}	
%	\end{subfigure}
%	%
%	\\[3ex]
%	%
%	\caption*{Figure A2: Comparison of intermediate regret $r_t$ performance, but plotted for \textit{mean} $r_t$ value of 500 repeated runs. }
%	\label{fig2a-a}
%\end{figure*} 

\section{Success Rate $\mathbb{P}(\tau \le t)$ Performance}

\begin{figure*}[!ht]
	
	\begin{subfigure}[b]{\textwidth}
		\vspace*{5ex}
		\includegraphics[width=.23\textwidth]{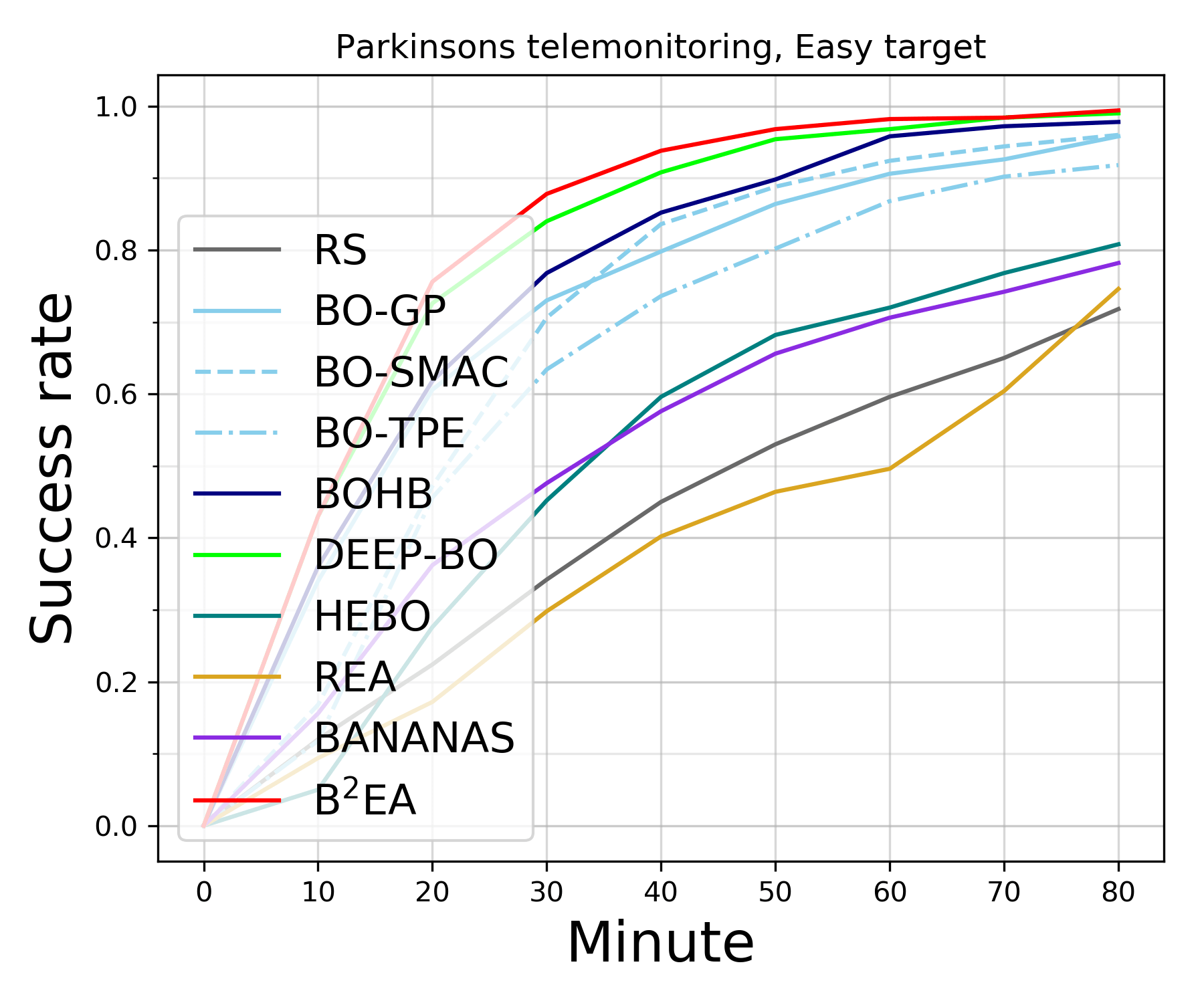}
		\hfill
		\includegraphics[width=.23\textwidth]{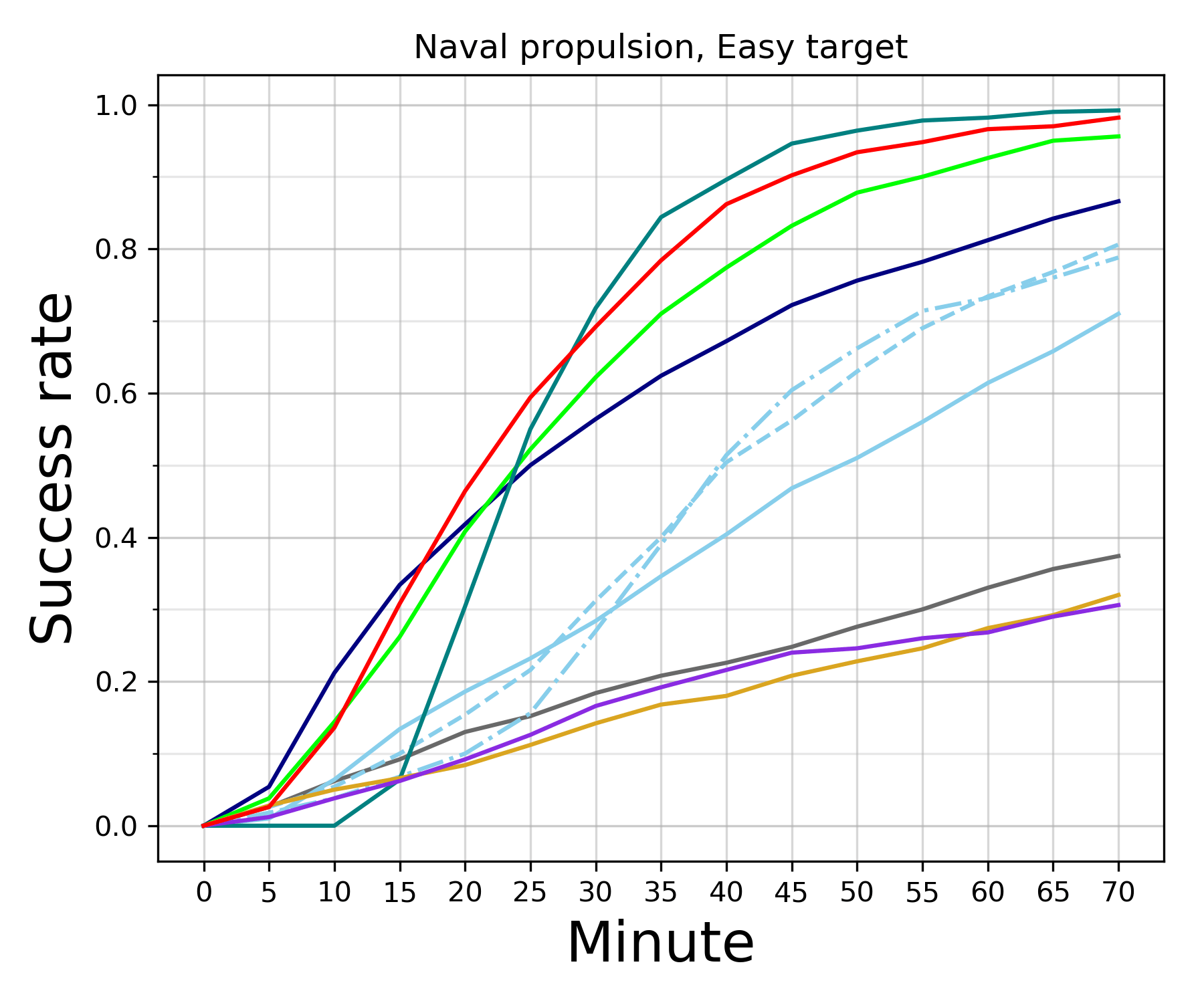}
		\hfill
		\includegraphics[width=.23\textwidth]{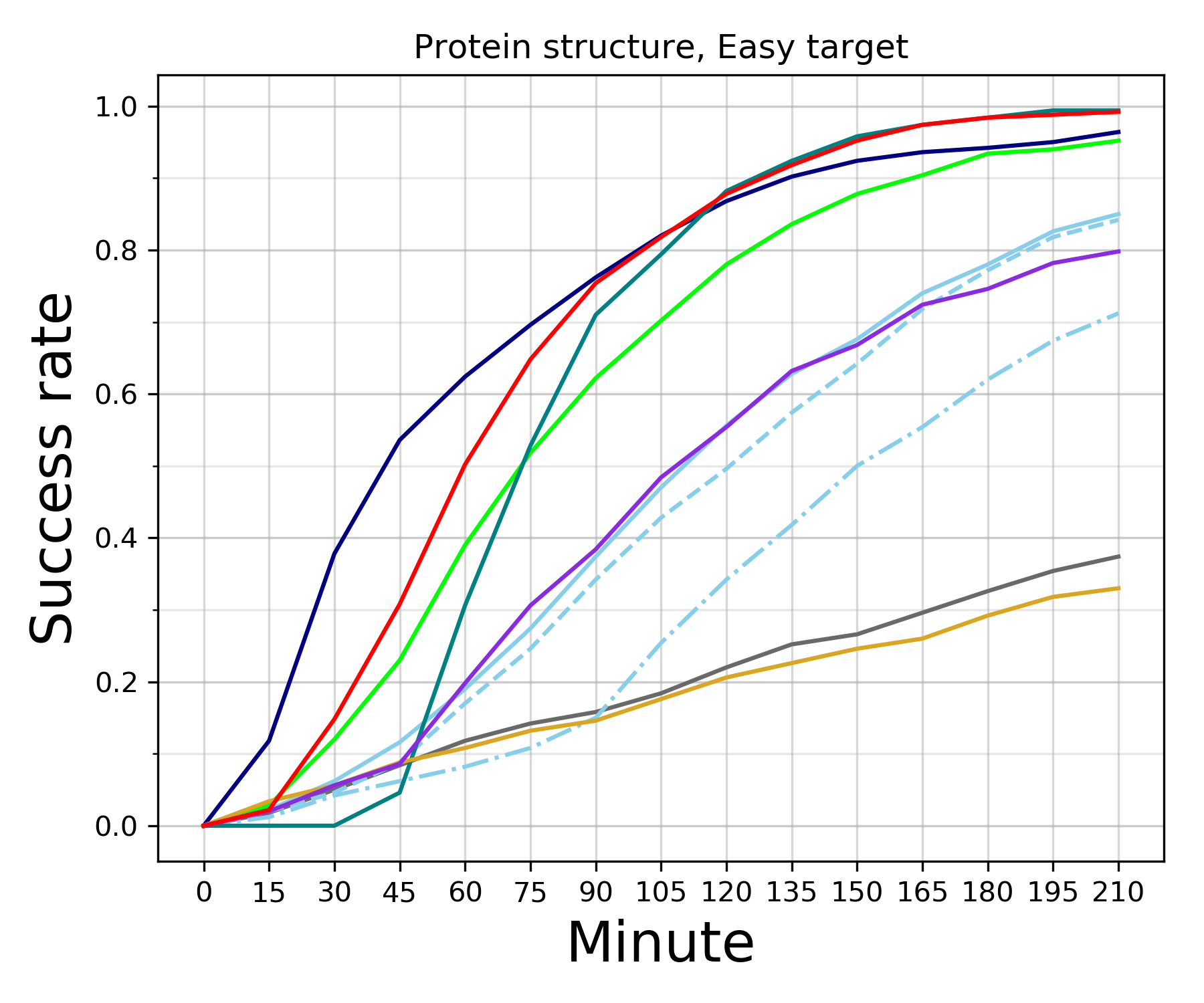}
		\hfill
		\includegraphics[width=.23\textwidth]{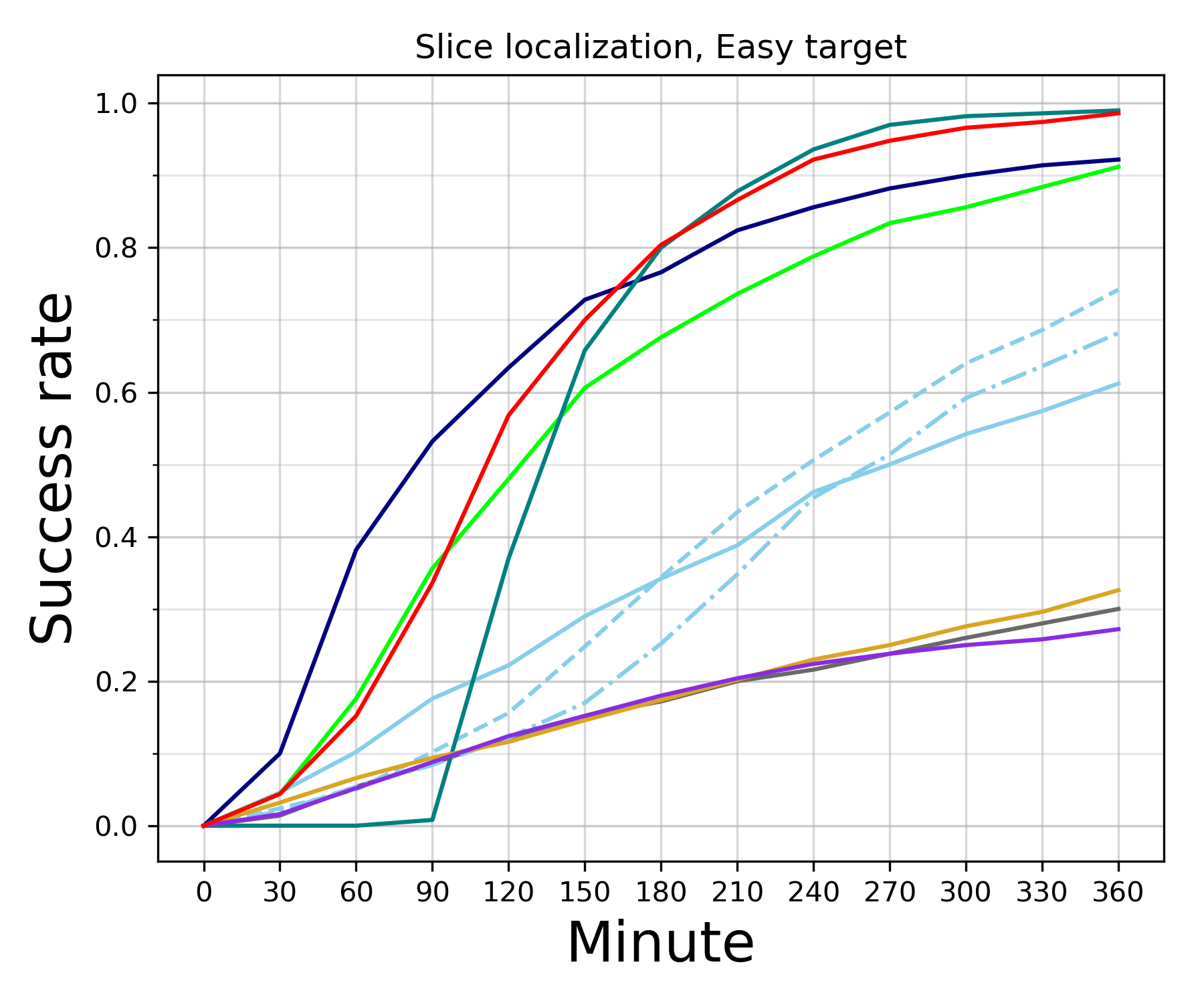}
		\hfill
		\caption{HPO-Bench tasks.}
		%\label{fig2c-a:a}
	\end{subfigure}
	\\[10ex]
	\begin{subfigure}[b]{\textwidth}		
		\includegraphics[width=.23\textwidth]{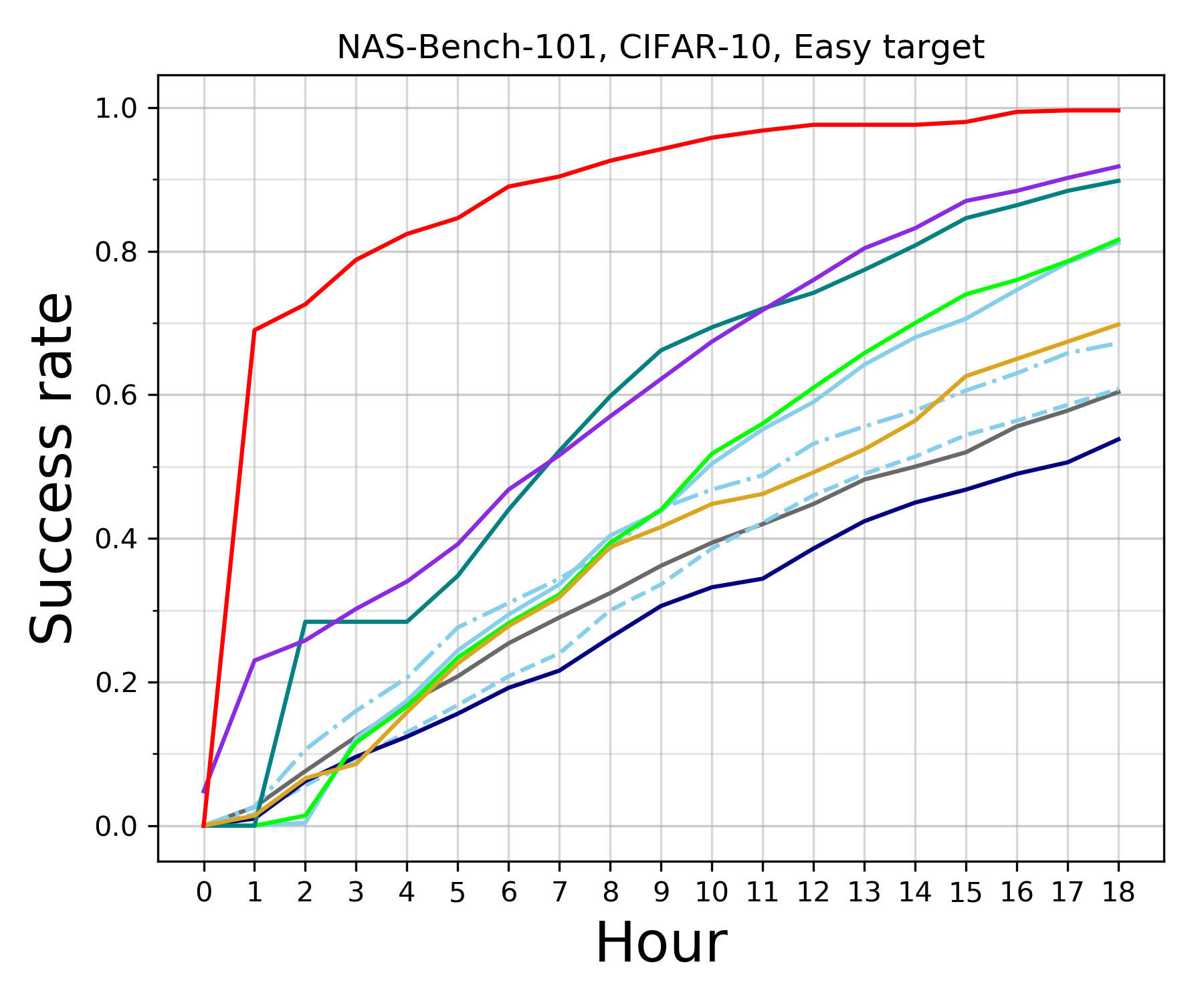}
		\hfill
		\includegraphics[width=.23\textwidth]{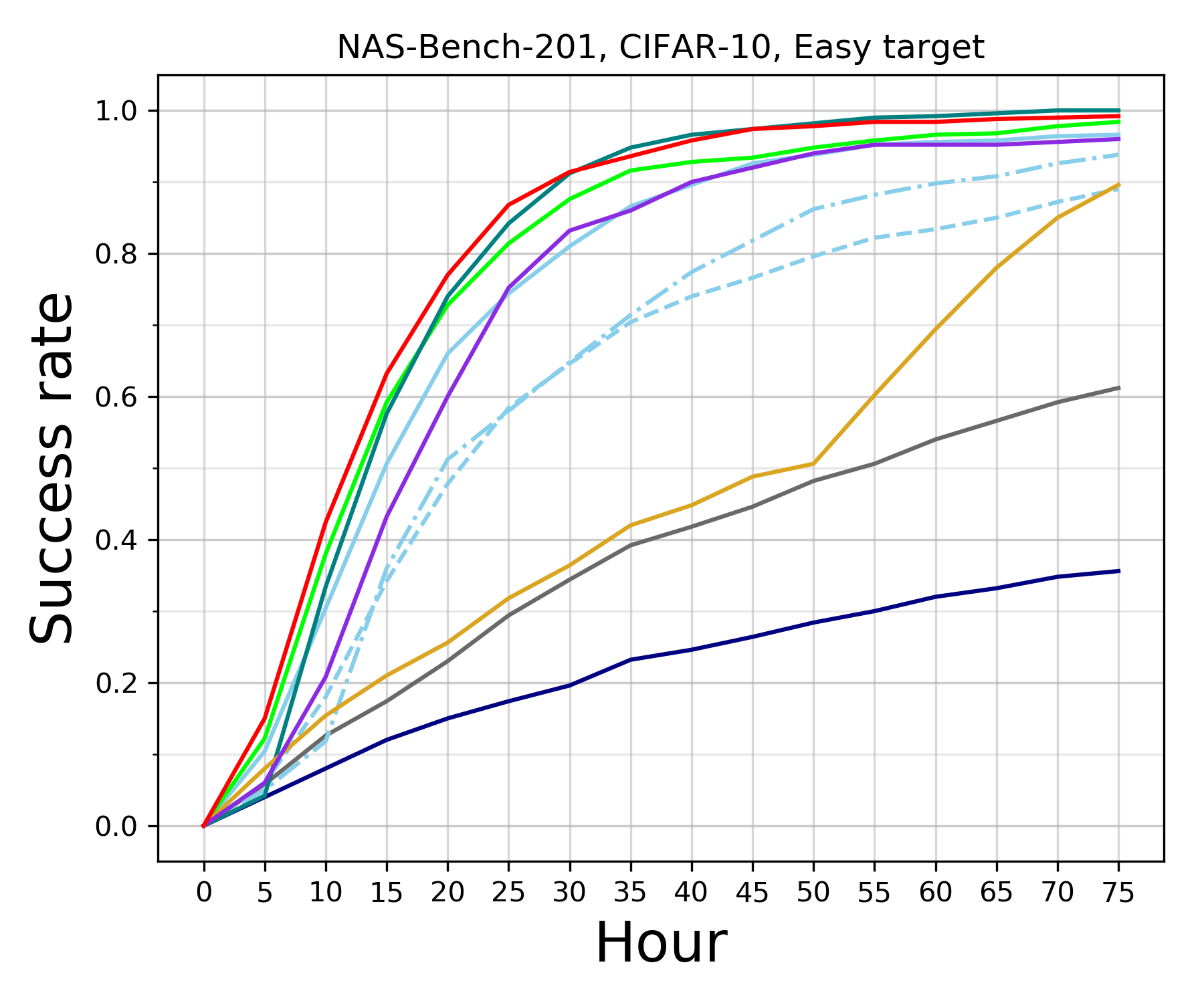}
		\hfill
		\includegraphics[width=.23\textwidth]{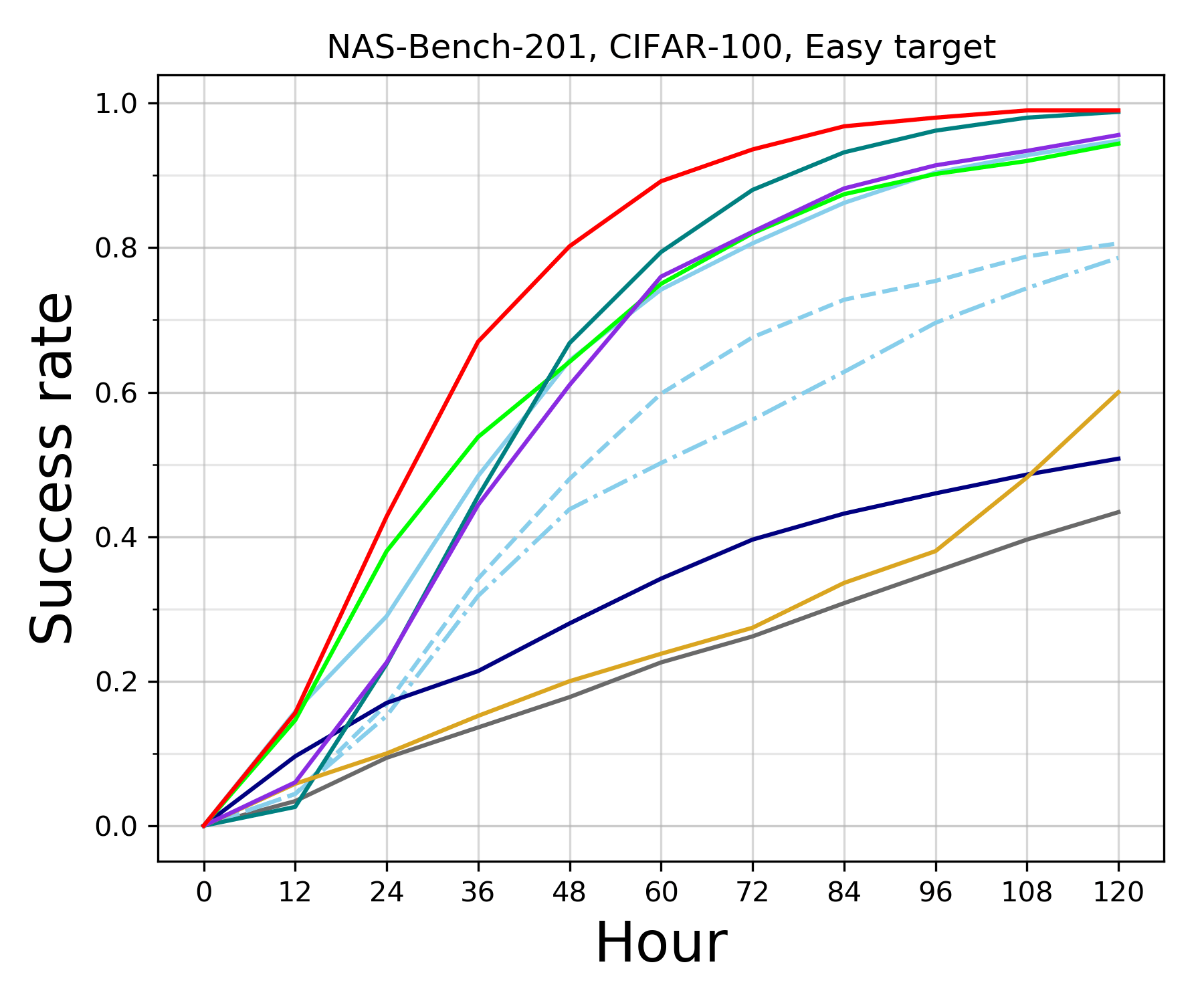}
		\hfill
		\includegraphics[width=.23\textwidth]{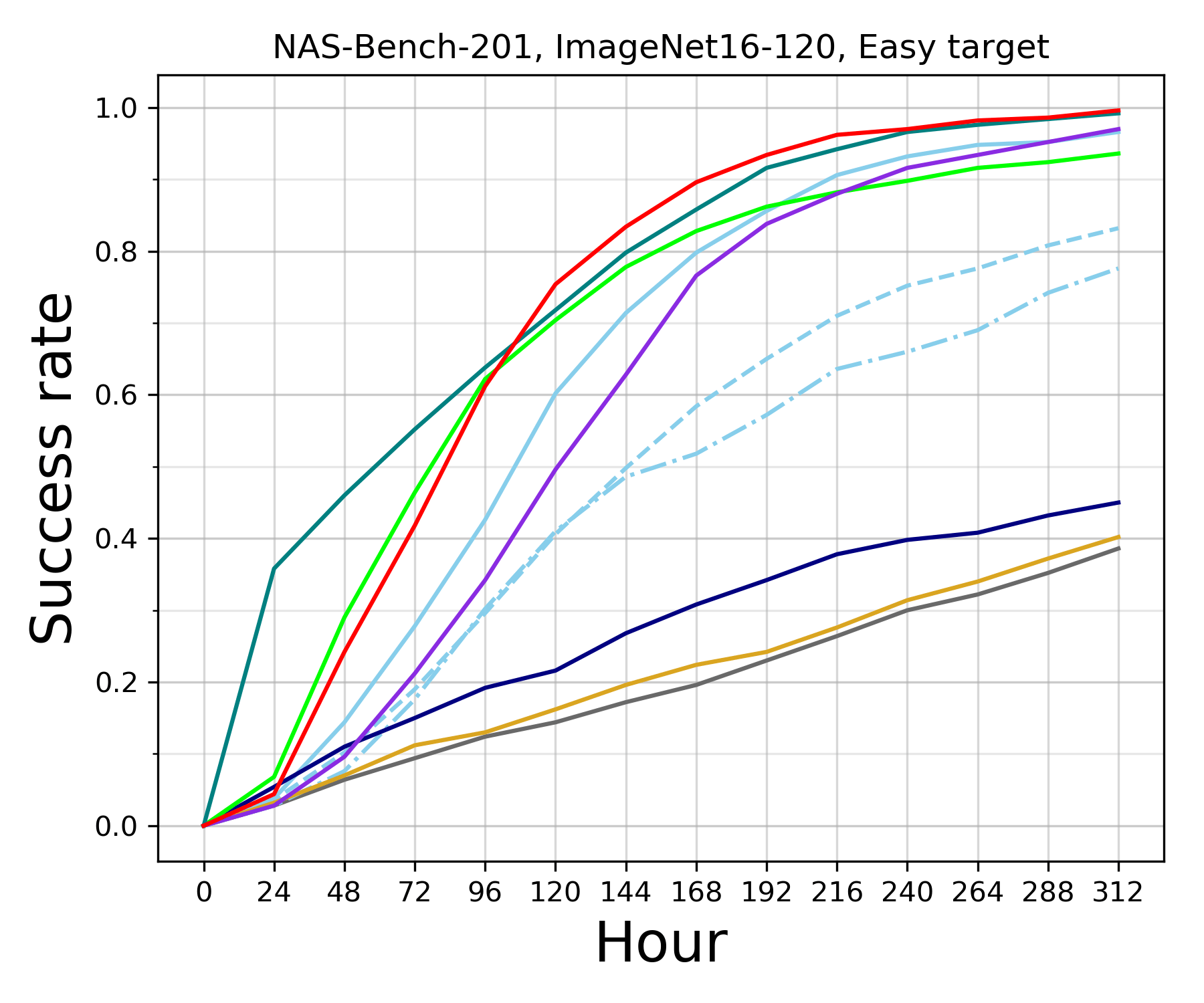}
		\hfill
		\caption{NAS-Bench tasks.}
		%\label{fig2c-a:b}	
	\end{subfigure}
	\\[10ex]
	\begin{subfigure}[b]{\textwidth}		
		\includegraphics[width=.23\textwidth]{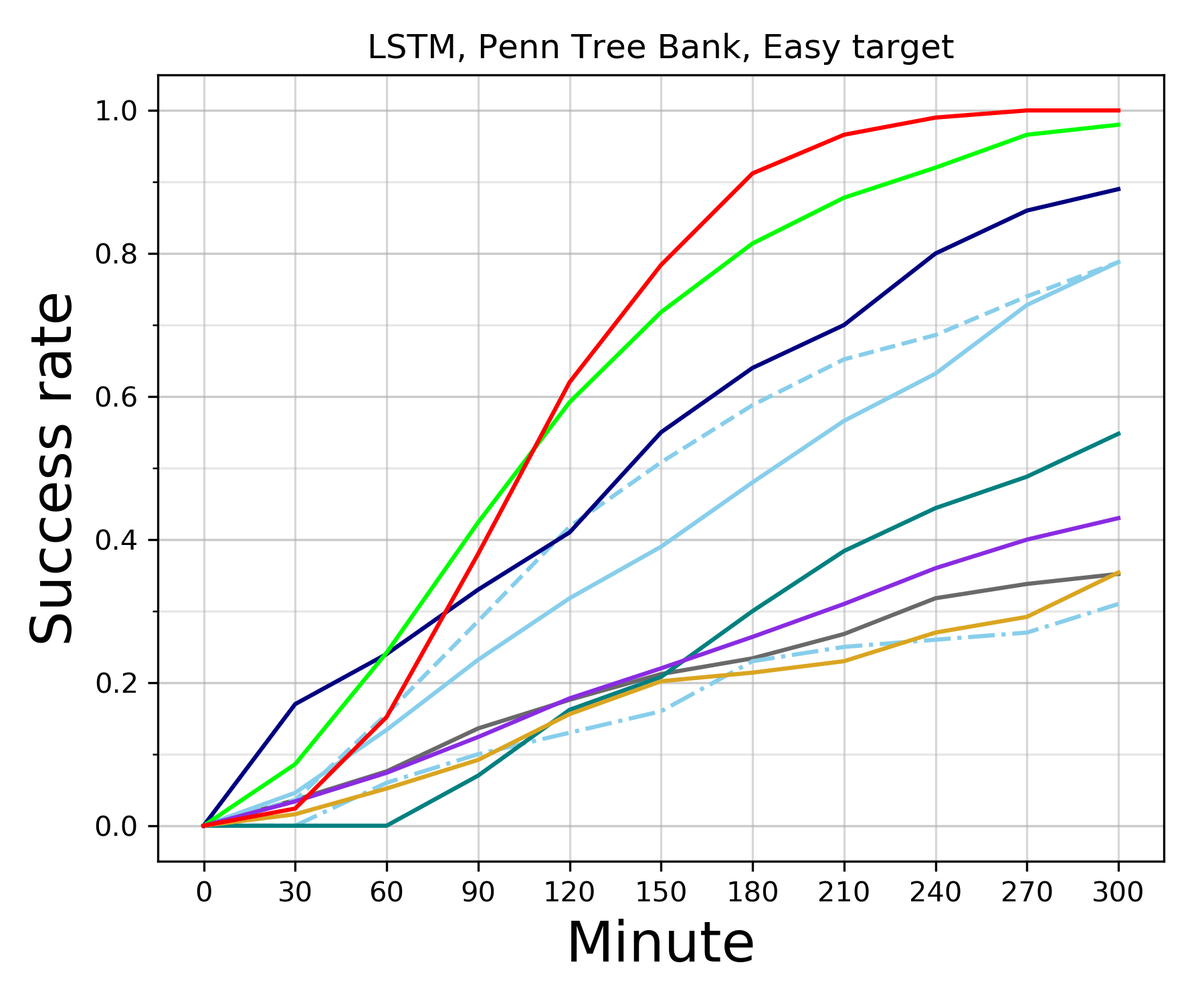}
		\hfill
		\includegraphics[width=.23\textwidth]{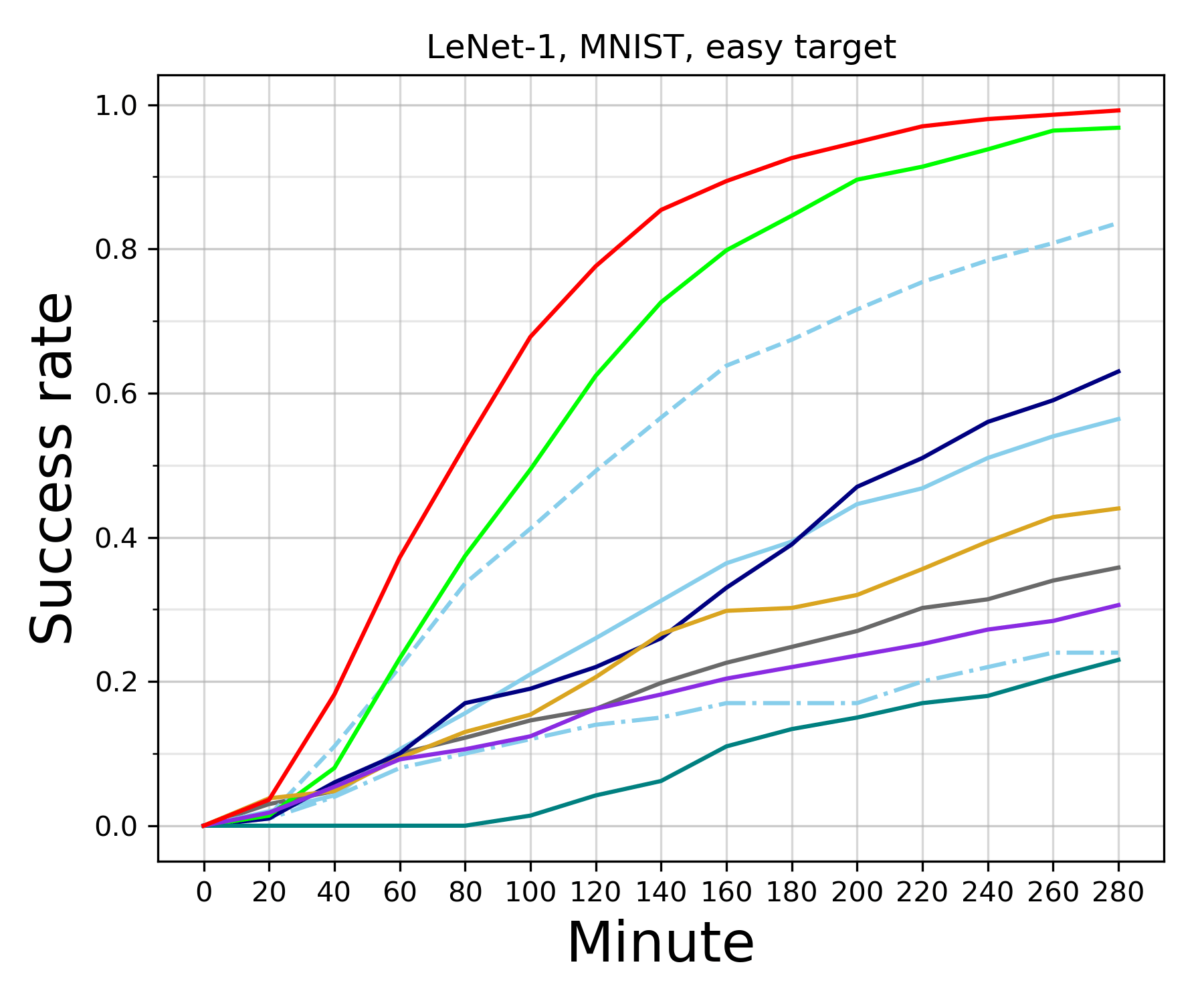}
		\hfill
		\includegraphics[width=.23\textwidth]{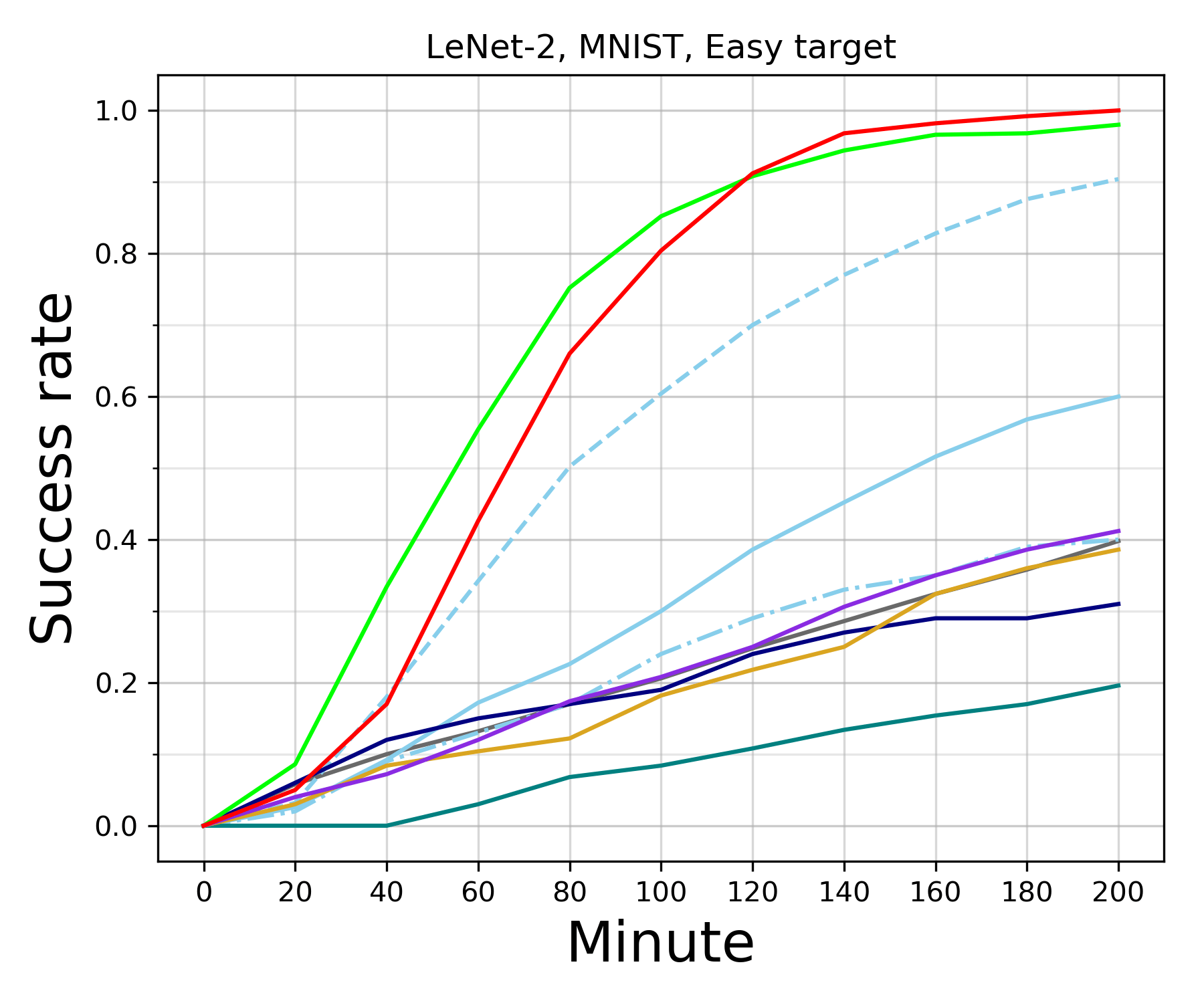}
		\hfill
		
		\vspace*{5ex}
		
		\includegraphics[width=.23\textwidth]{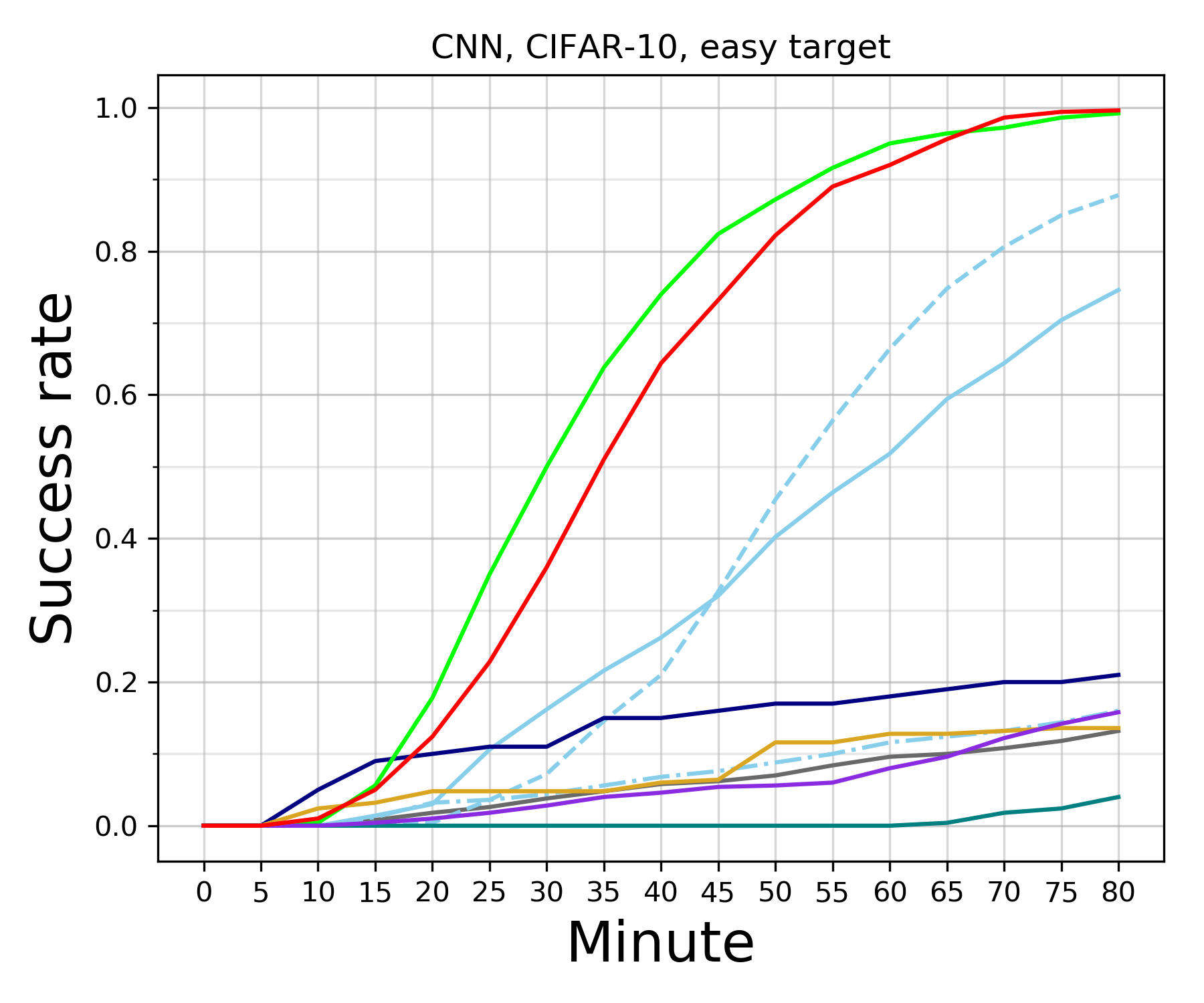}
		\hfill
		\includegraphics[width=.23\textwidth]{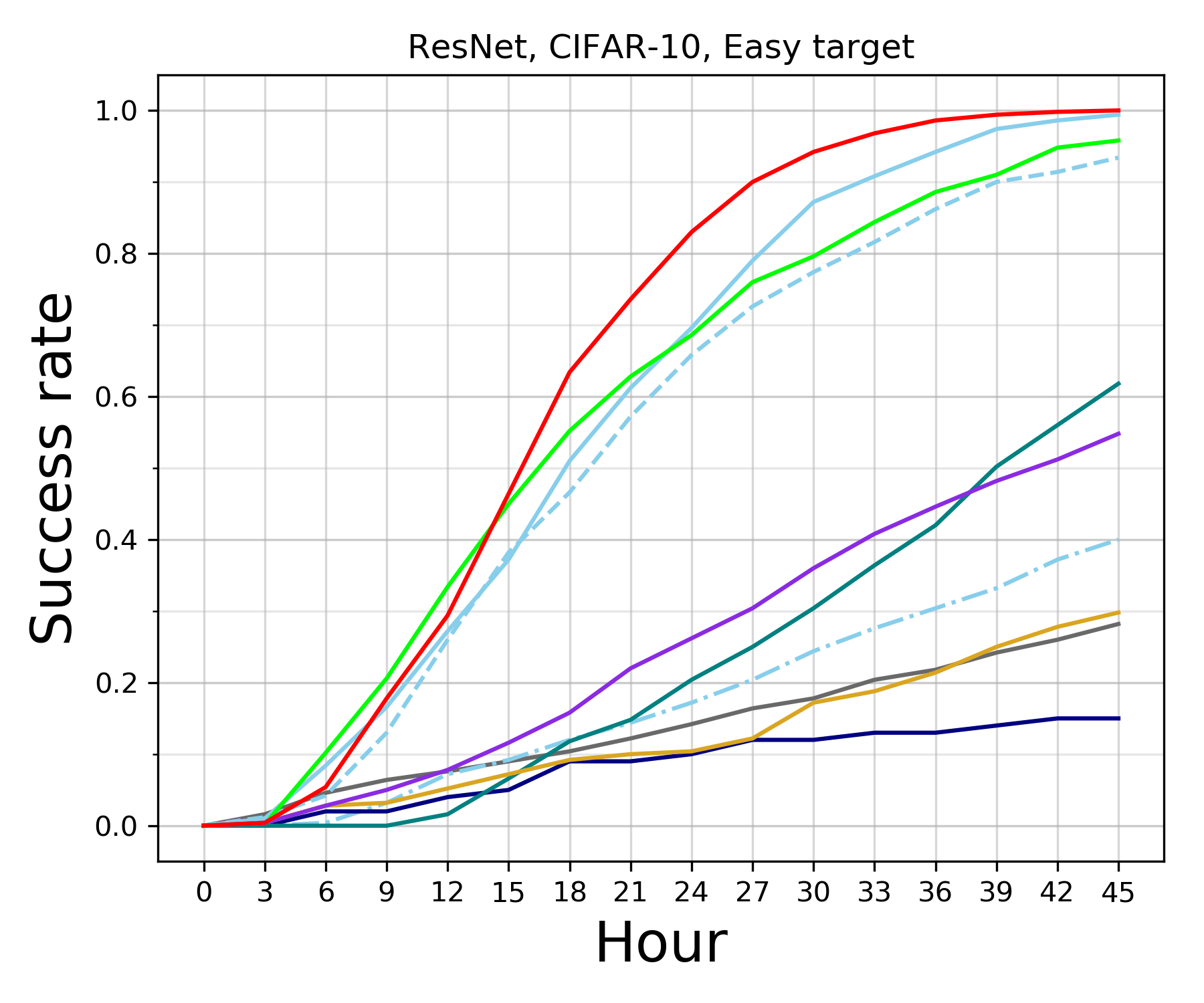}
		\hfill
		\includegraphics[width=.23\textwidth]{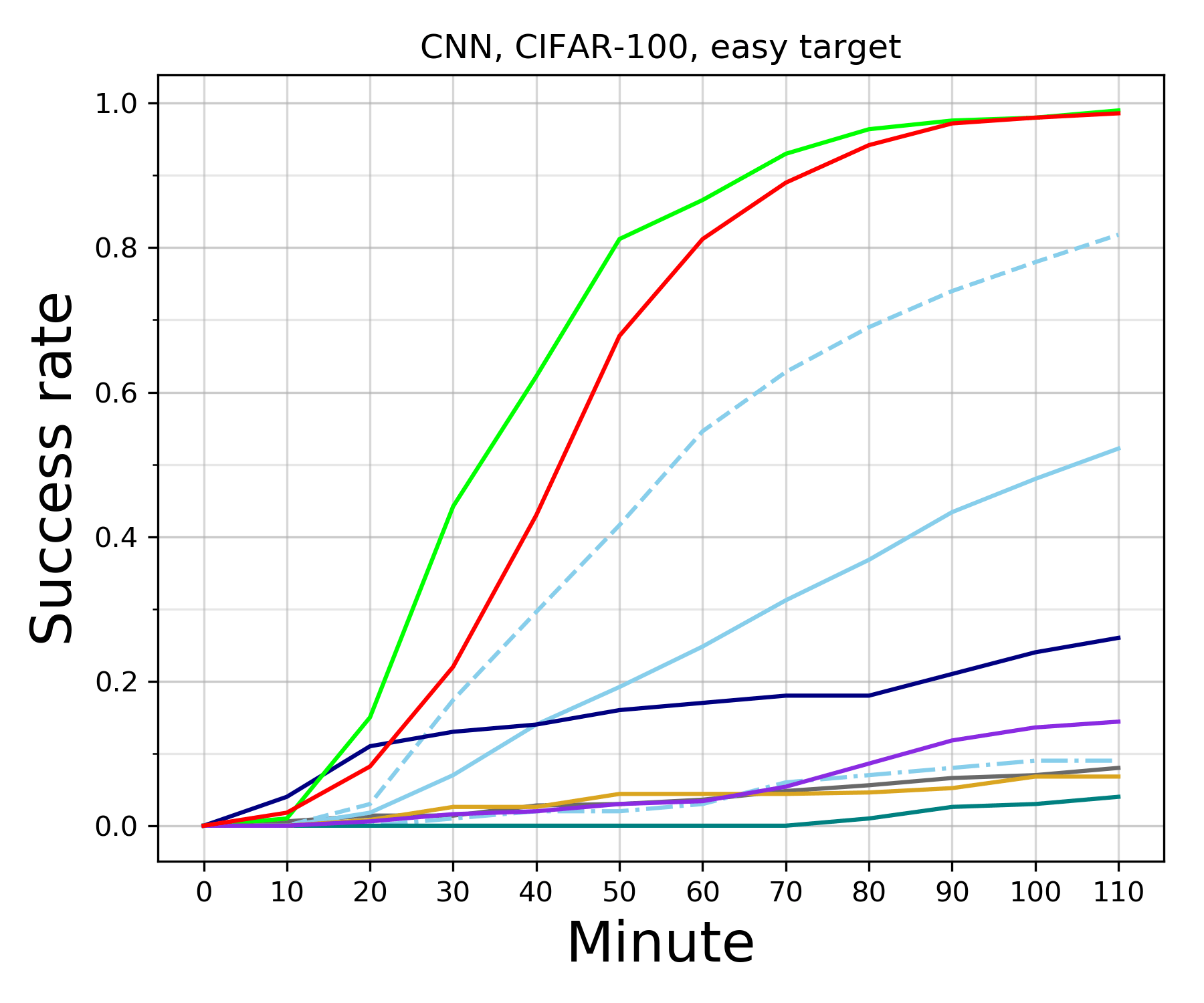}
		\hfill
		\caption{DNN-Bench tasks.}
		%\label{fig2c-a:c}	
	\end{subfigure}
	\\[5ex]
	\caption
	{Comparison of the success rate for the easy target $\mathbb{P}(\tau \le t_e)$, plotted with $t$ as the horizontal axis.}
	\label{fig2c-a}
\end{figure*}

\begin{figure*}[!ht]
	
	\begin{subfigure}[b]{\textwidth}
		\vspace*{5ex}
		\includegraphics[width=.23\textwidth]{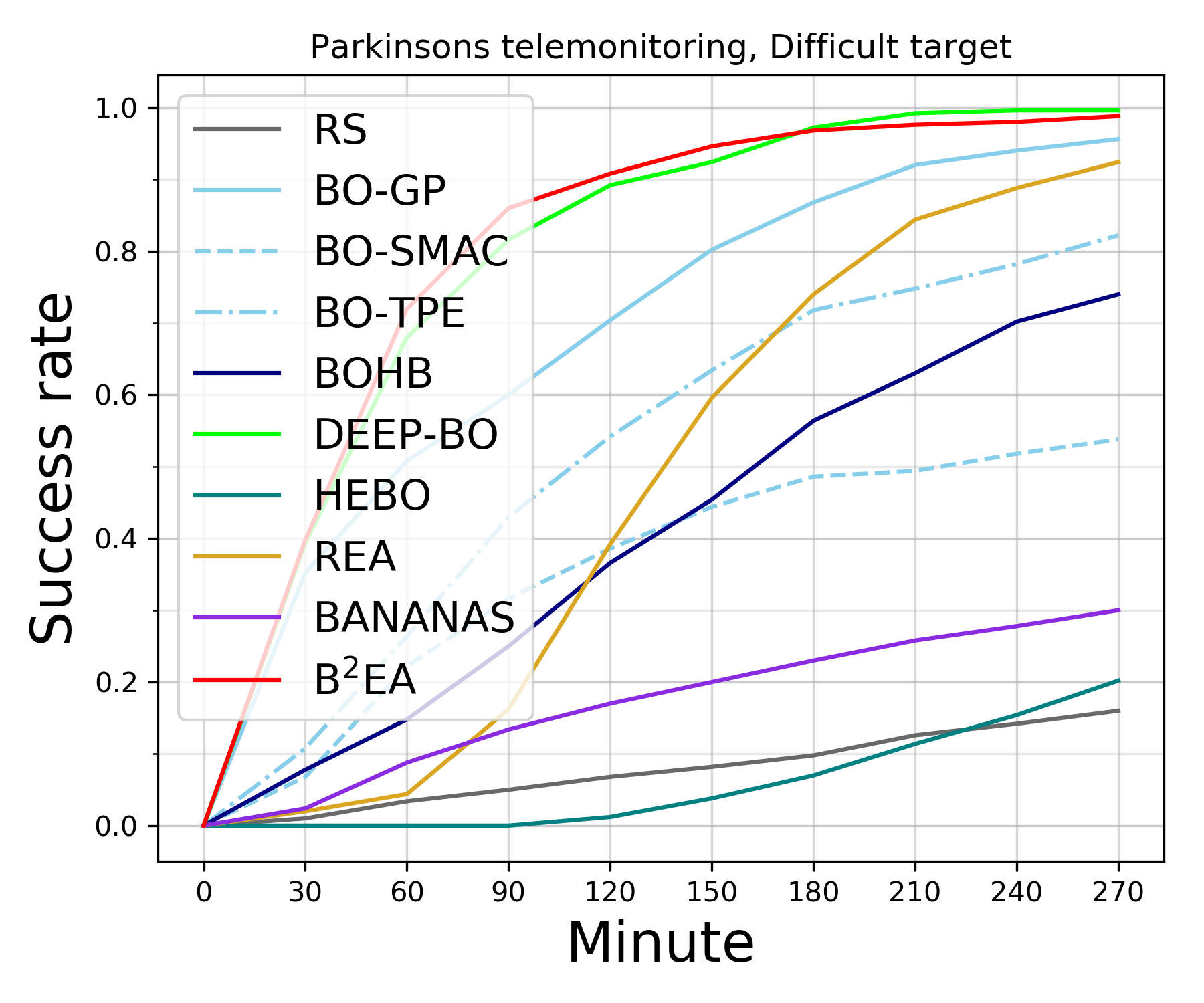}
		\hfill
		\includegraphics[width=.23\textwidth]{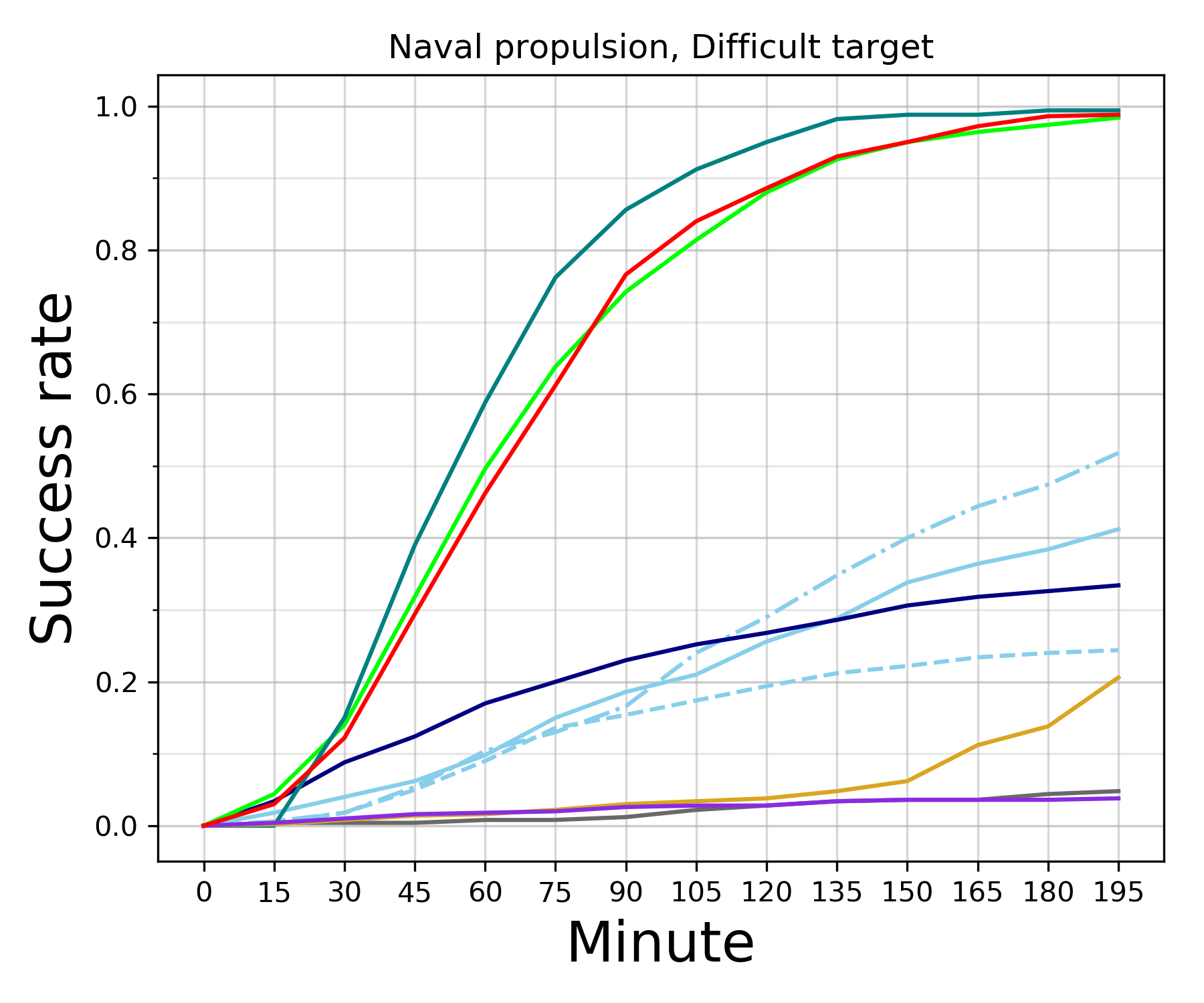}
		\hfill
		\includegraphics[width=.23\textwidth]{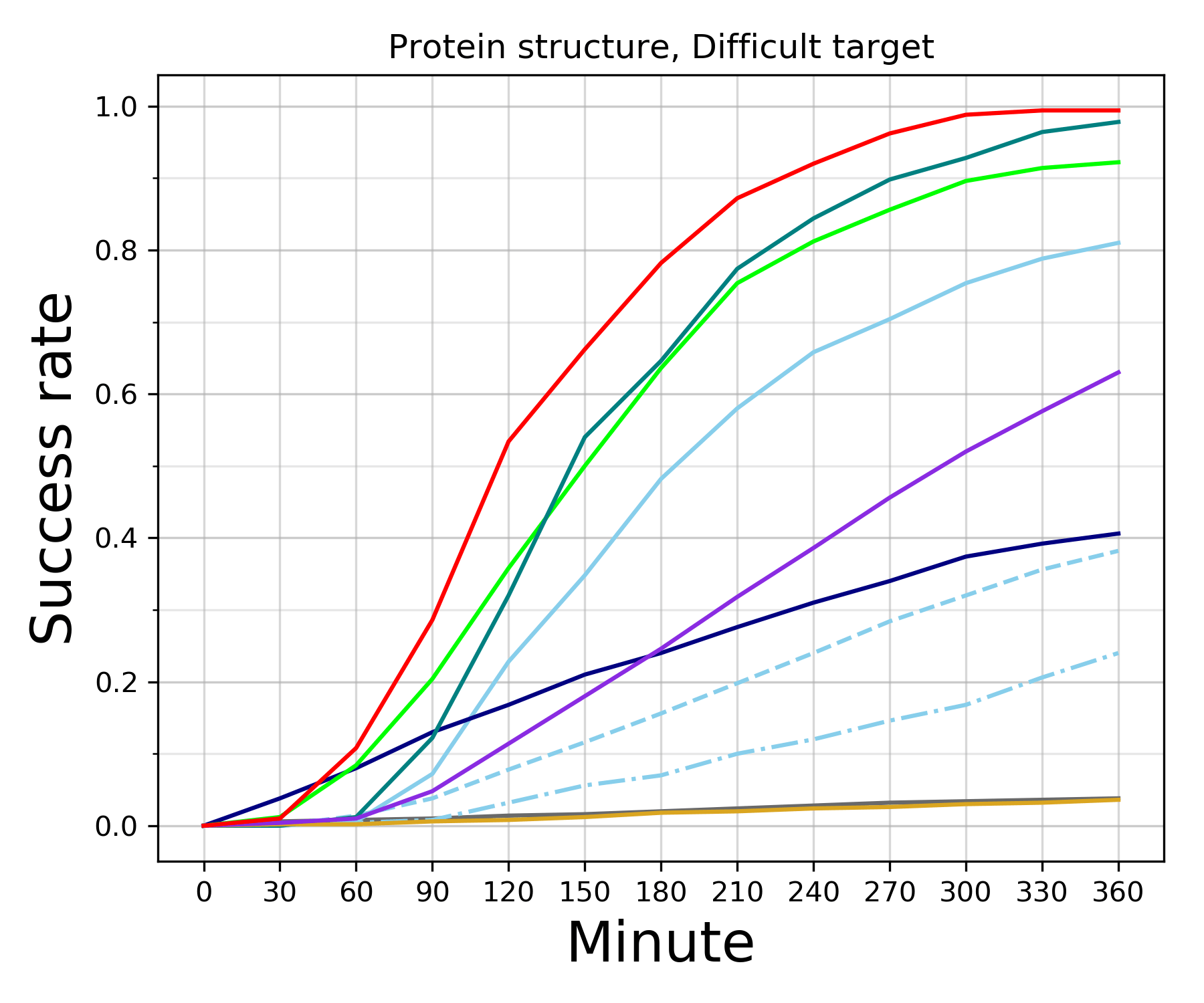}
		\hfill
		\includegraphics[width=.23\textwidth]{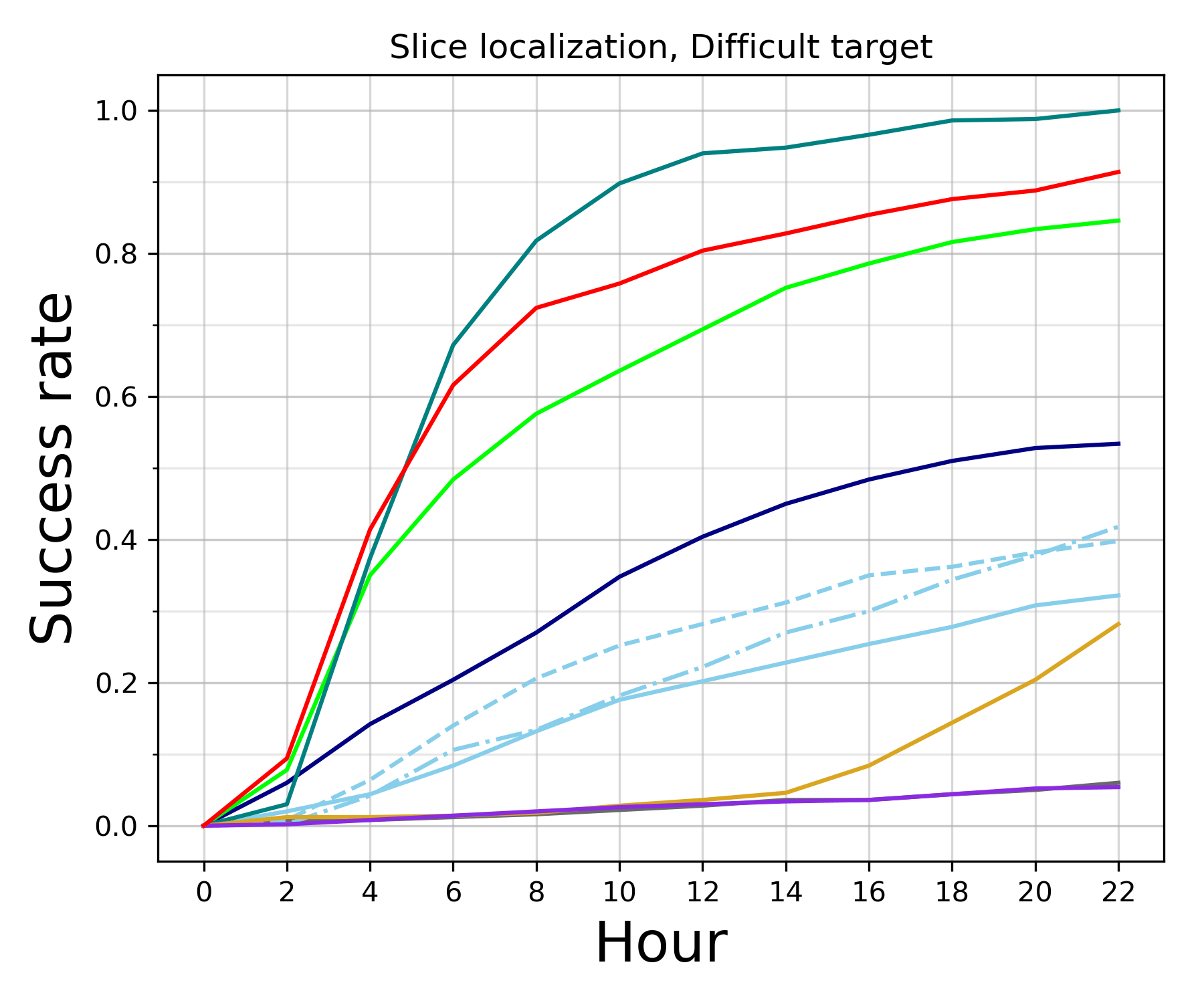}
		\hfill
		\caption{HPO-Bench tasks.}
		%\label{fig2d-a:a}
	\end{subfigure}
	\\[10ex]
	\begin{subfigure}[b]{\textwidth}		
		\includegraphics[width=.23\textwidth]{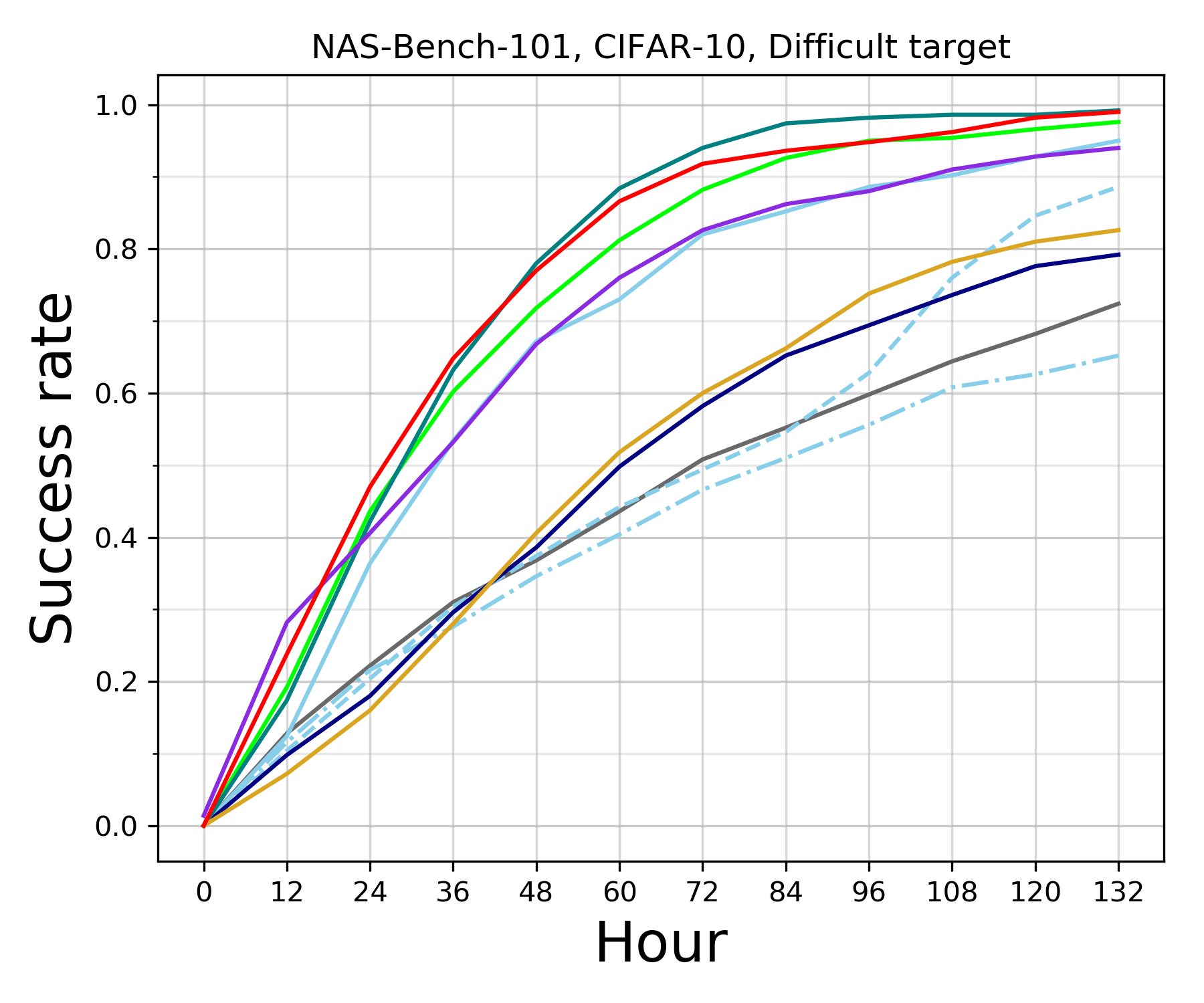}
		\hfill
		\includegraphics[width=.23\textwidth]{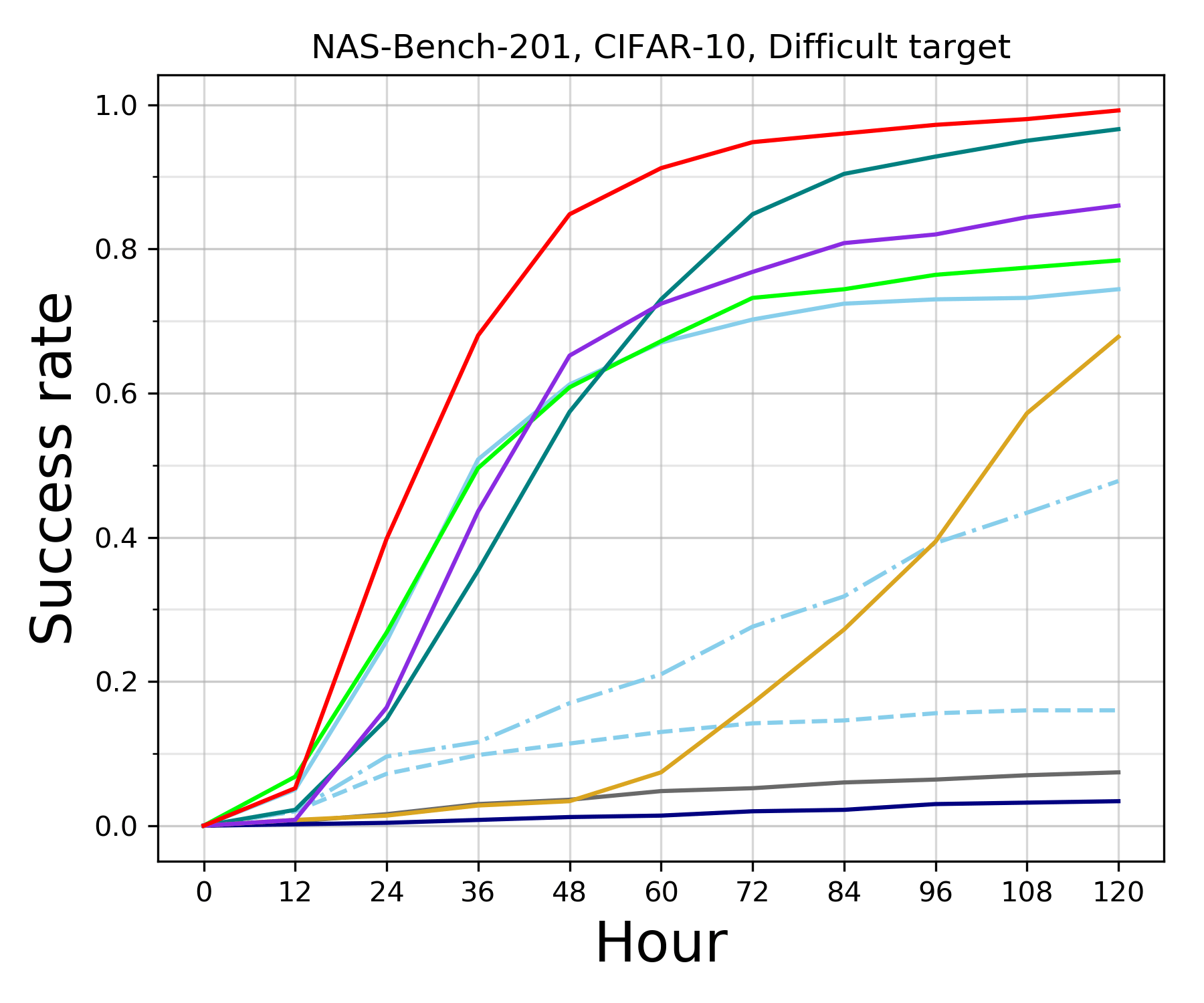}
		\hfill
		\includegraphics[width=.23\textwidth]{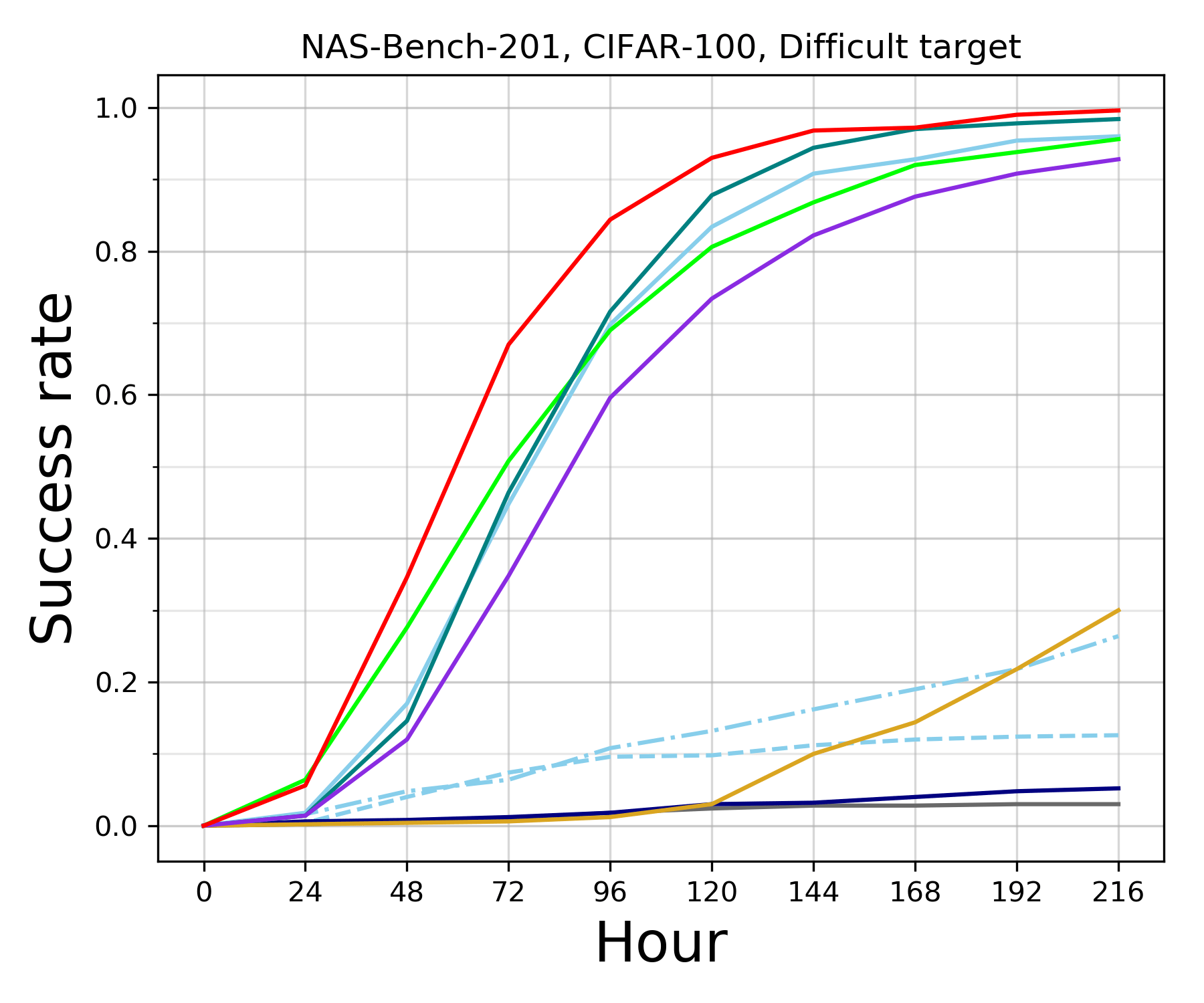}
		\hfill
		\includegraphics[width=.23\textwidth]{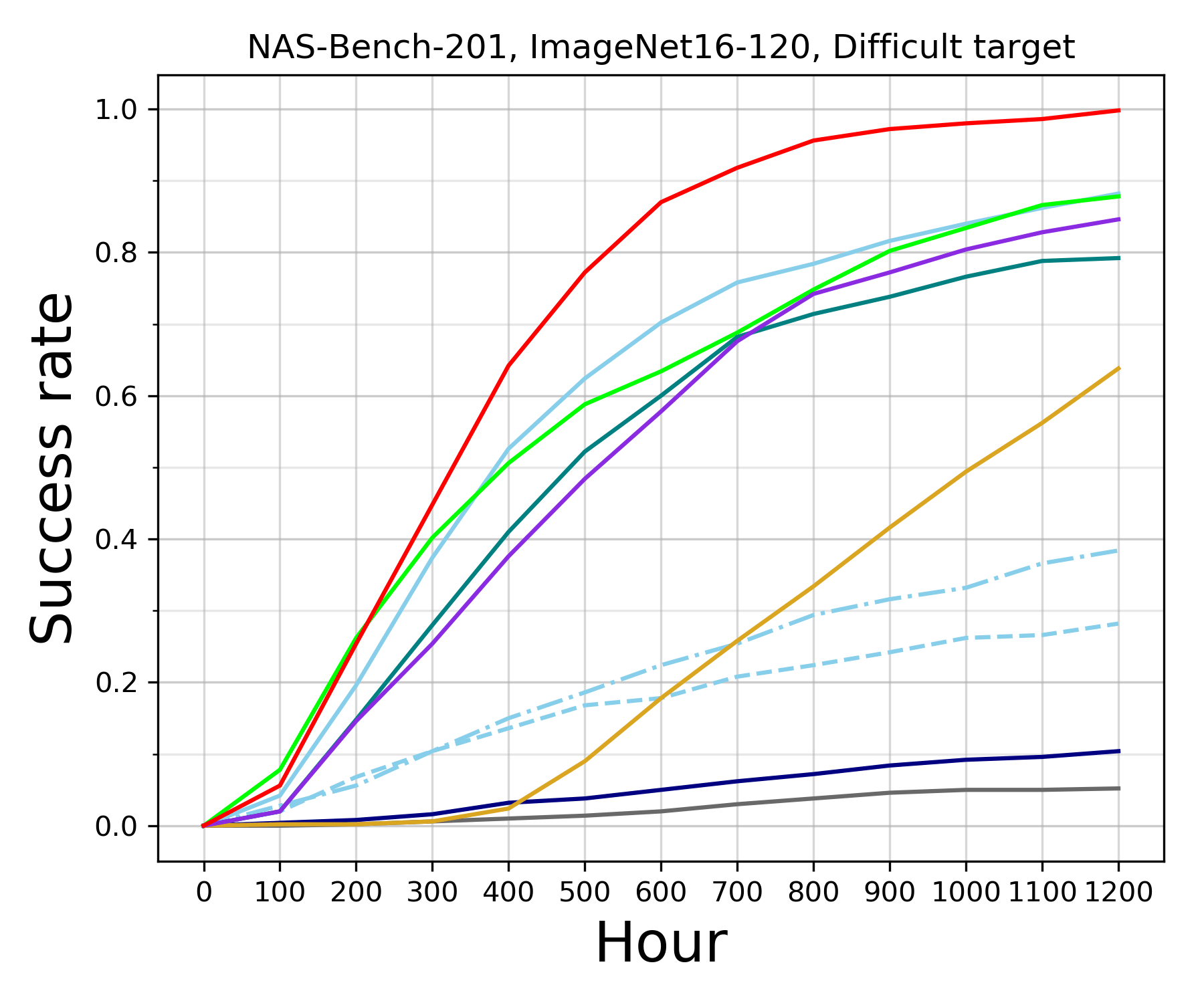}
		\hfill
		\caption{NAS-Bench tasks.}
		%\label{fig2d-a:b}	
	\end{subfigure}
	\\[10ex]
	\begin{subfigure}[b]{\textwidth}		
		\includegraphics[width=.23\textwidth]{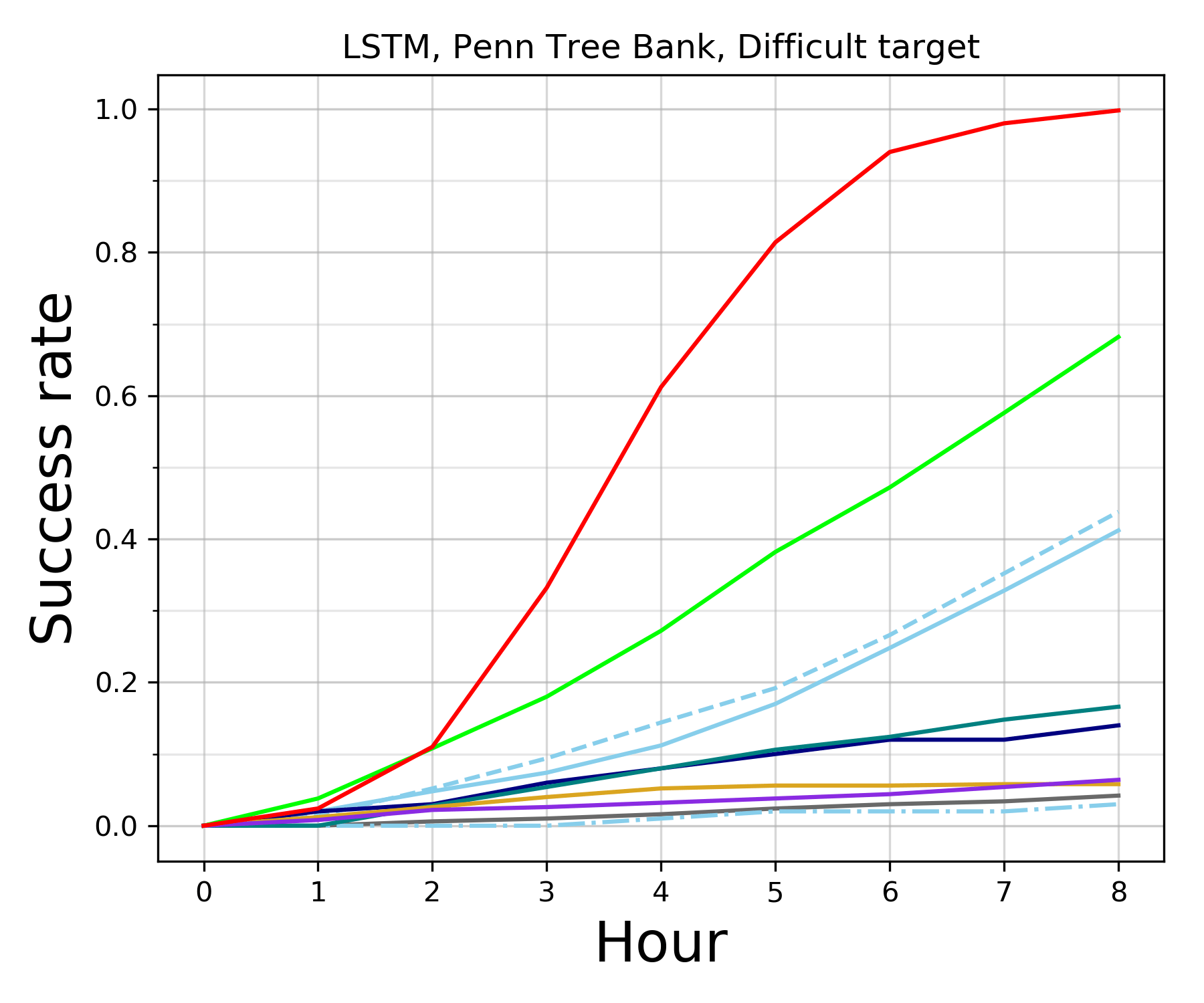}
		\hfill
		\includegraphics[width=.23\textwidth]{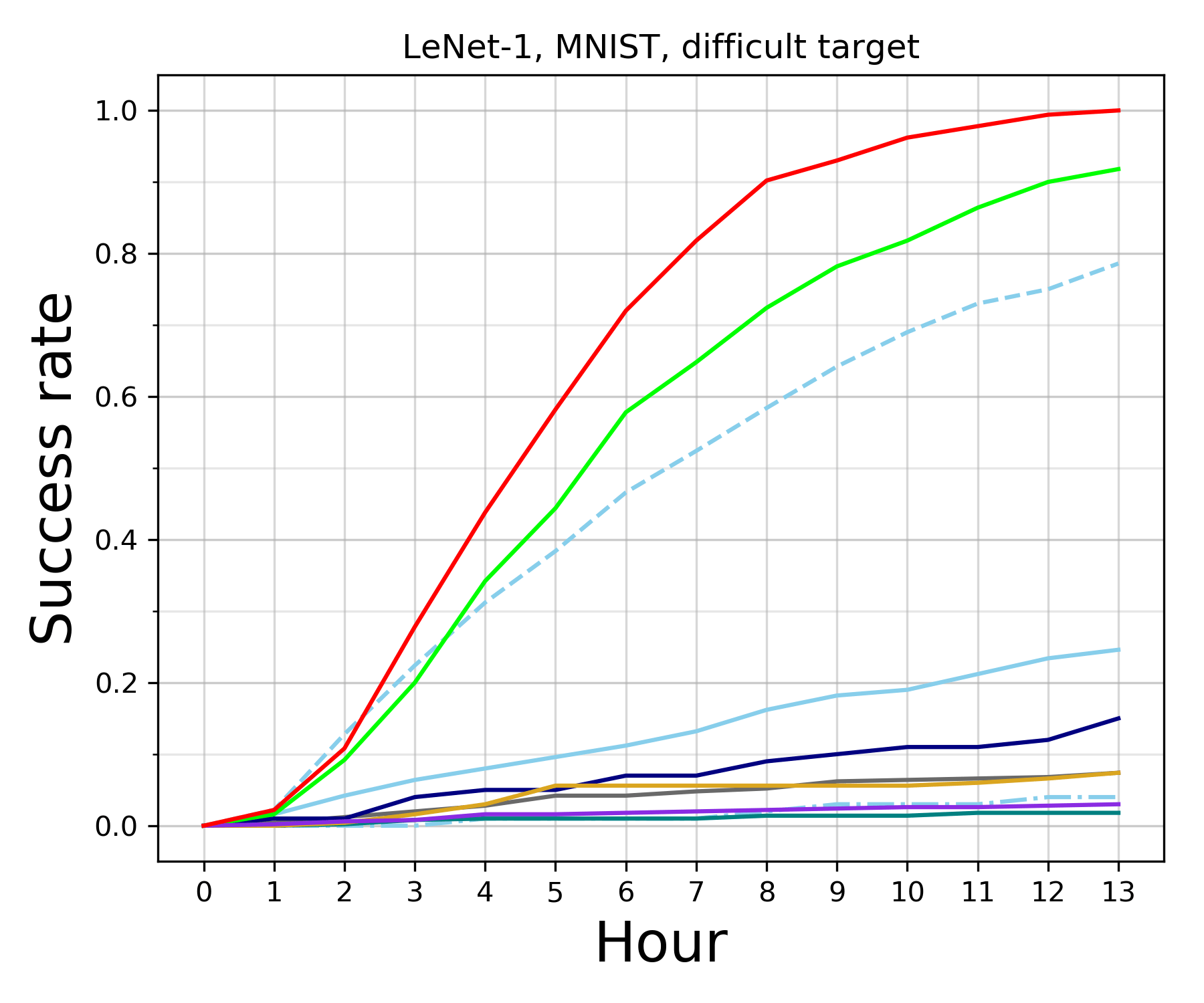}
		\hfill
		\includegraphics[width=.23\textwidth]{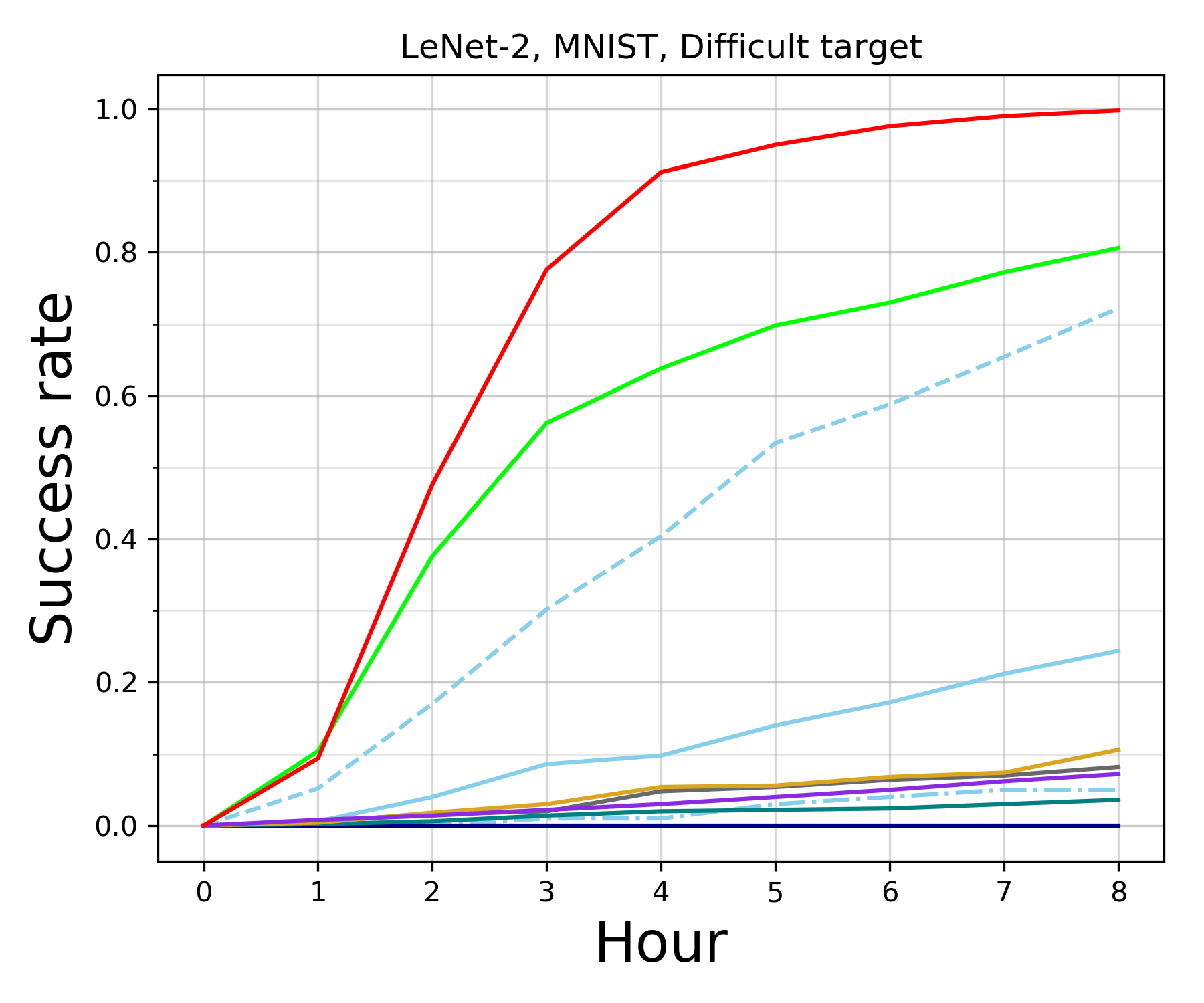}
		\hfill
		
		\vspace*{5ex}
		
		\includegraphics[width=.23\textwidth]{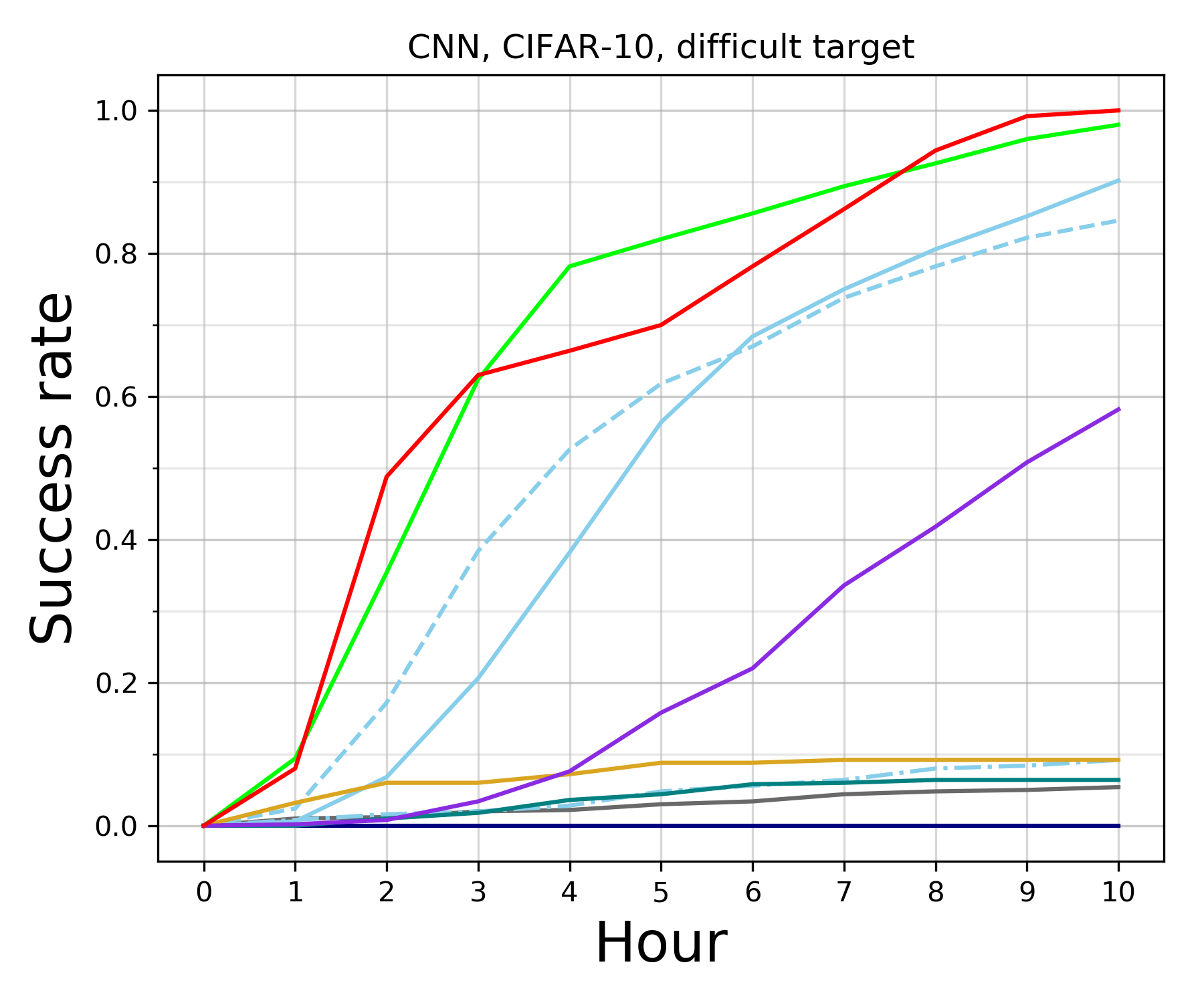}
		\hfill
		\includegraphics[width=.23\textwidth]{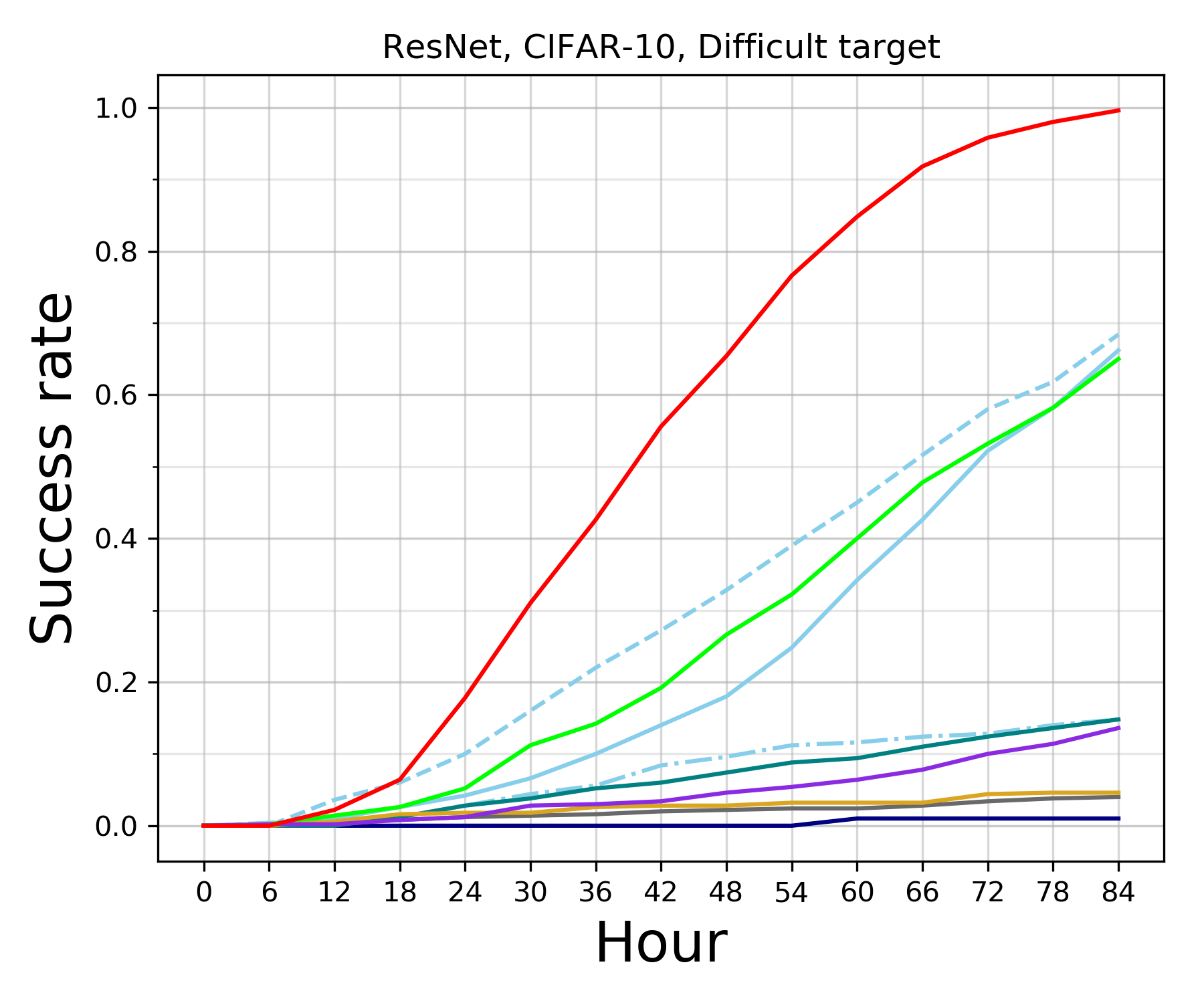}
		\hfill
		\includegraphics[width=.23\textwidth]{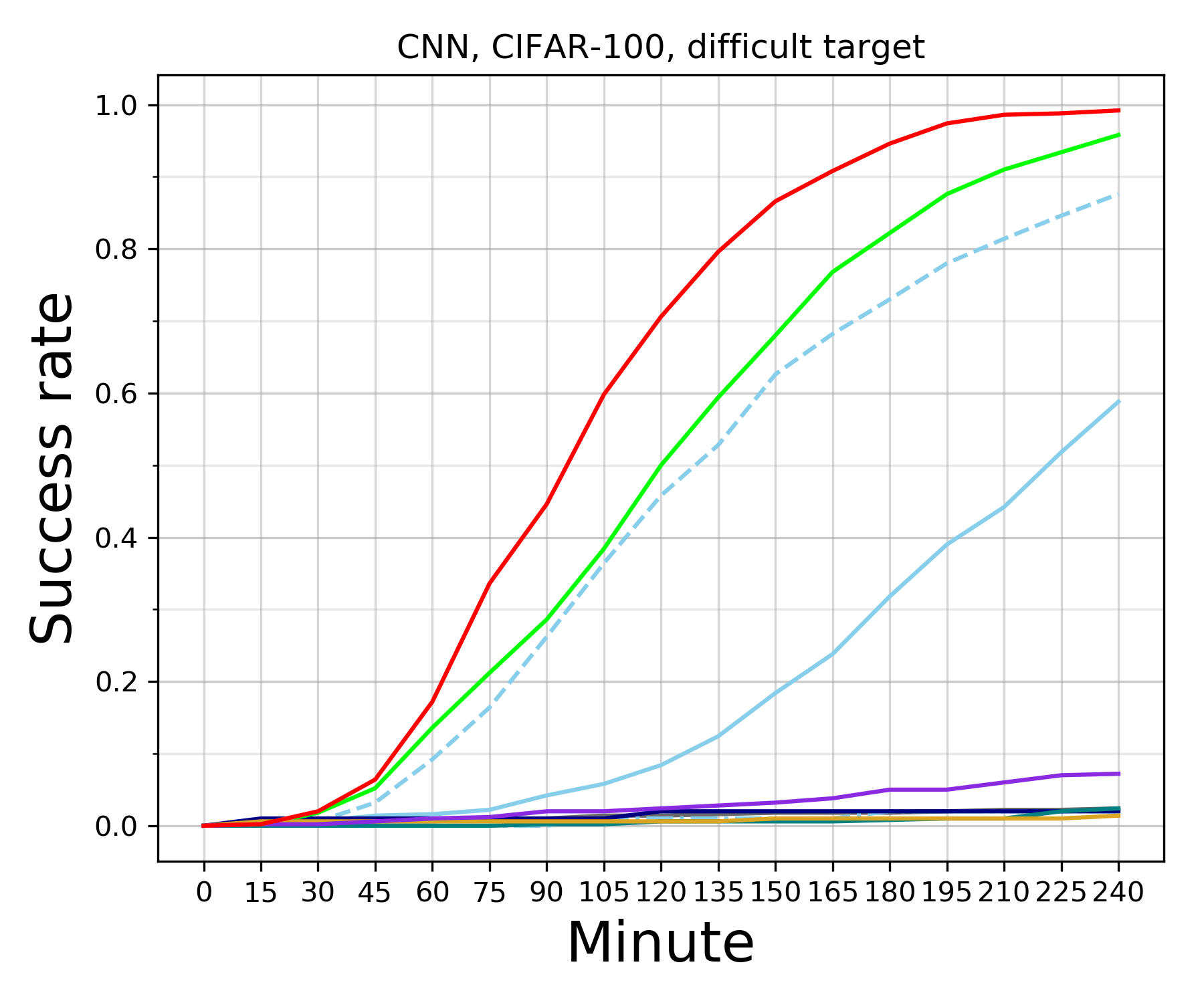}
		\hfill			
		\caption{DNN-Bench tasks.}
		%\label{fig2d-a:c}	
	\end{subfigure}
	\\[5ex]
	\caption
	{Comparison of the success rate for difficult target $\mathbb{P}(\tau \le t_d)$, plotted with $t$ as the horizontal axis.}
	\label{fig2d-a}
\end{figure*}

\newpage

\clearpage

\newpage

\begin{figure*}[!ht]
	
	\begin{subfigure}[b]{\textwidth}
		\vspace*{5ex}
		\includegraphics[width=.23\textwidth]{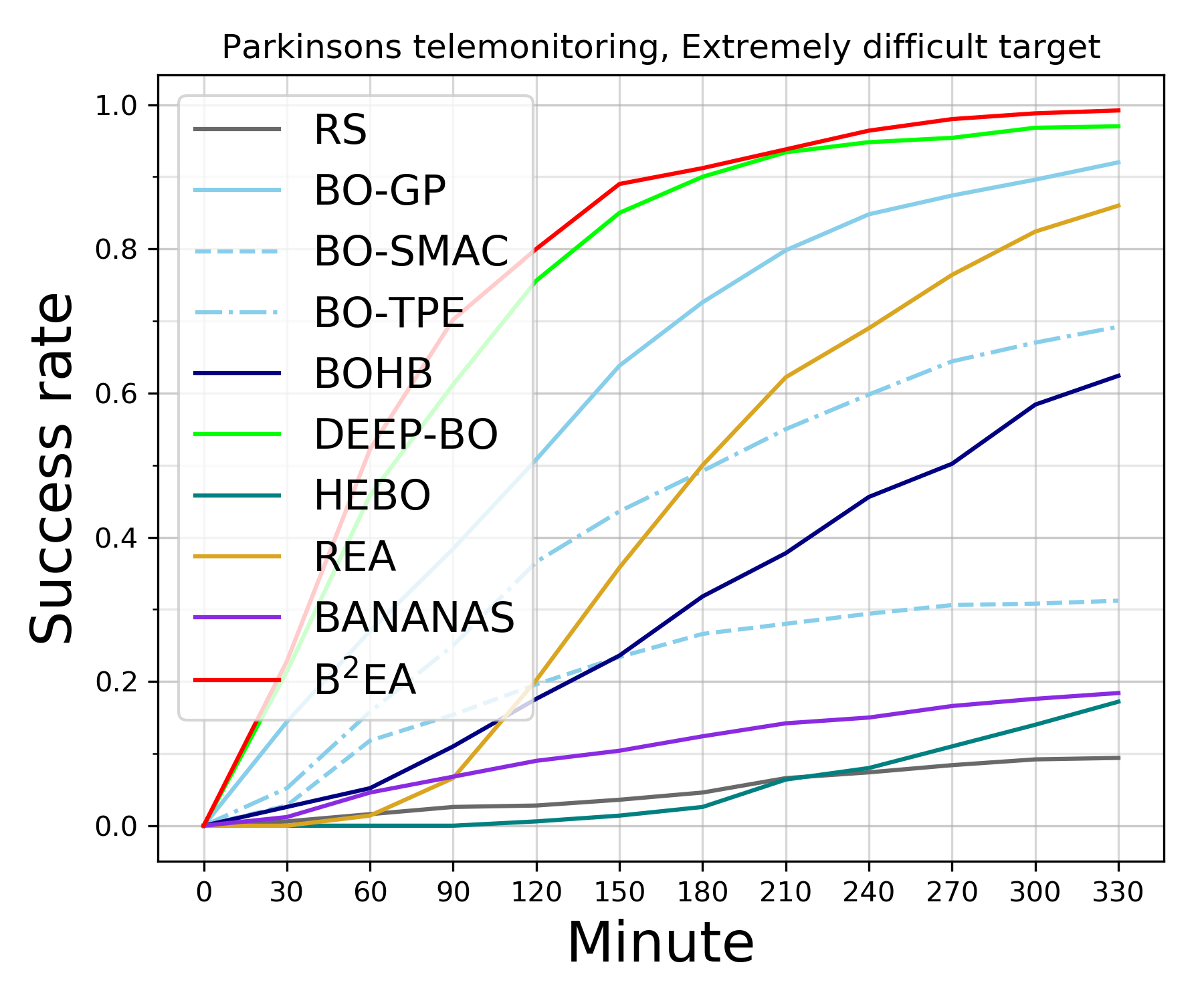}
		\hfill
		\includegraphics[width=.23\textwidth]{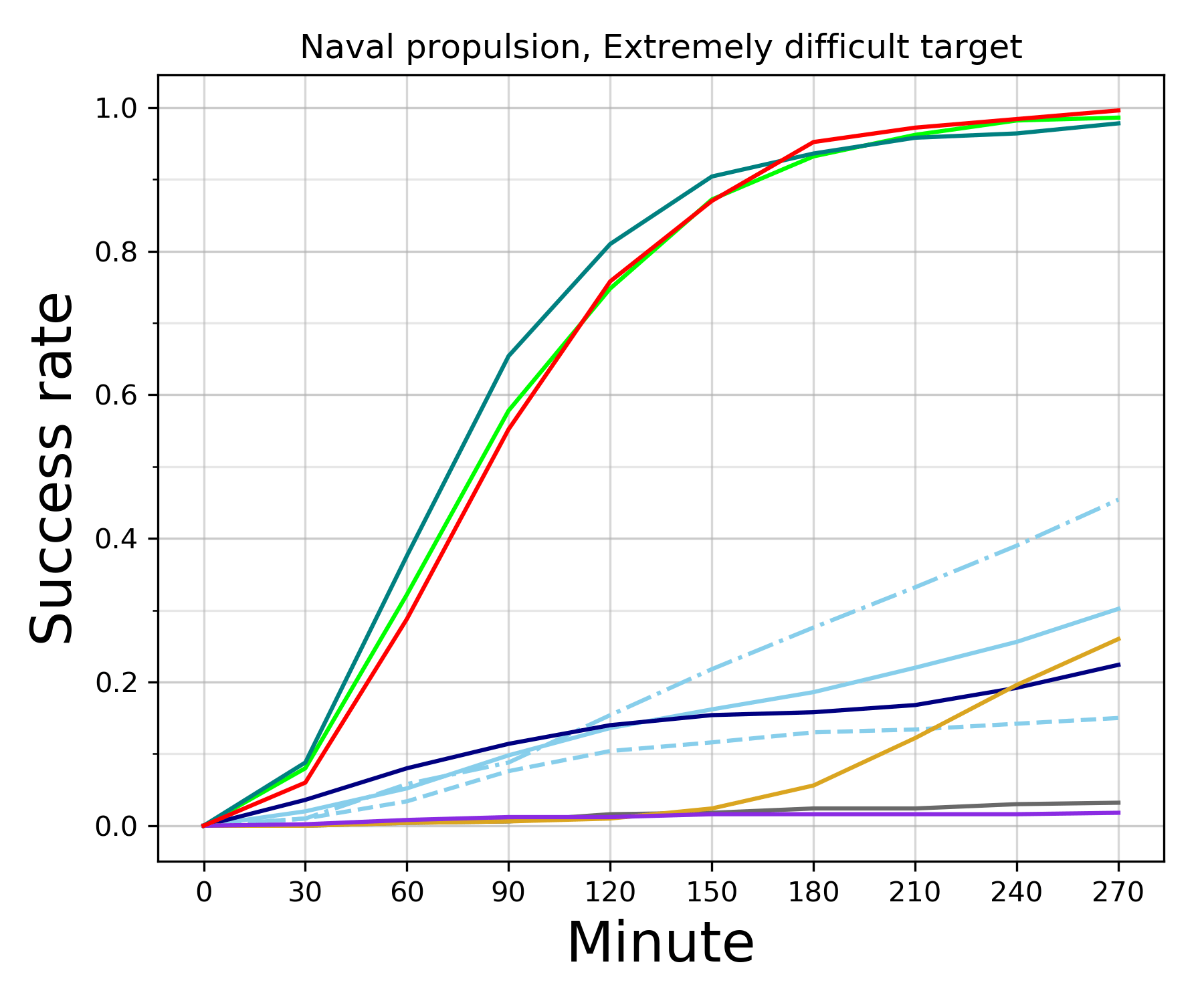}
		\hfill
		\includegraphics[width=.23\textwidth]{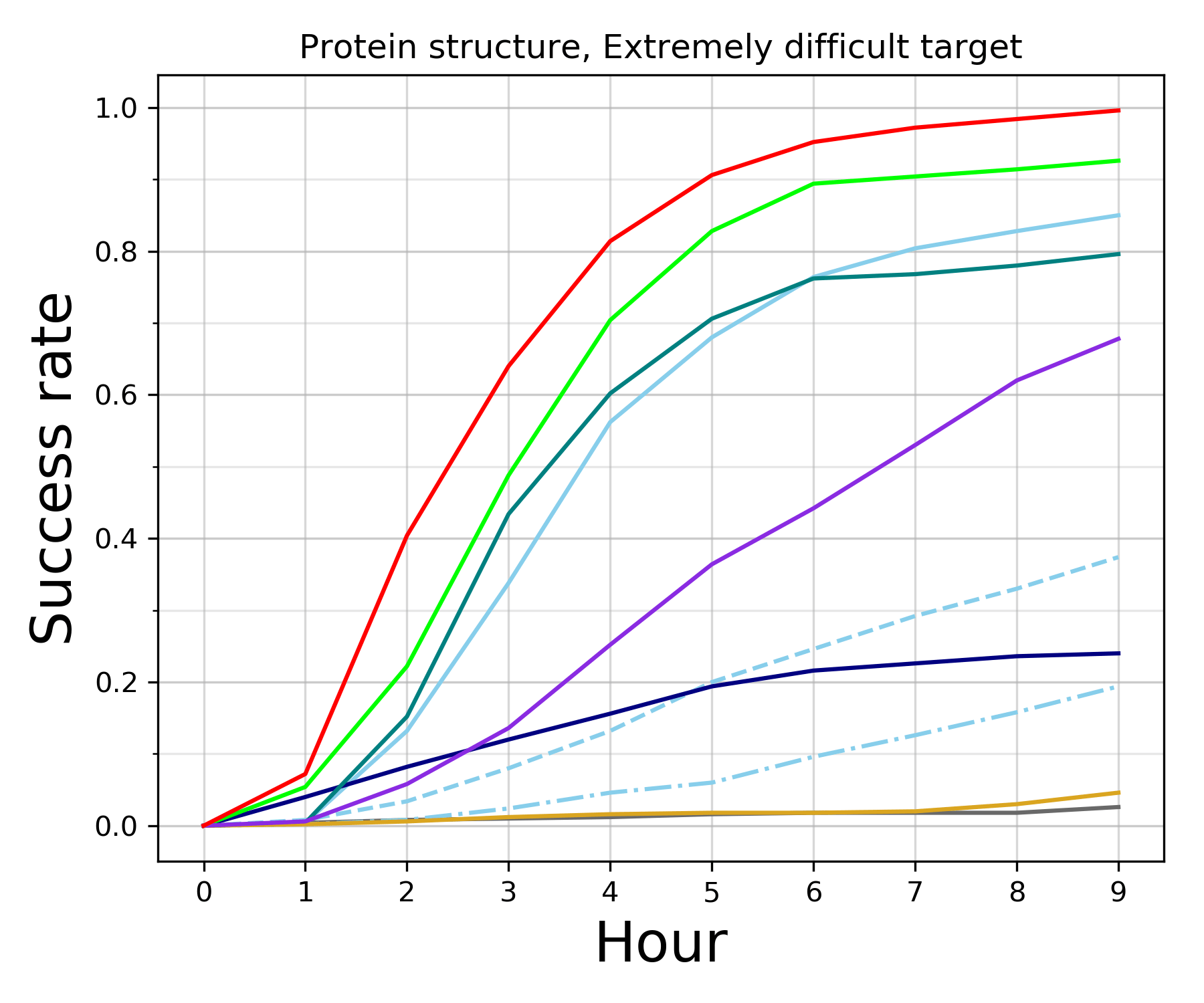}
		\hfill
		\includegraphics[width=.23\textwidth]{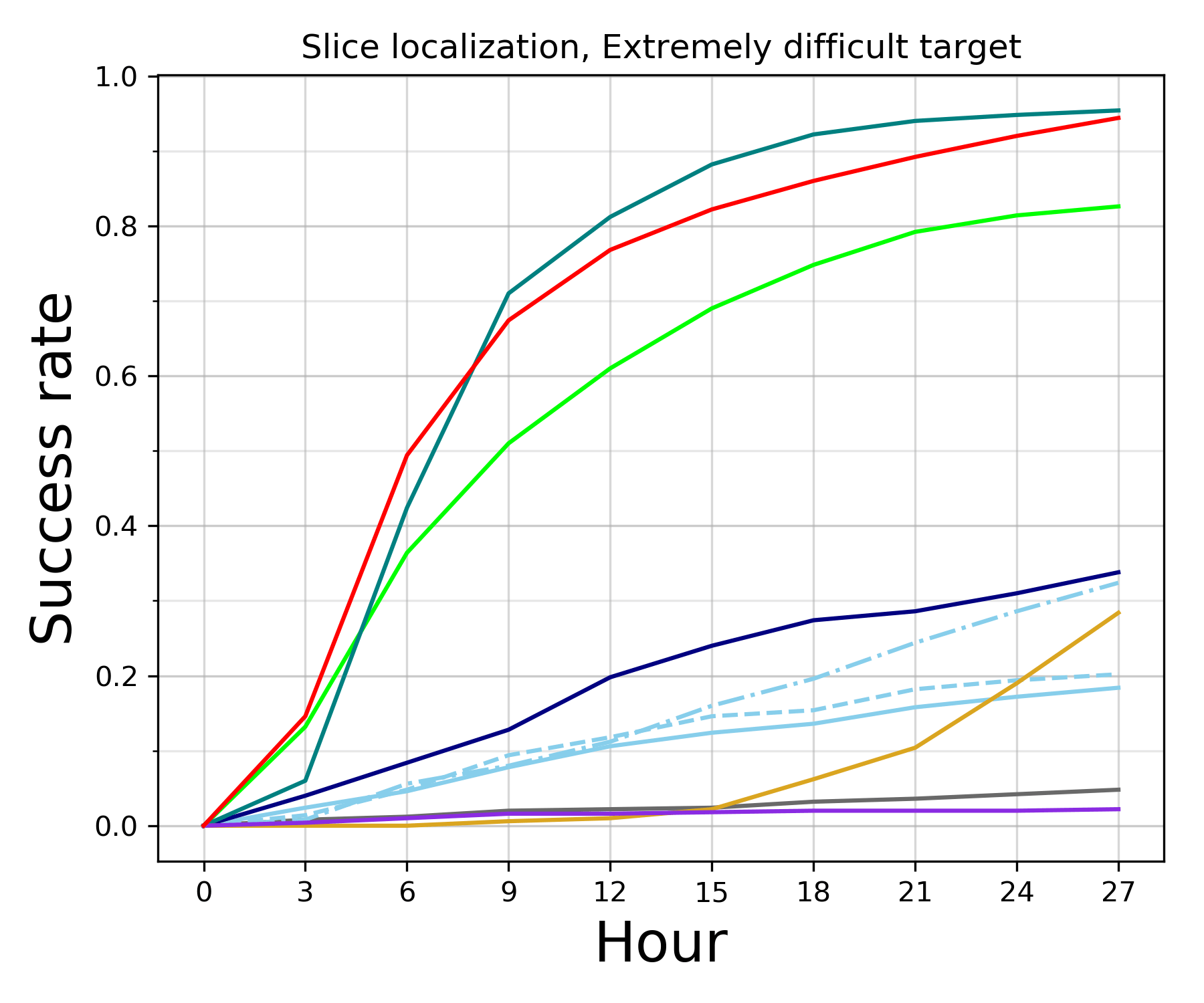}
		\hfill
		\caption{HPO-Bench tasks.}
		%\label{fig2d-a:a}
	\end{subfigure}
	\\[10ex]
	\begin{subfigure}[b]{\textwidth}		
		\includegraphics[width=.23\textwidth]{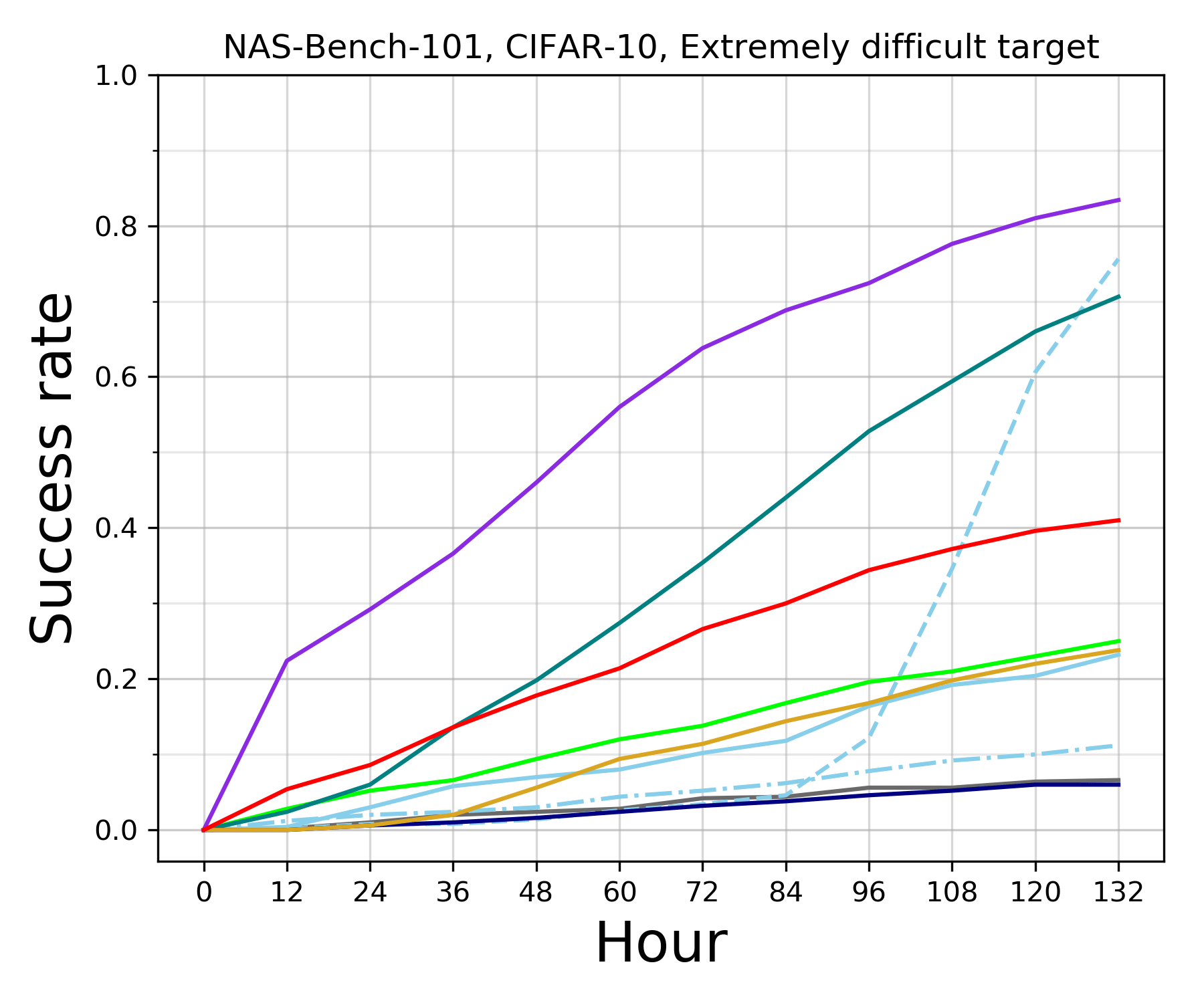}
		\hfill
		\includegraphics[width=.23\textwidth]{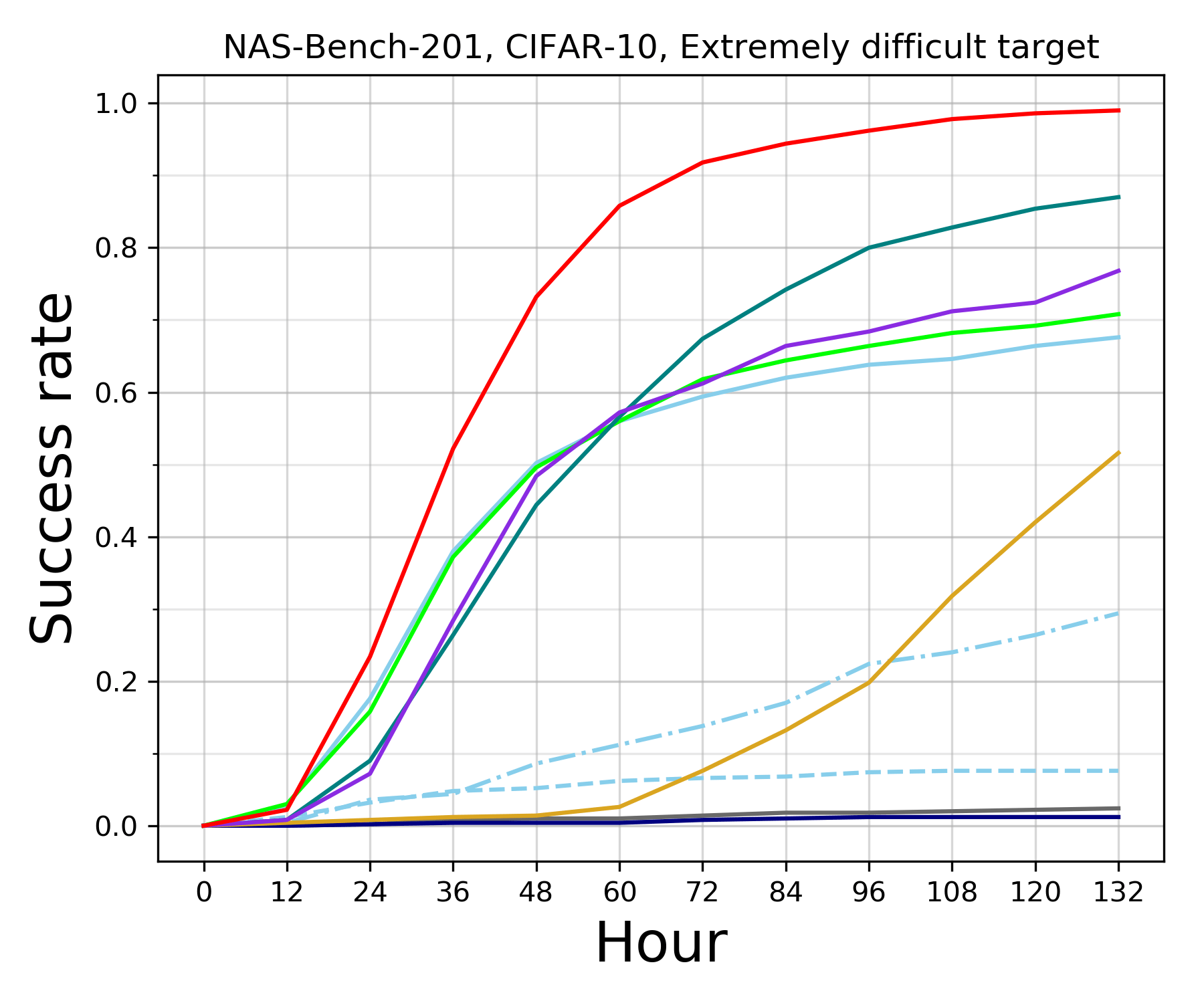}
		\hfill
		\includegraphics[width=.23\textwidth]{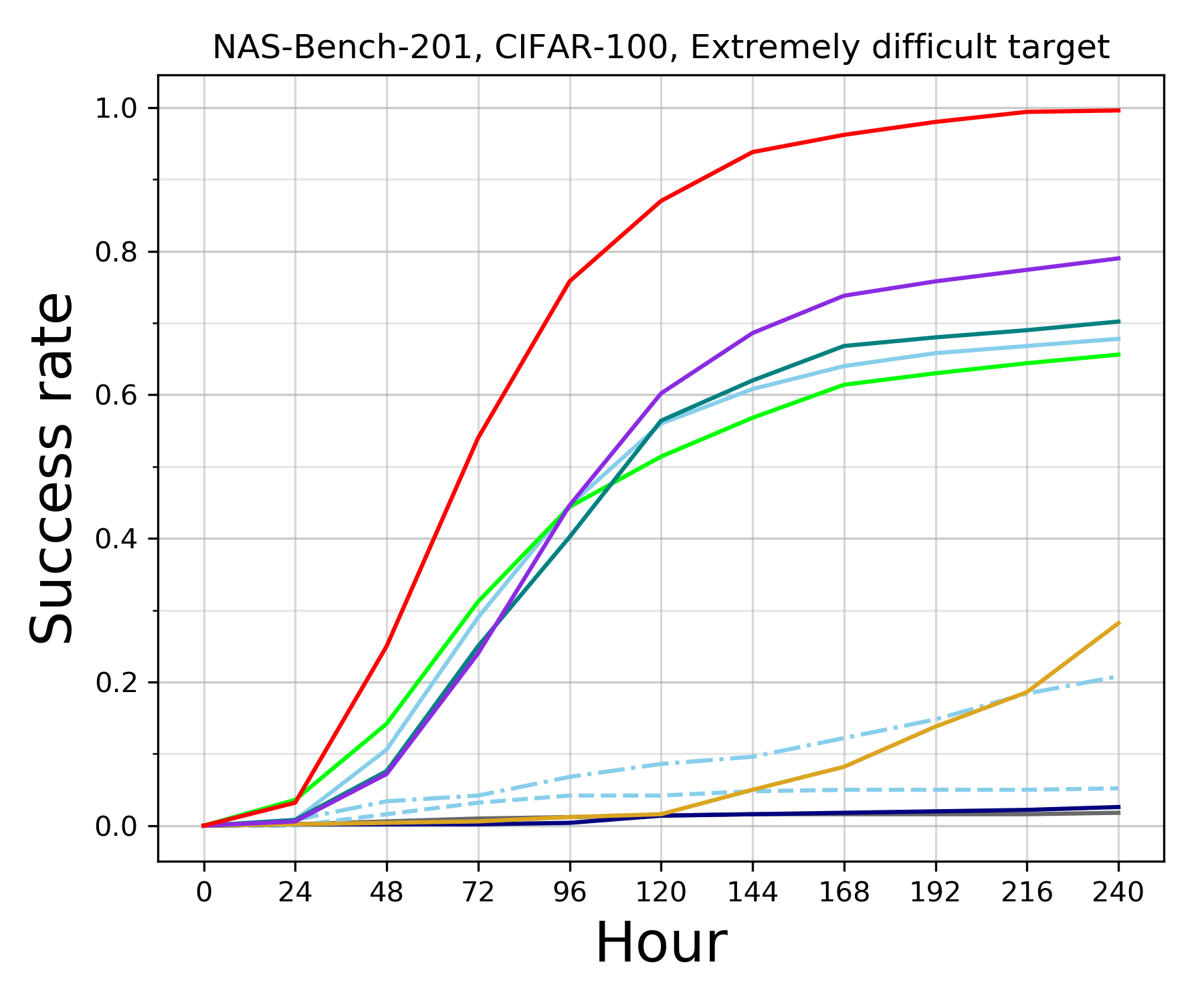}
		\hfill
		\includegraphics[width=.23\textwidth]{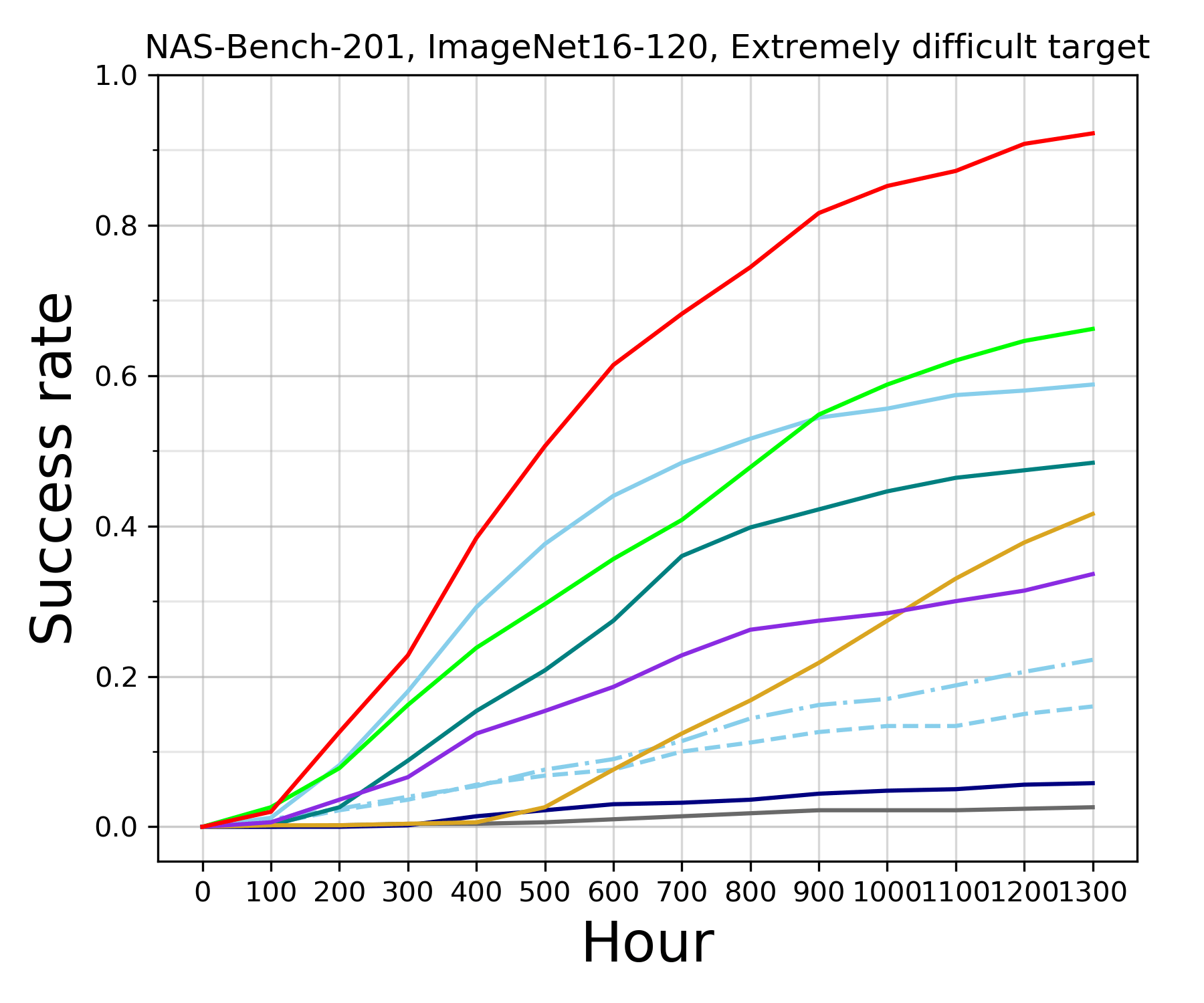}
		\hfill
		\caption{NAS-Bench tasks.}
		%\label{fig2d-a:b}	
	\end{subfigure}
	\\[10ex]
	\begin{subfigure}[b]{\textwidth}		
		\includegraphics[width=.23\textwidth]{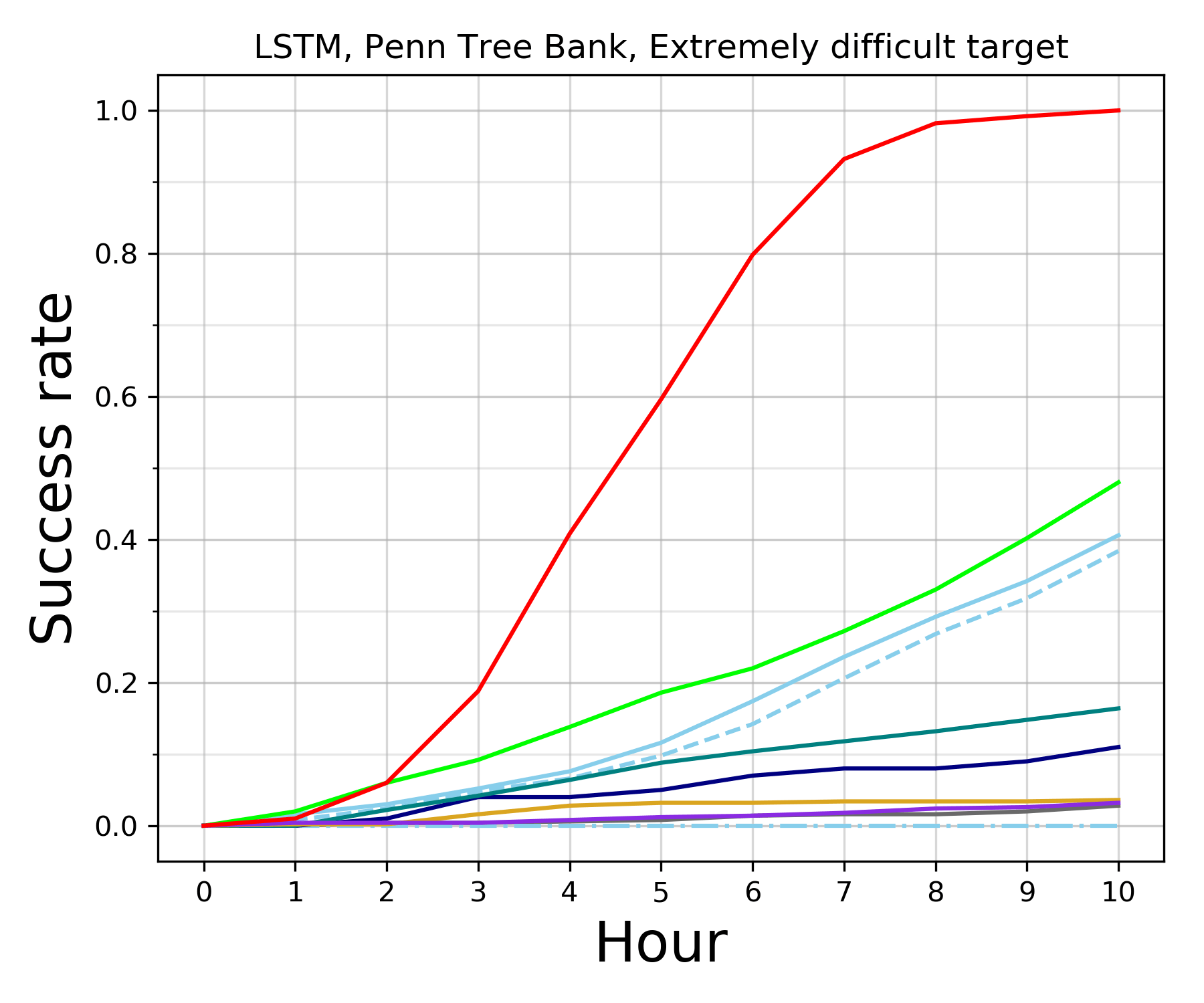}
		\hfill
		\includegraphics[width=.23\textwidth]{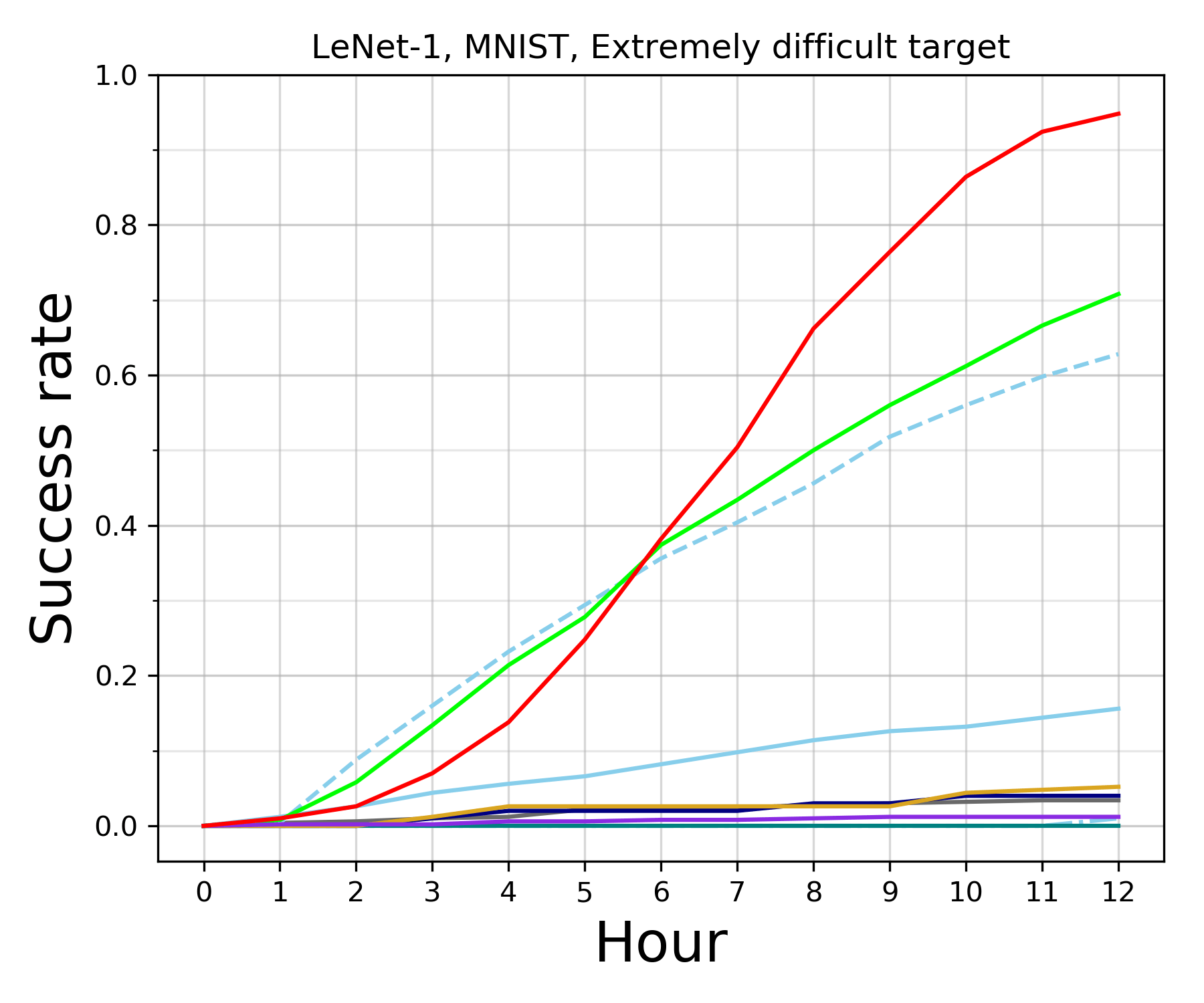}
		\hfill
		\includegraphics[width=.23\textwidth]{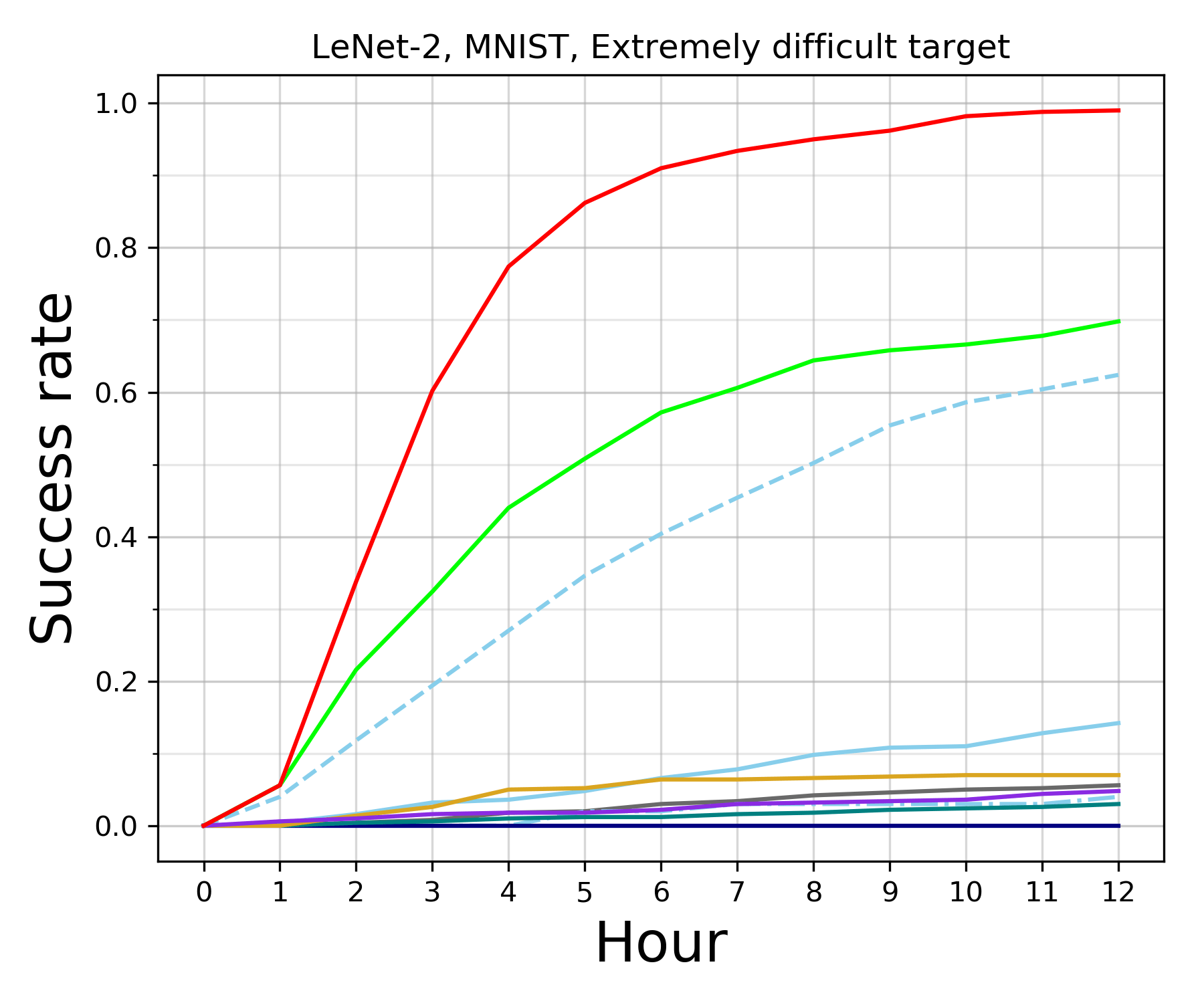}
		\hfill
		
		\vspace*{5ex}
		
		\includegraphics[width=.23\textwidth]{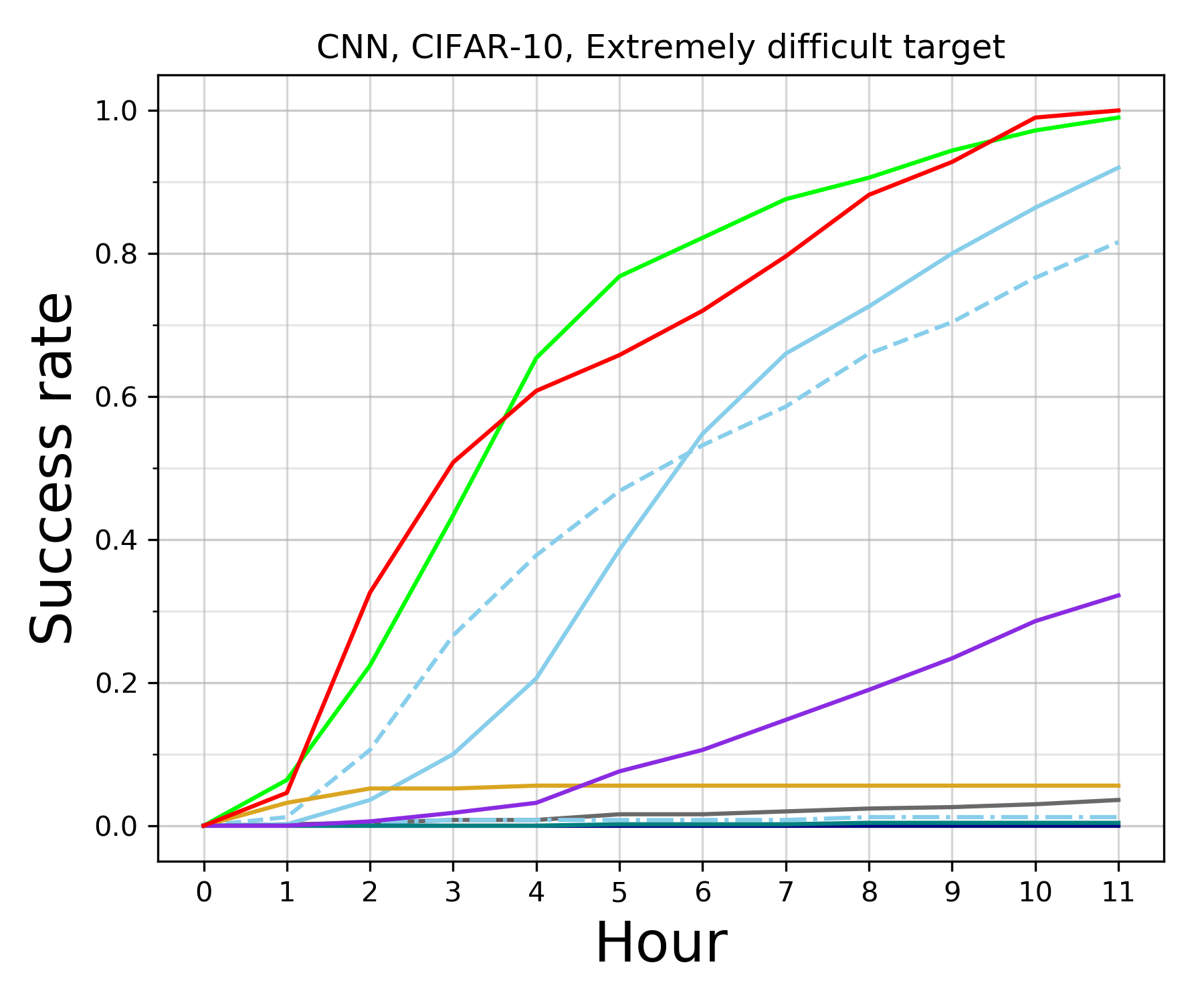}
		\hfill
		\includegraphics[width=.23\textwidth]{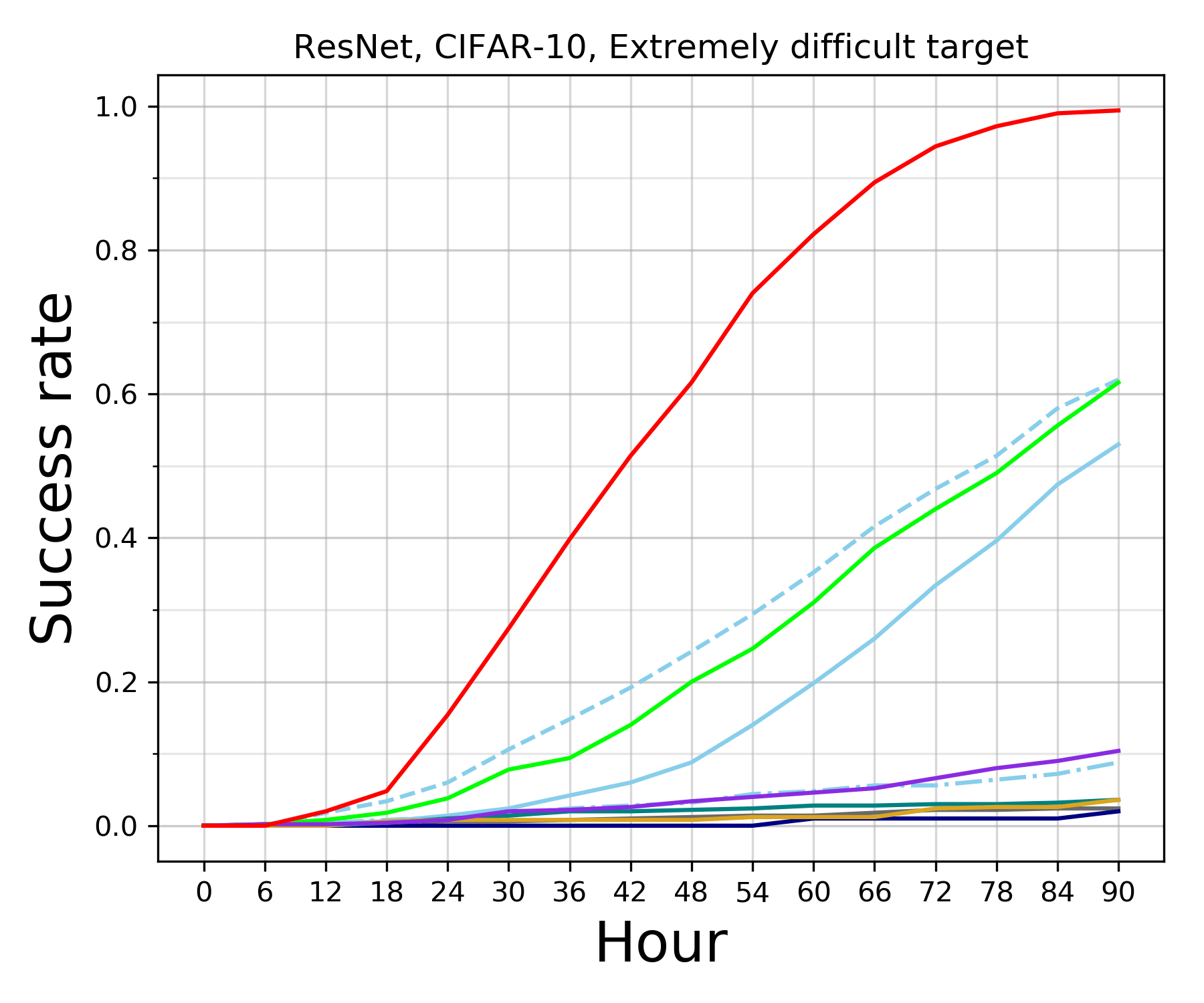}
		\hfill
		\includegraphics[width=.23\textwidth]{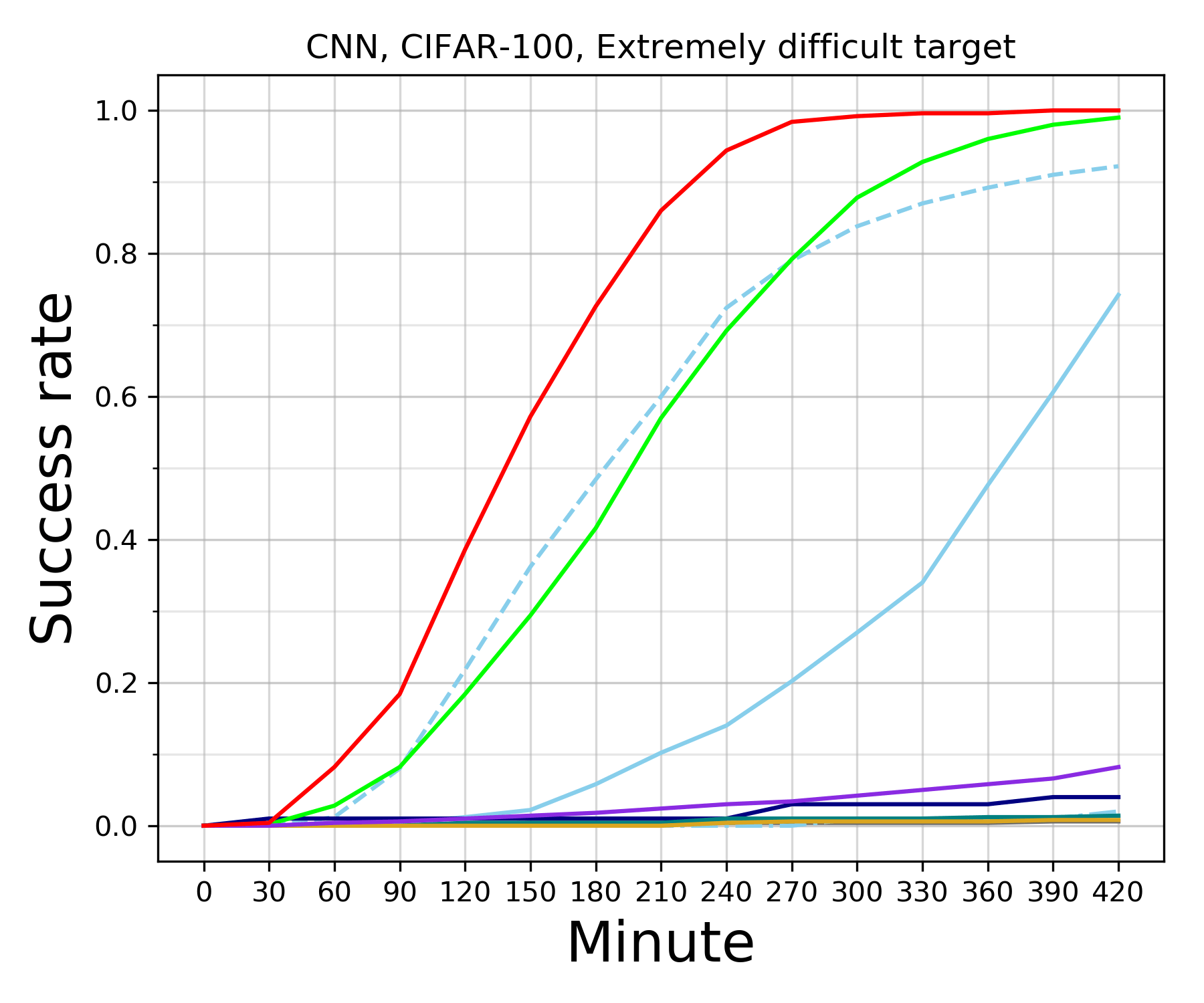}
		\hfill			
		\caption{DNN-Bench tasks.}
		%\label{fig2d-a:c}	
	\end{subfigure}
	\\[5ex]
	\caption
	{Comparison of the success rate for extremely difficult target $\mathbb{P}(\tau \le t_x)$, plotted with $t$ as the horizontal axis.}
	%\label{fig2e-a}
\end{figure*}

\bibliographystyle{unsrtnat}

\end{document}